\documentclass[11pt]{article}
\usepackage[margin=1in]{geometry}

\RequirePackage{amsthm,amsmath,amsfonts,amssymb}
\RequirePackage[numbers,sort&compress]{natbib}

\RequirePackage[colorlinks,citecolor=blue,urlcolor=blue]{hyperref}
\RequirePackage{graphicx}
\usepackage{xcolor}
\usepackage{mathtools}
\usepackage{enumerate}
\usepackage{makecell}
\usepackage{titletoc}
\usepackage{capt-of}
\usepackage{bm}
\usepackage{tikz}
\usetikzlibrary{arrows.meta,calc}

\theoremstyle{plain}

\newtheorem{theorem}{Theorem}[section]
\newtheorem{lemma}[theorem]{Lemma}
\newtheorem{proposition}[theorem]{Proposition}
\newtheorem{corollary}{Corollary}
\newtheorem{assumption}{Assumption}

\theoremstyle{remark}

\newtheorem*{remark}{Remark}

\newcommand{\ddel}{d_\delta}

\newcommand{\gw}[1]{\mathbb{W}({#1})}
\newcommand{\vary}{{\nu}_{Y}^2}
\newcommand{\varyinv}{{\nu}_Y^{-2}}
\newcommand{\varyinvone}{{\nu}_Y^{-1}}
\newcommand{\tsigmagap}{{\kappa}_{\text{d-gap}}}
\newcommand{\tsigmagapcip}{{\kappa}_{\text{c-gap}}}
\newcommand{\tsigmamin}{{\kappa}_{\text{min}}}

\newcommand{\teta}{\widetilde{\eta}}

\newcommand{\tsigma}{\widetilde{\sigma}}
\newcommand{\tsigmapool}{\tsigma_{\textnormal{Pool}}}

\newcommand{\rcip}{r_{\textsf{aw}}}
\newcommand{\Vcip}{V_{\textsf{aw}}}
\newcommand{\Qcip}{Q_{\textsf{aw}}}
\newcommand{\Qcipaug}{Q_{\textsf{aug}}}
\newcommand{\Qhcipaug}{\widehat{Q}_{\textsf{aug}}}
\newcommand{\Vcipaug}{V_{\textsf{aug}}}
\newcommand{\Vhcipaug}{\widehat{V}_{\textsf{aug}}}
\newcommand{\Qhcip}{\widehat{Q}_{\textsf{aw}}}
\newcommand{\Vhcip}{\widehat{V}_{\textsf{aw}}}
\newcommand{\betahftcip}{\widehat{\beta}_{\text{FT-CIP}}}
\newcommand{\betahftconn}{\widehat{\beta}_{\text{FT-OLS}}}
\newcommand{\rconn}{r_{\textsf{sc}}}
\newcommand{\rdip}{r_{\textsf{ca}}}
\newcommand{\Vdip}{V}
\newcommand{\Qdip}{Q}
\newcommand{\Vhdip}{\widehat{V}}
\newcommand{\Qhdip}{\widehat{Q}}
\newcommand{\betahftdip}{\widehat{\beta}_{\text{FT-DIP}}}
\newcommand{\dataset}{\mathcal{D}}
\newcommand{\distri}{\mathcal{P}}

\newcommand{\discrepancy}[2]{\mathfrak{D}\left({#1},{#2}\right)}
\newcommand {\probdiv}[2]{\mathfrak{D}\left({#1},{#2}\right)}

\newcommand{\tagm}[1]{{(#1)}}
\newcommand{\tagtar}{{(0)}}

\newcommand{\diam}[1]{\textnormal{diam}{\left({#1}\right)}}
\newcommand{\cover}[2]{\mathcal{N}\left({#1}, {#2}\right)}

\newcommand{\ntar}{n^\tagtar}
\newcommand{\ntarnew}{n_{\textsf{new}}^\tagtar}
\newcommand{\ntarval}{n^{\tagtar}_{\textnormal{val}}}
\newcommand{\ntaru}{n_{\textnormal{u}}^\tagtar}
\newcommand{\nsrc}{n^\tagm{m}}
\newcommand{\nsrcu}{n_{\textnormal{u}}^\tagm{m}}

\newcommand{\nsrcone}{n^{(1)}}
\newcommand{\nsrctwo}{n^{(2)}}
\newcommand{\nsrcend}{n^{(\mcip)}}
\newcommand{\nsrconeu}{n_{\textnormal{u}}^{(1)}}

\newcommand{\Thetasmall}{\Theta_{\textnormal{lo}}}
\newcommand{\Thetasmalli}[1]{\Theta_{\textnormal{lo},{#1}}}

\newcommand{\h}

\newcommand{\betahatfinal}{\widehat{\beta}_{\textnormal{final}}}
\newcommand{\thetahatfinal}{\widehat{\theta}_{\textnormal{final}}}

\newcommand{\Dtar}{\dataset^\tagtar}
\newcommand{\Dtarval}{\dataset^\tagtar_{\textnormal{val}}}
\newcommand{\Dsrc}{\dataset^\tagm{m}}
\newcommand{\Dsrcone}{\dataset^{(1)}}

\newcommand{\Dtaru}{\dataset_{\textnormal{u}}^\tagtar}

\newcommand{\Dsrcu}{\dataset_{\textnormal{u}}^\tagm{m}}
\newcommand{\Dsrconeu}{\dataset_{\textnormal{u}}^{(1)}}

\newcommand{\Ptar}{\distri^{\tagtar}}

\newcommand{\Psrc}{\distri^{(m)}}

\newcommand{\ones}[1]{\mathbf{1}_{{#1}}}

\makeatletter
\newcommand*{\rom}[1]{\expandafter\@slowromancap\romannumeral #1@}
\makeatother

\newcommand{\assumpref}[1]{Assumption~\ref{assump:#1}}
\newcommand{\assumpsref}[1]{Assumptions~\ref{assump:#1}}
\newcommand{\assumpssref}[1]{\ref{assump:#1}}
\newcommand{\figref}[1]{Figure~\ref{fig:#1}}

\newcommand{\secref}[1]{Section~\ref{sec:#1}}

\newcommand{\appref}[1]{Appendix~\ref{app:#1}}

\newcommand{\lemref}[1]{Lemma~\ref{lem:#1}}

\newcommand{\propref}[1]{Proposition~\ref{prop:#1}}

\newcommand{\thmref}[1]{Theorem~\ref{thm:#1}}

\newcommand{\thmsref}[1]{Theorems~\ref{thm:#1}}
\newcommand{\thmssref}[1]{\ref{thm:#1}}
 \newcommand{\corrref}[1]{Corollary~\ref{cor:#1}}
\newcommand{\tabref}[1]{Table~\ref{tab:#1}}

\newcommand{\eqnref}[1]{\eqref{eqn:#1}}

\DeclareMathOperator{\diag}{diag}

\DeclareMathOperator{\cov}{Cov}
\DeclareMathOperator{\var}{Var}

\DeclareMathOperator{\spn}{span}

\DeclareMathOperator*{\argmin}{arg\,min}

\newcommand{\defn}{\coloneqq}

\newcommand{\iid}[0]{i.i.d. }

\newcommand{\inner}[2]{\left\langle {#1}, {#2} \right\rangle}
\newcommand{\abs}[1]{\left| {#1} \right|}
\newcommand{\norm}[1]{\lVert{#1}\rVert}

\newcommand{\PP}[1]{\mathbb{P}\left\{{#1}\right\}} 
\newcommand{\EE}[1]{\mathbb{E}\left[{#1}\right]} 

\newcommand{\EEst}[2]{\mathbb{E}\left[{#1}\ \middle| \ {#2}\right]} 
\newcommand{\VV}[1]{\var\left({#1}\right)} 

\renewcommand{\O}[1]{\mathcal{O}\left({#1}\right)}

\def\R{\mathbb{R}}

\newcommand{\ident}{\mathbf{I}}

\def\independenT#1#2{\mathrel{\rlap{$#1#2$}\mkern2mu{#1#2}}}
\newcommand\independent{\protect\mathpalette{\protect\independenT}{\perp}}
\newcommand{\iidsim}{\stackrel{\mathrm{i.i.d.}}{\sim}}

\newcommand{\ignore}[1]{}

\newcommand{\eps}{\epsilon}

\newcommand{\Mset}{\mathcal{M}}
\newcommand{\Nset}{\mathcal{N}}
\newcommand{\Rset}{\mathcal{R}}

\newcommand{\betastar}{\beta^\star}

\newcommand{\Astar}{A^\star}

\newcommand{\Pfrak}{\mathfrak{P}}
\newcommand{\Pfraktar}{\mathfrak{P}^{\tagtar}}
\newcommand{\kl}[2]{\mathrm{D}_{\mathrm{KL}}\left(#1\,\|\,#2\right)}

\newcommand{\betatrue}{\beta^{\star}}
\newcommand{\alphatrue}{\alpha^{\star}}

\newcommand{\bstar}{b^\star}
\newcommand{\Atarcon}{A_{\textnormal{tar-con}}}

\newcommand{\alphahdip}{\widehat{\alpha}_{\text{FT-DIP}}}
\newcommand{\btarcon}{b_{\textnormal{tar-con}}}
\newcommand{\betatarcon}{\beta_{\textnormal{tar-con}}}
\newcommand{\alphatarcon}{\alpha_{\textnormal{tar-con}}}
\newcommand{\Uh}{\widehat{U}}
\newcommand{\Adiff}{A_{\textnormal{perb}}}

\newcommand{\betasrconedip}{\beta^{(1)}_{\textnormal{DIP}}}

\newcommand{\betasrconediph}{\widehat{\beta}^{(1)}_{\textnormal{DIP}}}

\newcommand{\betasrconedipt}{\widetilde{\beta}^{(1)}_{\textnormal{DIP}}}

\newcommand{\usrconedipt}{\widetilde{u}^{(1)}_{\textnormal{DIP}}}

\newcommand{\usrconediph}{\widehat{u}^{(1)}_{\textnormal{DIP}}}
\newcommand{\betacip}{\beta_{\textnormal{CIP}}}
\newcommand{\betahcip}{\widehat{\beta}_{\textnormal{CIP}}}
\newcommand{\betaciph}{\widehat{\beta}_{\textnormal{CIP}}}
\newcommand{\betaciptil}{\widetilde{\beta}_{\textnormal{CIP}}}

\newcommand{\ucip}{u_{\textnormal{CIP}}}
\newcommand{\uciptil}{\widetilde{u}_{\textnormal{CIP}}}
\newcommand{\uciph}{\widehat{u}_{\textnormal{CIP}}}
\newcommand{\mcip}{M}
\newcommand{\bhsrc}{\widehat{w}_{\textsf{aw}}^{(m)}}
\newcommand{\bhsrcone}{\widehat{w}_{\textsf{aw}}^{(1)}}
\newcommand{\bhsrctwo}{\widehat{w}_{\textsf{aw}}^{(2)}}

\newcommand{\bhsrcend}{\widehat{w}_{\textsf{aw}}^{(\mcip)}}
\newcommand{\bhsrcendprev}{\widehat{w}_{\textsf{aw}}^{(\mcip-1)}}

\newcommand{\Ph}{\widehat{P}_{\textsf{aw}}}
\newcommand{\Pw}{{P}_{\textsf{aw}}}
\newcommand{\Paug}{{P}_{\textsf{aug}}}

\newcommand{\SigmaXhbar}{\widehat{\overline{\Sigma}}_X}
\newcommand{\SigmaXbar}{\overline{{\Sigma}}_X}
\newcommand{\Gbar}{\overline{G}}
\newcommand{\Ghbar}{\widehat{\overline{G}}}

\newcommand{\thetada}{\theta_{\textnormal{UDA}}}

\newcommand{\thetahdai}[1]{\widehat{\theta}_{\textnormal{UDA},{#1}}}
\newcommand{\thetahfti}[1]{\widehat{\theta}_{\textnormal{FT},{#1}}}
\newcommand{\betahfti}[1]{\widehat{\beta}_{\textnormal{FT},{#1}}}

\newcommand{\fsrc}{f_{\textnormal{LS}}^{(m)}}

\newcommand{\epssrc}{\epsilon^{(m)}}
\newcommand{\epssrcone}{\epsilon^{(1)}}
\newcommand{\epssrctwo}{\epsilon^{(2)}}
\newcommand{\epssrcend}{\epsilon^{(\mcip)}}

\newcommand{\epstar}{\epsilon^{\tagtar}}

\newcommand{\Bsrc}{B^{(m)}}
\newcommand{\Bsrcone}{B^{(1)}}

\newcommand{\Btar}{B^{\tagtar}}
\newcommand{\bsrc}{b^{(m)}}
\newcommand{\bsrcone}{b^{(1)}}

\newcommand{\bsrctwo}{b^{(2)}}

\newcommand{\bsrcmp}{b^{(m')}}
\newcommand{\bsrcmprev}{b^{(m-1)}}
\newcommand{\bsrcprev}{b^{(m-1)}}
\newcommand{\bsrcend}{b^{(\mcip)}}
\newcommand{\bsrcendprev}{b^{(\mcip-1)}}
\newcommand{\tbtar}{\widetilde{b}^{\tagtar}}
\newcommand{\btar}{b^{\tagtar}}

\newcommand{\Hsrc}{H^{(m)}}

\newcommand{\Hsrcone}{H^{(1)}}

\newcommand{\Htar}{H^{\tagtar}}

\newcommand{\nbar}{\overline{n}}
\newcommand{\poly}{\mathrm{poly}}

\newcommand{\betasrc}{\beta^{(m)}_{\textnormal{LS}}}
\newcommand{\betasrcone}{\beta^{(1)}_{\textnormal{LS}}}

\newcommand{\betahsrcone}{\widehat{\beta}^{(1)}_{\textnormal{LS}}}

\newcommand{\fpool}{f_{\textnormal{Pool}}}
\newcommand{\betapool}{\beta_{\textnormal{Pool}}}

\newcommand{\fhpool}{\widehat{f}_{\textnormal{Pool}}}
\newcommand{\betahpool}{\widehat{\beta}_{\textnormal{Pool}}}

\newcommand{\SigmaXsrcone}{\Sigma_X^{(1)}}
\newcommand{\SigmaXsrc}{\Sigma_X^{(m)}}
\newcommand{\SigmaXtar}{\Sigma_X^{\tagtar}}

\newcommand{\SigmaXhsrcone}{\widehat{\Sigma}_X^{(1)}}

\newcommand{\SigmaXhtar}{\widehat{\Sigma}_X^{\tagtar}}
\newcommand{\SigmaW}{\Sigma_W}
\newcommand{\SigmaWh}{\widehat{\Sigma}_W}
\newcommand{\SigmaWt}{\Sigma_{\widetilde{W}}}
\newcommand{\SigmaWht}{\widehat{\Sigma}_{\widetilde{W}}}
\newcommand{\Wt}{\widetilde{W}}
\newcommand{\nt}{\widetilde{n}}
\newcommand{\lamminbar}{\overline{\lambda}_{\textnormal{min}}}
\newcommand{\lammaxbar}{\overline{\lambda}_{\textnormal{max}}}
\newcommand{\kappabar}{\overline{\kappa}}

\newcommand{\lamdipgap}{\lambda_{\textnormal{d-gap}}}
\newcommand{\lamcipgap}{\widetilde{\lambda}_{\textnormal{c-gap}}}
\newcommand{\lamcipauggap}{{\lambda}_{\textnormal{c-gap}}}

\newcommand{\lammax}{\lambda_{\rm max}}
\newcommand{\lammin}{\lambda_{\rm min}}
\newcommand{\lamgap}{\lambda_{\rm gap}}

\newcommand{\risksrc}[1]{\mathcal{R}^{(m)}({#1})}

\newcommand{\risktar}[1]{\mathcal{R}^{\tagtar}({#1})}

\newcommand{\riskhtar}[1]{\widehat{\mathcal{R}}^{\tagtar}({#1})}

\newcommand{\Wtar}{W^{\tagtar}}
\newcommand{\Xtar}{X^{\tagtar}}
\newcommand{\Ytar}{Y^{\tagtar}}
\newcommand{\Ztar}{Z^{\tagtar}}
\newcommand{\Xtarval}{\overline{X}^{\tagtar}}
\newcommand{\Ytarval}{\overline{Y}^{\tagtar}}

\newcommand{\Xtartil}{\widetilde{X}^{\tagtar}}
\newcommand{\Xsrc}{X^{(m)}}
\newcommand{\Xsrctil}{\widetilde{X}^{(m)}}
\newcommand{\Ysrc}{Y^{(m)}}
\newcommand{\Xsrcone}{X^{(1)}}
\newcommand{\Xsrctwo}{X^{(2)}}
\newcommand{\Xsrconetil}{\widetilde{X}^{(1)}}
\newcommand{\Ysrcone}{Y^{(1)}}
\newcommand{\Ysrctwo}{Y^{(2)}}
\newcommand{\Ysrcend}{Y^{(\mcip)}}

\newcommand{\bW}{\textbf{W}}
\newcommand{\bZ}{\textbf{Z}}
\newcommand{\bWtar}{\textbf{W}^{\tagtar}}
\newcommand{\bZtar}{\textbf{Z}^{\tagtar}}
\newcommand{\bYtar}{\textbf{Y}^{\tagtar}}
\newcommand{\bXtar}{\textbf{X}^{\tagtar}}
\newcommand{\bXsrc}{\textbf{X}^{(m)}}
\newcommand{\bYsrc}{\textbf{Y}^{(m)}}
\newcommand{\bXsrcone}{\textbf{X}^{(1)}}
\newcommand{\bYsrcone}{\textbf{Y}^{(1)}}

\newcommand{\bYsrctwo}{\textbf{Y}^{(2)}}

\newcommand{\bYsrcend}{\textbf{Y}^{(\mcip)}}

\newcommand{\Xsrcmprev}{{X^{(m-1)}}}

\newcommand{\Ysrcmprev}{{Y^{(m-1)}}}

\newcommand{\Pmartar}[1]{\distri^{\tagtar}_{{#1}}}

\newcommand{\Pmarsrc}[1]{\distri^{(m)}_{{#1}}}
\newcommand{\Pmarsrcone}[1]{\distri^{(1)}_{{#1}}}
\newcommand{\Pcondsrc}[2]{\distri^{(m)}_{{#1} \vert {#2}}}

\newcommand{\Pcondsrcmprev}[2]{\distri^{(m-1)}_{{#1} \vert {#2}}}

\newcommand{\Eset}{\mathcal{E}}
\newcommand{\Fset}{\mathcal{F}}

\newcommand{\Pset}{\mathcal{P}}

\newcommand{\Normal}{\mathcal{N}}
\newcommand{\Ind}{\ensuremath{\mathbb{I}}}

\newcommand{\tPset}{\widetilde{\mathcal{P}}}

\newcommand{\col}[1]{\textnormal{col}({#1})}

\newcommand{\kh}{\widehat{k}}
\newcommand{\Qh}{\widehat{Q}}
\newcommand{\Vh}{\widehat{V}}

\newcommand{\betah}{\widehat{\beta}}

\newcommand{\vecnorm}[2]{\left\| #1\right\|_{#2}}

\newcommand{\opnorm}[1]{\norm{{#1}}}
\newcommand{\vecopnorm}[1]{\left\| #1\right\|}
\newcommand{\vecfronorm}[1]{\left\| #1\right\|_{\textnormal{F}}}
\newcommand{\tdelta}{\widetilde{\delta}}

\newcommand{\binv}{b_{\text{inv}}}
\newcommand{\bnoninv}{b_{\text{noninv}}}
\newcommand{\bsrcnoninv}{b_{\text{non}}^{(m)}}

\long\def\comment#1{}
\definecolor{battleshipgrey}{rgb}{0.52, 0.52, 0.51}
\definecolor{darkgray}{rgb}{0.66, 0.66, 0.66}
\definecolor{darkgreen}{rgb}{0.0, 0.2, 0.13}
\definecolor{darkspringgreen}{rgb}{0.09, 0.45, 0.27}
\definecolor{dukeblue}{rgb}{0.0, 0.0, 0.61}
\definecolor{olivedrab7}{rgb}{0.24, 0.2, 0.12}
\definecolor{darkblue}{rgb}{0.0, 0.0, 0.55}
\definecolor{darkscarlet}{rgb}{0.34, 0.01, 0.1}
\definecolor{candyapplered}{rgb}{1.0, 0.03, 0.0}
\definecolor{ao(english)}{rgb}{0.0, 0.5, 0.0}
\definecolor{applegreen}{rgb}{0.55, 0.71, 0.0}

\newcommand{\brackets}[1]{\left[ #1 \right]}
\newcommand{\parenth}[1]{\left( #1 \right)}

\newcommand{\braces}[1]{\left\{ #1 \right \}}
\newcommand{\abss}[1]{\left| #1 \right |}

\newcommand{\tp}{^\top}

\begin{document}

	\title{When few labeled target data suffice: a theory of semi-supervised domain adaptation via fine-tuning from multiple adaptive starts}

	\author{Wooseok Ha\thanks{Department of Mathematical Sciences, KAIST} , 
	Yuansi Chen\thanks{Department of Mathematics, ETH Z\"{u}rich}}
	\date{}
	
	\maketitle
		
	\begin{abstract}
		Semi-supervised domain adaptation (SSDA) seeks 
		to achieve accurate predictions 
		in a target domain with limited labeled target data 
		by exploiting abundant source and unlabeled target data. 
		We study this problem under structural causal models (SCMs), 
		which provide a statistical framework to describe distribution shifts between source and target domains as interventions in the data-generating process rather than ad hoc changes in model parameters. 
		The central phenomenon is that, under low-dimensional interventions, source and unlabeled target data can help identify the high-dimensional shared structure, leaving only a low-dimensional target-specific correction to be learned from limited labeled target data. We formalize this principle for three canonical intervention models and propose the corresponding SSDA methods FT-DIP, FT-OLS-Src and FT-CIP. Under each intervention model, we demonstrate how extending an unsupervised domain adaptation (UDA) method to SSDA can achieve minimax-optimal target performance with limited target labels, with the labeled-target sample complexity scaling with the intervention dimension rather than the ambient dimension. When the distribution shift is underspecified, we propose the Multi-Adaptive-Start Fine-Tuning (MASFT) algorithm, which fine-tunes from multiple adaptive starts and selects among them using a small target validation set, incurring only logarithmic overhead in the number of starts. We validate the effectiveness of our proposed methods through simulated and real data experiments.
	\end{abstract}
	
	\section{Introduction}


	Classical statistical machine learning models are typically trained under the assumption that training and test data are drawn from the same probability distribution. This supervised learning paradigm is well-founded in statistical learning theory and serves as the basis for many modern applications. However, in real-world scenarios, this assumption often fails due to distribution shifts, where the data distribution in the training (or source) domain differs from that in the test (or target) domain. As a result, models trained under this assumption often experience a significant decline in model performance when deployed in new target domains. For example, modern image classification models, trained on large-scale datasets containing over a million labeled high-resolution images, achieve remarkable classification accuracy~\cite{krizhevsky2012imagenet}, sometimes surpassing human-level performance~\cite{he2015delving}. Nevertheless, these models often struggle to generalize to more challenging test scenarios such as those with a different background context~\cite{geirhos2020shortcut}. Acquiring and labeling millions of images for every possible deployment scenario is impractical and expensive. This motivates the development of domain adaptation (DA) methods, which aim to transfer knowledge from a source domain to a related but new target domain.
	
	A DA setting that has witnessed a recent surge of interest is unsupervised domain adaptation (UDA), where labeled source data is available, but only unlabeled data is observed from the target domain. UDA methods typically address distribution shifts by aligning feature distributions between source and target domains while optimizing source performance, with the hope that the learned common feature representation generalizes well to target data~\cite{baktashmotlagh2013unsupervised,ganin2016domain,courty2016optimal}. While this approach can be highly effective given strong beliefs on the assumed source-target relationship, the absence of target supervision makes UDA inherently unreliable in certain scenarios. There are settings where feature alignment does not necessarily lead to improved target performance~\cite{zhao2019learning}. More importantly, without labeled target data, there is no straightforward way to detect such a failure~\cite{wu2023prominent}. Another DA setting, called domain generalization (DG), trains models that are robust to distribution shifts without access to any labeled or unlabeled target data and prepares for worst-case distribution shifts. However, DG methods tend to be overly conservative, often leading to suboptimal performance when the target distribution is not as adversarial as assumed~\cite{heinze2021conditional,chen2021domain,duchi2021statistics,sagawa2019distributionally}. 
	
	Semi-supervised domain adaptation (SSDA)~\cite{donahue2013semi,saito2019semi} is an intermediate and more realistic DA setting by incorporating a few labeled target examples alongside the source and unlabeled target data. By leveraging a small amount of target supervision, SSDA bridges the gap between UDA, which lacks target labels, and fully supervised learning, which requires abundant labeled target data. For instance, in medical imaging~\citep{loizillon2023semi}, models trained on a large clinical routine dataset in one imaging setting may fail to generalize to data in a slightly different setting, due to differences in imaging protocols. While UDA methods may struggle to address such distribution shifts~\citep{zhao2019learning}, SSDA can leverage a few labeled target images to significantly improve adaptation and model performance.
	
	Intuitively, the advantages of SSDA are particularly pronounced in high-dimensional settings where distribution shifts exhibit a low-dimensional structure. In such cases, large amounts of labeled source data are essential to train a base model. At the same time, a few labeled target examples can guide model choices among candidate models that deviate from the base model in low-dimensional directions. However, effective use of the small labeled target dataset in training remains a key challenge in SSDA. Over-reliance on labeled source data can introduce significant bias, whereas depending too heavily on the small target dataset can lead to high variance. Striking a balance between these two sources of information in a structured fashion is crucial. While prior research has explored various strategies for using target labels, such as incorporating them into source model training, using them for model selection, or fine-tuning pre-trained models~\citep{ben2010theory,yu2023semi,yan2022multi,kumar2022fine}, a general guideline remains elusive.

	In this paper, we develop a theoretical framework to analyze the feasibility of SSDA under structural causal models (SCMs). The SCM viewpoint is useful for SSDA because it allows us to specify where the source and target distributions differ. Under this framework, SSDA methods can be quantitatively compared in various distribution-shift scenarios. Our goal is to rigorously address fundamental questions in SSDA such as: Can SSDA methods provably outperform empirical risk minimization (ERM) trained on target data alone? How many labeled target samples are required for SSDA to also outperform ERM trained on source data alone, or UDA and DG methods? What types of low-dimensional structures in distribution shifts can SSDA exploit? By addressing these questions, we hope to provide theoretical insights and practical guidelines for the effective use of limited labeled target data in SSDA.
	
	\subsection{Our contributions}
	Our contributions are three-fold. First, we formalize a statistical framework to study SSDA methods with limited labeled target data under anticausal linear SCMs. While the classical supervised learning theory assumes that training and test data are drawn from an identical underlying distribution, modeling source-target distribution shift requires an extension that goes beyond the identical distribution assumption. Second, we identify three canonical low-dimensional intervention models: confounded additive (CA) shifts, sparse connectivity (SC) shifts, and anticausal weight  (AW) shifts. For each intervention model, we introduce fine-tuning methods from corresponding UDA starts and establish minimax target excess risk upper and lower bounds for our proposed methods, showing that the labeled target sample complexity scales with the intervention dimension rather than the ambient dimension. Third, when the intervention model is underspecified, we introduce Multi Adaptive-Start Fine-Tuning (MASFT), which selects among multiple fine-tuned estimators using a small target validation set and achieves near-optimal performance.
	
	\subsection{Related work}
	As a subfield of transfer learning, domain adaptation, 
	also known as transductive transfer learning~\cite{redko2020survey}, 
	aims to develop effective methods when distribution shifts occur between source and target data, while the underlying task remains the same. There are three standard settings of DA: unsupervised domain adaptation (UDA), domain generalization (DG), and semi-supervised domain adaptation (SSDA). 
	As the name SSDA suggests, the literature on semi-supervised learning (SSL) and UDA constitute an important part of the prior work on SSDA. For a comprehensive review which treats SSL or UDA in an isolated context other than SSDA, we refer readers to recent surveys \cite{van2020survey} and \cite{wilson2020survey, zhou2022domain}. Here, we highlight the most relevant developments in SSDA.
	
	\textit{Empirical advances in SSDA.} 
	SSDA methods are shown to outperform UDA methods when even very few
labeled target samples are available. Saito et al.~\cite{saito2019semi} demonstrated 
the benefit of SSDA using a minimax entropy approach for invariant feature extraction in
image classification. Building on adversarial domain-invariant learning for
UDA~\cite{ganin2016domain}, Jiang et al.~\cite{jiang2020bidirectional}
leveraged labeled target data to refine feature alignment. Yang et
al.~\cite{yang2021deep} incorporated co-training~\cite{blum1998combining} by
jointly training an inter-domain UDA model and an intra-domain SSL model for
3-shot adaptation on DomainNet~\cite{peng2019moment}. Further improvements on
DomainNet have used stronger invariance regularization, such as multi-level
consistency learning~\cite{yan2022multi}, and improved feature adaptation
strategies, such as source label adaptation~\cite{yu2023semi}. These works
highlight the empirical value of limited target supervision in low-shot learning scenarios
in computer vision.

\textit{Relation to traditional fine-tuning.}
Fine-tuning (FT) adapts a pretrained model to a new domain or task 
by updating
some or all of its parameters on a smaller dataset.
It has been shown to be effective in many vision studies~\cite{kornblith2019better,chen2021empirical},
but can underperform under large distribution shifts~\cite{kumar2022fine}.
SSDA can be viewed as a structured form of fine-tuning, where labeled target samples guide
the update, while domain-specific assumptions and, when informative, unlabeled
data can determine the starting point and adaptation directions.

			
	\textit{SSDA ideas in statistics.} SSDA often appears in the literature under the broader framework of transfer learning, which may not explicitly mention the SSDA terminology. For instance, Bastani \cite{bastani2021predicting} studied SSDA---though they refer to it as transfer learning---in the context of high-dimensional prediction tasks, where the distribution shift between source and target data is captured by a sparse function of features. Li et al.~\cite{li2022transfer} analyzed a similar transfer learning problem in a high-dimensional linear regression setting, assuming that the regression coefficients change only in a small subset of coordinates between the source and target domains. Under this assumption, the problem reduces to the well-studied high-dimensional sparse regression (see e.g.,~\cite{wainwright2019high}), where effective adaptation can be achieved with far fewer target samples than the number of features. Additionally, Xiong et al.~\cite{xiong2023distributionally} considered a similar high-dimensional sparse linear regression problem under the distributionally robust optimization framework.
	
	Empirical Bayes methods \citep{robbins1992empirical} are also closely related to SSDA. They are Bayesian inference methods that place data-driven prior on regression coefficients rather than relying solely on expert knowledge. By learning priors from the source data, these methods effectively borrow source information for improved estimation and prediction on the target domain. Recent advances in empirical Bayes methods have extended their applicability to transfer learning~\cite{law2023distributional}.
	
	A fundamental distinction between these classical methods and our SCM-based SSDA approaches lies in how they model the relationship between source and target distributions:
	\begin{itemize}
		\item Classical methods, such as the two approaches above, assume that the distribution shift from source to target occurs directly in the regression coefficients to be estimated. In other words, they impose assumptions on how the true model parameters evolve across domains.
		\item  SSDA methods based on SCMs instead model distribution shifts as interventions
    	in the data generating process, 
		rather than defining them explicitly in terms
		of model parameters~\cite{pearl2009causality,peters2016causal,buhlmann2020invariance}.  
		As developed in Pearl's causality framework~\cite{pearl2009causality}, SCMs provide a powerful framework to model interventions and distributional shifts in a principled way and are complementary to the standard causal linear regression model $Y = X\tp \beta + \epsilon$ with $X$ independent of $\epsilon$.
		The induced changes in the observational distribution may then ultimately affect the regression parameter to be estimated,
    	but the primary assumption is placed on the underlying causal mechanism rather
    	than on the parameter itself.
	\end{itemize}

	This viewpoint of SCM-based SSDA methods is particularly useful in DA scenarios
	where direct modeling of parameter shifts is
	impractical. 
	For instance,
	in image classification tasks involving complex degradations like
	resolution loss, noise, and lighting changes, it is challenging to specify how these shifts directly affect model parameters.
	By modeling shifts in the data generation process instead, these methods offer a more flexible alternative framework for addressing such
	distribution shifts. This modeling-based formulation arises naturally
	in many applications, particularly in medical
	diagnosis~\citep{castro2020causality},
	scientific inverse problems such as cosmological parameter
	inference~\citep{ribli2019improved,ha2021adaptive},
	and spectroscopic calibration transfer~\citep{anderson2020achieving}.
	In this work, we build upon this SCM-based perspective to propose
	and analyze novel SSDA methods.



\section{Preliminaries and background}
In this section, we introduce the semi-supervised domain adaptation (SSDA) problem and structural causal models (SCMs) 
to specify the relationship between source and target data.
We then review several baseline UDA methods, which serve as baseline methods and also starting points for building SSDA methods. 

\subsection{SSDA problem setup}\label{sec:setup}

Consider the supervised learning problem with $M$ $(M \geq 1)$ source domains and a single target domain. Here each domain is associated with a data distribution and the \emph{labeled} and \emph{unlabeled} datasets drawn from that distribution. Specifically, for the $m$-th source domain, $m=1,\ldots,M$, we observe 
\begin{align*}
	\Dsrc=\{(\Xsrc_i,\Ysrc_i)\}_{i=1}^{\nsrc}\iidsim\Psrc \text{ and } \Dsrcu=\{\Xsrctil_i\}_{i=1}^{\nsrcu}\iidsim \Psrc_X,
\end{align*}
where $\Psrc$ denotes the $m$-th source distribution, and $\Psrc_X$ denotes its marginal distribution on the covariates $X$. For the target domain, we similarly observe 
\begin{align*}
	\Dtar=\{(\Xtar_i,\Ytar_i)\}_{i=1}^{\ntar}\iidsim \Ptar \text{ and } \Dtaru=\{\Xtartil_i\}_{i=1}^{\ntaru}\iidsim \Ptar_X,
\end{align*}
where $\Ptar$ denotes the target distribution, and $\Ptar_X$ denotes its marginal distribution on the covariates $X$. We assume that in all domains ($m\geq 0$), the covariates lie in $\R^d$, i.e., $\Xsrc\in \R^d$, and the responses (labels) are real-valued, $\Ysrc\in\R$.

Given a function $f:\R^d \to \R$ from a function class $\Fset$ that predicts labels from covariates, its prediction performance in the target domain is measured via the population target risk
\begin{align}\label{eqn:target_risk}
	\risktar{f} \coloneqq \EE{\ell(f(\Xtar),\Ytar )},
\end{align}
where $\ell:\R\times \R \to \R$ is a loss function. 
Throughout the paper, unless stated otherwise, we use the squared loss defined as $\ell(f(x),y) \defn (y-f(x))^2$ and restrict the function class to linear functions $\Fset = \{f_\beta: x \mapsto \beta^\top x \mid \beta \in \R^d\}$. 
The squared loss is used to simplify theoretical discussions, 
although several motivating examples below involve binary or multiclass classification problems, such as image classification. Extending the finite-sample analysis to logistic, cross-entropy, or other losses is left for future work.
The $m$-th source population risk ($m\geq 1$) is defined analogously as $\risksrc{f}  \coloneqq \EE{\ell(f(\Xsrc),\Ysrc )}$.  

In SSDA, one seeks to use both source and target data (unlabeled and limited labeled) to estimate a function $f$ such that its population target risk is small. This population target risk is compared with the optimal population target risk, and the difference is quantified as the excess risk, defined as
\begin{align}\label{eqn:excess_risk}
	\delta\risktar{f}  \coloneqq \risktar{f} - \min_{\widetilde{f}\in\Fset}\risktar{\widetilde{f}}.
\end{align}
When shifts between the source and target data can be arbitrary, it is generally impossible to achieve a small excess risk with only a limited number of labeled target samples, 
even with an infinite source and unlabeled target samples~\citep{chen2021domain}. 
Additional assumptions relating source and target data are therefore needed to achieve nontrivial target performance in SSDA. 

\paragraph*{Anticausal linear structural causal models} 
To enable rigorous comparison of UDA and SSDA methods, we first assume that 
the data in both source and target domains are generated according to linear structural causal models (SCMs)~\cite{pearl2009causality}. 
The algorithms introduced below can be implemented without specifying an SCM, but the SCM assumptions are used to prove excess risk guarantees and to interpret when each algorithm is appropriate.
\begin{assumption}[Anticausal linear SCMs]\label{assump:linear_SCMs}
	For $m=0,1,\ldots,M$, each data distribution $\Psrc$ corresponding to the $m$-th domain is specified by the structural equation
	\begin{align}\label{eqn:src_linear_SCM}
		\begin{bmatrix}
			\Xsrc \\ \Ysrc
		\end{bmatrix}= 		
		\begin{bmatrix}
			\Bsrc & \bsrc \\ 
			0 & 0
		\end{bmatrix}
		\begin{bmatrix}
			\Xsrc \\ \Ysrc
		\end{bmatrix} + 
	 \begin{bmatrix}
			\epssrc_X \\ \epssrc_Y
		\end{bmatrix}.
	\end{align}
	Here $\Bsrc\in\R^{d\times d}$ are the connectivity matrices encoding the causal relations among the covariates, and $\bsrc\in\R^d$ are the anticausal weights connecting labels and covariates. 
	We assume $\ident-\Bsrc$ is invertible for all $m\geq 0$.  
	At this level, $\epssrc_X$ and $\epssrc_Y$ are exogenous noise terms;
	additional conditions on these noises such as centering, tail, and independence conditions, when needed for finite-sample
	guarantees, are introduced in later assumptions.
\end{assumption}

Solving the structural equations in~\eqref{eqn:src_linear_SCM} gives
\[\Ysrc=\epssrc_Y, \; \text{ and } \; \Xsrc = (\ident-\Bsrc)^{-1}\bsrc\epssrc_Y
    +(\ident-\Bsrc)^{-1}\epssrc_X. \] 
As a consequence, each domain distribution $\Psrc$ is determined by the joint noise law, the connectivity matrix $\Bsrc$, and the anticausal weight vector $\bsrc$.  \assumpref{linear_SCMs} is a linear anticausal specialization of the general SCM framework~\citep{pearl2009causality}, where $Y$ causes $X$ and $\Bsrc$ describes the relations among the
coordinates of $X$. It is the anticausal analogue of the classical
linear regression model:
just as linear models provide a simple and tractable setting for studying causal prediction, 
linear anticausal SCMs provide a corresponding tractable setting 
for studying interventions across domains under anticausal structure.
The relation between classical linear regression and linear anticausal SCMs is made explicit below.
Extensions to mixed causal-anticausal
SCMs are possible, but the anticausal model suffices to illustrate our
main ideas.

\paragraph*{Relation to classic linear regression} At the level of a single domain observational distribution, \assumpref{linear_SCMs} includes the classic random-design linear regression model as a special case. 
Suppose
\begin{align*}
    Y=X \tp \beta+\varepsilon,\;\; \text{ and }
	 X\sim N(0,\Sigma_X),
    \;\; \varepsilon\sim N(0,\sigma_\varepsilon^2),
    \;\; \varepsilon\independent X.
\end{align*}
Let 
$\nu_Y^2=\beta^\top\Sigma_X\beta+\sigma_\varepsilon^2$ and
$b=\Sigma_X\beta/\nu_Y^2$. Then the joint distribution of $(X,Y)$ can be
generated via the equivalent model from \assumpref{linear_SCMs}
\begin{align*}
    Y=\eps_Y,
    \;\;
    X=bY+\eps_X,\text{ and}
    \;\;
    \eps_Y\sim N(0,\nu_Y^2),
    \;\;
    \eps_X\sim N(0,\Sigma_X-\nu_Y^2bb^\top),
    \;\;
    \eps_X\independent \eps_Y,
\end{align*}
where $\Sigma_X-\nu_Y^2bb^\top$ is positive semidefinite by the
Schur complement. Hence the causal and anticausal formulations are
observationally equivalent in a single domain. However, their difference arises when interventions are applied across domains. Classical linear regression
interprets $X\to Y$, whereas we study interventions that cause distribution shifts in the anticausal direction $Y\to X$.



\paragraph*{Anticausal prediction in practice}
Anticausal structure arises when a latent state or physical variable
generates observable measurements, while the prediction task is to recover it from the measurements. 
This setting appears in a broad range of real world problems.
For example, 
in medical imaging and diagnosis, a disease state $Y$ generates observable symptoms, biomarkers, or images $X$~\citep{castro2020causality}. Moreover, scientific and engineering inverse problems often have an anticausal structure, where an unknown physical parameter $Y$ generates observations $X$ through a forward process
$
X = \mathcal{G}(Y) + \eps_X,
$
where $\mathcal{G}$ encodes the underlying physics (e.g., a PDE-based forward model) and $\eps_X$ is measurement noise (e.g., see~\cite{dashti2013bayesian}). 
Examples include cosmological parameter inference from weak-lensing maps~\cite{ribli2019improved,ha2021adaptive}, and the prediction of chemical
concentrations from near-infrared (NIR) spectroscopy~\cite{zhao2019pls,anderson2020achieving}. Section~\ref{sec:numerical_exp}
includes experiments motivated by the NIR applications.
Despite its prevalence, anticausal prediction under interventions has received relatively less theoretical attention.

\begin{remark}[Why not causal SCMs for SSDA?]
\label{rmk:anticausal_rationale}
\assumpref{linear_SCMs} restricts attention to anticausal SCMs
($Y \to X$) rather than the causal setting ($X \to Y$) commonly
assumed in the literature. 
This restriction isolates a regime in which unlabeled target data can inform adaptation. 
In a causal prediction model, 
unlabeled target data reveal only the marginal distribution of $X$, and do not, by themselves, 
reveal additional information about the conditional distribution $Y \mid X$. 
Hence they are uninformative for adaptation.
In contrast, in an anticausal model, interventions on the mechanism generating $X$ from $Y$ can alter the conditional distribution $X \mid Y$, which in turn alters $Y \mid X$ by Bayes' rule. As a consequence, unlabeled target data  can reveal information about the distributional shifts and become useful for SSDA in anti-causal settings~\citep{scholkopf2012causal,castro2020causality,chen2021domain}. 
We focus on this important and relatively underdeveloped setting.
\end{remark}

\subsection{Baseline unsupervised domain adaptation (UDA) methods}\label{sec:baseline_DA_algorithms}
 We next introduce several UDA estimators which serve as baselines and starting points for our SSDA methods. 
We use the term UDA estimators broadly to refer to estimators that use source data and/or unlabeled target data but do \emph{not} rely on labeled target data. 
For clarity, we present these methods specialized to the linear function class; their generic forms can be found in, e.g.,~\cite{ganin2016domain,heinze2021conditional,wu2023prominent}.

Ordinary least squares (OLS) on individual source datasets is the most basic estimator in UDA: one trains on the source data, and hopes it works on the target data.
\begin{itemize}
	\item \textit{OLS-Src$^\tagm{m}$}: the population OLS estimator based on the $m$-th ($m\geq 1$) source dataset.
	\begin{align}\label{eqn:OLSSrc_pop}
		\fsrc(x) \defn \betasrc{}^\top x, \text{ where } \betasrc  \defn \argmin_{\beta}  \ \EE{(\Ysrc - \beta^\top \Xsrc)^2} .
	\end{align}	
	When the source and target domains are sufficiently different, OLS-Src$^\tagm{m}$ may have large target risk. Nevertheless, it serves as a baseline for risk comparison, and a starting point for fine-tuning.
\end{itemize}
 Next, we introduce two UDA estimators that use an invariance penalty to align source and target distributions in a learned feature representation, thereby reducing the bias of source-only training.

The first estimator is domain invariant projection (DIP)~\citep{baktashmotlagh2013unsupervised,ganin2016domain,sun2016deep}, also known as domain invariant representation learning, 
which learns a representation in which the source and target covariate marginals are matched.
This method has been popularized through its neural network implementation~\cite{ganin2016domain}.
Under the linear function class, DIP amounts to searching for a common linear projection.

\begin{itemize}
	\item \textit{DIP$^\tagm{m}$}: the population DIP$^\tagm{m}$ estimator based on the $m$-th ($m\geq 1$) source dataset and the marginal target covariates.
	\begin{align}\label{eqn:DIP_pop}
		\begin{split}
			f^{(m)}_{\text{gDIP}}(x) \defn \beta^{(m)}_{\text{gDIP}}{}^\top x, \text{ where } &\beta^{(m)}_{\text{gDIP}}{}^\top  \defn \argmin_{\beta}  \ \EE{(\Ysrc - \beta^\top \Xsrc)^2} ,  \\
			&\hspace{0.9in} \text{s.t. } \ \ \probdiv{\Pmarsrc{\beta^\top X}}{\Pmartar{\beta^\top X}}  = 0.
		\end{split}
	\end{align}
	The term $\probdiv{\cdot}{\cdot}$ denotes a divergence or a distance between two probability distributions, such as moment-based divergence~\cite{peng2019moment}, maximum mean discrepancy (MMD)~\cite{gretton2012kernel}, or adversarial-based domain classifier~\cite{ganin2016domain}. 
\end{itemize}

The second estimator is conditional invariant penalty (CIP)~\cite{gong2016domain,heinze2021conditional}, which aligns the conditional distributions of the feature representations given the labels $Y$. Unlike DIP which matches the unlabeled source and target data, CIP incorporates label information across multiple source datasets into distribution matching.
\begin{itemize}
	\item \textit{CIP}: the population CIP estimator based on the multiple source datasets.
	\begin{align}\label{eqn:CIP_pop}
		\begin{split}
			f_{\text{gCIP}}(x) \coloneqq\beta_{\text{gCIP}}^\top x, \text{ where } &\beta_{\text{gCIP}}  \coloneqq \argmin_{\beta}  \ \frac{1}{M}\sum_{m=1}^{M}\EE{(\Ysrc - \beta^\top \Xsrc)^2} , \\
			& \text{s.t. } \ \ \discrepancy{\Pcondsrc{\beta^\top X}{Y}}{\Pcondsrcmprev{\beta^\top X}{Y}}  = 0  \textit{ a.s.} \text{  $\forall m\in\{2,\ldots,M\}$},
		\end{split}
	\end{align}		
	where $\probdiv{\cdot}{\cdot}$ is a divergence between two probability distributions. The CIP penalty matches all conditional distributions of the covariates across all source domains. If this conditional invariance also holds in the target domain, then CIP operates as if source and target data share the same conditional distribution. 
\end{itemize}
The assumption behind CIP is the existence of features with invariant conditional distribution across all source and target domains. These features are also called \emph{conditionally invariant components} (CICs)~\cite{gong2016domain,heinze2021conditional}. While the use of CIP-type matching penalty ensures conditional invariance across all source domains and has various applications in DA~\cite{gong2016domain,li2018deep,heinze2021conditional}, whether it identifies the CICs which are also conditionally invariant in the target domain is generally far from trivial.
This problem under additive shift interventions was studied recently in~\cite{chen2021domain,wu2023prominent}, which quantifies the conditions and the number of heterogeneous source domains needed for identifiability.

\section{Minimax framework for SSDA and benefits of fine-tuning starting from UDA}\label{sec:proposed_algorithm} 
In this section, we present a statistical framework to study SSDA which extends the classical minimax framework in supervised learning. 
Then we provide a simple example to illustrate the benefits of fine-tuning starting from UDA methods in SSDA.

\subsection{Target-only lower bound and SSDA minimax criterion}\label{sec:minimax}

This subsection first establishes the target-only minimax lower bound under
anticausal linear SCMs and then defines the corresponding SSDA minimax criterion.  
Whenever a quantitative risk comparison is needed, we restrict our attention to the linear function class with data generated from linear SCMs in~\assumpref{linear_SCMs}. 

First, the oracle estimator is defined as the population target risk minimizer under linear function class and has an explicit form.
\begin{itemize}
	\item \textit{Oracle estimator}: the population OLS estimator, or oracle estimator, based on the target dataset.
	\begin{equation}\label{eqn:OLSTar}
		\betatrue\coloneqq\argmin_{\beta}  \EE{(\Ytar - \beta^\top \Xtar)^2} =  \SigmaXtar{}^{-1}\EE{\Xtar\Ytar},
	\end{equation}
	where \(\SigmaXtar \coloneqq \EE{\Xtar\Xtar{}^\top}\) is assumed to be invertible.	
\end{itemize}

Let $\Pfraktar$ denote the class of joint distributions over $\ntar$ data points from $\Dtar$ and $\ntaru$ data points from $\Dtaru$. The classical minimax excess risk is defined as
\begin{align}\label{eqn:minimax_excess_risk_target}
	\min_{\betah} \sup_{\Pset\in \Pfraktar}\EE{\delta\risktar{f_{\betah}} }=\EE{(\betah - \betatrue)^\top \SigmaXtar (\betah - \betatrue)},
\end{align}
where the minimum is over all estimators based on (both labeled and unlabeled) target datasets, and the expectation is over $\Pset\in\Pfraktar$.

When $\Pfraktar$ corresponds to the class of causal linear prediction models, the minimax excess risk has been extensively studied in the literature~\cite{hsu2012random,mourtada2022exact}. However, the minimax framework for anticausal models, with additional unlabeled data, has received relatively limited attention. In the following proposition, we establish a lower bound on the minimax excess risk~\eqnref{minimax_excess_risk_target} when $\Pfraktar$ is specified by the class of anticausal linear SCMs  in~\assumpref{linear_SCMs}. 
\begin{proposition}\label{prop:anticausal_minimax_lowerbound}
	Let $\Pfraktar$ denote the class of joint distributions over $\Dtar,\Dtaru$, where each labeled and unlabeled target sample is generated according to the anticausal linear SCMs in~\assumpref{linear_SCMs}. The minimax excess risk~\eqnref{minimax_excess_risk_target} over $\Pfraktar$ satisfies
	\begin{equation}\label{eqn:anticausal_minimax_lowerbound}
		\min_{\betah} \sup_{\Pset\in \Pfraktar}\EE{\delta\risktar{f_{\betah}} } \geq c\vary\min\left\{\frac{d}{\ntar},1 \right\}, 
	\end{equation}
	where $\vary\defn\VV{\Ytar}>0$ is the variance of $\Ytar$, and $c>0$ is a universal constant. The minimum is over all measurable estimators $\betah$ based only on the target labeled and unlabeled datasets $(\Dtar,\Dtaru)$.
\end{proposition}
The proof of~\propref{anticausal_minimax_lowerbound} follows from a
standard Fano argument by restricting to a least favorable Gaussian submodel,
and is given in~\appref{minimax_proof} for completeness.
The submodel is constructed so
that the marginal distribution of $X$ is identical across alternatives, 
so it does not reveal any information about the parameter of interest.
Consequently, unlabeled target covariates do
not reduce the minimax difficulty, and target-only estimators require $\ntar$ to be of order
at least $d$ to achieve nontrivial excess risk.

In SSDA, the labeled target sample size is often limited, with $\ntar\ll d$.
In this regime, the lower bound~\eqnref{anticausal_minimax_lowerbound}
implies that target-only estimators cannot reduce the excess risk below a
constant. This leads to a central question in SSDA: can additional source
data improve upon the target-only minimax rate
$\O{\min\{d/\ntar,1\}}$? 
To formalize this question, let $\Pfrak$ denote a class of joint distributions
over all source and target datasets, including both labeled and unlabeled
samples. We extend the minimax excess risk to the SSDA setting as
\begin{align}\label{eqn:minimax_excess_risk}
	\min_{\betah} \sup_{\Pset\in \Pfrak} \EE{\delta\risktar{f_{\betah}} },
\end{align}
where the minimum is over all estimators that may use both source and target
datasets, labeled and unlabeled, and the expectation is over $\Pset\in\Pfrak$.

In general, 
without explicitly specifying the relationship between source and target distributions in $\Pfrak$,
additional source data may be either unhelpful or as useful as additional
target data. For instance, (i) if $\bsrc=0$ for all $m\geq1$ while
$\btar\neq0$ under~\assumpref{linear_SCMs}, then source covariates are uninformative for label prediction
and the SSDA minimax problem reduces to the target-only problem. In contrast,
(ii) if all source and target distributions are identical, each source sample
is as informative as a target sample, and OLS on the pooled labeled data
achieves the corresponding supervised minimax rate of order
$d/\sum_{m=0}^M n^{(m)}$ when $\sum_{m=0}^M n^{(m)} > d$ (c.f. see~\propref{least_squares_bound}).

These two extremes show that 
understanding the usefulness of source data
requires careful modeling of the source-target relationship. In practice, most scenarios lie between these two extremes.
The challenge is that there is no ``one-size-fits-all'' solution: distribution shifts can take various forms, which impacts the choice of adaptation strategies.





\subsection{Illustrative example on benefits of fine-tuning starting from UDA}\label{sec:finetuning_example}

In this subsection, we illustrate how the choice of adaptation strategy depends
on the source-target relationship. In particular, we present a nontrivial
example showing that fine-tuning from a UDA estimator can improve target
performance when labeled target data are limited.

The intuition is as follows. Under a certain structured relationship between the source and target distributions,
UDA estimators in~\secref{baseline_DA_algorithms} can learn components shared across the source and target domains.
However, without labeled
target data in the UDA setting, it may fail to capture target-specific components that could further improve target prediction. 
When a few labeled target samples are also available in SSDA, we can fine-tune
the UDA estimator only along the low-dimensional directions where the target
differs from the source.

More generally, a fine-tuning method that starts from a UDA estimator can be formulated as follows. Suppose the function class $\Fset=\{f_\theta:\theta\in\Theta\}$ is parameterized by $\theta \in \Theta$, where $\Theta\subset\R^D$ is a subset of a $D$-dimensional Euclidean space. Let $\thetada\in\Theta$ denote a UDA estimator trained with source and/or unlabeled target data.
\begin{itemize}
	\item \textit{Generic fine-tuning formulation}: Given a subset $\Thetasmall\subset \Theta$, the model parameter $\theta$ is fine-tuned on the labeled target dataset $\Dtar$ via the following constrained optimization
	\begin{equation}\label{eqn:fine_tuning_generic}
		\begin{aligned}
			& \min_{\theta\in\Theta} \frac{1}{\ntar}\sum_{i=1}^{\ntar}\ell(f_{\theta} (\Xtar_i),\Ytar_i)  \text{ s.t. } \theta-\thetada \in \Thetasmall.
		\end{aligned}
	\end{equation}		
	Here, $\Thetasmall$ represents a low-dimensional or low-complexity set, such as a low-dimensional
	linear subspace or a set with small Gaussian width (e.g., measured by Gaussian width~\cite[Definition 7.5.1]{vershynin2018high}), tailored to specific problem setting.
\end{itemize}

In this formulation, unlabeled data 
enter the fine-tuning step indirectly through both the starting point $\thetada$ and through 
the data-dependent low-complexity set
$\Thetasmall$. The empirical objective uses the
labeled target samples, while source and unlabeled covariates determine the
subspace or constraint in which these few labels are allowed to modify
the UDA estimator. This separation is useful when $\ntar\ll d$: the unlabeled
and source data estimate high-dimensional nuisance structure, and the few target
labels estimate only the remaining low-dimensional correction.

Recent empirical studies have shown that fine-tuning in large neural networks
and language models often happens in a small subspace relative to the ambient
parameter dimension 
\citep{li2018measuring,gressmann2020improving,hu2022lora}. 
The formulation
\eqnref{fine_tuning_generic} captures this idea abstractly, but the choices of
$\thetada$ and $\Thetasmall$ are typically made in an ad hoc manner with limited justification.

Below, we show in
a linear SCM example how the appropriate starting point and fine-tuning 
subspace can be derived from the source-target relationship rather than chosen
ad hoc.

\begin{figure}[t!]
	\centering
	\includegraphics[width=0.6\textwidth]{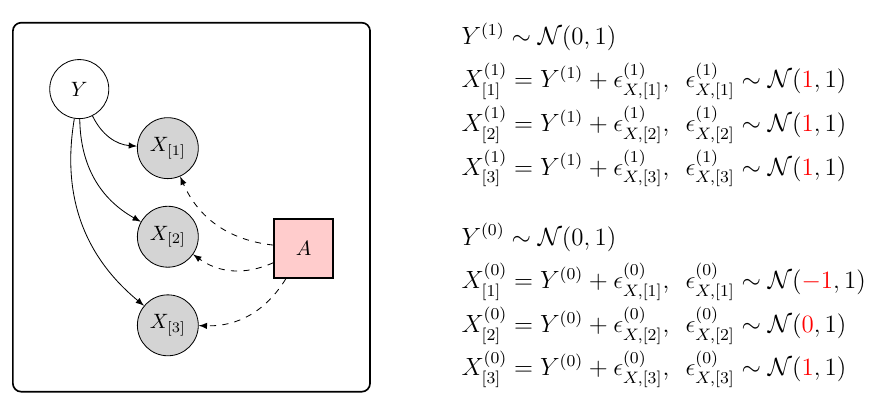} 
	\caption{The causal diagrams and structural equations illustrating the example of additive mean shift. The covariates $X$ exhibit mean shifts between the first source and target domains.}
	\label{fig:causal_mean_shift}
\end{figure}

\textit{Example (Additive mean shift).} 
As a concrete example, we consider source and target data generated 
from the linear SCMs in~\figref{causal_mean_shift}. 
Here, the shift between source and target
distributions arises from additive mean interventions, where the means of the
covariates are shifted by deterministic constants.

In this example, the DIP$^{(1)}$ estimator and the oracle estimator are explicitly derived as
\[
\betasrconedip = \frac{1}{272} \begin{bmatrix} 47, & 59, & 71 \end{bmatrix}^\top, \quad \text{and} \quad \betatrue = \frac{1}{4}\begin{bmatrix} 1, & 1, & 1 \end{bmatrix}^\top,
\]
where mean squared difference $\probdiv{\Pmarsrc{\beta^\top X}}{\Pmartar{\beta^\top X}} =(\EE{\beta^\top \Xsrcone} - \EE{\beta^\top \Xtar})^2$ is used in~\eqnref{DIP_pop}. 
At first glance, it is not evident how $\betasrconedip$ and $\betatrue$ are related. For the reasons that become clear later in~\secref{fine_tune_mixture_mean_shift}, we remark that $\SigmaXtar{}^{-1}(\EE{\Xsrcone}-\EE{\Xtar})= \frac{1}{12}\begin{bmatrix}
	7, & 3, & -1
\end{bmatrix}^\top$ is parallel to the direction $\betatrue - \betasrconedip$, i.e.,
\[
\betatrue - \betasrconedip = \frac{1}{272}\begin{bmatrix} 21, & 9, & -3 \end{bmatrix}^\top \propto v_{\textnormal{DIP}}\defn \SigmaXtar{}^{-1}\left(\EE{\Xsrcone}-\EE{\Xtar}\right).
\]
Crucially, the direction $v_{\textnormal{DIP}}$ is identifiable from source
and target covariates: source covariates estimate $\EE{\Xsrcone}$, while
unlabeled target covariates estimate $\EE{\Xtar}$ and $\SigmaXtar{}$. Hence
the unlabeled target sample reveals the direction in which fine-tuning should occur. Without unlabeled
target covariates, source data alone cannot determine the target shift
direction, and the low-dimensional fine-tuning constraint
$\Thetasmall=\spn(v_{\textnormal{DIP}})$ is not identifiable.

This example demonstrates that under additive interventions on the mean, 
DIP$^{(1)}$ can be an effective UDA starting point for fine-tuning, 
which settles the shared subspace from source and unlabeled target data and 
leaves only one direction to fine-tune. 
A generalization and its finite-sample guarantees are developed in~\secref{fine_tuning_mean_shift}.

\section{New SSDA methods and excess risk guarantees under linear SCMs}\label{sec:finite_sample_theory}

The anticausal linear SCM in~\assumpref{linear_SCMs} is parameterized by
three components: the noise law $(\epssrc_X,\epssrc_Y)$, the connectivity
matrix $\Bsrc$, and the anticausal weight vector $\bsrc$. Any distribution shifts
between source and target domains can therefore be viewed as interventions on
one or more of these components.

In this section, we study three canonical interventions, each corresponding to
one of these components: \emph{confounded additive (CA) shift} from
interventions on the noise law, \emph{sparse connectivity (SC) shift} from
interventions on $\Bsrc$, and \emph{anticausal weight (AW) shift} from
interventions on $\bsrc$. Each intervention induces a different form of
distribution shift and therefore requires a tailored adaptation strategy. 
In particular, for
each scenario, we propose a fine-tuning method initialized from a corresponding
UDA start. By carefully leveraging source data and, when informative, unlabeled
covariates, these methods improve upon the target-only minimax risk and achieve
the target-label minimax rates, up to constant or logarithmic factors.
\tabref{results} summarizes the main results.

\begin{table}[ht]
\centering
\begin{tabular}{||c | c c c||} 
 \hline
 Type of distribution shift & SSDA method & Data access needed & Target minimax rate \\ [0.5ex] 
 \hline\hline
No shift (target-only) & \makecell{-} & \makecell{-} & $d/\ntar$ \\
 \hline
 \makecell{Confounded additive (CA)~(\assumpssref{mean_shift_linear_SCM}) \\ on noise $\epssrc_X,\epssrc_Y$} & \makecell{FT-DIP \\ \thmref{dip_var_matching_finite}} & $\{\Dsrc,\Dsrcu\}_{m=0}^{1}$ & \makecell{$\rdip/\ntar$ \\ \thmref{dip_var_matching_finite_minimax}} \\ 
 \hline
 \makecell{Sparse connectivity (SC)~(\assumpssref{connectivity_shift_linear_SCM}) \\ on matrix $\Bsrc$} & \makecell{FT-OLS-Src \\ \thmref{connectivity_estimator_finite}} & $\{\Dsrc\}_{m=0}^{1}$ & \makecell{$\rconn/\ntar$ \\ \thmref{connectivity_estimator_finite_minimax}}\\
 \hline
 \makecell{Anticausal weight (AW)~(\assumpssref{coefficient_shift_linear_SCM}) \\ on vector $\bsrc$} & \makecell{FT-CIP \\ \thmref{cip_mean_matching_finite}} & $ \{\Dsrc\}_{m=0}^M,\Dsrconeu$ & \makecell{${}^*\rcip/\ntar$ \\ \thmref{cip_mean_matching_finite_minimax} } \\
 \hline
\end{tabular}
\caption{Summary of distribution shift scenarios, our proposed SSDA methods, and the minimax excess risk rates.
Here $\rdip$, $\rconn$, and $\rcip$ denote
the perturbation dimensions of the CA, SC, and AW interventions, respectively.
The entry marked with $*$ indicates the lower bound without access to unlabeled target data (see~\secref{fine_tuning_CIP_finite_sample}). 
}
\label{tab:results}
\end{table}

To simplify the following finite-sample risk analysis, we impose the following regularity
conditions on the noise variables and covariance matrices.
\begin{assumption}\label{assump:sub_gaussianity}
	In~\assumpref{linear_SCMs}, for all $m\geq 0$, the noise terms
	$\epssrc_X$ and $\epssrc_Y$ are zero-mean sub-Gaussian random vectors
	and random variables with parameters $\sigma_{\eps_X}^{(m)}$ and
	$\sigma_{\eps_Y}^{(m)}$, respectively.  Unless specified otherwise,
	$\epssrc_X$ and $\epssrc_Y$ are independent within each domain.
	Furthermore, the covariance matrices
	$\SigmaXsrc\coloneqq\EE{\Xsrc\Xsrc{}^\top}$ are invertible for all
	$m\geq0$.
\end{assumption}

Before proceeding, we use the following notation throughout this section. 
Under~\assumpref{sub_gaussianity}, the covariates $\Xsrc$ are centered
sub-Gaussian random vectors,
and we denote their sub-Gaussian parameter by $\sigma_X$ uniformly across all $m\geq 0$ (c.f. see~\lemref{bound_sigma_X}). 
Similarly,
we let $\sigma_Y=\max_{m\geq0}\sigma_{\eps_Y}^{(m)}$ denote the
sub-Gaussian parameter for $\Ysrc$ across all $m\geq0$; 
when~\assumpref{sub_gaussianity} holds, this equals $\max_{m\geq 0}\sigma_{\eps_Y}^{(m)}$.
Let $\lamminbar$ and $\lammaxbar$ be, respectively, the
minimum and maximum eigenvalues of $\SigmaXsrc$ over all $m\geq0$. We define
the associated condition numbers as
$
    \kappabar\coloneqq {\lammaxbar}/{\lamminbar},$
	and
    $\tsigmamin\coloneqq {\sigma_X^2}/{\lamminbar}.
$
Although we make the dependency on these parameters explicit in the upper bounds of our main theorems, 
to simplify all rate discussions, one may consider all these parameters to be of constant order.
We refer the reader to~\tabref{notation} in the Appendix for further details.

\subsection{Low-dimensional and confounded additive interventions under linear SCMs}\label{sec:fine_tuning_mean_shift}

We first consider a distribution shift setting 
in which interventions are additive,
arise due to unobserved confounding, and are restricted to a low-dimensional
subspace.  We refer to this setting as
\textit{confounded additive (CA) shift interventions}, following
the literature on invariance-based causal prediction
\citep{rothenhausler2018anchor,rothenhausler2019causal,shen2026causality}. 
This setting captures shifts in domain-specific nuisance or "style" factors
while the mechanism connecting $Y$ to $X$ is shared across domains.
Examples include changes in lighting or noise level, background, imaging protocol, scanner
effects, or batch effects, which may perturb many observed features through a
low-dimensional additive component
\citep{heinze2021conditional,hendrycks2019benchmarking,guan2022domain,
loizillon2023semi}.
This is also related to anchor regression and related frameworks~\citep{rothenhausler2018anchor,shen2026causality},
where distributional changes are modeled through additive shift interventions on the noise. 
The key role of unlabeled target covariates in this regime is to identify the
low-dimensional shift subspace that can be used for fine-tuning.

\begin{assumption}[Confounded additive (CA) shift interventions]\label{assump:mean_shift_linear_SCM}
	In~\assumpref{linear_SCMs}, the first source and target domains share the same connectivity matrices and the anticausal weights, i.e., $B\coloneqq \Bsrcone=\Btar$ and $b\coloneqq \bsrcone=\btar$. The noise distributions are shifted as follows: in source, $\epssrcone_{X}\sim \Pset_{\eps_X}$ and $\epssrcone_{Y}\sim \Pset_{\eps_Y}$; while in target, $(\epstar_X, \epstar_Y)$ are additively intervened in the $\rdip$-dimensional subspace ($\rdip < d$), i.e., for independent $\xi_X\sim \Pset_{\eps_X}$ and $\xi_Y\sim \Pset_{\eps_Y}$,
  \begin{align*}
    \epstar_X=WZ + \xi_X, \;\; \epstar_Y = w_Y^\top Z + \xi_Y,
  \end{align*}
	where $Z$ is a $\rdip$-dimensional centered random vector with covariance $\ident_{\rdip}$ that is independent of $\xi_X,\xi_Y$, and $W\in\R^{d\times \rdip}$ and $w_Y\in\R^{\rdip}$ are fixed. We assume that $Z,\xi_X,\xi_Y$ are mutually independent. In words, $Z$ is a shared hidden confounder that affects both the covariate and label noise in the target domain. The essential rank condition is $\operatorname{rank}(W+bw_Y^\top)=\rdip$.
\end{assumption}  
\assumpref{mean_shift_linear_SCM} is the centered covariance-shift
analogue of the deterministic additive mean shift setting 
considered in~\secref{finetuning_example}. Under \assumpref{mean_shift_linear_SCM}, 
the target noise components are generally dependent because they share $Z$. In particular, 
with $\VV{Z}=\ident_{\rdip}$ and independent idiosyncratic noises, $\cov(\epstar_X,\epstar_Y)=Ww_Y$. 

Before introducing our SSDA method, we first discuss how UDA estimators can handle this distribution shift. 
For mean shift interventions as in~\secref{finetuning_example}, DIP with mean matching penalty 
is known to achieve low target risk~\cite{chen2021domain}. However, under~\assumpref{mean_shift_linear_SCM} of CA shift interventions, 
the mean matching penalty no longer provides any meaningful constraint to align source and target covariates due to the centering of covariates.
As a result, we consider a DIP variant that incorporates second-moment matching:
\begin{equation}\label{eqn:cov_distance}
	\probdiv{\Pmarsrcone{\beta^\top X}}{\Pmartar{\beta^\top X}}   = \left(\cov(\beta^\top \Xtar) - \cov(\beta^\top \Xsrcone)\right)^2 = \left(\beta^\top (\SigmaXtar - \SigmaXsrcone )\beta\right)^2.
\end{equation}
Under~\assumpref{mean_shift_linear_SCM}, the matrix $\SigmaXtar - \SigmaXsrcone$ is positive semi-definite and has rank $\rdip$, so the covariance matching penalty encourages the prediction model to use covariates in the remaining $(d-\rdip)$-dimensional subspace. 

The corresponding DIP is defined as follows. Let $\Vdip\in\R^{d\times \rdip}$ denote the matrix of eigenvectors corresponding to the $\rdip$ nonzero eigenvalues of $\SigmaXtar - \SigmaXsrcone$.
\begin{itemize}
	\item \textit{DIP$^{(1)}$-cov}: the population DIP$^{(1)}$ with covariance matching penalty is
	\begin{equation}\label{eqn:dip_cov_matching_pop}
		\betasrconedip \defn \argmin_{\beta} \EE{\left(\Ysrcone - \beta^\top \Xsrcone\right)^2} \text{ s.t. } \ \ \vecnorm{\Pi_{\Vdip} \beta}{2}=0,
	\end{equation}
	where $\Pi_{\Vdip}$ is the projection operator onto $\col{\Vdip}$. Since $\SigmaXtar - \SigmaXsrcone$ is positive semi-definite, the constraint $\vecnorm{\Pi_{\Vdip} \beta}{2}=0$ is equivalent to $\left(\beta^\top (\SigmaXtar - \SigmaXsrcone)\beta\right)^2=0$. 
	
	\item \textit{DIP$^{(1)}$-cov in finite-sample:} the finite-sample DIP$^{(1)}$-cov based on the first source (labeled and unlabeled) datasets $\Dsrcone,\Dsrconeu$, and the unlabeled target dataset $\Dtaru$, is
	\begin{align}\label{eqn:dip_cov_matching_finite}
		\betasrconediph &\defn \argmin_{\beta} \frac{1}{\nsrcone}\sum_{i=1}^{\nsrcone} \left(\Ysrcone_i - \beta^\top \Xsrcone_i\right)^2  \text{ s.t. } \ \ \vecnorm{\Pi_{\Vhdip} \beta}{2} =0,
	\end{align}
	where $\Vhdip\in\R^{d\times \rdip}$ is 
	formed by the top $\rdip$ eigenvectors (in absolute value) of $\SigmaXhtar - \SigmaXhsrcone$, where $\SigmaXhtar,\SigmaXhsrcone$ are the empirical covariance matrices from $\Dtaru, \Dsrconeu$, respectively.
\end{itemize}

In the formulation~\eqnref{dip_cov_matching_finite}, $\widehat\Sigma_X^{(0)}$ and
$\widehat\Sigma_X^{(1)}$ are computed from unlabeled samples to simplify the
analysis. This separation is not essential in practice: since covariance
estimation uses only marginal information about $X$, one may also use the
covariates from labeled samples.
When the source and unlabeled target sample sizes are large, 
the difference between the finite-sample estimator and its population counterpart is expected to be small.
We quantify this error for
DIP$^{(1)}$-cov under CA shift interventions in~\appref{finite_sample_DIP}.


\subsubsection{Fine-tuning method starting from DIP}\label{sec:fine_tune_mixture_mean_shift}

In SSDA with CA shift interventions, the limited labeled target samples can be used to fine-tune
DIP$^{(1)}$-cov while achieving lower excess risk in the target domain. 
The key observation is that DIP$^{(1)}$-cov already solves
the source risk minimization problem on the $(d-\rdip)$-dimensional common
subspace, so only the remaining $\rdip$ directions need to be adjusted for
the target domain.

Specifically, let $\Qdip\in\R^{d\times (d-\rdip)}$ denote an orthonormal basis for the orthogonal complement of $\col{\Vdip}$. 
Under the CA shift intervention in~\assumpref{mean_shift_linear_SCM}, we can arrive at the following alternative expression for the population DIP$^{(1)}$-cov estimator:
\begin{equation}\label{eqn:dip_alternative}
	\begin{aligned}
		\betasrconedip =  \Qdip(\Qdip^\top \SigmaXsrcone \Qdip)^{-1}  \Qdip^\top \EE{\Xsrcone\Ysrcone}  
		&=\Qdip(\Qdip^\top \SigmaXtar \Qdip)^{-1} \Qdip^\top\EE{\Xtar\Ytar} \\
		&= \betatrue-\SigmaXtar{}^{-1}\Vdip(\Vdip^\top \SigmaXtar{}^{-1} \Vdip)^{-1}\Vdip^\top \betatrue.
	\end{aligned}
\end{equation}
The first equality follows directly from solving the constrained problem~\eqnref{dip_cov_matching_pop} in closed form. 
The second equality uses the fact that, on the common subspace $\col{\Qdip}$, the source and target have matching second moments for both covariates and cross-moments of $X$ and $Y$. 
The last equality follows from a standard identity for projection operators. We provide additional details of this derivation in~\appref{pop_dip_estimator}. 

Equation~\eqnref{dip_alternative} shows that the difference
$\betatrue-\betasrconedip$ lies in the $\rdip$-dimensional subspace
$\col{\SigmaXtar{}^{-1}\Vdip}$. This suggests a natural fine-tuning strategy that fine-tunes DIP$^{(1)}$-cov only
in this subspace. When $\rdip\ll d$, this can reduce the labeled target sample
complexity from the ambient dimension $d$ to the intervention dimension
$\rdip$.


To formalize this approach in finite-sample, let $\Qhdip\in\R^{d\times (d-\rdip)}$ be an orthonormal basis for the orthogonal complement of $\col{\Vhdip}$. 
\begin{itemize}
	\item \textit{FT-DIP$^{(1)}$}: the fine-tuning estimator under CA shift interventions initiated from DIP$^{(1)}$-cov based on the labeled target dataset $\Dtar$ is
	\begin{equation}\label{eqn:dip_var_matching_finite}
		\begin{aligned}
			\betahftdip\defn\argmin_{\beta: \norm{ \Vhdip^\top \SigmaXhtar \beta}_2\leq \varrho} \frac{1}{\ntar}\sum_{i=1}^{\ntar} \left(\Ytar_i - \beta^\top \Xtar_i\right)^2 \text{ s.t. }\vecnorm{\Qhdip^\top \SigmaXhtar (\beta-\betasrconediph)}{2} = 0,
		\end{aligned}
	\end{equation}
	where $\varrho>0$ is a regularization parameter to be chosen.
\end{itemize}
At the population level, \eqnref{dip_alternative} implies
$\Qdip^\top\SigmaXtar(\betatrue-\betasrconedip)=0$. FT-DIP$^{(1)}$ replaces
this population constraint with its empirical analogue, using
$\SigmaXhtar$ and $\Vhdip$ estimated from unlabeled source and target covariates. Thus,
FT-DIP$^{(1)}$ is a two-stage SSDA method where
unlabeled covariates determine the fine-tuning subspace, while labeled target
samples estimate the correction within that subspace.

\subsubsection{Finite-sample risk guarantees of FT-DIP}\label{sec:fine_tune_finite_sample_guarantee}

Under CA shift interventions, our first main theorem establishes a finite-sample target excess risk upper bound for FT-DIP$^{(1)}$. 
To quantify this bound, we introduce the eigen-gap condition that controls estimation of the 
perturbation subspace. Letting $\Adiff= (bw_Y^\top + W)(bw_Y^\top + W)^\top$, we define
\begin{equation}\label{eqn:ftdip_eigengap}
	\lamdipgap \coloneqq  \lambda_{\rdip}(\Adiff) \cdot \lammin((\ident-B)^{-\top }(\ident - B)^{-1}),
\end{equation}
where $\lambda_{\rdip}(\Adiff)$ denotes the $\rdip$-th largest eigenvalue of $\Adiff$ (i.e., the smallest nonzero eigenvalue under rank-$\rdip$), 
and $\lammin((\ident-B)^{-\top }(\ident - B)^{-1})$ is the minimum eigenvalue of $(\ident-B)^{-\top }(\ident - B)^{-1}$. 
Note that the quantity \(\lamdipgap\) increases when (i) the columns of $W$ become more linearly independent, (ii) the vector $b$ becomes more linearly independent of the column space of $W$, or (iii) the magnitudes of $W$, $b$, and $w_Y$ become larger. 
Therefore, the condition $\lamdipgap>0$ in \thmref{dip_var_matching_finite} is precisely the finite-sample eigen-gap condition that allows the perturbation subspace to be estimated from unlabeled data.

To formalize the dependence of excess risk upper bound on this eigen-gap, we define the condition number $\tsigmagap\coloneqq \lamdipgap^{-1}\sigma_X^2$, and for any $\delta>0$, define the shorthand $\ddel \defn d + \log(1/\delta)$. We are ready to state the theorem.

\begin{theorem}[FT-DIP excess risk upper bound in CA shift]\label{thm:dip_var_matching_finite}
	Under~\assumpsref{linear_SCMs},~\assumpssref{sub_gaussianity},~\assumpssref{mean_shift_linear_SCM}, there exist universal constants $c,c_1,c_2>0$ such that for $\delta \in (0, 1/10)$, if $\ntar\geq \rdip \log (1/\delta)$, $\nsrcone\geq c_1 \tsigmamin^2 d\log(1/\delta),\nsrconeu \geq c_1 \tsigmagap^2  \ddel,\ntaru \geq c_1 \max\{1,\tsigmamin^2,\tsigmagap^2 \} \ddel$,
	\sloppy and if we choose $\varrho \geq c_2 \lammaxbar\parenth{\vecnorm{\betatrue}{2} + \kappabar^{3/2}\norm{\betasrcone}_{2}+\kappabar^{3/2}\frac{\sigma_Y}{\sigma_X} }$ in~\eqnref{dip_var_matching_finite}, then with probability at least $1-10\delta$, the estimator $\betahftdip$ satisfies
	\begin{align*}
	\delta\risktar{f_{\betahftdip}} \leq c \eta  {\frac{\rdip}{\ntar}}\log(1/\delta)
		+ C \brackets{\frac{d\log(1/\delta)}{\nsrcone}
		+  {\frac{\ddel}{\nsrconeu}} + {\frac{\ddel}{\ntaru}} },
	\end{align*}
	where the randomness is over $\Dtar,\Dtaru,\Dsrcone,\Dsrconeu$, and $\eta=\lammaxbar\brackets{\varrho\lamminbar^{-1}+\frac{\sigma_Y}{\sigma_X} }^2 \kappabar^2\tsigmamin^2 $, and $C = \O{\lammaxbar \parenth{\norm{\betatrue}_2 + \norm{\betasrcone}_2 + \norm{\betasrconedip}_2+\frac{\sigma_Y}{\sigma_X}}^2 \poly(\kappabar,\tsigmamin, \tsigmagap) }$.
\end{theorem}

The proof of~\thmref{dip_var_matching_finite} is given
in~\appref{dip_var_matching_finite}. The bound in~\thmref{dip_var_matching_finite} has two main components. The
first term is the error from fine-tuning in the $\rdip$-dimensional target
subspace using $\ntar$ labeled target samples. The second term is the nuisance
estimation error from estimating DIP$^{(1)}$-cov and the subspace
$\col{\SigmaXtar{}^{-1}\Vdip}$ using labeled source samples and unlabeled
source and target covariates. Thus FT-DIP$^{(1)}$ achieves small excess risk
with $\ntar=\O{\rdip}$, provided that the source and unlabeled sample sizes are
large enough, whereas target-only methods require $\ntar=\O{d}$ in general as shown in~\propref{anticausal_minimax_lowerbound}.

To characterize the tightness of the above upper bound, we provide a lower bound on the minimax excess risk rate in CA shift interventions. 
\begin{theorem}[Minimax excess risk lower bound in CA shift]\label{thm:dip_var_matching_finite_minimax}
	Let $\Pfrak$ denote the class of joint distributions over $\Dsrc,\Dsrcu$, $m=0,1$, where each source and target (labeled and unlabeled) sample is generated according to the linear SCMs with CA shift interventions in~\assumpref{mean_shift_linear_SCM}. For any $\nsrcone,\nsrconeu,\ntaru>0$, the minimax excess risk~\eqnref{minimax_excess_risk} over $\Pfrak$ satisfies
	 \begin{equation}\label{eqn:dip_var_matching_finite_minimax}
	 	\min_{\betah}  \sup_{\Pset \in \Pfrak} \EE{\delta\risktar{f_{\betah}} } \geq c\sigma_Y^2 \frac{\rdip}{\ntar}, 
	 \end{equation}
	where $c>0$ is a universal constant.  The minimum is over all measurable estimators $\betah$ based on $(\Dsrcone,\Dsrconeu,\Dtar,\Dtaru)$.
\end{theorem}
The proof of~\thmref{dip_var_matching_finite_minimax} is given
in~\appref{minimax_ftdip_proof}. The lower bound is constructed in the
constant-order regime where
$\kappabar,\tsigmamin,\tsigmagap$ and
$\vecnorm{\betatrue}{2},\norm{\betasrcone}_{2},\norm{\betasrconedip}_{2}$
are bounded, and $\sigma_X$ is a constant multiple of $\sigma_Y$. In this
regime, since $\lammaxbar\leq\sigma_X^2$, the quantity $\eta$ in
\thmref{dip_var_matching_finite} satisfies $\eta=\O{\sigma_Y^2}$.
Therefore, \thmref{dip_var_matching_finite} and
\thmref{dip_var_matching_finite_minimax} establish that FT-DIP$^{(1)}$
achieves the target-label minimax rate up to constant factors whenever,
up to logarithmic factors, the source and unlabeled-data errors satisfy
$$
	\frac{d}{\nsrcone}
	+
	\frac{d}{\nsrconeu}
	+
	\frac{d}{\ntaru}
	\lesssim
	\frac{\rdip}{\ntar}.
$$
Thus, when source and unlabeled covariate samples are sufficiently large,
no other estimator can substantially improve the dependence on the number of
labeled target samples under CA shift interventions.

\begin{remark}\label{rmk:naive_OLS}
	If there is no confounder, i.e.,$w_Y=0$ in~\assumpref{mean_shift_linear_SCM}, the additive shift affects
	only the covariate noise. In this covariate-only intervention case, the
	upper bound in~\thmref{dip_var_matching_finite} remains valid, but the
	target-label minimax interpretation is no longer tight. Indeed,
	$\EE{\Xsrcone\Ysrcone}=\EE{\Xtar\Ytar}$, and hence
	$\betatrue=\SigmaXtar{}^{-1}\EE{\Xsrcone\Ysrcone}$. Thus the naive plug-in
	estimator
	$
		\widehat{\beta}_{\textnormal{naive}}
		=
		\SigmaXhtar{}^{-1}
		\left[
		\frac{1}{\nsrcone}
		\sum_{i=1}^{\nsrcone}\Xsrcone_i\Ysrcone_i
		\right]
	$
	can estimate the oracle without labeled target samples. 
	Nevertheless, when there is confounder with $w_Y\neq 0$,  
	the target cross-moment generally changes, so this plug-in estimator is no
	longer consistent and the low-dimensional fine-tuning step becomes necessary.
\end{remark}

\subsection{Sparse connectivity shift interventions under linear SCMs}\label{sec:fine_tuning_connectivity_shift}

Next, we consider a distribution shift scenario in which the causal
connections among covariates change between source and target domains.
We call this type of intervention \textit{sparse connectivity (SC) shift
interventions}.
This type of intervention induces sparse changes 
between the source and
target regression coefficients,
connecting our SCM formulation to the sparse transfer learning literature
\citep{li2022transfer,bastani2021predicting}.


\begin{assumption}[Sparse connectivity (SC) shift interventions]
\label{assump:connectivity_shift_linear_SCM}
	In~\assumpref{linear_SCMs}, the first source and target domains share
	the same anticausal weight vector and noise distributions, i.e.,
	$b\defn \bsrcone=\btar$,
	$\epssrcone_X,\epstar_X\sim\Pset_{\eps_X}$, and
	$\epssrcone_Y,\epstar_Y\sim\Pset_{\eps_Y}$, while their connectivity
	matrices $\Bsrcone$ and $\Btar$ differ. Specifically, the intervention
	modifies $\Btar$ in $\rconn$ columns indexed by
	$S=\{j_1,\ldots,j_{\rconn}\}\subseteq[d]$, with $\rconn<d$, so that
	$\Bsrcone_{*S}\neq\Btar_{*S}$ while
	$\Bsrcone_{*S^c}=\Btar_{*S^c}$.
\end{assumption}

Under~\assumpref{connectivity_shift_linear_SCM},
the intervention modifies the connections pointing towards $\rconn$ covariates, 
$X_{[j_1]},X_{[j_2]},\ldots,X_{[j_{\rconn}]}$, 
with respect to the other covariates, where $X_{[j]}$ denotes the $j$-th
coordinate of $X$. 
This effect propagates to all descendants of the intervened covariates 
but does not affect non-descendants. In particular, since $Y$ is an ancestor of 
all covariates in anticausal models, the marginal distribution of $Y$ is unaffected.

To see how SC shift interventions partially affect the population OLS estimators,
define the shared quantities
$\vary\defn\VV{\Ysrc}$ and
$\Sigma_{\eps_X}\defn\EE{\epssrc_X\epssrc_X{}^\top}$ across $m=0,1$.
Applying the Sherman-Morrison formula to the explicit forms of $\betasrcone,\betatrue$ then gives
(see~\appref{formula_olssrc} for details)
\begin{equation}\label{eqn:sc_beta}
			\betasrcone = \frac{\vary}{1+\vary b^\top \Sigma_{\eps_X}^{-1}b} (\ident-\Bsrcone{}^\top)\Sigma_{\eps_X}^{-1}b,\text{ and }
			\betatrue = \frac{\vary}{1+\vary b^\top \Sigma_{\eps_X}^{-1}b} (\ident-\Btar{}^\top)\Sigma_{\eps_X}^{-1}b.
\end{equation}
Inspecting equations in~\eqnref{sc_beta}, we see that since $\Btar-\Bsrcone$ differs only in the columns indexed by $S$,
the vector $\betatrue-\betasrcone$ is supported on $S$ and is therefore
$\rconn$-sparse. This observation naturally leads to a new fine-tuning strategy based on sparsity constraint.

\subsubsection{Fine-tuning OLS-Src with finite-sample guarantees}

The sparsity structure between OLS-Src$^{(1)}$ and the oracle estimator 
highlights that
target adaptation can be performed through a sparse update of
the source OLS estimator. We therefore introduce FT-OLS-Src$^{(1)}$, which
fine-tunes OLS-Src$^{(1)}$ on the labeled target data under an $\ell_1$
constraint.
\begin{itemize}
	\item \textit{FT-OLS-Src$^{(1)}$}: let
	$
	\betahsrcone
	\defn
	\argmin_{\beta}
	\frac{1}{\nsrcone}\sum_{i=1}^{\nsrcone}
	\left(\Ysrcone_i-\beta^\top\Xsrcone_i\right)^2
	$
	denote the finite-sample OLS estimator on the first source dataset.
	The fine-tuning estimator under SC shift interventions initiated from OLS-Src$^{(1)}$ is
	\begin{equation}\label{eqn:connectivity_finite}
		\betahftconn
		\defn
		\argmin_{\beta:\norm{\beta}_2\leq \rho}
		\frac{1}{\ntar}\sum_{i=1}^{\ntar}
		\left(\Ytar_i-\beta^\top\Xtar_i\right)^2
		\quad
		\text{s.t. }
		\vecnorm{\beta-\betahsrcone}{1}\leq \gamma,
	\end{equation}
	where $\gamma>0$ controls the $\ell_1$ distance from $\betahsrcone$,
	and $\rho>0$ is a regularization parameter.
\end{itemize}
We note that a method with similar $\ell_1$ norm regularization appeared 
in~\cite{li2022transfer}, but their motivation is substantially different 
from ours. In particular, their method was proposed to tackle the 
assumed sparse changes in the parameters of interest. 
However, we do not directly impose sparsity on $\betatrue-\betasrcone$. 
The necessity of a sparse structure arises as a consequence of tackling 
a specific kind of interventions in data generation models.
 The point of view that a fine-tuning strategy should be tailor-made 
 for underlying assumptions on the type of interventions that 
 cause distribution shifts is a common theme across all our SSDA methods.  

The following theorem upper bounds the excess risk of FT-OLS-Src$^{(1)}$. 
Recall that $\ddel \defn d + \log(1/\delta)$.
\begin{theorem}[FT-OLS-Src excess risk upper bound in SC shift]\label{thm:connectivity_estimator_finite}
	Under~\assumpsref{linear_SCMs},~\assumpssref{sub_gaussianity},~\assumpssref{connectivity_shift_linear_SCM}, there exist universal constants $c, c_1>0$ such that for $\delta \in (0, 1/5)$ and $d\geq 3$, if $\ntar\geq \log d + \log (1/\delta)$ and $\nsrcone \geq c_1 \max\braces{\ddel \tsigmamin^2,d\log (1/\delta)} $,
	and if we choose $\rho \geq \vecnorm{\betastar}{2}$ and $\gamma=\norm{\betatrue - \betahsrcone}_1$ in~\eqnref{connectivity_finite}, with probability at least $1-5\delta $, the estimator $\betahftconn$ satisfies
	\begin{align*}
		\delta\risktar{f_{\betahftconn}} \leq c \eta \frac{\rconn \left(\log d+\log(1/\delta)\right)  }{\ntar} \log (1/\delta) + C \max \braces{\frac{{d\log d}}{\sqrt{\ntar\nsrcone}}, \frac{d^2\log d}{{\ntar\nsrcone}} }, 
	\end{align*}	
	\sloppy where the randomness is over $\Dtar$ and $\Dsrcone$, $\eta = \lammaxbar  \parenth{\rho + \frac{\sigma_Y}{\sigma_X}}^2 \tsigmamin^2 \kappabar$, and $
    C = \O{\lammaxbar \parenth{ \norm{\betatrue}_2 + \norm{\betasrcone}_{2} + \frac{\sigma_Y}{\sigma_X}}^2 \poly( \kappabar,\tsigmamin) \cdot  \log^2(1/\delta) }$.
\end{theorem}
The proof of~\thmref{connectivity_estimator_finite} is given
in~\appref{proof_connectivity_theorem}. In~\thmref{connectivity_estimator_finite}, 
the bound has two main components.
The first term, of order $\rconn\log d/\ntar$, is the finite-sample error from fine-tuning
with $\ntar$ labeled target samples under the $\ell_1$ constraint, while the second
term is the finite-sample error from estimating $\betasrcone$ using source data.
In particular, the excess risk becomes small when $\ntar=\O{ \rconn}$, given $\ntar\nsrcone \gg  d^2$. 
In the regime where $\rconn \ll d$, it suffices to have $\nsrcone \gtrsim d^2/\rconn$ and $\ntar\gtrsim \rconn$ to guarantee nontrivial excess risk. This improves upon the order $d$ labeled target samples needed for the target-only methods in~\propref{anticausal_minimax_lowerbound}. 


To show that the upper bound is tight, we provide a lower bound of the minimax excess risk rate under SC shift.
\begin{theorem}[Minimax excess risk lower bound in SC shift]\label{thm:connectivity_estimator_finite_minimax}
	Let $\Pfrak$ denote the class of joint distributions over $\Dsrc,\Dsrcu$, $m=0,1$, where each source and target (labeled and unlabeled) sample is generated according to the linear SCMs with SC shift interventions in~\assumpref{connectivity_shift_linear_SCM}. For any $\nsrcone,\nsrconeu,\ntaru>0$, the minimax excess risk~\eqnref{minimax_excess_risk} over $\Pfrak$ satisfies
	\begin{equation}\label{eqn:connectivity_estimator_finite_minimax}
		\min_{\betah}  \sup_{\Pset \in \Pfrak} \EE{\delta\risktar{f_{\betah}} } \geq c\sigma_Y^2 \frac{\rconn}{\ntar}, 
	\end{equation}
	where $c>0$ is a universal constant.  
	The minimum is over all measurable estimators $\betah$ 
	based on $(\Dsrcone,\Dsrconeu,\Dtar,\Dtaru)$.
\end{theorem}
The proof of~\thmref{connectivity_estimator_finite_minimax} is given in~\appref{minimax_ftsparse_proof}. 
Analogous to the discussion following~\thmref{dip_var_matching_finite_minimax}, 
we focus on the regime where 
all condition numbers and the magnitudes of the population parameters, $\kappabar,\tsigmamin,\vecnorm{\betatrue}{2},\norm{\betasrcone}_{2}$,
are bounded and $\sigma_X$ is proportional to $\sigma_Y$.
In this regime,
FT-OLS-Src$^{(1)}$ matches the target-label minimax rate up to logarithmic factors whenever the source estimation term in \thmref{connectivity_estimator_finite} is dominated by the sparse fine-tuning term, for instance when
\[
\max\left\{\frac{d}{\sqrt{\ntar\nsrcone}},\frac{d^2}{\ntar\nsrcone}\right\}\lesssim \frac{\rconn}{\ntar},
\]
up to logarithmic factors.
Therefore, when the source sample size $\nsrcone$ is sufficiently large, no other
estimator can substantially improve the dependence on $\ntar$ under SC shift.
Importantly, FT-OLS-Src$^{(1)}$ achieves this guarantee 
without making use of unlabeled target data,  
indicating that unlabeled target data are not
essential for achieving the target-label minimax rate in this regime.

\subsection{Anticausal weight shift interventions under linear SCMs}\label{sec:cip_finetune}

The third distribution shift scenario we consider is induced by changes in the
anticausal weight vector $b$, which determines how the label $Y$ generates
the covariates $X$. We call this type of intervention
\textit{anticausal weight (AW) shift interventions}.
This setting captures label-dependent shifts in the generation
mechanism of $X$: the way a label manifests in the observed covariates
changes across domains. This is the regime designed for invariant methods such
as IRM~\citep{arjovsky2019invariant} and CIP~\citep{gong2016domain,heinze2021conditional,li2018deep}, 
and the same label-dependent shifts have also been used in
theoretical analyses of IRM~\citep{rosenfeld2020risks} and CIP~\citep{chen2021domain}.
Under AW shift, marginal alignment methods such as DIP can severely fail, a phenomenon
well known in the causal domain adaptation literature
\citep{zhao2019learning,johansson2019support,wu2023prominent}, and we provide a
concrete example in~\appref{failure_DIP_AW_shift} for completeness.

\begin{assumption}[Anticausal weight (AW) shift interventions]\label{assump:coefficient_shift_linear_SCM}
	In~\assumpref{linear_SCMs}, all source and target domains share the same connectivity matrices and the noise distributions, i.e., $B\defn\Bsrc$ for all $m\geq 0$, and $\epssrc_X\sim \Pset_{\eps_X}$ and $\epssrc_Y\sim \Pset_{\eps_Y}$ are independently drawn from the same distributions across all $m\geq 0$. The intervention modifies the anticausal weight vector in a way that $\btar\neq \bsrc$, $m=1,\ldots,M$, while 
	\begin{equation}\label{eqn:cic_assumption}
		\btar - \bsrcone \in \spn(b^{(2)}-\bsrcone, \ldots, b^{(M)}-\bsrcone),
	\end{equation}
	where $\dim(\spn(b^{(2)}-\bsrcone, \ldots, b^{(M)}-\bsrcone))=\rcip$ with $1 \leq \rcip < d$. We further assume that $\bsrcone\notin \spn(b^{(2)}-\bsrcone, \ldots, b^{(M)}-\bsrcone)$ and $\rcip+1 \leq M \leq d$.
\end{assumption}  
Unlike~\assumpsref{mean_shift_linear_SCM} and
\assumpssref{connectivity_shift_linear_SCM}, this assumption uses multiple
source domains instead of a single one. Condition~\eqnref{cic_assumption} requires the target shift
$\btar-\bsrcone$ to lie in the span of the perturbation directions observed
in the source domains. This condition becomes more plausible when the source
domains are sufficiently heterogeneous.
It also guarantees the existence of a $(d-\rcip)$-dimensional subspace
orthogonal to
$\spn((\ident-B)^{-1}(\bsrc-\bsrcone),\,m=2,\ldots,M)$, such that
the conditional distribution of the covariates $X$ projected onto this subspace, given the label $Y$, is invariant across all source and target domains~\citep{chen2021domain}.
These projected covariates represent the conditionally
invariant components (CICs), which play an important role in domain adaptation
\citep{gong2016domain,heinze2021conditional,wu2023prominent}.

As discussed above, under AW shift interventions, DIP-type marginal alignment
methods can fail drastically (see~\appref{failure_DIP_AW_shift}).
On the other hand, CIP with conditional mean matching penalty can identify the CICs given many heterogeneous source domains and can provide guarantees on the target population risk~\cite{wu2023prominent}.
The formulation of CIP with conditional mean matching penalty is given below.
\begin{itemize}
	\item \textit{CIP-mean}: the population CIP estimator with conditional mean matching penalty is
	\begin{equation}\label{eqn:cip_mean_matching_pop}
		\begin{split}
			&\betacip = \argmin_{\beta} \frac{1}{M}\sum_{m=1}^{M}\EE{\left(\Ysrc - \beta^\top \Xsrc\right)^2}, \\
			&\hspace{0.1in} \text{s.t. } \ \ \EEst{\beta^\top \Xsrc}{\Ysrc} = \EEst{\beta^\top \Xsrcmprev}{\Ysrcmprev}  \text{ a.s. for all $m\geq 2$}.
		\end{split}
	\end{equation}
\end{itemize}
Under~\assumpref{coefficient_shift_linear_SCM},
$\EEst{\Xsrc}{\Ysrc}=(\ident-B)^{-1}\bsrc\Ysrc$. Hence the conditional
mean-matching constraint is equivalent to requiring $\beta$ to be orthogonal
to the subspace spanned by
$(\ident-B)^{-1}(\bsrc-\bsrcmprev)$, $m\geq2$. Let
$\Vcip\in\R^{d\times\rcip}$ be an orthonormal basis for this subspace. Then
the population constraint can be written as
$\norm{\Pi_{\Vcip}\beta}_2=0$, where $\Pi_{\Vcip}$ is the projection operator onto
$\col{\Vcip}$.

For the finite-sample version, we can estimate $(\ident-B)^{-1}\bsrc$ by
regressing $\Xsrc$ onto $\Ysrc$ in each source domain; see
\appref{finite_sample_CIP}.
Let $\Vhcip$ be the empirical analogue of $\Vcip$, formed by the $\rcip$
dominant left singular vectors of the estimated subspace
$
\spn\left\{
\widehat{(\ident-B)^{-1}\bsrc}
-
\widehat{(\ident-B)^{-1}\bsrcmprev}
:\; m\geq2
\right\}.
$

\begin{itemize}
	\item \textit{CIP-mean in finite-sample:} the finite-sample CIP-mean based on multiple source datasets $\Dsrc$, $m=1,\ldots,M$, is
	\begin{equation}\label{eqn:cip_mean_matching_finite_sample}
		\betahcip = \argmin_{\beta} \frac{1}{M}\sum_{m=1}^M\frac{1}{\nsrc}\sum_{i=1}^{\nsrc}{\left(\Ysrc_i - \beta^\top \Xsrc_i\right)^2}  \text{ s.t. } \ \ \vecnorm{\Pi_{\Vhcip} \beta}{2} =0.
	\end{equation}	
\end{itemize}
When the source sample sizes are large, $\betahcip$ is expected to be close to its population
version $\betacip$. We quantify this error under AW shift interventions in
\appref{finite_sample_CIP}.

\subsubsection{Fine-tuning method starting from CIP}\label{sec:fine_tuning_CIP_finite_sample}

We now introduce a method that fine-tunes CIP-mean for improved target prediction. 
Since CIP can already identify 
$(d-\rcip)$-dimensional conditionally invariant features, 
our strategy is to fine-tune in the remaining $\rcip$ directions with additional labeled target samples.

Recall that the conditional mean matching penalty in~\eqnref{cip_mean_matching_pop} looks for $\beta$ which is orthogonal to $\Vcip$. 
This constraint resembles the DIP penalty in~\eqnref{dip_cov_matching_finite} with $\Vcip$ in place of $\Vdip$. Following the same reasoning used in~\eqnref{dip_alternative}, 
we can then arrive at the equation (see~\appref{pop_cip_estimator} for detailed derivations)
\begin{equation}\label{eqn:cip_alternative}
	\betacip   =\betatrue  -\SigmaXtar{}^{-1}\Vcip(\Vcip^\top \SigmaXtar{}^{-1} \Vcip)^{-1}\Vcip^\top \betatrue \ \Longrightarrow {\Qcip^\top\SigmaXtar (\betatrue-\betacip)}=0,
\end{equation}
where $\Qcip$ is an orthonormal basis for the orthogonal complement of $\col{\Vcip}$. 
If unlabeled target data are available, the constraint in~\eqnref{cip_alternative}
can be directly estimated using the target covariance matrix, leading to the fine-tuning method analogous to FT-DIP~\eqnref{dip_var_matching_finite} (we refer to this method as FT-CIP-Tar in~\secref{numerical_exp}).

On the other hand, a useful feature of the AW shift setting is that the heterogeneous source domains
can identify a fine-tuning subspace even without unlabeled target covariates.
To see this, in AW shift interventions, simple calculations can verify that 
(without loss of generality, the first source domain is used as the reference domain)
\begin{equation}\label{eqn:cip_inv_cov_relation}\SigmaXtar{}^{-1} = \SigmaXsrcone{}^{-1} - \SigmaXsrcone{}^{-1}\begin{bmatrix}
	(\ident-B)^{-1}\bsrcone \  ; & \  \Vcip \alpha
\end{bmatrix}\cdot S, \end{equation}
for some $\alpha\in\R^{\rcip}$ and $S\in\R^{2\times d}$ (see~\appref{pop_cip_estimator} for details). Substituting this into~\eqnref{cip_alternative}, rearranging terms, and multiplying both sides by $\SigmaXsrcone$, we can deduce
\[\SigmaXsrcone (\betacip - \betatrue) =-\left( \ident - \begin{bmatrix}
	(\ident-B)^{-1}\bsrcone \ ; & \ \Vcip \alpha
\end{bmatrix}\cdot S \right)\Vcip(\Vcip^\top \SigmaXtar{}^{-1} \Vcip)^{-1}\Vcip^\top \betatrue. \]
Now let $\Qcipaug\in\R^{d\times (d-\rcip-1)}$ be an orthonormal basis for the
orthogonal complement of
$\spn\{(\ident-B)^{-1}\bsrcone,\Vcip\}$. Then it follows
\begin{equation}\label{eqn:cip_alternative2}
	\Qcipaug^\top \SigmaXsrcone (\betacip -\betatrue)  = 0.
\end{equation}
Comparing with~\eqnref{cip_alternative}, the constraint
\eqnref{cip_alternative2} uses the source covariance matrix rather
than the target covariance matrix. The price of not using unlabeled target
covariates is one additional fine-tuning direction: the difference
$\betatrue-\betacip$ is now constrained to an $(\rcip+1)$-dimensional subspace
instead of an $\rcip$-dimensional subspace. 
In the remainder of this section, we therefore focus on the setting where large source data (both labeled and unlabeled) and
a small set of labeled target samples are available but unlabeled target covariates are
not.


To formally introduce our fine-tuning method under this setup, 
let $\Qhcipaug$ be an orthonormal basis
for the subspace orthogonal to both $\Vhcip$ and
$\widehat{(\ident-B)^{-1}\bsrcone}$, and let $\SigmaXhsrcone$ denote the
empirical covariance matrix computed from the unlabeled first source dataset $\Dsrconeu$.
\begin{itemize}
	\item \textit{FT-CIP}: the fine-tuning estimator under AW shift interventions initiated from CIP-mean based on the labeled target dataset $\Dtar$ is
	\begin{equation}\label{eqn:cip_flne_tuning}
		\betahftcip= \argmin_{\beta: \norm{ \Vhcipaug^\top \SigmaXhsrcone \beta}_2\leq \varrho} \frac{1}{\ntar}\sum_{i=1}^{\ntar} \left(\Ytar_i - \beta^\top \Xtar_i\right)^2 \text{ s.t. }\vecnorm{\Qhcipaug^\top \SigmaXhsrcone (\beta-\betaciph)}{2} = 0,
	\end{equation}
	where $\varrho>0$ is a regularization parameter, and
$\Vhcipaug\in\R^{d\times(\rcip+1)}$ is an orthonormal basis for the
orthogonal complement of $\col{\Qhcipaug}$.
\end{itemize}
Note that both FT-CIP~\eqnref{cip_flne_tuning} and FT-DIP$^{(1)}$~\eqnref{dip_var_matching_finite} 
are formulated as empirical risk minimizer constrained on a low-dimensional subspace; the only difference is how the fine-tuning
subspace is formed. Since in AW shift interventions all quantities in the constraint of FT-CIP can be estimated from multiple source datasets,
FT-CIP provides an effective two-stage SSDA method which
does not need any unlabeled target data.

\subsubsection{Finite-sample risk guarantees of FT-CIP}\label{sec:finite_sample_guarantees_FT_CIP}
Our theoretical results establish the optimality of FT-CIP under AW shift interventions 
when there is no access to large unlabeled target dataset. 
The upper bound on the excess risk of FT-CIP involves a parameter that plays a role similar to the eigen-gap $\lamdipgap$~\eqnref{ftdip_eigengap} for FT-DIP. 
Specifically, define the matrix 
\begin{equation}\label{eqn:def_Paug}
	\Paug\defn\begin{pmatrix}
		\left(\ident - B\right)^{-1}\bsrcone, \left(\ident - B\right)^{-1}\left(\bsrctwo -\bsrcone\right),\cdots, \left(\ident - B\right)^{-1}\left( \bsrcend - \bsrcendprev\right)
	\end{pmatrix} .
\end{equation}
Under~\assumpref{coefficient_shift_linear_SCM}, $\Paug$ has rank $\rcip+1$, and $\Qcipaug$ spans the subspace orthogonal to the (left) singular space of $\Paug$. 
We then define $\lamcipauggap$ as 
\begin{equation}\label{eqn:singular_gap_aug}
	\lamcipauggap \coloneqq \min_{i:\sigma_i(\Paug)\neq 0}\sigma_i(\Paug),
\end{equation}
where $\sigma_i(\Paug)$ denotes the $i$-th largest singular value of $\Paug$. 
Note that $\lamcipauggap$ measures how well the nonzero singular values of $\Paug$ are separated from zero. It increases when the perturbation directions $\left(\ident -B\right)^{-1}(\bsrc-\bsrcprev)$ are more diverse across source domains, when these directions are more linearly independent of the source direction $\left(\ident -B\right)^{-1}\bsrcone$, and when all these directions have larger magnitudes. In particular, a larger $\lamcipauggap$ improves the identifiability of the singular space of $\Paug$, which leads to tighter control and improved risk bounds for FT-CIP.

To formally state the excess risk bound for FT-CIP, 
let $\nu_Y^2\defn\VV{\Ysrc}$ denote the variance of $\Ysrc$ 
under AW shift (which is common across all source domains), 
and let $\nbar\defn\min_{m=1,\ldots,M}\nsrc$ denote the minimum source sample size across all source domains. 
Define the condition number $\tsigmagapcip\defn \frac{\sigma_X}{\lamcipauggap\nu_Y}$ that involves $\nu_Y$ and $\lamcipauggap$, and 
recall the shorthand $d_\delta\defn d+\log(1/\delta)$.
The following is the main excess risk bound regarding FT-CIP. 
\begin{theorem}[FT-CIP excess risk upper bound in AW shift]\label{thm:cip_mean_matching_finite}
	Under~\assumpsref{linear_SCMs},~\assumpssref{sub_gaussianity},~\assumpssref{coefficient_shift_linear_SCM}, there exist universal constants $c,c_1,c_2, c_3>0$ such that for $\delta \in (0,1/(c_1M))$, if $\ntar\geq \rcip \log (1/\delta)$, $\nbar\geq c_2 \max \braces{\tsigmamin^2 \ddel,  \tsigmamin^2 d\log(1/\delta),\frac{\sigma_Y^4}{\nu_Y^4}\log (1/\delta) }$, $\nsrconeu\geq c_2\max\{1,\tsigmamin^2\}d_\delta$, and if we choose $\varrho \geq c_3\lammaxbar\left(\vecnorm{\betatrue}{2} + \kappabar^{3/2}\norm{\betapool}_{2}+\kappabar^{3/2}\frac{\sigma_Y}{\sigma_X}\right)$, with probability at least $1- c_1 M \delta$, the estimator $\betahftcip$ satisfies
	\begin{align*}
		\delta\risktar{f_{\betahftcip}} \leq c \eta \frac{\rcip}{\ntar} \log(1/\delta)
		+ C \brackets{{\frac{\ddel}{\nbar}} 
		+   {\frac{\ddel}{\nsrconeu}} },
	\end{align*}
	where the randomness is over $\parenth{\{\Dsrc\}_{m=0}^{M},\Dsrconeu}$, $\eta=\lammaxbar \brackets{\varrho\lamminbar^{-1}+\frac{\sigma_Y}{\sigma_X}  }^2 \kappabar^2\tsigmamin^2 $, and $C = \O{\lammaxbar \parenth{\norm{\betatrue}_2 + \norm{\betacip}_2 + \norm{\betapool}_2 + \frac{\sigma_Y}{\sigma_X}}^2 \poly(\kappabar,\tsigmamin, \tsigmagapcip)\cdot \log(1/\delta) }$.\footnote{$\betapool$ denotes the pooled estimator trained on the pooled set of source datasets. See~\appref{OLS_pool} for its formal definition and the finite-sample guarantee.}
\end{theorem}
The proof of~\thmref{cip_mean_matching_finite} is given
in~\appref{cip_theorem_proof}. The theorem shows that, for
$M=\rcip+1$, choosing $\delta=\O{1/\rcip}$ yields small excess risk with high
probability as long as $\ntar\gtrsim \rcip\log\rcip$ and the source sample sizes
satisfy $\nbar,\nsrconeu\gg d$. Importantly,
unlike
FT-DIP which uses unlabeled target
covariates to estimate the fine-tuning subspace, FT-CIP attains this guarantee
using (labeled and unlabeled) source data and labeled target samples only.

To characterize the tightness of the above upper bound, we provide a lower bound of the minimax excess risk rate in AW shift interventions.
\begin{theorem}[Minimax excess risk lower bound in AW shift]\label{thm:cip_mean_matching_finite_minimax}
	Let $\Pfrak$ denote the class of joint distributions over $\Dtar$ and $\{\Dsrc,\Dsrcu\}_{m=1}^{M}$ where each source and target (labeled and unlabeled) sample is generated according to the linear SCMs with AW shift interventions in~\assumpref{coefficient_shift_linear_SCM}. 
	For any $\nsrcone,\ldots,\nsrcend,\nsrconeu,\ldots,n^{(M)}_\textnormal{u}>0$, the minimax excess risk~\eqnref{minimax_excess_risk} over $\Pfrak$ satisfies
	\begin{equation}\label{eqn:cip_mean_matching_finite_minimax}
		\min_{\betah}  \sup_{\Pset \in \Pfrak} \EE{\delta\risktar{f_{\betah}} } \geq c\sigma_Y^2 \frac{\rcip}{\ntar}, 
	\end{equation}
	where $c>0$ is a universal constant. 
	The minimum is over all measurable
	estimators $\betah$ based on $(\Dtar,\{\Dsrc,\Dsrcu\}_{m=1}^{M})$; the estimator has no access to unlabeled target data.

\end{theorem} 
The proof of~\thmref{cip_mean_matching_finite_minimax} is provided in~\appref{minimax_ftcip_proof}. 
In the setting where all condition numbers $\kappabar,\tsigmamin,\tsigmagapcip$ and 
the magnitudes of the parameters, $\vecnorm{\betatrue}{2},\norm{\betacip}_{2},\norm{\betapool}_2$, 
are bounded, and $X$ and $Y$ are on the same scale ($\sigma_X$ is proportional to $\sigma_Y$),
\thmref{cip_mean_matching_finite_minimax} shows that with no access to unlabeled target data, 
FT-CIP is target-label minimax-optimal up to logarithmic factors 
whenever the finite source terms in the upper bound of~\thmref{cip_mean_matching_finite} are dominated by the fine-tuning term, 
i.e.,
$\frac{d}{\nbar}+\frac{d}{\nsrconeu}\lesssim \frac{\rcip}{\ntar}.$ 
In particular, the lower bound in~\thmref{cip_mean_matching_finite_minimax} applies to 
estimators that may use all source unlabeled datasets,
whereas FT-CIP needs only a single source unlabeled dataset to achieve the minimax rate.

\section{Selecting from multiple fine-tuning strategies in SSDA}\label{sec:model_selection}
Our theoretical guarantees in \secref{finite_sample_theory} are shift-specific: 
FT-DIP, FT-OLS-Src, and FT-CIP are tailored to CA, SC, and AW shift interventions, respectively. 
In practice, the analyst may not know which intervention model is most appropriate. 
We therefore introduce \textit{multi adaptive-start fine-tuning (MASFT)}, 
which constructs several fine-tuned candidates and uses a small labeled target validation set to select among them.


Let 
$
	\Dtarval \defn \{(\Xtarval_i,\Ytarval_i) \}_{i=1}^{\ntarval} \iidsim \Ptar
$
be an additional independent hold-out validation dataset from the target domain.
MASFT proceeds in three steps:
\begin{enumerate}
	\item[1)] Train $L$ many UDA estimators  $\{\thetahdai{1},\ldots,\thetahdai{L} \}$ from the available source and unlabeled target datasets $(\{\Dsrc,\Dsrcu\}_{m=1}^M,\Dtaru)$.
	\item[2)] For each $k=1,\ldots,L$, fine-tune $\thetahdai{k}$ on the labeled target dataset $\Dtar$ by solving~\eqnref{fine_tuning_generic} with $\thetada=\thetahdai{k}$ and a candidate-specific low-complexity set $\Thetasmall=\Thetasmalli{k}$. Denote the resulting estimators by $\{\thetahfti{1},\ldots,\thetahfti{L}\}$.
	\item[3)] Select the fine-tuned estimator with the smallest validation risk on $\Dtarval$:
    \begin{align}\label{eqn:model_selection_step_3}
        \begin{cases}
            \kh = \argmin_{k\in\{1,\ldots,L\}} \frac{1}{\ntarval}\sum_{i=1}^{\ntarval} \ell\bigl(f_{\thetahfti{k}}(\Xtarval_i),\Ytarval_i\bigr),\\
            \thetahatfinal = \widehat\theta_{\textnormal{FT},\kh}.
        \end{cases}
    \end{align}
\end{enumerate}

MASFT succeeds when at least one candidate fine-tuning strategy captures the correct low-dimensional structure of the source-target shift. The validation set is then used only to choose among finitely many candidates, so its sample-size cost is logarithmic in the number of candidates $L$. The excess risk bound of MASFT only occurs a logarithmic overhead from model selection, which is a standard model selection result (see e.g.,~\appref{model_selection_MASFT}).

For the three shift scenarios in \secref{finite_sample_theory}, we instantiate MASFT as follows. For each source domain $m=1,\ldots,M$, define FT-DIP$^{(m)}$ and FT-OLS-Src$^{(m)}$ analogously to FT-DIP$^{(1)}$~\eqnref{dip_var_matching_finite} and FT-OLS-Src$^{(1)}$~\eqnref{connectivity_finite}. Running these two procedures for each source gives $2M$ candidates. We also run one FT-CIP estimator using all source domains. Hence $L=2M+1$, with candidates denoted by $\{\betahfti{1},\ldots,\betahfti{L}\}$, and MASFT returns the validation-selected estimator $\betahatfinal$.

\begin{corollary}\label{cor:MSFT_combined}
Suppose that at least one candidate model is correctly specified: either (i) for some source domain $m\in\{1,\ldots,M\}$, the source-target pair satisfies the CA shift interventions (\assumpref{mean_shift_linear_SCM}) or the SC shift interventions (\assumpref{connectivity_shift_linear_SCM}); or (ii) the source and target domains satisfy the AW shift interventions (\assumpref{coefficient_shift_linear_SCM}). Under the conditions of~\thmsref{dip_var_matching_finite},~\thmssref{connectivity_estimator_finite},~\thmssref{cip_mean_matching_finite}, there exist universal constants $c,c',c''>0$ such that for $\delta\in(0,1/(cM))$, if $\ntarval\ge c'\log(M/\delta)$, then with probability at least $1-cM\delta$,
\[
\delta\Rset(\betahatfinal) \le c''\eta\sqrt{\frac{\log(M/\delta)}{\ntarval}} + \eps_{\textnormal{type}},
\]
where $\eta=\sigma_X^2\left(\frac{\sigma_Y}{\sigma_X}+\max_{j=1,\ldots,2M+1}\vecnorm{\betahfti{j}}{2}\right)^2$,
and $\eps_{\textnormal{type}}$ equals $\eps_{\textnormal{ca}}$, $\eps_{\textnormal{sc}}$, or $\eps_{\textnormal{aw}}$ according to whether the correct shift intervention is CA, SC, or AW. 
Here $\eps_{\textnormal{ca}}$, $\eps_{\textnormal{sc}}$, and $\eps_{\textnormal{aw}}$ are the upper bounds in~\thmref{dip_var_matching_finite},~\thmref{connectivity_estimator_finite}, and~\thmref{cip_mean_matching_finite}, respectively.
\end{corollary}

\corrref{MSFT_combined} follows directly by combining the model selection guarantee in~\propref{model_selection} 
with the individual excess risk bounds from \secref{finite_sample_theory}. 
It shows that MASFT performs nearly as well as the best correctly specified fine-tuning strategy, up to the validation overhead $\sqrt{\log(M)/\ntarval}$. 
Consequently, $\ntarval\gtrsim\log M$ labeled validation target samples suffice to adaptively select among the candidate SSDA strategies.

\section{Numerical experiments}\label{sec:numerical_exp}

In this section, we evaluate the empirical performance of our proposed SSDA
methods on both synthetic and real datasets. 

To compare with our SSDA methods, we implement several
existing UDA and SSDA algorithms. 
For UDA methods, we include OLS-Src, DIP and CIP introduced in~\secref{baseline_DA_algorithms}. 
Additionally, we implement OLS-Pool, which runs OLS on the pooled source datasets (see~\appref{OLS_pool}), 
and several widely used DA methods such as groupDRO~\cite{sagawa2019distributionally}, Invariant Risk Minimization (IRM)~\cite{arjovsky2019invariant}, and V-REx~\cite{krueger2021out}. 
For SSDA, we also evaluate an $\ell_2$-based fine-tuning method, denoted FT-OLS-Src-L2. To differentiate from FT-OLS-Src-L2, FT-OLS-Src-L1 is used to denote FT-OLS-Src with $\ell_1$-constraint~\eqnref{connectivity_finite}.
Whenever unlabeled target data are available,
we also consider FT-CIP-Tar, the variant of FT-CIP 
that incorporates this unlabeled target data for fine-tuning (see the discussion in~\secref{fine_tuning_CIP_finite_sample}). 
For a target-only baseline, we include OLS-Tar, which performs OLS on the labeled target samples only.

\paragraph*{Synthetic experiments}
We first generate synthetic datasets from anticausal linear SCMs under the
three intervention scenarios given in~\secref{finite_sample_theory}, i.e., CA, SC, and AW
shifts. In all simulations, we set $d=100$ and vary the intervention
dimension as $\rdip\in\{5,10,15\}$ for CA shift, $\rconn\in\{5,10,15\}$ for SC
shift, and $\rcip\in\{4,6,8\}$ for AW shift. The labeled target sample size is
fixed at $\ntar=100$, and an additional held-out labeled target validation set
of size $\ntarval=100$ is used for hyperparameter tuning and MASFT selection.
Additional details on the data generation procedures are deferred to~\appref{simulation_details}.

\begin{table}[ht!]
\centering
\footnotesize
\setlength{\tabcolsep}{3pt}
\renewcommand{\arraystretch}{1.1}
\resizebox{\textwidth}{!}{%
\begin{tabular}{|l|ccc|ccc|ccc|}
\hline
& \multicolumn{3}{c|}{Simulation 1: CA shift} & \multicolumn{3}{c|}{Simulation 2: SC shift} & \multicolumn{3}{c|}{Simulation 3: AW shift} \\
\hline
Method & $\rdip{=}5$ & $\rdip{=}10$ & $\rdip{=}15$ & $\rconn{=}5$ & $\rconn{=}10$ & $\rconn{=}15$ & $\rcip{=}4$ & $\rcip{=}6$ & $\rcip{=}8$ \\
\hline
OLS-Src           & $53.6\pm34.5$ & $64.9\pm27.0$ & $94.4\pm40.9$ & $1.82\pm0.85$ & $3.87\pm2.15$ & $5.02\pm2.27$ & $17.3\pm2.97$ & $17.1\pm2.85$ & $16.6\pm3.88$ \\
OLS-Pool          & $145\pm85$    & $151\pm59$    & $171\pm49$    & $2.67\pm1.09$ & $4.58\pm1.70$ & $5.67\pm2.39$ & $5.68\pm2.26$ & $9.66\pm2.71$ & $9.22\pm2.48$ \\
DIP               & $24.9\pm15.1$ & $27.9\pm13.6$ & $24.7\pm7.83$ & $21.5\pm3.01$ & $24.1\pm0.56$ & $24.4\pm0.73$ & $22.7\pm1.16$ & $23.1\pm0.73$ & $23.1\pm0.95$ \\
CIP               & $147\pm91$    & $151\pm58$    & $170\pm44$    & $2.68\pm1.10$ & $4.57\pm1.67$ & $5.64\pm2.33$ & $0.84\pm0.42$ & $0.98\pm0.51$ & $1.06\pm0.55$ \\
groupDRO          & $139\pm90$    & $157\pm70$    & $169\pm57$    & $2.65\pm1.15$ & $4.54\pm1.69$ & $5.66\pm2.36$ & $2.75\pm1.80$ & $6.65\pm3.03$ & $6.48\pm2.74$ \\
IRM               & $63.1\pm23.2$ & $73.5\pm25.6$ & $63.3\pm15.1$ & $2.67\pm1.11$ & $4.60\pm1.76$ & $5.69\pm2.32$ & $1.44\pm0.58$ & $2.89\pm1.63$ & $2.72\pm1.33$ \\
V-REx              & $111\pm69$    & $143\pm70$    & $143\pm66$    & $2.66\pm1.10$ & $4.50\pm1.60$ & $5.57\pm2.03$ & $1.72\pm0.85$ & $3.49\pm1.86$ & $3.70\pm1.91$ \\
\hline
FT-DIP            & $\boldsymbol{0.28\pm0.13}$ & $\boldsymbol{0.57\pm0.24}$ & $\boldsymbol{0.82\pm0.26}$ & $1.12\pm0.38$ & $1.11\pm0.39$ & $1.16\pm0.37$ & $1.97\pm0.97$ & $1.51\pm0.29$ & $1.45\pm0.37$ \\
FT-CIP            & $54.6\pm46.9$ & $97.2\pm29.2$ & $124\pm44$    & $1.35\pm0.25$ & $2.56\pm0.58$ & $3.55\pm1.03$ & $\boldsymbol{0.15\pm0.07}$ & $\boldsymbol{0.21\pm0.09}$ & $\boldsymbol{0.26\pm0.14}$ \\
FT-CIP-Tar        & $148\pm92$    & $151\pm59$    & $170\pm44$    & $1.95\pm0.89$ & $3.84\pm1.63$ & $4.96\pm2.12$ & $\boldsymbol{0.13\pm0.07}$ & $\boldsymbol{0.20\pm0.09}$ & $\boldsymbol{0.24\pm0.12}$ \\
FT-OLS-Src-L1     & $1.14\pm0.43$ & $1.92\pm0.73$ & $2.96\pm1.39$ & $\boldsymbol{0.075\pm0.028}$ & $\boldsymbol{0.16\pm0.06}$ & $\boldsymbol{0.28\pm0.11}$ & $2.07\pm0.67$ & $1.77\pm0.64$ & $2.04\pm0.85$ \\
FT-OLS-Src-L2     & $1.07\pm0.37$ & $1.86\pm0.73$ & $2.84\pm1.44$ & $0.43\pm0.06$ & $0.65\pm0.12$ & $0.80\pm0.13$ & $1.95\pm0.62$ & $1.76\pm0.64$ & $1.96\pm0.78$ \\
MASFT             & $0.43\pm0.45$ & $0.58\pm0.24$ & $0.82\pm0.25$ & $0.075\pm0.028$ & $0.16\pm0.06$ & $0.28\pm0.11$ & $0.13\pm0.07$ & $0.19\pm0.09$ & $0.24\pm0.13$ \\
\hline
OLS-Tar           & $>1000$ & $>1000$ & $>10000$ & $>1000$ & $>100$ & $>100$ & $>10000$ & $>100$ & $>1000$ \\
OLS-Tar-MoreData  & $0.17\pm0.04$ & $0.40\pm0.10$ & $0.66\pm0.10$ & $0.013\pm0.003$ & $0.029\pm0.006$ & $0.040\pm0.010$ & $0.09\pm0.02$ & $0.13\pm0.04$ & $0.19\pm0.06$ \\
\hline
\end{tabular}%
}
\caption{
Empirical target excess risks on linear SCMs under three different interventions, averaged over $20$ trials.
Bold indicates the matching fine-tuning methods, which achieve the best or comparable performance among all other methods:
FT-DIP under CA shift, FT-OLS-Src-L1 under SC shift, and FT-CIP/FT-CIP-Tar under AW shift.}
\label{tab:simulation_results}
\end{table}

\tabref{simulation_results} summarizes the empirical target excess risks averaged over
$20$ trials.
In each shift setting, 
the fine-tuning method tailored to the corresponding shift type achieves the best or nearly best performance, 
consistently outperforming all other UDA baselines and OLS-Tar:
FT-DIP under CA shift, FT-OLS-Src-L1 under SC shift, and FT-CIP/FT-CIP-Tar under AW shift.
Additionally, 
we include OLS-Tar-MoreData, which runs OLS on a larger labeled target
dataset, where the sample size is chosen to match the target-label theoretical bound
of the tailored fine-tuning method in each shift setting, i.e., $d/\ntarnew = r/\ntar$ (ignoring constants),
where $r\in\{\rdip,\rconn,\rcip\}$ is the shift
dimension of each shift setting.
As shown by \tabref{simulation_results}, FT-DIP and FT-CIP/FT-CIP-Tar 
achieve performance comparable to OLS-Tar-MoreData, while 
FT-OLS-Src-L1 performs slightly worse than OLS-Tar-MoreData under SC shift. Nevertheless, it substantially 
outperforms all remaining methods, including the fine-tuning method with $\ell_2$ norm penalty.
Finally, without prior knowledge of the shift type, 
MASFT which selects among our proposed fine-tuning candidates by target validation loss
achieves performance comparable to the corresponding tailored method in each shift setting.

\paragraph*{Application I: SSDA in a real physical system}


We next evaluate our SSDA methods on data generated by a real physical system: the Light Tunnel Mk2 from the Causal Chambers benchmark~\cite{gamella2025causal}. 
The device consists of a controllable RGB light source, two motorized polarizers,
and light-intensity sensors placed along the optical path. 
The \texttt{linked\_leds} configuration, a variation of the standard light-tunnel configuration, exposes light-source controls (\texttt{R}, \texttt{G}, \texttt{B}), 
polarizer settings and angle measurements (e.g., \texttt{$\theta_1$}, \texttt{$\theta_2$}), 
sensor measurements (\texttt{$\widetilde I_1$}, \texttt{$\widetilde I_2$}, \texttt{$\widetilde I_3$}, \texttt{$\widetilde V_1$}, \texttt{$\widetilde V_2$}, \texttt{$\widetilde V_3$}), and sensor parameters such as exposure time and photodiode choice.
Additionally, it comes with a ground-truth causal graph 
(\figref{light_tunnel_graph})
whose edges have an interventional interpretation.
For our experiments, we treat the light-source control \texttt{R} as the response $Y$ and 
use ten selected light-source, sensor and polarizer variables (\texttt{G}, \texttt{B}, \texttt{$\widetilde I_1$}, \texttt{$\widetilde I_2$}, \texttt{$\widetilde I_3$}, \texttt{$\widetilde V_1$}, \texttt{$\widetilde V_2$}, \texttt{$\widetilde V_3$}, \texttt{$\theta_1$}, \texttt{$\theta_2$}) as covariates $X$. 
Physically, the light-source setting generates the downstream sensor readings 
through the optical system, whereas prediction reverses this direction by estimating the source setting from its measured consequences. 
This makes the Light Tunnel a natural real system to test the anticausal SCMs considered in \secref{setup}: 
the source and target domains are produced by actual interventions on a laboratory device rather than by simulated perturbations of a linear model.

We use three light-tunnel datasets, each consisting of multiple source and target domains and chosen
to highlight one of the three intervention types studied in~\secref{finite_sample_theory}.
In \texttt{lt\_CA}, the source-to-target shift is induced by a weak intervention
on \texttt{$\widetilde V_1$}, which can be approximated as an additive shift.
In \texttt{lt\_SC}, the source-to-target intervention changes the relationship
between \texttt{$\widetilde I_2$} and \texttt{$\widetilde V_3$} in a nonlinear way, leading to a
sparse-connectivity-type shift.
In \texttt{lt\_AW}, the target domain is generated by a strong intervention on
\texttt{$\widetilde I_2$}, which can be approximated as a large anticausal-weight shift.
See~\appref{light_tunnel_details} for further details on the data collection and
intervention procedures. In all collections, each environment contains $5000$ observations.
From the target environment, we reveal only $\ntar=40$ labeled samples for
fine-tuning and $\ntarval=40$ labeled samples for validation.
For comparison, we also include OLS-Tar-MoreData, a target-only
baseline trained on the target training set of size $4000$.

\begin{figure}[t]
\centering
\begin{minipage}[t]{0.3\textwidth}
	\centering
	\vspace{0pt}
	\resizebox{\linewidth}{!}{\begin{tikzpicture}[
  x=0.72cm,
  y=1.12cm,
  >=Latex,
  causal/.style={-{Latex[length=2.0mm,width=1.35mm]}, line width=0.62pt},
  intervention/.style={rectangle, draw=red, text=red, line width=0.8pt, minimum width=0.48cm, minimum height=0.48cm, inner sep=0pt, outer sep=0pt, font=\Large\bfseries},
  intervention path/.style={red, line width=0.7pt},
  intervention arrow/.style={-{Latex[length=2.0mm,width=1.35mm]}, red, line width=0.7pt},
  var/.style={inner sep=1.0pt, outer sep=0pt, font=\Large},
  measured/.style={var},
  control/.style={var, font=\Large\bfseries},
  title/.style={font=\sffamily\normalsize, anchor=west, align=left, text width=3.8cm}
]

\node[title] at (-5.45,4.35) {Light Tunnel \texttt{linked\_leds}};

\node[control] (rgb) at (-1.25,2.10) {$\bm{(R,G,B)}$};

\node[control]  (th1)  at (1.95,2.55) {$\bm{\theta_1}$};
\node[control]  (th2)  at (3.45,2.55) {$\bm{\theta_2}$};

\node[measured] (i1) at (-5.10,1.35) {$\widetilde I_1$};
\node[measured] (v1) at (-4.00,-0.95) {$\widetilde V_1$};
\node[measured] (i2) at (-1.85,-2.15) {$\widetilde I_2$};
\node[measured] (v2) at ( 0.80,-2.15) {$\widetilde V_2$};
\node[measured] (i3) at ( 2.35,-1.15) {$\widetilde I_3$};
\node[measured] (v3) at ( 4.55, 0.90) {$\widetilde V_3$};

\foreach \target in {i1,i2,v1,v2,i3,v3}{
  \draw[causal] (rgb) -- (\target);
}

\foreach \source in {th1,th2}{
  \draw[causal] (\source) -- (i3);
  \draw[causal] (\source) -- (v3);
}

\draw[causal] (i2) -- (v3);

\node[intervention] (a) at (-4.85,-3.25) {A};

\coordinate (i2join) at ($(rgb)!0.68!(i2)$);
\coordinate (v3join) at ($(i2)!0.78!(v3)$);

\draw[intervention arrow] (a) to[out=80,in=230] (v1);

\draw[intervention arrow] (a) to[out=80,in=230] (i2);


\draw[intervention arrow] (a) to[out=-8,in=250] (v3);

\draw[causal] (i1) -- (v2);

\end{tikzpicture}}
	\vspace{0.7cm}
	\caption{A ground-truth causal graph provided by the Causal Chambers benchmark}
	\label{fig:light_tunnel_graph}
\end{minipage}\hfill
\begin{minipage}[t]{0.68\textwidth}
	\centering
	\vspace{0pt}
	\footnotesize
	\setlength{\tabcolsep}{3pt}
	\renewcommand{\arraystretch}{1.0}
	\begin{tabular}{|l|ccc|}
	\hline
	& \multicolumn{3}{c|}{Light-tunnel experiments} \\
	\hline
	Method & \texttt{lt\_CA} & \texttt{lt\_SC} & \texttt{lt\_AW}\\
	\hline
	OLS-Src           & 0.056 $\pm$ 0.003 & 0.115 $\pm$ 0.009 & 0.281 $\pm$ 0.011 \\
	OLS-Pool          & 0.066 $\pm$ 0.003 & 43.860 $\pm$ 2.015 & 0.130 $\pm$ 0.005 \\
	DIP               & 0.065 $\pm$ 0.005 & 0.076 $\pm$ 0.007 & 0.181 $\pm$ 0.011 \\
	CIP               & 0.065 $\pm$ 0.003 & 29.217 $\pm$ 2.223 & 0.079 $\pm$ 0.004 \\
	groupDRO          & 0.059 $\pm$ 0.004 & 1.650 $\pm$ 0.078 & 0.107 $\pm$ 0.004 \\
	IRM               & 0.056 $\pm$ 0.003 & 0.146 $\pm$ 0.012 & 0.113 $\pm$ 0.004 \\
	V-REx              & 0.058 $\pm$ 0.004 & 0.757 $\pm$ 0.032 & 0.118 $\pm$ 0.004 \\
	\hline
	FT-DIP            & \textbf{0.043 $\pm$ 0.003} & 0.053 $\pm$ 0.010 & 0.059 $\pm$ 0.008 \\
	FT-CIP            & 0.048 $\pm$ 0.007 & 5.250 $\pm$ 0.799 & \textbf{0.059 $\pm$ 0.004} \\
	FT-CIP-Tar        & 0.047 $\pm$ 0.007 & 29.579 $\pm$ 2.270 & \textbf{0.057 $\pm$ 0.005} \\
	FT-OLS-Src-L1     & 0.050 $\pm$ 0.003 & \textbf{0.051 $\pm$ 0.006} & 0.062 $\pm$ 0.005 \\
	FT-OLS-Src-L2     & 0.049 $\pm$ 0.003 & 0.053 $\pm$ 0.005 & 0.061 $\pm$ 0.007 \\
	MASFT             & 0.046 $\pm$ 0.003 & 0.050 $\pm$ 0.005 & 0.057 $\pm$ 0.006 \\
	\hline
	OLS-Tar           & 0.054 $\pm$ 0.010 & 0.054 $\pm$ 0.011 & 0.066 $\pm$ 0.008 \\
	OLS-Tar-MoreData  & 0.041 $\pm$ 0.003 & 0.039 $\pm$ 0.001 & 0.050 $\pm$ 0.003 \\
	\hline
	\end{tabular}
	\captionof{table}{
Target test MSEs in the light-tunnel experiments, averaged over $10$ independent random splits.
Bold indicates the matching fine-tuning methods 
that achieve the best performance among 
all UDA and SSDA methods.
}
	\label{tab:lightchamber_results}
\end{minipage}
\end{figure}

\tabref{lightchamber_results} shows the target test MSEs of different
methods, averaged over $10$ independent random splits.
Across the three light-tunnel collections, the fine-tuning method tailored to the corresponding intervention type 
achieves the best performance among all UDA and SSDA methods:
FT-DIP for \texttt{lt\_CA}, FT-OLS-Src-L1 for \texttt{lt\_SC}, and
FT-CIP/FT-CIP-Tar for \texttt{lt\_AW}.
In particular, these tailored fine-tuning methods show clear improvements over their corresponding UDA counterparts,
namely DIP, OLS-Src, and CIP, respectively, which do not use labeled target data
for fine-tuning. 
MASFT also achieves comparable performance without requiring prior knowledge of the intervention type.
Additionally, for \texttt{lt\_CA} the performance of FT-DIP is comparable to that of OLS-Tar-MoreData,
highlighting the effectiveness of the proposed fine-tuning approach in the
small target-label regime. 

\paragraph*{Application II: SSDA in near-infrared spectroscopy (NIR)}

Finally, we apply our methods to near-infrared spectroscopy (NIR) datasets in calibration transfer problems, which is
a central problem for model maintenance in chemometrics~\citep{roger2025cloning}. 
In NIR, a calibration model predicts a chemical or physical attribute $Y$, 
such as dry matter, moisture, protein content, or a chemical concentration, 
from a high-dimensional spectrum $X$. The data-generating direction, however,
is the opposite: the chemical property $Y$ determines the absorption and scattering 
pattern $X$ observed by the spectrometer, so this fits naturally into an
anticausal structure.
In the idealized setting, by the Beer-Lambert law, 
absorbance is approximately linear in concentration at each wavelength, 
so the generation mechanism can be approximated by a linear SCM. In practice, 
other factors such as  
scattering, baseline effects, and instrument responses make 
this relation only an approximation. 
Nevertheless, this approximate linear structure motivates 
the use of linear calibration models, such as partial least squares (PLS) regression,
which are standard in NIR chemometrics~\citep{zhao2019pls,anderson2020achieving,roger2025cloning}.

In NIR, calibration transfer is needed because a calibration model 
trained in one measurement domain often degrades when deployed in another. 
Such domain changes include different spectrometers, temperatures, 
seasons, locations, and physical forms of the 
same material~\citep{anderson2020achieving,sun2020achieving,sun2020nirs,zhao2019pls}. 
This makes NIR calibration transfer a natural real data setting for our linear SCM-based SSDA methods, where
source domains provide existing labeled calibration data, target spectra are comparatively easy to collect, and only a small number of target reference measurements are available for fine-tuning and validation. 

Here we use two NIR benchmark datasets. The first is the corn dataset used
in calibration transfer studies~\citep{zhao2019pls,zhao2019calibration}, where
the task is to predict moisture content from spectra of the same corn samples
measured on different spectrometers. 
Among the three instrument domains, 
we use \texttt{m5} as the source instrument, set either \texttt{mp6} or \texttt{mp5} as the target, 
and use the remaining instrument as an additional source domain.
For each task, we use $\ntar=24$ labeled target samples for fine-tuning and $\ntarval=12$ for validation, while OLS-Tar-MoreData is trained on $56$ labeled target samples.
The second is the mango dataset~\citep{anderson2020achieving}, where the task is to predict dry-matter content from intact-fruit spectra. 
We consider cultivar shift, where one cultivar is held out as the target,
and ripening-stage shift, where the source and target correspond to different physiological stages.
In~\tabref{nir_results}, the cultivar indices \texttt{0}, \texttt{1}, \texttt{2}, and \texttt{3} denote \texttt{Caly}, \texttt{KP}, \texttt{HG}, and \texttt{R2E2}, respectively,
 while for the ripening-stage shift, \texttt{0} denotes Hard Green and \texttt{1} denotes Ripen.
For all four mango tasks, we use $\ntar=60$ labeled target samples for fine-tuning and $\ntarval=120$ labeled target samples for validation, 
while OLS-Tar-MoreData is trained on a larger target training set, with sizes
$993$, $800$, $693$, and $3398$ for the four mango columns in~\tabref{nir_results}, respectively. 
Further details on preprocessing and additional NIR experiments are given in~\appref{nir_details}.

\begin{table}[t!]
\centering
\footnotesize
\setlength{\tabcolsep}{3pt}
\renewcommand{\arraystretch}{1.0}
\resizebox{\textwidth}{!}{%
\begin{tabular}{|l|cc|cc|cc|}
\hline
& \multicolumn{2}{c|}{Corn} & \multicolumn{2}{c|}{Mango (cultivar shift)} & \multicolumn{2}{c|}{Mango (ripening shift)} \\
\hline
Method & $\texttt{m5}\to\texttt{mp6}$ & $\texttt{m5}\to\texttt{mp5}$ & $\texttt{1,2,3}\to\texttt{0}$ & $\texttt{0,2,3}\to\texttt{1}$ & $\texttt{0}\to\texttt{1}$ & $\texttt{1}\to\texttt{0}$  \\
\hline
OLS-Src           & $0.079\pm0.029$ & $0.060\pm0.028$ & $1.208\pm0.085$ & $2.011\pm0.155$ & $1.821\pm0.124$ & $1.908\pm0.269$ \\
OLS-Pool          & $0.028\pm0.008$ & $0.020\pm0.006$ & $0.944\pm0.078$ & $1.499\pm0.091$ & - & - \\
DIP               & $0.039\pm0.021$ & $0.033\pm0.009$ & $1.689\pm0.193$ & $2.238\pm0.264$ & $1.929\pm0.089$ & $1.649\pm0.158$ \\
CIP               & $0.028\pm0.008$ & $0.021\pm0.007$ & $1.008\pm0.101$ & $1.513\pm0.107$ & - & - \\
groupDRO          & $0.026\pm0.006$ & $0.024\pm0.007$ & $1.378\pm0.082$ & $1.456\pm0.083$ & - & - \\
IRM               & $0.027\pm0.007$ & $0.025\pm0.007$ & $1.274\pm0.080$ & $1.466\pm0.088$ & - & - \\
VREx              & $0.027\pm0.007$ & $0.024\pm0.007$ & $1.406\pm0.065$ & $1.493\pm0.094$ & - & - \\
\hline
FT-DIP            & $\boldsymbol{0.022\pm0.007}$ & $0.024\pm0.012$ & $0.634\pm0.057$ & $1.531\pm0.194$ & $1.611\pm0.072$ & $\boldsymbol{1.100\pm0.093}$ \\
FT-CIP            & $0.025\pm0.012$ & $0.020\pm0.006$ & $0.664\pm0.068$ & $\boldsymbol{1.405\pm0.083}$ & - & - \\
FT-CIP-Tar        & $0.023\pm0.004$ & $\boldsymbol{0.018\pm0.009}$ & $0.638\pm0.065$ & $1.485\pm0.142$ & - & - \\
FT-OLS-Src-L1     & $0.026\pm0.011$ & $0.028\pm0.009$ & $\boldsymbol{0.619\pm0.057}$ & $1.459\pm0.176$ & $1.531\pm0.214$ & $1.148\pm0.123$ \\
FT-OLS-Src-L2     & $0.026\pm0.014$ & $0.029\pm0.009$ & $0.623\pm0.050$ & $1.434\pm0.163$ & $\boldsymbol{1.526\pm0.211}$ & $1.150\pm0.113$ \\
MASFT             & $0.020\pm0.007$ & $0.022\pm0.008$ & $0.642\pm0.065$ & $1.398\pm0.124$ & $1.505\pm0.098$ & $1.091\pm0.090$ \\
\hline
OLS-Tar           & $0.036\pm0.030$ & $0.032\pm0.010$ & $0.652\pm0.067$ & $1.551\pm0.179$ & $2.062\pm1.033$ & $1.188\pm0.161$ \\
OLS-Tar-MoreData  & $0.020\pm0.008$ & $0.018\pm0.005$ & $0.535\pm0.057$ & $1.216\pm0.033$ & $1.256\pm0.085$ & $0.890\pm0.034$ \\
\hline
\end{tabular}%
}
\caption{
Target test MSEs in the NIR experiments, averaged over $10$ random splits.
Dashes indicate methods that are not applicable in the single-source setting.
Bold indicates the best UDA or SSDA method in each column.
}
\label{tab:nir_results}
\end{table}

\tabref{nir_results} presents target test MSEs for the NIR experiments.
Overall, the fine-tuning methods that use a small number of labeled target samples improve over the UDA methods that do not use labeled target data.
This suggests that UDA methods can identify useful source-target structure, but the best performance is obtained after adapting this structure with limited target labels.
On the corn instrument-shift tasks, the fine-tuned DIP/CIP variants attain the smallest errors among UDA and SSDA methods and perform close to OLS-Tar-MoreData.
On the mango cultivar-shift tasks, FT-OLS-Src-L1 and FT-CIP give the strongest performance.
For the mango ripening-stage shift,
only single-source methods are applicable
and FT-OLS-Src-L2 and FT-DIP perform best, while FT-OLS-Src-L1 
also performs well and substantially improves over its untuned counterpart.
MASFT also gives
strong performance in most NIR tasks, except for \texttt{1,2,3}$\to$\texttt{0} in the mango cultivar-shift, 
where it is somewhat worse than the best method but still improves over the OLS-Tar baseline.
Overall, the NIR experiments show that our SSDA methods can use a small amount of target labels to
effectively adapt calibration models across measurement
conditions.

\section{Discussion} 


In this paper, we develop an SCM-based framework for SSDA with limited labeled target data, 
showing that different low-dimensional shifts require different fine-tuning strategies 
initialized from UDA methods, while MASFT adaptively selects among them when the intervention type is unknown.
Looking ahead, extending our analysis beyond anticausal linear SCMs 
to mixed causal-anticausal or nonlinear SCMs 
is a promising direction. In practical scenarios, 
data often arise from latent representations transformed via nonlinear mixing, 
making it unclear how to apply fine-tuning derived from linear models effectively. 
Integrating ideas from causal representation learning to recover latent structure 
and adapt in representation space remains an open challenge.

\newpage

\begin{appendix}
\startcontents[appendix]

	\vskip 0.1in

			\noindent\textbf{Table of contents.}\par\smallskip
		\begingroup
		\hypersetup{linkcolor=black}
		\contentsmargin{2em}
		\titlecontents{section}
			[1.5em]
			{\small}
			{\contentslabel{1.5em}}
			{}
			{\titlerule*[0.5pc]{.}\contentspage}
		\titlecontents{subsection}
			[3.5em]
			{\small}
			{\contentslabel{2em}}
			{}
			{\titlerule*[0.5pc]{.}\contentspage}
		\printcontents[appendix]{}{1}[2]{}
		\endgroup

			\section{Additional details for numerical experiments}\label{app:experiment_details}

			In this section, we provide additional details for the numerical experiments in~\secref{numerical_exp}.  

			\subsection{Experimental setting and implementation details}\label{app:experiment_protocol}
			This subsection describes the experimental setting and the implementation details for the numerical experiments.

			\paragraph*{Use of target data}
			We use the target data according to the following protocol.
			After preprocessing the data, 
			in each experiment, we split the available labeled target data into three disjoint subsets:
			a training set used for OLS-Tar and fine-tuning,
			a validation set used for hyperparameter tuning and MASFT selection,
			and a test set used for the final evaluation.
			Unlabeled target covariates are used by methods that require marginal target distributional information, such as DIP, FT-DIP and FT-CIP-Tar,
			but their labels are not used for training or tuning.
			Specifically, these covariates are used to estimate target covariance matrices and the corresponding projection subspaces; the labeled target training set is used only after these subspaces or adaptive starts have been constructed.
			The numbers of labeled and unlabeled target samples used are as follows:
			\begin{itemize}
				\item synthetic simulations: $\ntar=100$ and $\ntarval=100$, with $\ntaru=6000$ for the CA and AW shifts and $\ntaru=10000$ for the SC shift;
				\item light-tunnel experiments: $\ntar=40$, $\ntarval=40$, and $\ntaru=4000$;
				\item NIR experiments: $\ntar=24$, $\ntarval=12$, and $\ntaru=56$ for corn; $\ntar=60$, $\ntarval=120$, and $\ntaru\in\{993,800,693,3398\}$ for the four mango columns in~\tabref{nir_results}, corresponding to the larger target training pools used by OLS-Tar-MoreData; and $\ntar=24$, $\ntarval=24$, and $\ntaru=170$ for the additional wheat dataset in~\appref{nir_details}.
			\end{itemize}
			The synthetic simulations are averaged over $20$ independent trials, while the light-tunnel and NIR experiments are averaged over $10$ random data splits.

			\paragraph*{Method implementation}
			Throughout the experiments, we compare target-only, source-only, 
			pooled-source, UDA, and SSDA estimators. 
			OLS-Tar fits ordinary least squares using only the small labeled 
			target training set. 
			OLS-Tar-MoreData is a stronger target-only baseline that fits OLS 
			on a larger labeled target pool; it is included to indicate
			the performance attainable with substantially more target supervision 
			or to provide a comparison point that uses as much labeled target data 
			as available.
			OLS-Src uses one designated source domain, 
			whereas OLS-Pool pools all source domains. 
			DIP and CIP are implemented using the DIP-cov and CIP-mean formulations, 
			respectively, where the subspace dimension for projection is treated as a hyperparameter,
			while groupDRO, IRM, and V-REx are trained by SGD with mini-batches. 
			Each fine-tuned method initializes from its corresponding source or UDA estimator and 
			then uses the small labeled target training set to estimate a target-specific correction.
			In particular, for FT-DIP, FT-CIP, and FT-CIP-Tar, we use 
			penalized variants that constrain the fine-tuning in a low-dimensional subspace 
			while regularizing its deviation from the adaptive start. 
			For FT-OLS-Src-L1 and FT-OLS-Src-L2, we use penalized versions that regularize deviations from the source OLS estimator with $\ell_1$ and $\ell_2$ penalties, respectively.

			\paragraph*{Hyperparameter tuning}
			For FT-DIP, FT-CIP and FT-CIP-Tar, we tune the fine-tuning subspace dimension $k$ and the penalty parameter $\lambda$,
			where $k$ determines how many target-specific directions are fine-tuned, while $\lambda$ controls the strength of the $\ell_2$ regularization toward the corresponding adaptive start.
			For FT-OLS-Src-L1 and FT-OLS-Src-L2, we tune the penalty strength $\lambda$ for the deviation from the source OLS estimator under an $\ell_1$ or $\ell_2$ penalty, respectively.
			In the NIR experiments, we additionally run soft relaxation variants of 
			FT-DIP, FT-CIP, and FT-CIP-Tar, which keep the same subspace dimension $k$ 
			and the same decomposition into the corresponding fine-tuning subspace and its complement 
			(e.g., $(\Vhdip,\Qhdip)$ for FT-DIP, $(\Vhcipaug,\Qhcipaug)$ for FT-CIP, and $(\Vhcip,\Qhcip)$ for FT-CIP-Tar),
			but replace the hard constraint on the $Q$ component by penalties controlled by $\lambda_V$ and $\lambda_Q$.
			Here $\lambda_V$ controls shrinkage within the fine-tuned subspace, while $\lambda_Q$ controls shrinkage within the complementary subspace.
			For groupDRO, we tune the group-weight learning rate, while for IRM and V-REx we tune the penalty strength and the annealing step used in SGD training.
			

		\subsection{Synthetic simulations}\label{app:simulation_details}

		This subsection gives additional details on how the three synthetic experiments in~\secref{numerical_exp} are generated and presents the risk plots that visualize the numerical results given in~\tabref{simulation_results}.  

\paragraph*{Simulation 1: CA shift interventions}

We generate three datasets under CA shift interventions with $d=100$ while varying the dimension of the unobserved confounders $\rdip$ with $\rdip\in\{5,10,15\}$. We consider $M=4$ source domains. For each $m=1,\ldots,4$, the matrix $\Bsrc$ is a lower triangular matrix with diagonal entries zeros and off-diagonal entries drawn \iid from $\Nset(0,9/d)$, and the vector $\bsrc$ has entries drawn \iid from $\Nset(0,0.25/d)$. Each source domain is generated with sample size $\nsrc=\nsrcu=6000$.
For the target domain, we set $\Btar=\Bsrcone$ and $\btar=\bsrcone$, and draw the entries of the matrix $W$ and the vector $w_Y$ \iid from $\Nset(0,25/\rdip)$. We generate $\ntar=100$ labeled target samples for training the SSDA methods and the OLS-Tar baseline method. 
The additional $\ntarval=100$ labeled samples are generated for hyperparameter tuning and for selecting the best fine-tuning method in MASFT. 
We also generate two independent sets of $50000$ labeled samples from the target domain: one to compute the oracle estimator and one to estimate the target test MSE, 
from which the empirical target excess risk is obtained 
by subtracting the test MSE of the oracle estimator. 
For methods that take advantage of unlabeled target data, 
we generate $\ntaru=6000$ unlabeled target samples.

\begin{figure}[t]
		\centering
		\includegraphics[height=0.32\textheight]{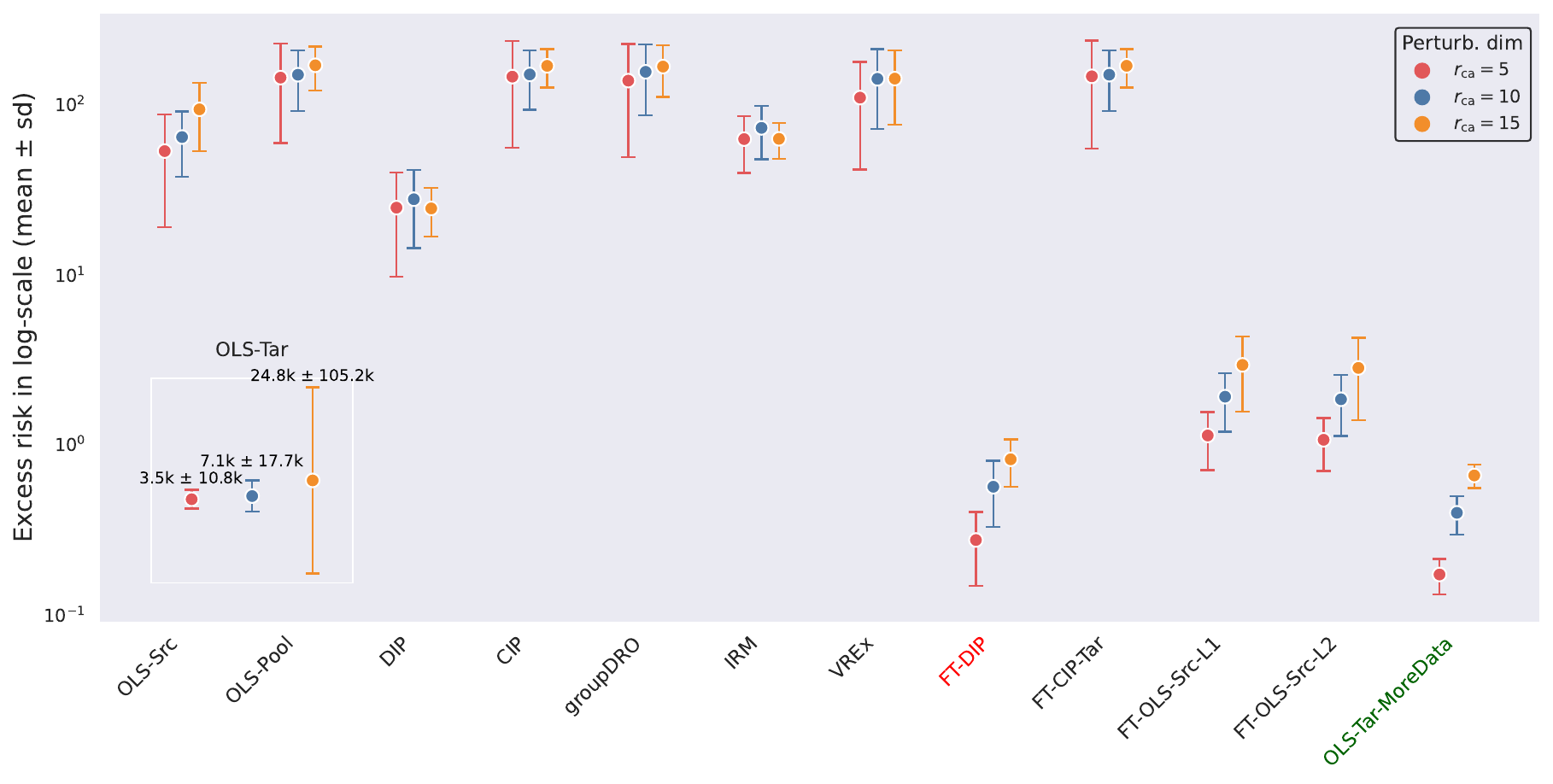}
		\caption{Comparison of empirical target excess risks for Simulation 1 under CA shift interventions. The error bars represent the mean $\pm$ one standard deviation over $20$ trials. 
		Red, blue, and orange points correspond to $\rdip=5,10,15$, respectively. The method highlighted in red indicates the best fine-tuning method, and OLS-Tar-MoreData is highlighted in green. The y-axis is on a logarithmic scale.
		}
		\label{fig:plot_CA}
	\end{figure}

Figure~\ref{fig:plot_CA} visualizes the CA-shift results given in~\tabref{simulation_results}, 
showing the empirical target excess risk averaged over $20$ independent trials with error bars indicating one standard deviation. 
The figure also includes OLS-Tar separately to show the larger error scale of the target-only baseline.

\paragraph*{Simulation 2: SC shift interventions}

Next we generate three datasets under SC shift interventions with $d=100$, varying the number of intervened columns $\rconn$ in the connectivity matrix with $\rconn\in \{5,10,15\}$. We consider $M=4$ source domains. For each $m=1,\ldots,4$, the matrix $\Bsrc$ is a lower triangular matrix with diagonal entries zeros and off-diagonal entries drawn \iid from $\Nset(0,0.25/d)$, while the vector $\bsrc$ has entries drawn \iid from $\Nset(0,4/d)$. For the target domain, we set $\btar=\bsrcone$, and construct the connectivity matrix $\Btar$ as  follows: for $j=1,\ldots,\rconn$ and $i>j$, the $(i,j)$-th entry $\Btar_{[i,j]}$ of $\Btar$ is set to $\Bsrcone_{[i,j]}+\btar_{[i]} + \eta_{ij}$, where $\eta_{ij}\iidsim \Nset(0,1/d)$; all other entries remain unchanged from $\Bsrcone$. The sample sizes for the labeled target datasets are the same as Simulation 1 while for the source and unlabeled target datasets, we use $\nsrc=\nsrcu=\ntaru=10000$.

\begin{figure}[t]
		\centering
		\includegraphics[height=0.32\textheight]{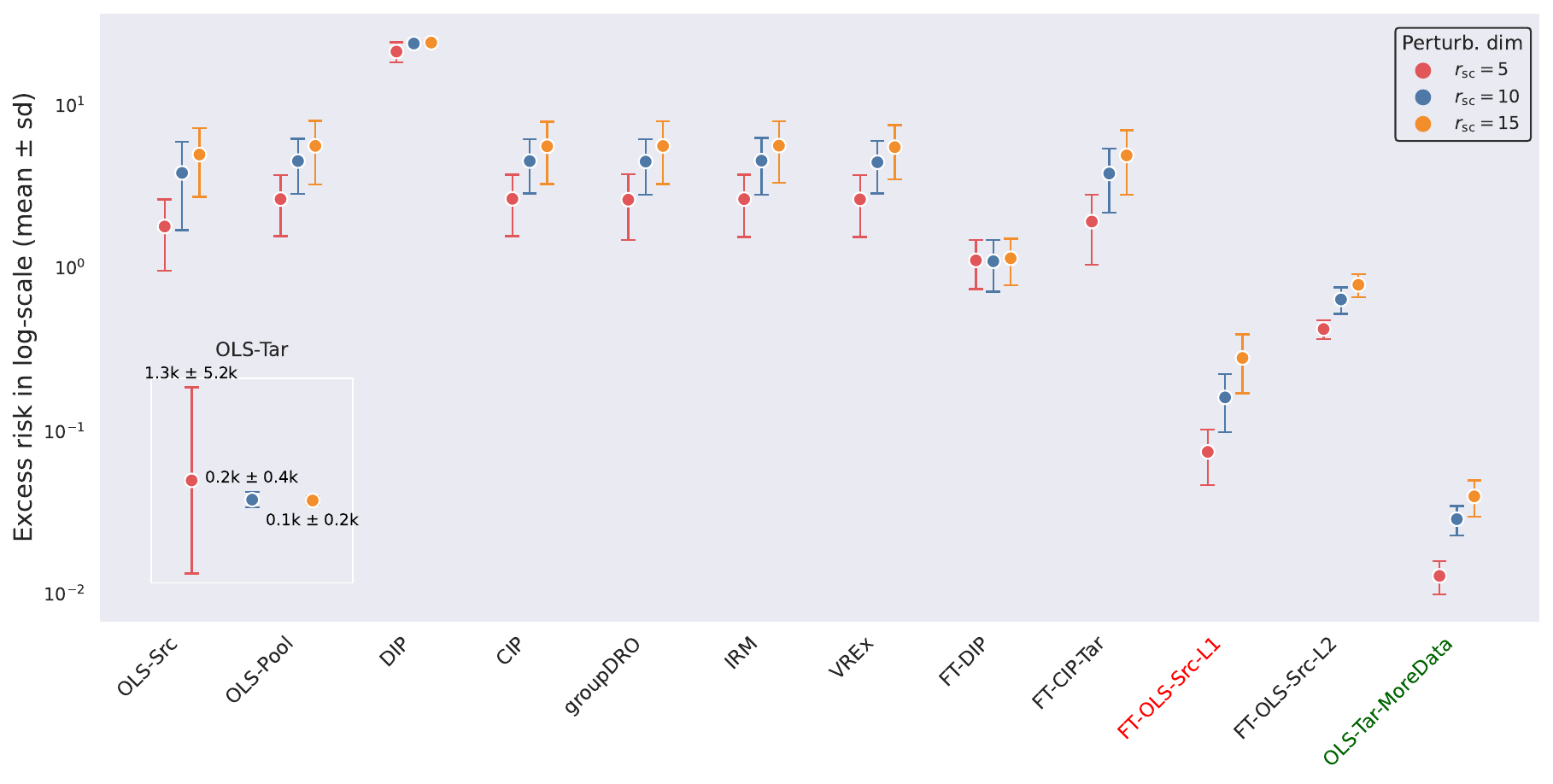}
			\caption{Comparison of empirical target excess risks for Simulation 2 under SC shift interventions. The error bars represent the mean $\pm$ one standard deviation over $20$ trials. Red, blue, and orange points correspond to $\rconn=5,10,15$, respectively. The method highlighted in red indicates the best fine-tuning method, and OLS-Tar-MoreData is highlighted in green. The y-axis is on a logarithmic scale.
		}
		\label{fig:plot_SC}
	\end{figure}

Figure~\ref{fig:plot_SC} visualizes the SC-shift results given in~\tabref{simulation_results}, showing the empirical target excess risk averaged over $20$ independent trials.
The figure also includes OLS-Tar separately to show the larger error scale of the target-only baseline.

\paragraph*{Simulation 3: AW shift interventions}

Finally, we generate three datasets under AW shift interventions with $d=100$ while varying the dimension of non-invariant features $\rcip$ with $\rcip\in \{4,6,8\}$. We consider $M=11$ source domains. The matrix $B$ is a lower triangular matrix with diagonal zeros and off-diagonal entries drawn \iid from $\Nset(0,9/d)$. Once generated, $B$ is fixed across all source domains, i.e., $\Bsrc=B$ for all $m\geq 1$. To construct $\bsrc$, we first generate $U\in\R^{70\times \rcip}$ with entries drawn \iid from $\Nset(0,1/d)$, and set $\bnoninv^{(m)}=U\zeta^{(m)}$, where each entry of $\zeta^{(m)}\in\R^{\rcip}$ is drawn \iid from $\abs{\Nset(0,16/\rcip)}$. We then generate $\binv\sim \Nset(0,\ident_{30}/d)$ and define $\bsrc=\begin{bmatrix}\bnoninv^{(m)}{}^\top; & \binv^\top \end{bmatrix}^\top\in\R^d$. For the target domain, we set $\Btar=B$, and define $\btar=\begin{bmatrix}\bnoninv^{\tagtar}{}^\top; & \binv^\top \end{bmatrix}^\top\in\R^d$ where $\bnoninv^{\tagtar}=U\zeta^{\tagtar}$, and each entry of $\zeta^{\tagtar}\in\R^{\rcip}$ is drawn \iid from $\abs{\Nset(0,0.25/\rcip)}$. The sample sizes for the source and target datasets are the same as Simulation 1.

\begin{figure}[t]
		\centering
		\includegraphics[height=0.32\textheight]{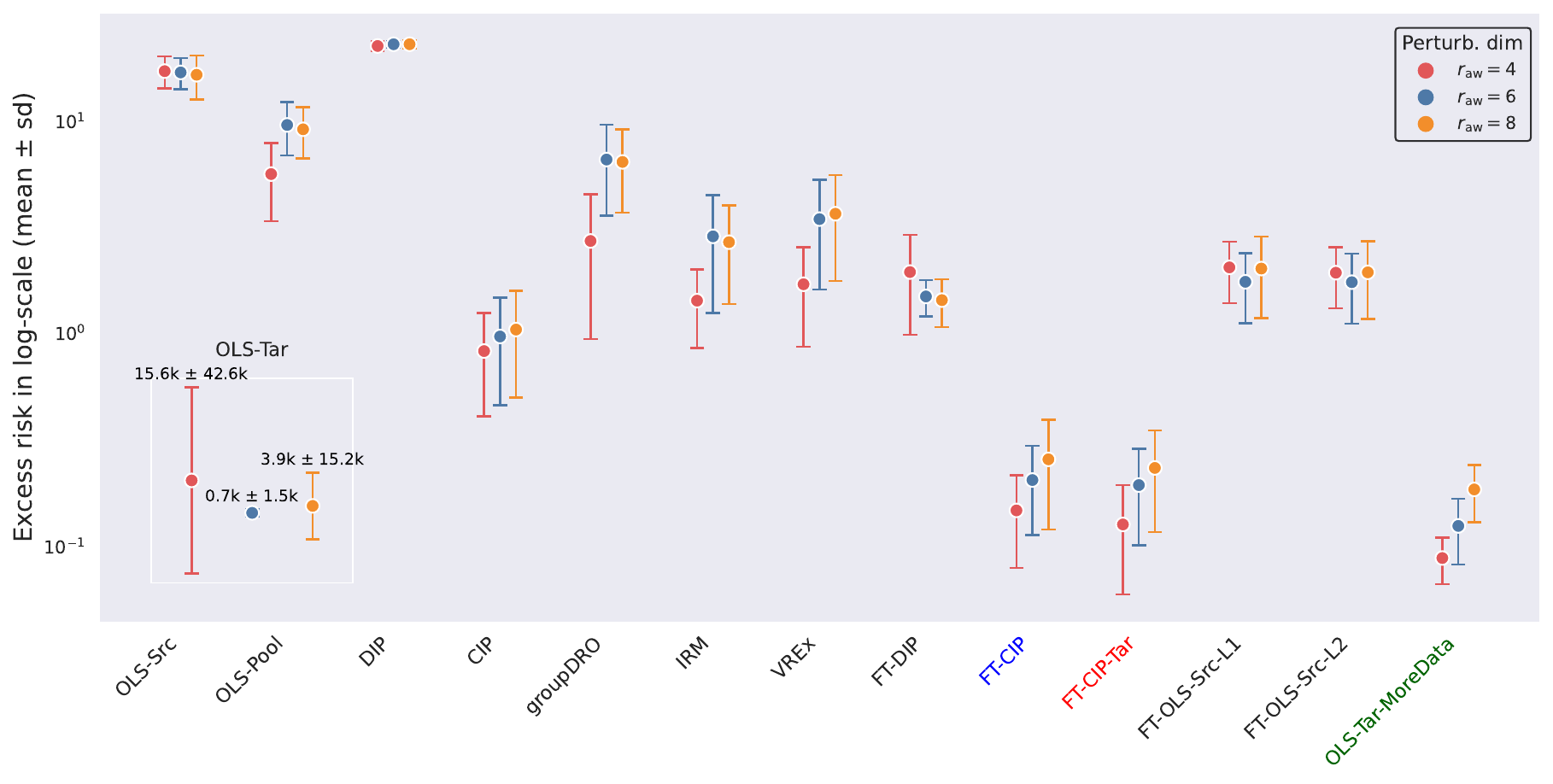}
			\caption{Comparison of empirical target excess risks for Simulation 3 under AW shift interventions. The error bars represent the mean $\pm$ one standard deviation over $20$ trials. Red, blue, and orange points correspond to $\rcip=4,6,8$, respectively. The methods highlighted in red and blue indicate FT-CIP-Tar and FT-CIP, respectively, and OLS-Tar-MoreData is highlighted in green. The y-axis is on a logarithmic scale.
		}
		\label{fig:plot_AW}
	\end{figure}

Figure~\ref{fig:plot_AW} visualizes the AW-shift results summarized in~\tabref{simulation_results}, showing the empirical target excess risk averaged over $20$ independent trials.
The figure also includes OLS-Tar separately to show the larger error scale of the target-only baseline.

			\subsection{Light-tunnel experiments}\label{app:light_tunnel_details}\label{app:application_I}


	Here we provide additional details for the light-tunnel experiments.
	We collect data from the Light Tunnel Mk2 in the Causal Chambers benchmark~\cite{gamella2025causal}.
	The \texttt{linked\_leds} light-tunnel configuration, a variation of the standard configuration, contains light-source controls (\texttt{R}, \texttt{G}, \texttt{B}), polarizer settings and angle measurements (e.g., \texttt{$\theta_1$}, \texttt{$\theta_2$}), sensor measurements (\texttt{$\widetilde I_1$}, \texttt{$\widetilde I_2$}, \texttt{$\widetilde I_3$}, \texttt{$\widetilde V_1$}, \texttt{$\widetilde V_2$}, \texttt{$\widetilde V_3$}), and sensor parameters such as exposure time, photodiode choice and ADC settings.
	In our experiments, we use the light-source control \texttt{R} as response, 
	and the remaining light-source controls \texttt{G} and \texttt{B}, the polarizer settings \texttt{$\theta_1$} and \texttt{$\theta_2$}, and the six sensor readings \texttt{$\widetilde I_1$}, \texttt{$\widetilde I_2$}, \texttt{$\widetilde I_3$}, \texttt{$\widetilde V_1$}, \texttt{$\widetilde V_2$}, \texttt{$\widetilde V_3$} as covariates.
	The configuration provides a ground-truth causal graph, for which we show a simplified graph in~\figref{light_tunnel_graph},
	where an edge $X_1 \to X_2$ in this graph 
	can be interpreted interventionally:
	an intervention on $X_1$ changes the distribution of subsequent measurements of $X_2$.
	This information makes it possible to generate a wide range of interventional datasets for causal discovery,
	causal inference, and domain adaptation~\cite{gamella2025causal}.

	\paragraph*{Data and preprocessing} The three datasets are collected by applying weak and strong interventions to the light-tunnel configuration as follows. 
	The parameters that are not mentioned are set to their default values in the \texttt{linked\_leds} configuration.
	\begin{itemize}
		\item In \texttt{lt\_CA}, for the target domain, we apply a weak intervention to \texttt{$\widetilde V_1$} by changing the brightness of the external light source \texttt{led\_1\_ir} above the sensor for \texttt{$\widetilde V_1$}. For the first source domain, there is no intervention applied. For the other five source domains, a weak intervention is applied to \texttt{$\widetilde I_1$} and \texttt{$\widetilde I_2$}; a strong intervention is applied to \texttt{$\widetilde I_1$}, \texttt{$\widetilde I_2$}, and \texttt{$\widetilde V_1$}.
		\item In \texttt{lt\_SC}, for the target domain, we apply an intervention to make the input of \texttt{$\widetilde I_2$} going into \texttt{$\widetilde V_3$} change from an increasing to a decreasing function. For the first source domain, an intervention is also applied to the connection between \texttt{$\widetilde I_1$} and \texttt{$\widetilde V_2$}. Two random interventions are applied to the other two source domains.
		\item In \texttt{lt\_AW}, for the target domain, we apply a strong intervention to \texttt{$\widetilde I_2$} by changing the brightness of the external light source \texttt{led\_2\_ir} above the sensor for \texttt{$\widetilde I_2$}. For the first source domain, there is no intervention applied.  For the other five source domains, a weak intervention is applied to \texttt{$\widetilde I_1$}, \texttt{$\widetilde I_2$} and \texttt{$\widetilde V_1$}; a strong intervention is applied to \texttt{$\widetilde I_1$} and \texttt{$\widetilde V_2$}.
	\end{itemize}

	For each environment in these configurations, we use $5000$ observations.
	For preprocessing, we apply centering and scaling to the variables,
	and after training, the target prediction performance is measured by the target MSE.
	The use of target data and the implementation of the methods follow the general protocol given in~\appref{experiment_protocol}.

	\paragraph*{Empirical intervention diagnostics}
	
\begin{figure}[t!]
		\centering
		\includegraphics[width=\textwidth]{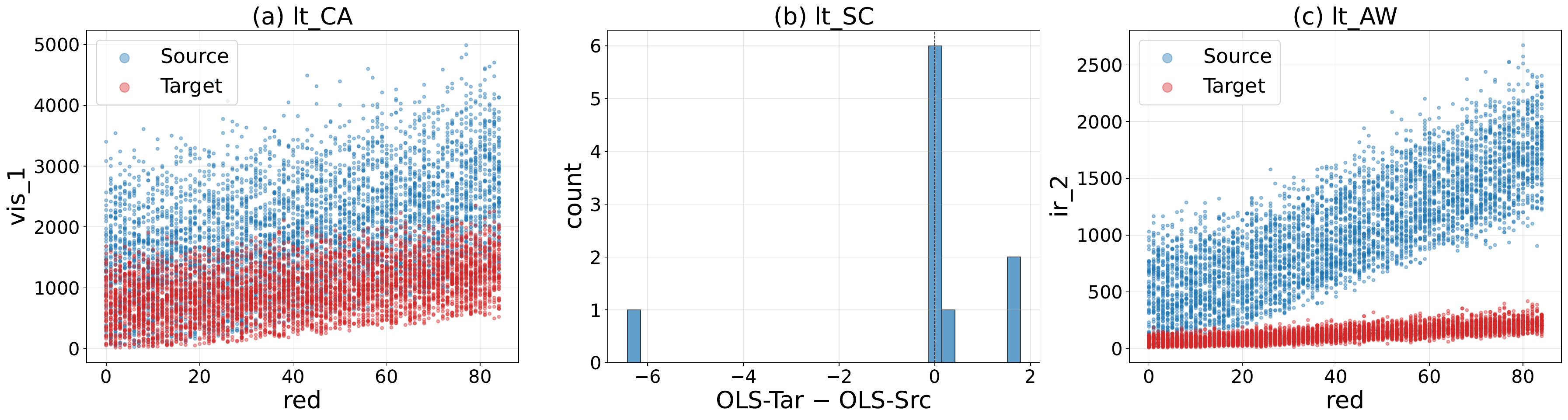}
		\caption{Empirical diagnostics for the light-tunnel experiments. In our notation, \texttt{red} = \texttt{R}, \texttt{vis\_1} = \texttt{$\widetilde V_1$}, \texttt{ir\_2} = \texttt{$\widetilde I_2$}. The three panels illustrate the interventions in \texttt{lt\_CA}, \texttt{lt\_SC}, and \texttt{lt\_AW}, respectively.}
		\label{fig:light_tunnel_diagnostics}
		\end{figure}

	Figure~\ref{fig:light_tunnel_diagnostics} provides visual diagnostics 
	for the three light-tunnel interventional settings. 
	\begin{itemize}
		\item Panel~(a) shows a scatter plot of \texttt{R} against \texttt{$\widetilde V_1$}, which is intervened in the target domain of \texttt{lt\_CA}.
		The plot shows an additive shift in \texttt{$\widetilde V_1$}, 
		while the linear relationship between \texttt{R} and \texttt{$\widetilde V_1$} remains visually similar. 
		This suggests that the weak intervention on \texttt{$\widetilde V_1$} induces an approximately 
		additive shift which is consistent with the CA shift setting. 
		\item 	Panel~(b) shows a histogram of the coefficient differences between OLS-Tar-MoreData and 
		OLS-Src. Most coefficients remain close to zero, with only a few visible changes in OLS coefficients,
		which is favorable for sparsity-inducing regularization in the SC shift setting. 
		\item Panel~(c) shows a scatterplot of \texttt{R} against \texttt{$\widetilde I_2$}. In contrast to Panel~(a), the mechanism relating \texttt{R} to \texttt{$\widetilde I_2$} changes more substantially across domains. This indicates that the strong intervention on \texttt{$\widetilde I_2$} affects the anticausal mechanism itself, which is consistent with an AW-type shift. 
	\end{itemize}
	

			\subsection{NIR experiments}\label{app:nir_details}

			Near-infrared (NIR) spectroscopy provides a fast and non-destructive
			way to measure agricultural and food samples
			~\citep{zhao2019pls,zhao2019calibration,anderson2020achieving,sun2020nirs}.
			In a typical NIR measurement, a sample is illuminated by near-infrared
			light, and the instrument records a wavelength-indexed spectrum from
			the light interacting with the sample through absorption, reflection,
			transmission, and scattering. The resulting spectrum contains indirect
			information about the chemical composition and physical structure of
			the sample. A calibration model then maps this high-dimensional
			spectrum to a reference measurement, such as moisture, dry
			matter, or protein content. 
			Thus, NIR calibration is an anticausal
			prediction problem, where the underlying sample property affects the observed
			spectrum through the measurement process, while the statistical task is
			to predict that property from the measured spectrum.

			Because spectra are sensitive to instrument response, temperature,
			season, and sample condition, a calibration model trained in
			one measurement domain often loses accuracy when deployed in another.
			Maintaining calibration performance across such domain changes is therefore a
			central problem in chemometrics, commonly studied as calibration transfer
			and model maintenance~\citep{roger2025cloning,zhao2019pls,anderson2020achieving,sun2020nirs}.
			This setting naturally connects NIR calibration transfer to domain
			adaptation, and recent chemometric work has used DA algorithms to align source and target spectral distributions and improve
			target domain prediction~\citep{roger2025cloning,nikzad2018domain,mishra2021brief}.

			\paragraph*{Data and preprocessing}
			In our NIR experiments, we use three datasets: corn, wheat, and mango.
			The corn and wheat datasets are standard NIR benchmarks used in
			calibration-transfer studies~\citep{zhao2019pls,zhao2019calibration}.
			The corn data contain $80$ corn samples measured on three instruments,
			\texttt{m5}, \texttt{mp5}, and \texttt{mp6}; the response is moisture
			content, and each raw spectrum has $700$ wavelength channels from
			$1100$ to $2498$ nm. For corn, we use \texttt{m5} as the
			source instrument and consider two transfer tasks, \texttt{m5}$\to$\texttt{mp6}
			and \texttt{m5}$\to$\texttt{mp5}. The wheat data contain $248$ wheat
			samples measured on three manufacturer-B instruments, denoted
			\texttt{B1}, \texttt{B2}, and \texttt{B3}; the response is protein content,
			and each raw spectrum has $741$ wavelength channels from $730$ to
			$1100$ nm. For wheat, we use \texttt{B1} as the source
			instrument and consider two transfer tasks, \texttt{B1}$\to$\texttt{B2}
			and \texttt{B1}$\to$\texttt{B3}. In both datasets, the third instrument
			that is not used as the target in a given task is included as an additional
			source domain.

			The mango data~\citep{anderson2020achieving} contain $11691$
			intact-fruit spectra collected with an F750 Produce Quality Meter across
			four seasons, ten cultivars, and two physiological stages, Hard Green
			and Ripen. The response is dry-matter content, and each raw spectrum
			has $306$ wavelength channels covering approximately $309$-$1149$ nm.
			We use this dataset in two ways. For cultivar shift, we restrict the data
			to Hard Green fruits and define domains by cultivar. The four largest
			cultivar domains are indexed as \texttt{0}=\texttt{Caly},
			\texttt{1}=\texttt{KP}, \texttt{2}=\texttt{HG}, and
			\texttt{3}=\texttt{R2E2}. In~\tabref{nir_results}, we consider two transfer tasks, i.e.,
			\texttt{1,2,3}$\to$\texttt{0}, where the target is \texttt{Caly}, and
			\texttt{0,2,3}$\to$\texttt{1}, where the target is \texttt{KP}. For
			ripening-stage shift, we use all fruits and define domains by
			physiological stage, where \texttt{0} denotes Hard Green and \texttt{1} denotes Ripen.
			This yields two single-source tasks, \texttt{Type\_0\_1} (Hard Green
			$\to$ Ripen) and \texttt{Type\_1\_0} (Ripen $\to$ Hard Green). Since the ripening-stage tasks
			have only one source domain, multi-source methods such as OLS-Pool,
			CIP, FT-CIP, groupDRO, IRM, and V-REx are not applicable in this setting.

			For preprocessing, for corn and wheat, we first apply the centering, and then apply partial least squares (PLS) to extract features, 
			following the preprocessing steps used in~\citep{zhao2019pls,zhao2019calibration}. 
			PLS is commonly used
			in NIR calibration because raw spectra are high-dimensional and strongly
			collinear, and it constructs a low-dimensional representation that is
			predictive of the response. In our experiments, the PLS projection is fitted
			on the source domain and then applied to all source and target domains,
			yielding $13$ PLS components for corn and $14$ PLS components for
			wheat. For mango, we follow the preprocessing used in the original mango
			study~\citep{anderson2020achieving}. Specifically, we first apply a
			Savitzky-Golay second-derivative filter with window length $17$ and
			polynomial order $2$, and then restrict the spectra to the wavelength
			range $684$-$990$ nm. After this spectral preprocessing, we 
			apply centering and then PLS for
			feature extraction with $12$ components.

			\begin{table}[t]
			\centering
			\footnotesize
			\setlength{\tabcolsep}{3pt}
			\renewcommand{\arraystretch}{1.1}
			\begin{tabular}{|l|cc|}
			\hline
			& \multicolumn{2}{c|}{Wheat} \\
			\hline
			Method & $\texttt{B1}\to\texttt{B2}$ & $\texttt{B1}\to\texttt{B3}$ \\
			\hline
			OLS-Src           & $0.186\pm0.043$ & $0.772\pm0.139$ \\
			OLS-Pool          & $\boldsymbol{0.141\pm0.037}$ & $0.308\pm0.126$ \\
			DIP               & $0.222\pm0.112$ & $0.211\pm0.111$ \\
			CIP               & $0.259\pm0.058$ & $0.309\pm0.125$ \\
			groupDRO          & $1.434\pm0.318$ & $1.557\pm0.352$ \\
			IRM               & $1.472\pm0.336$ & $1.581\pm0.357$ \\
			VREx              & $1.458\pm0.324$ & $1.579\pm0.356$ \\
			\hline
			FT-DIP            & $0.179\pm0.052$ & $0.207\pm0.111$ \\
			FT-CIP            & $0.151\pm0.043$ & $\boldsymbol{0.188\pm0.133}$ \\
			FT-CIP-Tar        & $0.260\pm0.058$ & $0.308\pm0.125$ \\
			FT-OLS-Src-L1     & $0.163\pm0.042$ & $0.233\pm0.122$ \\
			FT-OLS-Src-L2     & $0.158\pm0.043$ & $0.236\pm0.122$ \\
			MASFT             & $0.161\pm0.041$ & $0.192\pm0.127$ \\
			\hline
			OLS-Tar           & $0.762\pm0.918$ & $0.449\pm0.376$ \\
			OLS-Tar-MoreData  & $0.116\pm0.041$ & $0.157\pm0.152$ \\
			\hline
			\end{tabular}
			\caption{
			Target test MSEs on the additional wheat NIR dataset, averaged over $10$ random splits.
			Bold indicates the best UDA or SSDA method in each column.}
			\label{tab:wheat_results}
			\end{table}

			\paragraph*{Target sample sizes}
			For corn, each target instrument contains $80$ spectra. We split them
			into $56$ spectra for training, $12$ spectra for validation, and $12$ spectra for testing.
			OLS-Tar-MoreData is trained using all $56$ labeled target training
			samples, whereas OLS-Tar and the fine-tuning methods use only $24$ labeled target training samples. 
			For mango, OLS-Tar-MoreData is trained on a larger target training set.
			For the four mango columns in~\tabref{nir_results}, the corresponding
			target training sizes are $993$, $800$, $693$ and $3398$, respectively;
			these are the larger target pools for the two cultivar-shift and two
			ripening-shift tasks.
			In contrast, OLS-Tar and the fine-tuning methods use only $60$ labeled
			target training samples and $120$ labeled target validation samples in
			each task.
			For wheat, OLS-Tar-MoreData is trained
			on $170$ labeled target samples, whereas OLS-Tar and the fine-tuning
			methods use only $24$ labeled target training samples and $24$ labeled
			target validation samples.

			\paragraph*{Additional results for wheat dataset}

			\tabref{wheat_results} presents additional results on the wheat dataset.
			For the \texttt{B1}$\to$\texttt{B2} task, OLS-Pool achieves the smallest
			error among the UDA and SSDA methods, while FT-CIP gives the next best
			performance. This shows that, for this task, simply pooling source datasets is
			already effective, so additional invariance penalties and fine-tuning
			provide limited further benefit. For the \texttt{B1}$\to$\texttt{B3} task,
			FT-CIP achieves the best performance among the UDA and SSDA methods,
			substantially improving over OLS-Pool and the other baselines.

	\section{Preliminaries and technical lemmas}
	
	In this section, we present technical lemmas that will be used in our proofs. Some of these lemmas follow directly from adaptations of existing results while others will be proved in~\appref{proof_of_lemmas}.
		
	\begin{table}\centering
		\begin{tabular}{|c|c|}
			\hline
			Notation&Definition\\\hline
			$\ident$ & identity matrix (with subscript, e.g., $\ident_k$, to indicate dimension $k$ when needed) \\\hline
			$X_{[j]}$ & $j$-th coordinate of a random vector $X$  \\\hline
			$\vecnorm{x}{1}$ & $\ell_1$ norm of a vector $x$ \\\hline
			$\vecnorm{x}{2}$ & $\ell_2$ norm of a vector $x$ \\\hline
			$\spn(x_1,\ldots,x_k)$ & linear subspace spanned by $x_1,\ldots,x_k$ \\\hline
			$\col{A}$ & column space of a matrix $A$ \\\hline
			$A_{*S}$ & submatrix of $A$ consisting of the columns indexed by $S$ \\\hline
			$\lammin(A)$& minimum eigenvalue of symmetric matrix $A$\\\hline
			$\lammax(A)$& maximum eigenvalue of symmetric matrix $A$\\\hline
			$\kappa(A)$& $\lammax(A)/\lammin(A)$ (the condition number of $A$)  \\\hline
			$\vecopnorm{A}$& spectral norm/operator norm ($=$maximum singular value of $A$) \\\hline
			$\vecnorm{A}{\text{F}}$& Frobenius norm of $A$ \\\hline
			$\sigma_{\text{min}}(A)$& minimum singular value of $A$\\\hline
			$\sigma_X$& sub-Gaussian parameter of $\Xsrc$ for all $m$ (see Equation~\eqnref{def_sigma_X})\\\hline
			$\sigma_Y$& sub-Gaussian parameter of $\Ysrc$ for all $m$ ($=\max_{m\geq 0}\sigma_{\eps_Y}^{(m)}$)\\\hline

			$\bXsrc$& design matrix of the $m$-th domain ($\nsrc\times d$) \\
			&(the $i$-th row is given by $\Xsrc_i$)\\\hline
			$\bYsrc$& response vector of the $m$-th domain ($\nsrc\times 1$) \\
			&(the $i$-th component is given by $\Ysrc_i$)\\\hline      		
			$\Hsrc$ or $H$& $(\ident-\Bsrc)^{-1}$ or $(\ident-B)^{-1}$\\\hline
			$\lamminbar$& $\min_{m=0,1,\ldots,M}\lammin(\SigmaXsrc)$\\\hline
			$\lammaxbar$& $\max_{m=0,1,\ldots,M}\lammax(\SigmaXsrc)$\\\hline
			$\kappabar$& $\max_{m=0,1,\ldots,M}\kappa(\SigmaXsrc)$\\\hline
			$\nbar$& $\min_{m=1,\ldots,M}\nsrc$\\\hline
			$\tsigmamin$& condition number $\lamminbar^{-1}\sigma_X^2$ that involves $\sigma_X$ and $\lamminbar$\\\hline
			$\Theta(U,V)$ & vector $(\theta_1,\ldots,\theta_p)$ of principal angles between subspaces $U$ and $V$;\\
			& $\theta_k=\arccos\sigma_k(U^{\top}V)$ where $\sigma_k(\cdot)$ is the $k$-th singular value\\\hline
			$\sin\Theta(U,V)$ & $\diag(\sin\theta_1,\ldots,\sin\theta_p)$\\\hline
			$\kl{\Pset_1}{\Pset_2}$ & Kullback--Leibler (KL) divergence between probability measures $\Pset_1$ and $\Pset_2$ \\\hline
			$\Pset_1 \otimes \Pset_2$ & product measure of probability distributions $\Pset_1$ and $\Pset_2$ \\\hline
			\makecell{$c, c', c'', \dots$ \\ or $c_1, c_2, \dots$} & universal constants (whose definitions may change from one result to another)\\\hline
		\end{tabular}
		\caption{Notation used throughout the proofs.}
		\label{tab:notation}
	\end{table}
	\subsection{Sub-Gaussian and sub-exponential norms}\label{app:sub_gaussian_sub_exp}
	We begin with describing the properties of sub-Gaussian and sub-exponential norms. Recall that for a zero-mean random variable $Z$, the sub-Gaussian norm~\cite[Definition 2.5.6]{vershynin2018high} is defined as
	\[\vecnorm{Z}{\psi_2} \defn \inf\{t>0:\EE{\exp(Z^2/t^2)}\leq 2 \},\]
	and the sub-exponential norm~\cite[Definition 2.7.5]{vershynin2018high} is defined as
	\[\vecnorm{Z}{\psi_1} \defn \inf\{t>0:\EE{\exp(\abs{Z}/t)}\leq 2 \}. \]
	The sub-Gaussian and sub-exponential parameters of $Z$ are given by $\vecnorm{Z}{\psi_2}$ and $\vecnorm{Z}{\psi_1}$, respectively, up to universal constants~\cite[Proposition 2.5.2, Proposition 2.7.1]{vershynin2018high}.
	For a zero-mean random vector $W$, the sub-Gaussian and sub-exponential norm are defined as the maximum of the sub-Gaussian norms of one-dimensional projections:
	\[\vecnorm{W}{\psi_2} \defn \sup_{u:\norm{u}_2=1} \vecnorm{u^\top W}{\psi_2}, \text{ and }  \vecnorm{W}{\psi_1} \defn \sup_{u:\norm{u}_2=1} \vecnorm{u^\top W}{\psi_1}.\]
	
	We present two lemmas describing the properties of sub-Gaussian and sub-exponential norms. The first lemma shows how the sub-Gaussian norm behaves under linear transformations of sub-Gaussian random vectors. This result follows directly from~\cite[Exercise 2.7.11]{vershynin2018high}, and the proof is omitted.
	\begin{lemma}[Sub-Gaussianity under linear transformations]\label{lem:sub_gaussian}
		Let $W \in \mathbb{R}^d$ and $Z \in \mathbb{R}^k$ be zero-mean sub-Gaussian random vectors with parameters $\sigma_W$ and $\sigma_Z$, respectively. Then, for any matrix $A \in \mathbb{R}^{k \times d}$, 
		there exists a universal constant $c>0$ such that
		\[ \vecnorm{AW+Z}{\psi_2} \leq c (\sigma_W\opnorm{A} + \sigma_Z).\]
	\end{lemma}
	
	The second lemma provides a bound on the sub-exponential norm of the product of a sub-Gaussian random vector and a sub-Gaussian random variable. This is a consequence of the property of the sub-exponential norm~\cite[Lemma 2.7.6]{vershynin2018high}, and the proof is omitted.
	\begin{lemma}[Sub-exponential norm of the product of sub-Gaussian variables]\label{lem:sub_exp}
		Let $W \in \mathbb{R}^d$ be a zero-mean sub-Gaussian random vector with parameter $\sigma_W$, and let $Z \in \mathbb{R}$ be a zero-mean sub-Gaussian random variable with parameter $\sigma_Z$. 
		Then there exists a universal constant $c>0$ such that
		\[\vecnorm{WZ}{\psi_1} \leq c\sigma_W\sigma_Z. \]
	\end{lemma}
		Consider the source and target covariates $\Xsrc$, $m=0,1,\ldots,M$, generated from linear SCMs under~\assumpref{linear_SCMs}. Solving $\Xsrc$ from linear structural equations~\eqnref{src_linear_SCM} yields the expression
		\begin{align}\label{eqn:Xsrc_equation}\Xsrc = \Hsrc \bsrc\epssrc_Y +\Hsrc\epssrc_X, \end{align}
		where $\Hsrc\coloneqq(\ident-\Bsrc)^{-1}$ (see~\tabref{notation} for notation). Applying~\lemref{sub_gaussian} and the inequality $\norm{Ax}_2\leq\opnorm{A}\norm{x}_2$, the following result immediately follows; the proof is omitted.
		\begin{lemma}[Sub-Gaussian parameter of the covariates]\label{lem:bound_sigma_X}
			Suppose~\assumpsref{linear_SCMs} and~\assumpssref{sub_gaussianity} hold. Then the sub-Gaussian norm of $\Xsrc$ is bounded as
			\[\norm{\Xsrc}_{\psi_2} \leq c\opnorm{\Hsrc}\left(\norm{\bsrc}_2\cdot\sigma^{(m)}_{\eps_Y}+\sigma^{(m)}_{\eps_X}\right)\text{ for all $m=0,1,\ldots,M$},\]
			for some universal constant $c>0$. Therefore, if we define 
		\begin{equation}\label{eqn:def_sigma_X}
			\sigma_X\coloneqq \max_{m=0,1,\ldots,M} \left\{c\opnorm{\Hsrc}\left(\norm{\bsrc}_2\cdot\sigma^{(m)}_{\eps_Y}+\sigma^{(m)}_{\eps_X}\right)\right\},
		\end{equation}
		then for any $m=0,1,\ldots,M$, $\Xsrc$ is a sub-Gaussian vector with parameter $\sigma_X$.
		\end{lemma}

		\subsection{Concentration inequalities}
		
		To facilitate the use of concentration inequalities for functions of sub-Gaussian and sub-exponential random vectors, we recall them here.  Proofs are deferred to~\appref{proof_of_lemmas}.
		
		The first lemma states a standard concentration of a sample covariance matrix, and its proof is omitted.
		\begin{lemma}[Adaptation of Theorem 6.5 of~\cite{wainwright2019high}]\label{lem:bounds_cov_matrices}
			Let $W_i \in \mathbb{R}^d$, $i=1,\ldots,n$, be $\iid$ zero-mean sub-Gaussian random vectors with parameter $\sigma_W$. Let $\SigmaW=\EE{W_1W_1^\top}$ denote the population covariance matrix, and $\SigmaWh=\frac{1}{n}\sum_{i=1}^{n}W_iW_i^\top$ the sample covariance matrix. There exists a universal $c>0$ such that for $\delta \in (0,1)$, if $n\geq \max\braces{d, \log(1/\delta)}$, with probability at least $1- \delta$,
			\begin{align*}
				\vecopnorm{\SigmaW - \SigmaWh }\leq c\sigma_W^2 \parenth{\sqrt{\frac{d}{n}} + \sqrt{\frac{\log(1/\delta)}{n}}}.
			\end{align*}
		\end{lemma}
		A direct consequence of~\lemref{bounds_cov_matrices} together with Weyl's inequality is upper and lower bounds on the minimum and maximum eigenvalues of a sample covariance matrix. For $\delta \in (0, 1)$, there exists a universal constant $c>0$ such that if $n \geq c \sigma_W^4\lammin^{-2}(\SigmaW) \cdot \max\braces{d, \log(1/\delta)}$, then with probability at least $1-\delta$,
    \begin{align}\label{eqn:bounds_eig_values}
      \lammin(\SigmaW)/2 \leq \lammin(\SigmaWh) \leq \lammax(\SigmaWh) \leq 2\lammax(\SigmaW).
    \end{align}
	 	The next lemma considers the setting where the leading eigenvalues (in absolute value) of the difference between two covariance matrices are well separated from the remaining eigenvalues. In particular, it shows that with a sufficiently large sample size, the eigenvectors corresponding to these leading eigenvalues can be accurately estimated by those of the corresponding finite-sample quantities.
				
		\begin{lemma}\label{lem:error_orthogonal_space}
			Under the settings of~\lemref{bounds_cov_matrices}, let $\Wt_i\in\mathbb{R}^d$, $i=1,\ldots,\nt$, be another set of $\iid$ zero-mean sub-Gaussian random vectors with parameter $\sigma_W$, and define $\SigmaWt=\EE{\Wt_1\Wt_1^\top}$ and $\SigmaWht=\frac{1}{\nt}\sum_{i=1}^{\nt}\Wt_i\Wt_i^\top$. Suppose there exists $\lambda_{\text{gap}} > 0$ such that the largest $r$ eigenvalues in absolute value of $\SigmaW - \SigmaWt$ are greater than $\lambda_{\text{gap}}$, and the remaining eigenvalues have absolute values less than $\frac{1}{2}\lambda_{\text{gap}}$. Let $V \in \mathbb{R}^{d \times r}$ be the matrix of eigenvectors corresponding to the top $r$ eigenvalues in absolute value, and let $Q \in \mathbb{R}^{d \times (d - r)}$ be an orthogonal complement to the column space of $V$. Similarly, let $\widehat{V} \in \mathbb{R}^{d \times r}$ and $\widehat{Q} \in \mathbb{R}^{d \times (d - r)}$ be the corresponding matrices from the finite-sample estimate $\widehat{\Sigma}_W - \widehat{\Sigma}_{\widetilde{W}}$. There exists universal constants $c_0, c_1 > 0$ such that for $\delta\in (0,1/2)$, if $n,\nt \geq  c_0 (\sigma_W^4 \lamgap^{-2}\vee 1) \max\braces{d, \log(1/\delta)}$, with probability at least $1- \delta$, 
			\begin{align*}
				&\inf_{R:R^\top R=\ident}\vecopnorm{\widehat{V} - VR} \leq \frac{c_1 \sigma_W^2}{\lamgap}\parenth{ \sqrt{\frac{d}{n}} + \sqrt{\frac{\log(1/\delta)}{n}} + \sqrt{\frac{d}{\nt}} + \sqrt{\frac{\log(1/\delta)}{\nt}} }, \text{ and } \\
				&\inf_{R:R^\top R=\ident}\vecopnorm{\widehat{Q} - QR} \leq \frac{c_1 \sigma_W^2}{\lamgap}\parenth{ \sqrt{\frac{d}{n}} + \sqrt{\frac{\log(1/\delta)}{n}} + \sqrt{\frac{d}{\nt}} + \sqrt{\frac{\log(1/\delta)}{\nt}} }.
			\end{align*}
		\end{lemma}

		Let $T\subset \R^d$ be a general set. The Gaussian width of $T$~\cite[Definition 7.5.1]{vershynin2018high} is defined as 
		\begin{align*}
			\gw{T} \defn \EE{\sup_{x \in T} \inner{g}{x}},
		\end{align*}
		where $g\sim \mathcal{N}(0,\ident_d)$ is a standard Gaussian random vector. The Gaussian width is a measure that captures the ``size" or ``complexity" of a set in terms of how it interacts with Gaussian noise. In our setting, the Gaussian width provides a tool in establishing the restricted eigenvalue condition (\cite{bickel2009simultaneous}) for a sub-Gaussian ensemble over a set $T$. Using the arguments from~\cite[Remark 1.2]{liaw2017simple}, we can obtain the following result.
		\begin{lemma}[Adaptation of Remark 1.2 of~\cite{liaw2017simple}]\label{lem:restricted_eigenvalue}
			Let $W_i \in \mathbb{R}^d$, $i=1,\ldots,n$, be $\iid$ zero-mean sub-Gaussian random vectors with parameter $\sigma_W$ and covariance matrix $\Sigma_W=\EE{W_1W_1^\top}$. Let $\bW\in\R^{n\times d}$ denote the matrix with rows $W_i$. Then, there exists a universal constant $c_0>0$ such that for any subset $T\subset\R^d$ that contains the origin $0\in T$ and for $\delta \in (0, 1/2)$, with probability at least $1-2\delta$,
			\begin{align*}
				\frac{1}{n}\vecnorm{\bW x}{2}^2 \geq \frac{\lammin(\Sigma_W)}{2}\vecnorm{x}{2}^2  - c_0 \frac{ \sigma_W^4 \lammin^{-2}(\Sigma_W) \log(1/\delta) \vecopnorm{\Sigma_W} \mathbb{W}^2(T)}{{n}} \text{ for all $x\in T$}.
			\end{align*}
		\end{lemma}
				
		The following two lemmas provide concentration bounds on the norms of the product of a sub-Gaussian random vector and a sub-Gaussian random variable. Their proofs are provided in~\appref{proof_of_lemmas}. The $\ell_2$ norm bound follows from the concentration inequality under sub-exponential condition. 
						
		\begin{lemma}\label{lem:l2_norm_bound}
			Let \( W_1, \ldots, W_n \in \mathbb{R}^d \) be i.i.d.\ zero-mean sub-Gaussian random vectors with parameter \( \sigma_W \), and let \( Z_1, \ldots, Z_n \in \mathbb{R} \) be i.i.d.\ zero-mean sub-Gaussian random variables with parameter \( \sigma_Z \). Define \( \bW \in \mathbb{R}^{n \times d} \) with rows \( W_i \) and \( \bZ \in \mathbb{R}^n \) with entries \( Z_i \). Then, for any $\delta \in (0, 1/2)$, with probability at least \( 1 - \delta \),
			\[
			\left\| \frac{1}{n} \left( \bW^\top \bZ - \mathbb{E}[\bW^\top \bZ] \right) \right\|_2 \leq c \sigma_W \sigma_Z \left\{ \sqrt{ \frac{d \log(1/\delta)}{n} } + \frac{d \log(1/\delta)}{n} \right\}.
			\]
		\end{lemma}
		
		The $\ell_\infty$ norm bound follows from applying Bernstein's inequality to each coordinate and taking a union bound. 
		\begin{lemma}\label{lem:l_infty_norm_bound}
			Let \( W_1, \ldots, W_n \in \mathbb{R}^d \) be i.i.d.\ zero-mean sub-Gaussian random vectors with parameter \( \sigma_W \), and let \( Z_1, \ldots, Z_n \in \mathbb{R} \) be i.i.d.\ zero-mean sub-Gaussian random variables with parameter \( \sigma_Z \). Define \( \bW \in \mathbb{R}^{n \times d} \) with rows \( W_i \) and \( \bZ \in \mathbb{R}^n \) with entries \( Z_i \). Then, for any \( \delta \in (0,1) \), with probability at least \( 1 - \delta \),
			\[
			\left\| \frac{1}{n} \left( \bW^\top \bZ - \mathbb{E}[\bW^\top \bZ] \right) \right\|_\infty \leq c \sigma_W \sigma_Z \left( \sqrt{ \frac{ \log d + \log(1/\delta) }{n} } + \frac{ \log d + \log(1/\delta) }{n} \right).
			\]
		\end{lemma}
		
		\section{Relations between UDA and oracle estimators}
		
		In this section, we provide details relating the UDA estimators to the oracle estimator discussed in~\secref{finite_sample_theory}.
		
		\subsection{DIP$^{(1)}$-cov under CA shift interventions}\label{app:pop_dip_estimator}
		
		We consider the data generation according to~\assumpref{mean_shift_linear_SCM}. Letting $H=(\ident - B)^{-1}$, we can calculate
		\begin{align}\label{eqn:dip_sigma_diff} \SigmaXsrcone-\SigmaXtar = -H (b w_Y^\top +W)(b w_Y^\top +W)^\top H^\top.\end{align}
		Recall the subspace notation given in~\secref{fine_tuning_mean_shift}, i.e., $\Vdip\in\R^{d\times \rdip}$ is an orthonormal basis for the column space of $\SigmaXsrcone-\SigmaXtar$, and $\Qdip\in\R^{d\times (d-\rdip)}$ is an orthonormal basis for the orthogonal complement of $\col{\Vdip}$.
		
		Since the covariance matching penalty $\beta^\top (\SigmaXsrcone-\SigmaXtar)\beta=0$ is equivalent to $\vecnorm{\Pi_{\Vdip} \beta}{2} = 0$, any feasible $\beta$ under covariance matching penalty lies in the orthogonal complement of $\col{\Vdip}$ and can be written as $\beta=\Qdip u$ for some $u\in\R^{d-\rdip}$. Substituting this into Equation~\eqnref{dip_cov_matching_pop}, the population DIP$^{(1)}$-cov is the minimizer of the following unconstrained problem
		\[u_{\text{DIP}}^{(1)} :=\argmin_u \EE{\left(\Ysrcone - u^\top \Qdip^\top \Xsrcone\right)^2} . \]
		Setting the gradient of the objective to zero yields the explicit form 
		$$u_{\text{DIP}}^{(1)}=\left(\Qdip^\top \SigmaXsrcone \Qdip\right)^{-1} \Qdip^\top\EE{\Xsrcone\Ysrcone}.$$ Transforming the variable back, we get
		\begin{align}\label{eqn:pop_dip_estimator}
			\betasrconedip=\Qdip u_{\text{DIP}}^{(1)} = \Qdip\left(\Qdip^\top \SigmaXsrcone \Qdip\right)^{-1} \Qdip^\top \EE{\Xsrcone\Ysrcone}.
		\end{align}
		By construction, $\Qdip$ is orthogonal to the column space of $H(bw_Y^\top + W)$, and thus
		\begin{align*}
			\EE{\Qdip^\top \Xtar \Ytar} &= \EE{\Qdip^\top Hb\epstar_Y{}^2 + \Qdip^\top H\epstar_X\epstar_Y}  \nonumber \\
			&= \Qdip^\top Hb (w_Y^\top w_Y + \VV{\xi_Y})  + \Qdip^\top H W w_Y\\
			&=  \Qdip^\top Hb \VV{\xi_Y} \\
			&= \Qdip^\top Hb \VV{\epssrcone_Y}  = \EE{\Qdip^\top \Xsrcone\Ysrcone}, 
		\end{align*}
		where the first step uses~\eqnref{Xsrc_equation}, the second step follows from~\assumpref{mean_shift_linear_SCM}, and the last step again uses~\eqnref{Xsrc_equation}. Together with the covariance matching $\Qdip^\top\SigmaXsrcone \Qdip=\Qdip^\top \SigmaXtar \Qdip$, it follows that the DIP estimator~\eqnref{pop_dip_estimator} is equivalently given by
		\begin{align*}
			\betasrconedip = \Qdip\left(\Qdip^\top \SigmaXtar \Qdip\right)^{-1} \Qdip^\top \EE{\Xtar\Ytar} \overset{(i)}{=} \Qdip\left(\Qdip^\top \SigmaXtar \Qdip\right)^{-1} \Qdip^\top \SigmaXtar \betatrue,
		\end{align*}
		where step $(i)$ uses $\betatrue=\SigmaXtar{}^{-1}\EE{\Xtar\Ytar}$. Finally, applying the identity for the projection operator~\cite{mccullagh2009marginal}
		$$\Qdip\left(\Qdip^\top \SigmaXtar \Qdip\right)^{-1} \Qdip^\top \SigmaXtar=\ident - \SigmaXtar{}^{-1}\Vdip(\Vdip^\top \SigmaXtar{}^{-1} \Vdip)^{-1}\Vdip^\top, $$ we conclude 
		\[\betasrconedip = \left(\ident - \SigmaXtar{}^{-1}\Vdip(\Vdip^\top \SigmaXtar{}^{-1} \Vdip)^{-1}\Vdip^\top \right)\betatrue.\]		
		
		\subsection{OLS-Src$^{(1)}$ under SC shift interventions}\label{app:formula_olssrc}

		We consider the data generation according to~\assumpref{connectivity_shift_linear_SCM}. For $m=0,1$, let \(\Hsrc = (\ident - \Bsrc)^{-1}\), $\vary=\VV{\epssrc_Y}$, and $\Sigma_{\eps_X} =\EE{\epssrc_X\epssrc_X{}^\top} $. Since $\Xsrc=\Hsrc b\epssrc_Y + \Hsrc \epssrc_X$, we can compute
		\[\SigmaXsrc = \Hsrc(\vary b b^\top + \Sigma_{\eps_X})\Hsrc{}^\top, \text{ for $m=0,1$}. \]
		Using the Sherman-Morrison inversion formula, the inverse of $\SigmaXsrc$ is 
		\begin{align*}
			\SigmaXsrc{}^{-1} &= \left(\vary\Hsrc b b^\top \Hsrc{}^\top +\Hsrc \Sigma_{\eps_X}\Hsrc{}^\top \right)^{-1}\\
			&= \Hsrc{}^{-\top} \left[  \Sigma_{\eps_X}^{-1} - \frac{\vary  \Sigma_{\eps_X}^{-1} b b^\top \Sigma_{\eps_X}^{-1}}{1 + \vary {b}^\top \Sigma_{\eps_X}^{-1} {b}}\right] \Hsrc{}^{-1}.
		\end{align*}
		Additionally, we have 
		\[\EE{\Xsrc\Ysrc} = \EE{\Hsrc b \epssrc_Y{}^2 + \Hsrc \epssrc_X\epssrc_Y} = \vary \Hsrc b ,  \text{ for $m=0,1$}. \]
		Combining these, the population OLS-Src$^{(1)}$ estimator and the oracle estimator are given by
		\[
		\begin{cases}
			\betasrcone = \SigmaXsrcone{}^{-1} \EE{\Xsrcone\Ysrcone} = \frac{\vary}{1+\vary b^\top \Sigma_{\eps_X}^{-1}b} \Hsrcone{}^{-\top} \Sigma_{\eps_X}^{-1}b, \\
			\betatrue = \SigmaXtar{}^{-1} \EE{\Xtar\Ytar} = \frac{\vary}{1+\vary b^\top \Sigma_{\eps_X}^{-1}b} \Htar{}^{-\top} \Sigma_{\eps_X}^{-1}b.
		\end{cases}
		\]
		The difference between the OLS-Src$^{(1)}$ and oracle estimators is therefore
		\[
		\betasrcone - \betatrue = \frac{\vary}{1+\vary b^\top \Sigma_{\eps_X}^{-1}b}  \left(\Btar - \Bsrcone\right)^\top \Sigma_{\eps_X}^{-1}b.
		\]
		Since \((\Btar - \Bsrcone)_{*S^c} = 0\) by~\assumpref{connectivity_shift_linear_SCM}, it follows that \((\betasrcone - \betatrue)_{S^c} = 0\), that is, \(\betasrcone - \betatrue\) is $\rconn$-sparse with support $S$.

		\subsection{CIP-mean under AW shift interventions}\label{app:pop_cip_estimator}
		
		We consider the data generation according to~\assumpref{coefficient_shift_linear_SCM}. Let $H=(\ident-B)^{-1}$. Since $\epssrc_X$ and $\epssrc_Y$ are independent for all $m=0,1,\ldots,M$, we have
		\begin{align*}
			\EEst{\Xsrc}{\Ysrc=y} = \EEst{H\bsrc \Ysrc + H\epssrc_X}{\Ysrc=y} = H\bsrc y.
		\end{align*}
		The CIP-mean matching penalty~\eqnref{cip_mean_matching_pop} therefore becomes
		\begin{align}\label{eqn:pop_cip_constraint}
			\beta^\top \left(H\bsrc -H\bsrcmprev \right)=0 \text{ for all $m\geq 2$}.
		\end{align}
		Recall the subspace notation given in~\secref{cip_finetune}, i.e., $\Vcip\in\R^{d\times \rcip}$ represent an orthonormal basis for the subspace $\spn\{ H\bsrc -H\bsrcone, m\geq 2\}$, and $\Qcip$ represent an orthonormal matrix satisfying $\Qcip^\top \Vcip=0$. From~\eqnref{pop_cip_constraint}, any feasible $\beta$ must be orthogonal to $\Vcip$, so we can reparameterize $\beta=\Qcip u$ for some $u\in\R^{d-\rcip}$. Substituting this into the original CIP-mean problem~\eqnref{cip_mean_matching_pop}, we get
		\[u_{\text{CIP}} \defn\argmin_u \frac{1}{M}\sum_{m=1}^M \EE{\left(\Ysrc - u^\top \Qcip^\top \Xsrc\right)^2} . \]
		By definition of $\Qcip$, we have $\Qcip^\top H\bsrc = \Qcip^\top H\bsrcmp$ for all $m\neq m'\geq 1$. Further, the assumption~\eqnref{cic_assumption} implies $\Qcip^\top H\btar = \Qcip^\top H\bsrcone$, and therefore $\Qcip^\top H\btar =\Qcip^\top H\bsrc$ for all $m=1,\ldots,M$. Thus, for any $m\geq 1$, by~\assumpref{coefficient_shift_linear_SCM}, we have
		\begin{align*}
			\Qcip^\top \Xsrc &= \Qcip^\top H\bsrc \epssrc_Y+  \Qcip^\top H \epssrc_X\nonumber \\
			&= \Qcip^\top H\btar \epssrc_Y +  \Qcip^\top H \epssrc_X\nonumber \\
			&\overset{D}{=}  \Qcip^\top H\btar \epstar_Y +  \Qcip^\top H \epstar_X\nonumber\\
			&= \Qcip^\top \Xtar,
		\end{align*}
		and additionally the joint distribution of $(\Qcip^\top \Xsrc,\Ysrc)$ is identical to that of $(\Qcip^\top \Xtar,\Ytar)$.
		It follows that $u_{\text{CIP}}$ is also the minimizer of the least squares problem in the target domain, i.e.,
		\[u_{\text{CIP}} =\argmin_{u}\EE{\left(\Ytar - u^\top \Qcip^\top \Xtar\right)^2}=(\Qcip^\top \SigmaXtar \Qcip)^{-1}\Qcip^\top \EE{\Xtar\Ytar},\]
		which yields the population CIP
		\begin{equation}\label{eqn:cip_pop_expression}
			\betacip=\Qcip\ucip = \Qcip(\Qcip^\top \SigmaXtar \Qcip)^{-1}\Qcip^\top \EE{\Xtar\Ytar}.
		\end{equation}
		Following the same argument as in~\appref{pop_dip_estimator}, this leads to
		\[\betacip = \left(\ident - \SigmaXtar{}^{-1}\Vcip(\Vcip^\top \SigmaXtar{}^{-1} \Vcip)^{-1}\Vcip^\top \right)\betatrue. \]
		
		\paragraph*{Relation between inverses of source and target covariance matrices} Since $H\btar - H\bsrcone \in \spn\{ H\bsrc -H\bsrcone, m\geq 2\}$ by~\eqnref{cic_assumption}, we can write 
		\[H\btar - H\bsrcone=\Vcip\alpha,\]
		for some $\alpha\in\R^{\rcip}$. Letting $\vary=\VV{\epssrcone_Y}=\VV{\epstar_Y}$, we observe that 
		\begin{align*}
			\SigmaXtar - \SigmaXsrcone &= \vary \left( H\btar\btar{}^\top H^\top - H\bsrcone\bsrcone{}^\top H^\top\right) \\
			&=\vary\left( \Vcip\alpha \bsrcone{}^\top H^\top + H\bsrcone \alpha^\top \Vcip^\top + \Vcip\alpha\alpha^\top \Vcip^\top \right). 
		\end{align*}
		By the Woodbury matrix identity, it follows
		\begin{align*}
			\SigmaXtar{}^{-1} &= \left(\SigmaXsrcone +\vary \Vcip\alpha \btar{}^\top H^\top + \vary H\btar \alpha^\top \Vcip^\top +\vary \Vcip\alpha\alpha^\top \Vcip^\top  \right)^{-1} \\
			&= \SigmaXsrcone{}^{-1} -\vary \SigmaXsrcone{}^{-1}\bigl[\,H\bsrcone \;\; \Vcip\alpha\bigr]\,
			S^{-1}\,
			\bigl[\,H\bsrcone \;\; \Vcip\alpha\bigr]^{\top}
			\SigmaXsrcone{}^{-1},
		\end{align*}
		where 
		\[ S= \begin{pmatrix}
			\vary\alpha^\top \Vcip^\top \SigmaXsrcone{}^{-1}\Vcip\alpha & -(1+\vary\bsrcone{}^\top H^\top \SigmaXsrcone{}^{-1} \Vcip\alpha) \\ -(1+\vary\bsrcone{}^\top H^\top \SigmaXsrcone{}^{-1} \Vcip\alpha) & -1+\vary\bsrcone{}^\top H^\top \SigmaXsrcone{}^{-1}H\bsrcone
		\end{pmatrix}. \]
		This proves Equation~\eqnref{cip_inv_cov_relation}.

		\section{Finite-sample guarantees of UDA estimators}\label{app:UDA_estimator_finite_sample_guarantees}
		
		In this section, we quantify the finite-sample errors for (pooled) OLS-Src, DIP$^{(1)}$-cov, and CIP-mean.
		
		\subsection{Finite-sample analysis for OLS-Pool}\label{app:OLS_pool}
		
		We begin by introducing OLS-Pool that performs OLS on a pooled set of source datasets. This estimator is arguably the most common baseline method when working with multiple source datasets.
		\begin{itemize}
			\item \textit{OLS-Pool in population:} the population OLS estimator based on the pooled source distributions is
			\begin{align*}
				\fpool(x) \defn\betapool^\top x, \text{ where }\betapool  \defn\argmin_{\beta}  \ \frac{1}{M}\sum_{m=1}^{M}\EE{(\Ysrc - \beta^\top \Xsrc)^2} .
			\end{align*}
			In the current formulation of OLS-Pool, we use uniform weights for each source distribution, but other weighting schemes can also be applied. For example, the weights can be determined based on the similarity of each source distribution to the target distribution~\cite{yao2024improving}. 
		\end{itemize}
		By pooling all source datasets, OLS-Pool can be more robust to the perturbations observed across multiple source domains compared to individual OLS-Src$^{(m)}$. This estimator is often used in the literature as a baseline to evaluate the effectiveness of more advanced UDA methods~\citep{gulrajani2020search}. However, OLS-Pool does not leverage any labeled target samples or exploit structural relationships between the source and target, so it does not always guarantee low target risk.
		\begin{itemize}
			\item \textit{OLS-Pool in finite-sample:} the finite-sample version of OLS-Pool based on pooled multiple source datasets is
			\begin{align}\label{eqn:OLSPool_finite}
				\fhpool(x) \defn\betahpool^\top x, \text{ where }\betahpool  \defn\argmin_{\beta}  \ \frac{1}{M}\sum_{m=1}^{M} \frac{1}{\nsrc}\sum_{i=1}^{\nsrc}\left(\Ysrc_i - \beta^\top \Xsrc_i\right)^2.
			\end{align}		
		\end{itemize}
		
		The following proposition provides finite-sample guarantees for $\betahpool$ assuming that the data are generated from anticausal linear SCMs in~\assumpsref{linear_SCMs}. For the case of a single source domain ($M = 1$), such results are well-established in the context of causal linear prediction. However, analogous results for anticausal linear SCMs with multiple source environments are not readily available in the literature. 
		
		\begin{proposition}\label{prop:least_squares_bound}
			Under~\assumpsref{linear_SCMs},~\assumpssref{sub_gaussianity}, there exist universal constants $c_0, c_1 >0$ such that for $\delta \in (0, 1/(2M))$, if $\nbar \geq c_1 \max\braces{ d_\delta\tsigmamin^2 , d\log(1/\delta) }$ with $d_\delta=d+\log(1/\delta)$, with probability at least $1-2M\delta$, the estimator $\betahpool$ satisfies 
			\begin{align*}
				\vecnorm{\betahpool-\betapool}{2}\leq 
				c_0 \eta \max_{m=1,\ldots,M}\sqrt{\frac{d\log(1/\delta)}{\nsrc}},
			\end{align*}	
			where $\eta=  \tsigmamin \cdot ( \vecnorm{\betapool}{2} + \frac{\sigma_Y}{\sigma_X})$.
		\end{proposition}
		The proof is given in~\appref{ols_pool_proof}. When $\eta$ is of constant order, this result recovers the standard error rate of $\sqrt{d/\nsrc}$ for a single source domain ($M=1$), and achieves a rate of $\max_{m\geq 1}\sqrt{d/\nsrc}$ when multiple source domains are used.
		
		\subsection{Finite-sample analysis for DIP$^{(1)}$-cov}\label{app:finite_sample_DIP}
		
		We assume the CA shift interventions in~\assumpref{mean_shift_linear_SCM} and use the notation given in~\secref{fine_tuning_mean_shift}. Recall $\Vhdip\in\R^{d\times \rdip}$ is the matrix of eigenvectors corresponding to the top $\rdip$ eigenvalues (in absolute value) of $\SigmaXhtar-\SigmaXhsrcone$, where $\SigmaXhsrcone$ and $\SigmaXhtar$ are the empirical covariance matrices given by
		\begin{equation}\label{eqn:unlabeled_sample_cov_mats}
			\SigmaXhsrcone = \frac{1}{\nsrconeu}\sum_{i=1}^{\nsrconeu}\Xsrconetil_i\Xsrconetil_i{}^\top \text{ and } \ \SigmaXhtar = \frac{1}{\ntaru}\sum_{i=1}^{\ntaru}\Xtartil_i\Xtartil_i{}^\top.
		\end{equation}
		While $\Vhdip$ serves as a finite-sample estimate of $\Vdip$, its accuracy depends on the size of the eigen-gap, i.e., the separation between the $\rdip$ nonzero eigenvalues of $\SigmaXtar-\SigmaXsrcone$ and zero. 
		
		To illustrate how $\lamdipgap$ defined in~\eqnref{ftdip_eigengap} relates to this eigen-gap, we can write from~\eqnref{dip_sigma_diff} that $\SigmaXtar - \SigmaXsrcone  = H\Adiff H^\top$,
		where recall from~\secref{fine_tune_finite_sample_guarantee} that $\Adiff=(bw_Y^\top + W)(bw_Y^\top + W)^\top$. Since both $\SigmaXtar - \SigmaXsrcone $ and $\Adiff$ are positive semi-definite with rank $\rdip$, let $\lambda_1(\SigmaXtar - \SigmaXsrcone)\geq \ldots\geq \lambda_{\rdip}(\SigmaXtar - \SigmaXsrcone) > 0$ denote the $\rdip$ nonzero eigenvalues of $\SigmaXtar-\SigmaXsrcone $, and let $\lambda_1(\Adiff)\geq\ldots\geq \lambda_{\rdip}(\Adiff)> 0$ denote the $\rdip$ nonzero eigenvalues of $\Adiff$. By~\cite[Theorem 1]{ostrowski1959quantitative}, we then have 
		\[\frac{\lambda_i(\SigmaXtar - \SigmaXsrcone) }{\lambda_i(\Adiff )} \geq \lammin(H^\top H) \text{ for all $i=1,\ldots,\rdip$}.\]
		This inequality implies that the nonzero eigenvalues of $\SigmaXtar-\SigmaXsrcone $ are always lower bounded by $\lamdipgap$, and thereby characterizing the eigen-gap condition for $\SigmaXtar-\SigmaXsrcone$.
				
		In the following, we formalize how the finite-sample error bound of DIP$^{{(1)}}$-cov depends on this eigen-gap and the sizes of the source and unlabeled target samples.
		\begin{proposition}\label{prop:DIP_finite_sample_error_bound}
			Under~\assumpsref{linear_SCMs},~\assumpssref{sub_gaussianity},~\assumpssref{mean_shift_linear_SCM}, there exist universal constants $c,c',c''>0$ such that for $\delta \in (0, 1/3)$, if $\nsrcone\geq  c\max\braces{\tsigmamin^2 \ddel,d\log(1/\delta)}$ and $\nsrconeu,\ntaru\geq c' \tsigmagap^2 \ddel$ with $\ddel = d + \log(1/\delta)$,
		    with probability at least $1-3 \delta$, the estimator $\betasrconediph$ satisfies 
			\begin{align*}
				\vecnorm{\betasrconediph-\betasrconedip}{2}\leq\eta_1\sqrt{\frac{d\log(1/\delta)}{\nsrcone}} + \eta_2\parenth{\sqrt{\frac{\ddel}{\nsrconeu}} + \sqrt{\frac{\ddel}{\ntaru}} },
			\end{align*}
			where $\eta_1 =c'' \tsigmamin \cdot \left(\sqrt{\kappabar}\vecnorm{\betasrcone}{2} + \frac{\sigma_Y}{\sigma_X}\right)$ and $\eta_2 = c''\tsigmagap \cdot {\kappabar^{3/2}\vecnorm{\betasrcone}{2}}$.
		\end{proposition}
		The proof is given in~\appref{dip_finite_proof}. \propref{DIP_finite_sample_error_bound} shows that when $\eta_1,\eta_2$ are all constant orders, and the number of labeled source samples $\nsrcone$ and unlabeled source/target samples $(\nsrconeu,\ntaru)$ are at least $\gtrsim d$, the estimation error scales as 
		\[\sqrt{\frac{d}{\nsrcone}} \text{ (labeled) } + \sqrt{\frac{d}{\nsrconeu}} + \sqrt{\frac{d}{\ntaru}} \text{ (unlabeled)}. \]
		The bound further shows that a larger eigen-gap $\lamdipgap$ leads to more accurate estimation of DIP$^{(1)}$-cov, which is consistent with the fact that estimating $\Vdip$ reliably requires well-separated eigenvalues.
		
		\subsection{Finite-sample analysis for CIP-mean}\label{app:finite_sample_CIP}
		We assume the AW shift interventions in~\assumpref{coefficient_shift_linear_SCM}. To simplify the analysis, each source dataset \( \Dsrc \) is divided into two subsets of equal size, \( \Dsrc = \Dsrc_1 \cup \Dsrc_2 \), \( m = 1, \ldots, \mcip \), where
		\[
		\Dsrc_\ell \defn \{ (\Xsrc_{\ell i}, \Ysrc_{\ell i}) \}_{i=1}^{\nsrc}, \quad \ell = 1, 2.
		\]
		Note that each subset $\Dsrc_\ell$ contains $\nsrc$ samples. When  it is clear from context the specific subset being used, we omit the subscript $\ell$ and denote the covariates and labels simply as \( \Xsrc_i \) and \( \Ysrc_i \). In the finite-sample CIP-mean~\eqnref{cip_mean_matching_finite_sample}, the first subset $\Dsrc_1$ is used to estimate $\Vhcip$ for the empirical conditional mean-matching penalty, and the second subset $\Dsrc_2$ is used to compute the least-squares objective.
		
		Before presenting the finite-sample guarantees of CIP-mean, we introduce additional notation than what was given in~\secref{cip_finetune}. Let $\Pw\in\R^{d\times (\mcip-1)}$ be the matrix whose columns are the differences $H\bsrc - H\bsrcprev$ across source domains
		\begin{equation}\label{eqn:def_Pw}\Pw \defn \begin{pmatrix}
			H\bsrctwo - H\bsrcone,\cdots, H\bsrcend - H\bsrcendprev
		\end{pmatrix} .\end{equation}
		According to~\assumpref{coefficient_shift_linear_SCM}, the matrix $\Pw$ has rank $\rcip$, and further, $\Vcip$ corresponds to the left-singular vectors of $\Pw$ corresponding to its $\rcip$ nonzero singular values.

		To construct the finite-sample estimate of $\Vcip$, we first estimate each $H\bsrc$ via regressing $\Xsrc$ on $\Ysrc$ using $\Dsrc_1$, i.e.,
		\begin{equation}\label{eqn:cip_ols_estimator}
			\bhsrc \defn \argmin_w \frac{1}{\nsrc}\sum_{i=1}^{\nsrc}  \norm{ \Xsrc_{1i} - \Ysrc_{1i} \cdot w}_2^2, \text{ for $m \geq 1$}.
		\end{equation}
		We then form a plug-in estimator of $\Pw$ as
		\begin{equation}\label{eqn:def_Pwh}\Ph \defn\begin{pmatrix}
			\bhsrctwo - \bhsrcone,\cdots, \bhsrcend - \bhsrcendprev
		\end{pmatrix}. \end{equation}
		The matrix $\Vhcip$ is then defined as the left-singular vectors of \( \Ph \) corresponding to its top $\rcip$ singular values. 

		To control the finite-sample error of $\Vhcip$---and hence of $\betacip$---it is necessary to make sure there is a sufficient separation between the nonzero singular values of $\Pw$ and zero. Comparing $\Pw$ with $\Paug$ in~\eqnref{def_Paug}, we observe that the smallest singular value of $\Pw$ is lower bounded by that of $\Paug$, i.e., $\lamcipauggap$ defined in~\eqnref{singular_gap_aug}. Therefore, $\lamcipauggap$ quantifies the separation of singular values and plays a critical role in controlling the finite-sample error of $\betaciph$. 

		The following proposition establishes the finite-sample guarantees for CIP-mean (recall $\nu_Y^2=\VV{\Ysrc}$ for all $m=0,1,\ldots,M$).
		\begin{proposition}\label{prop:CIP_finite_sample_error_bound}
			Under~\assumpsref{linear_SCMs},~\assumpssref{sub_gaussianity},~\assumpssref{coefficient_shift_linear_SCM}, there exist universal constants $c,c',c''>0$ such that for $\delta \in (0, 1/(4\mcip+4))$, if $\nbar\geq  c\max \parenth{\tsigmamin^2 d_\delta, d\log (1/\delta), \frac{\sigma_Y^4}{\nu_Y^4}\log (1/\delta) }$,
			 with probability at least  $1-(4\mcip+4)\delta$, the estimator $\betaciph$ satisfies 
			\begin{align*}
				\vecnorm{\betaciph-\betacip}{2}\leq \eta_1\sqrt{\frac{d\log(1/\delta)}{\nbar}} +  \eta_2\sqrt{\frac{\ddel}{\nbar}},
			\end{align*}
			where $\eta_1 =c''\tsigmamin\cdot \left(\sqrt{\kappabar}\vecnorm{\betapool}{2}+\frac{\sigma_Y}{\sigma_X}\right)$ and $\eta_2=c''\tsigmagapcip\cdot {\kappabar^{3/2}}\vecnorm{\betapool}{2}  $.
		\end{proposition}
		
		Its proof is provided in \appref{proof_of_CIP_finite_sample_error_bound}. \propref{CIP_finite_sample_error_bound} shows that when $\eta_1,\eta_2$ are constant orders, the error of \( \betahcip \) scales as \( \max_{m=1,\ldots,\mcip}\sqrt{d / \nsrc} \). This rate matches the convergence rate of the OLS-Pool estimator (see~\propref{least_squares_bound}). Furthermore, the second term in the bound shows that more diversity in the perturbation directions \( H\bsrc - H\bsrcprev \) leads to improved estimation accuracy, highlighting the importance of heterogeneity across source domains for achieving better CIP-mean estimates.
		
		\section{Risk upper bounds with model selection}\label{app:model_selection_MASFT}
		
		We provide target excess risk bound of MASFT described in~\secref{model_selection} when the loss function is given by the squared loss. In particular, we show that the model selection step~\eqnref{model_selection_step_3} in MASFT only introduces a mild overhead in the labeled target sample complexity beyond that needed for individual fine-tuning.
		\begin{proposition}\label{prop:model_selection}
			Suppose that the loss function is the squared loss, and for all $\theta\in \{\thetahfti{1}, \ldots, \thetahfti{L}\}$, $\Ytar-f_{\theta}(\Xtar)$ is zero-mean sub-Gaussian with parameter $\sigma_\ell$. Then, for any $\delta\in (0,1)$, if $\ntarval\geq c\log(2L/\delta)$, with probability at least $1-\delta$ over the randomness of the target validation dataset $\Dtarval$, we have
			\begin{align*}
				\delta\risktar{f_{\thetahatfinal}} \leq \min_{j=1,\ldots,L}\delta\risktar{f_{\thetahfti{j}}} + c'\sigma_\ell^2 \sqrt{\frac{\log(2L/\delta)}{\ntarval}},
			\end{align*}
			where $c,c'>0$ are universal constants.
		\end{proposition}
		
		The proof follows from classical model selection results (e.g.,~\cite[Section 4.6]{bach2024learning}) and is provided in~\appref{model_selection_proof} for completeness. The assumption that \(\Ytar - f_{\theta}(\Xtar)\) is a sub-Gaussian random variable is a mild condition that holds under general settings, such as when both \(\Xtar\) and \(\Ytar\) are sub-Gaussian and $f_\theta$, for \(\theta \in \{\thetahfti{1}, \ldots, \thetahfti{L}\}\), satisfies the Lipschitz condition. In this case, the sub-Gaussian parameter $\sigma_\ell$ depends on the sub-Gaussian parameters of \(\Xtar\) and \(\Ytar\), as well as the Lipschitz constant of $f_\theta$. 
		
		\propref{model_selection} shows that the target excess risk of the estimator $\thetahatfinal$ returned by MASFT is at most the minimum excess risk among all fine-tuned estimators, with an additional term of $\sqrt{\log L/\ntarval}$. Therefore, MASFT guarantees that only an extra $\ntarval\approx \log L$ labeled target samples are needed to select a near-optimal fine-tuned estimator from the candidate models.
		
		\section{Proof of main theorems on risk upper bounds}
		In this section, we prove~\thmref{dip_var_matching_finite},~\thmref{connectivity_estimator_finite}, and~\thmref{cip_mean_matching_finite}, which provide upper bounds for the fine-tuning methods presented in~\secref{finite_sample_theory}.

		\subsection{Proof of~\thmref{dip_var_matching_finite}}\label{app:dip_var_matching_finite}
		We begin by observing that the oracle estimator~\eqnref{OLSTar} satisfies the constraint $\Qdip^\top \SigmaXtar \betasrconedip = \Qdip^\top \SigmaXtar \betatrue$ after multiplying $\Qdip^\top \SigmaXtar$ on both sides of~\eqnref{dip_alternative}. This implies that $\betatrue$ is an optimal solution to the following constrained optimization problem
		\begin{equation}\label{eqn:fine_tune_dip_betatrue_constrained}
			\begin{aligned}
				&\betatrue= \argmin_{\beta} \EE{(\Ytar - \beta^\top \Xtar)^2}  \text{ s.t. } \Qdip^\top \SigmaXtar \betasrconedip = \Qdip^\top \SigmaXtar \beta.
			\end{aligned}
		\end{equation}
		Comparing with FT-DIP$^{(1)}$~\eqnref{dip_var_matching_finite}, the above problem uses population quantities in both its objective and constraint. To deal with finite-sample errors that arise from the objective and constraint separately, we introduce an intermediate estimator, where the objective is still at population level while the constraint is from empirical estimates,
		\begin{equation}\label{eqn:fine_tune_dip_betatarcon_constrained}
			\begin{aligned}
				&\betatarcon= \argmin_{\beta} \EE{(\Ytar - \beta^\top \Xtar)^2}  \text{ s.t. } \Qhdip^\top \SigmaXhtar \betasrconediph = \Qhdip^\top \SigmaXhtar \beta.
			\end{aligned}
		\end{equation}
		Therefore, our proof strategy is to decompose the error into two terms by writing $\betatrue - \betahftdip = (\betatrue - \betatarcon) + (\betatarcon - \betahftdip)$, and bound each term separately. Specifically, we name the two terms as follows:
		\begin{equation}\label{eqn:ft_dip_decompose}
			\vecnorm{\betatrue - \betahftdip}{2} \leq \underbrace{\vecnorm{\betatrue-\betatarcon}{2}}_{\eqqcolon T_1} + \underbrace{\vecnorm{\betatarcon - \betahftdip}{2}}_{\eqqcolon T_2}.
		\end{equation}
		
		The first term $T_1$ arises from using the finite-sample estimates $\Qdip, \SigmaXtar$, $\betasrconedip$ in the constraint. We expect $T_1$ to depend on the source and unlabeled target sample sizes $\nsrcone,\nsrconeu, \ntaru$, which are assumed to be large enough to ensure $T_1$ is small. The second term $T_2$ captures finite-sample error of least squares on a constrained subspace of dimension $\rdip$, so we expect it to scale as $\sqrt{\rdip/\ntar}$. In particular, the introduction of $\betatarcon$ allows us to isolate  the finite-sample error due to the labeled target samples $\Dtar$ while conditioning on $\Qhdip$, $\SigmaXhtar$, $\betasrconediph$ being fixed. Since $\Dtar$ is independent of $\Dsrcone$, $\Dsrconeu$, $\Dtaru$, combining $T_1$ and $T_2$ via a union bound suffices to conclude.

		Before we proceed to bound $T_1$ and $T_2$, given the subspace notations in Section~\ref{sec:fine_tune_mixture_mean_shift}, we define high probability events under which the empirical estimates based on source and unlabeled target data concentrate around their population counterparts. Define orthonormal matrices \(R_q \in \R^{(d-\rdip) \times (d-\rdip)}\) and \(R_v \in \R^{\rdip \times \rdip}\) as 
		\[
		R_q \defn \inf_{\widetilde{R}^\top\widetilde{R} = \ident} \vecopnorm{\Qhdip - \Qdip\widetilde{R}}, \quad \text{and} \quad R_v \defn \inf_{\widetilde{R}^\top\widetilde{R} = \ident} \vecopnorm{\Vhdip - \Vdip\widetilde{R}}.
		\]
		The first event $\Eset_1$ ensures the covariance matrix and subspaces are well aligned with their population counterparts.
		\begin{align*}
			\Eset_1 \defn \Bigg\{ \,
			& \vecopnorm{\SigmaXhtar - \SigmaXtar} \leq c_1\sigma_X^2\sqrt{\frac{\ddel}{\ntaru}}; \ \ \frac{\lamminbar}{2} \leq \lammin(\SigmaXhtar) \leq \lammax(\SigmaXhtar) \leq 2\lammaxbar; \\
			& \vecopnorm{\Qhdip - \Qdip R_q} \leq c_1 \tsigmagap \left( \sqrt{\frac{\ddel}{\nsrconeu}} + \sqrt{\frac{\ddel}{\ntaru}} \right); \\
			& \vecopnorm{\Vhdip - \Vdip R_v} \leq c_1 \tsigmagap \left( \sqrt{\frac{\ddel}{\nsrconeu}} + \sqrt{\frac{\ddel}{\ntaru}} \right)
			\; \Bigg\}.
		\end{align*}
		Since $\nsrconeu\geq c_2\tsigmagap^2 \ddel$ and $\ntaru\geq c_2 \max\{1,\tsigmagap^2,\tsigmamin^2\} \ddel $, by~\lemref{bounds_cov_matrices}, Equation~\eqnref{bounds_eig_values}, and~\lemref{error_orthogonal_space}, the event $\Eset_1$ holds with probability at least $1-2 \delta$ over the randomness of $\Dsrconeu,\Dtaru$. 
		
		Immediate consequences of being in $\Eset_1$, together with the sub-multiplicative property of the operator norm and the triangle inequality, are the following bounds
			\begin{align}\label{eqn:ftdip_ineq4}
			\vecopnorm{\ident - \SigmaXhtar{}^{-1} \SigmaXtar}\leq \vecopnorm{\SigmaXhtar{}^{-1}}\vecopnorm{\SigmaXhtar - \SigmaXtar}\leq  2c_1 \tsigmamin\sqrt{\frac{\ddel}{\ntaru}};
		\end{align}
		\begin{align}\label{eqn:ftdip_ineq1}
			\vecopnorm{\SigmaXhtar{}^{-1} - \SigmaXtar{}^{-1}}\leq \vecopnorm{\SigmaXtar{}^{-1}}\vecopnorm{\ident - \SigmaXhtar{}^{-1} \SigmaXtar} \leq  2c_1 \lamminbar^{-1}\tsigmamin\sqrt{\frac{\ddel}{\ntaru}};
		\end{align}
		\begin{align}\label{eqn:ftdip_ineq2}
			&\vecopnorm{\SigmaXhtar{}^{-1}\Vhdip - \SigmaXtar{}^{-1}\Vdip R_v}  \\
			&\leq \vecopnorm{\SigmaXhtar{}^{-1}\Vhdip - \SigmaXhtar{}^{-1}\Vdip R_v} + \vecopnorm{\SigmaXhtar{}^{-1}\Vdip R_v - \SigmaXtar{}^{-1}\Vdip R_v} \nonumber \\
			&\leq \vecopnorm{\SigmaXhtar{}^{-1}}\vecopnorm{\Vhdip-\Vdip R_v} + \vecopnorm{\SigmaXhtar{}^{-1} - \SigmaXtar{}^{-1}} \nonumber \\
			&\leq 2c_1\lamminbar^{-1} \tsigmagap\sqrt{\frac{\ddel}{\nsrconeu}} +    2c_1\lamminbar^{-1}\left(\tsigmamin + \tsigmagap\right)\sqrt{\frac{\ddel}{\ntaru}}; \nonumber
		\end{align}
    \begin{align}\label{eqn:ftdip_ineq3}
			&\vecopnorm{\SigmaXhtar{}^{-1}\Qhdip - \SigmaXtar{}^{-1}\Qdip R_q}   \\
			&\hspace{.2in}\leq 2c_1\lamminbar^{-1} \tsigmagap\sqrt{\frac{\ddel}{\nsrconeu}} +    2c_1\lamminbar^{-1}\left(\tsigmamin + \tsigmagap\right)\sqrt{\frac{\ddel}{\ntaru}}. \nonumber
		\end{align}
		
		Next, define the event \(\Eset_2\) as
		\begin{align*}
			\Eset_2 \defn \left\{\vecnorm{\betasrconediph - \betasrconedip}{2} \leq \eta_1\sqrt{\frac{d\log (1/\delta)}{\nsrcone}} + \eta_2\left(\sqrt{\frac{\ddel}{\nsrconeu}}  + \sqrt{\frac{\ddel}{\ntaru}}\right) \right\},
		\end{align*}
		where 
    \begin{align}\label{eqn:eta1_2_def}
      \eta_1 = c_3\tsigmamin\left(\frac{\sigma_Y}{\sigma_X}+\sqrt{\kappabar}\vecnorm{\betasrcone}{2}\right), \quad
      \eta_2 = c_3\tsigmagap\kappabar\sqrt{\kappabar}\vecnorm{\betasrcone}{2}.
    \end{align}
    By~\propref{DIP_finite_sample_error_bound} and the assumption $\nsrcone\geq c_2\max\{\tsigmamin^2 \ddel,d\log(1/\delta) \}$, $\nsrconeu,\ntaru\geq c_2\tsigmagap^2  \ddel$, the event $\Eset_2$ holds with probability at least $1-3\delta$. 

    Define
    \begin{align} \label{eqn:event_thm41_3}
      \Eset_3 \defn \braces{\vecnorm{\betasrcone - \betahsrcone}{2} \leq c_3 \eta_{\text{LS}} \sqrt{\frac{d\log(1/\delta)}{\nsrcone}}},
    \end{align}
		where $\eta_{\text{LS}} \defn \tsigmamin \parenth{\frac{\sigma_Y}{\sigma_X} + \vecnorm{\betasrcone}{2}}$. By~\propref{least_squares_bound} with $M=1$ and the assumption $\nsrcone\geq c_2 \max\{\ddel \tsigmamin^2  ,d\log (1/\delta)\}$, the event $\Eset_3$ holds with probability at least $1-2\delta$ over the randomness of $\Dsrcone$. 

    Define the adjusted response and projected covariate 
    \begin{align}\label{eqn:adj_resp_proj_covariate}
    	\begin{cases}
			     \Ztar_i \defn \Ytar_i -\Xtar_i{}^\top  \SigmaXhtar{}^{-1} \Qhdip\Qhdip^\top \SigmaXhtar \betasrconediph, \\
			\Wtar_i \defn \Vhdip^\top \SigmaXhtar{}^{-1} \Xtar_i.
    	\end{cases}
    \end{align}
    Let $\bWtar\in\R^{\ntar\times \rdip}$ be the matrix whose $i$-th row is $\Wtar_i$, and $\bZtar\in\R^{\ntar}$ be the vector with entries $\Ztar_i$. Define
    \begin{align}\label{eqn:event_thm41_4}
      \Eset_4 \defn \Bigg\{\frac{2}{\ntar}\vecnorm{\bWtar{}^\top ( \bWtar\Vhdip^\top \SigmaXhtar\betatarcon - \bZtar) - \EE{\bWtar{}^\top ( \bWtar\Vhdip^\top \SigmaXhtar\betatarcon - \bZtar)}}{2}   \\\leq c_4\sigma_W\sigma_Z \sqrt{\frac{\rdip\log(1/\delta)}{\ntar}} \Bigg\}. \nonumber
    \end{align}

    Since $\ntar\geq \rdip\log(1/\delta)$, applying~\lemref{l2_norm_bound} for concentration of product of sub-Gaussian vectors around its mean, event~$\Eset_4$ holds with probability at least $1-\delta$, where $\sigma_W$ and $\sigma_Z$ represent the sub-Gaussian parameters of $\Wtar_i$ and $\Wtar_i{}^\top \Vhdip^\top \SigmaXhtar\betatarcon - \Ztar_i=\Xtar_i{}^\top \betatarcon - \Ytar_i$, respectively.

    Let
		\begin{align}\label{eqn:event_thm41_5_high_prob_set}
			\Lambda \defn \braces{x:\vecnorm{\Qhdip^\top\SigmaXhtar x}{2}=0,\  \vecnorm{\Vhdip^\top\SigmaXhtar x}{2}\leq 2\varrho },
		\end{align}
		and define 
    \begin{align} \label{eqn:event_thm41_5}
      \Eset_5 \defn \braces{\frac{1}{\ntar}\vecnorm{\bXtar x}{2}^2 \geq  \frac{\lamminbar}{2}\vecnorm{x}{2}^2 - \frac{c_5  \lamminbar^{-1}\sigma_X^4\kappabar\gw{\Lambda}^2  \log (1/\delta)}{\ntar},\quad \forall x \in \Lambda}.
    \end{align}
    Since $\Lambda$ contains origin, applying~\lemref{restricted_eigenvalue}, it follows $\Eset_5$ holds with probability at least $1-2\delta$ over the randomness of $\Dtar$. 
		
    We define the intersection of the events
    \[\Eset \defn \Eset_1 \cap \Eset_2 \cap \Eset_3 \cap \Eset_4 \cap \Eset_5. \] 
    In the following, we prove our result under the event $\Eset$. 
		
		\subsubsection{Bounding $T_1$} 
		
		The main steps to bound $T_1$ in~\eqnref{ft_dip_decompose} are as follows.
		\begin{enumerate}[(a)]
			\item (Decomposition of error) First, we decompose $\betatrue$ in the directions of $\SigmaXtar{}^{-1}\Qdip$ and $\SigmaXtar{}^{-1}\Vdip$, and similarly decompose $\betatarcon$ in the directions of $\SigmaXhtar{}^{-1}\Qhdip$ and $\SigmaXhtar{}^{-1}\Vhdip$.
			\item (Error in $V$ directions) Second, we express the components of $\betatrue$ and $\betatarcon$ in the directions of $\SigmaXtar{}^{-1}\Vdip$ and $\SigmaXhtar{}^{-1}\Vhdip$, respectively, as solutions to linear programs and use this to bound the difference between them.
			\item (Error in $Q$ directions) Third, the components in the directions of $\SigmaXtar{}^{-1}\Qdip$ and $\SigmaXhtar{}^{-1}\Qhdip$, are given by $\Qdip^\top \SigmaXtar \betasrconedip$ and $\Qhdip^\top \SigmaXhtar \betasrconediph$, respectively, which we control using the inequalities under the event $\Eset$. 
		\end{enumerate}
		
		\paragraph*{(a) Decomposition of error} Since $\betatrue$ satisfies the constraint $\Qdip^\top \SigmaXtar \betasrconedip = \Qdip^\top \SigmaXtar \betatrue$ and $(\Qdip, \Vdip)$ together form an orthonormal basis, we can write \(\betatrue\) as
		\begin{align}\label{eqn:ft_dip_betastar}
			\betatrue &= \SigmaXtar{}^{-1}(\Qdip R_q R_q^\top \Qdip^\top + \Vdip R_v R_v^\top \Vdip^\top)\SigmaXtar \betatrue  \\
			&= \SigmaXtar{}^{-1} \Qdip R_q R_q^\top \Qdip^\top \SigmaXtar \betasrconedip +  \SigmaXtar{}^{-1}\Vdip R_v  \underbrace{R_v^\top \Vdip^\top\SigmaXtar \betatrue}_{\eqqcolon \alphatrue}.\nonumber
		\end{align}
		For notational convenience, we overwrite $\Qdip=\Qdip R_q$ and $\Vdip=\Vdip R_v$ for the remainder of the proof. Now consider the objective in~\eqnref{fine_tune_dip_betatrue_constrained}.
		\begin{align*}
			&\EE{(\Ytar - \beta^\top \Xtar)^2}  \\
			&= \EE{(\Ytar - \beta^\top \SigmaXtar (\Qdip \Qdip^\top + \Vdip  \Vdip^\top)\SigmaXtar{}^{-1} \Xtar)^2} \\
			&\overset{(i)}{=} \EE{(\Ytar - \betasrconedip{}^\top \SigmaXtar \Qdip  \Qdip^\top\SigmaXtar{}^{-1} \Xtar + \underbrace{ \beta^\top \SigmaXtar \Vdip }_{\eqqcolon \alpha} \Vdip^\top \SigmaXtar{}^{-1} \Xtar )^2},
		\end{align*}
		where step $(i)$ holds due to the constraint $\Qdip^\top \SigmaXtar \betasrconedip = \Qdip^\top \SigmaXtar \beta$. Note that \(\alphatrue\) minimizes the above objective seen as a function of $\alpha$. Therefore, it satisfies the first-order optimality condition,
		\begin{align}\label{eqn:ft_dip_lin_system_pop}
			&\underbrace{\EE{\Vdip^\top \SigmaXtar{}^{-1}\Xtar(\Ytar - \betasrconedip{}^\top \SigmaXtar \Qdip  \Qdip^\top\SigmaXtar{}^{-1} \Xtar )}}_{\eqqcolon \bstar}  \\
      &\quad\quad = {\underbrace{\EE{\Vdip^\top \SigmaXtar{}^{-1}\Xtar\Xtar{}^{\top}\SigmaXtar{}^{-1} \Vdip }}_{\eqqcolon \Astar}}   \alphatrue. \nonumber
		\end{align}
		
		Similarly, $\betatarcon$ satisfies the constraint $\Qh^\top \SigmaXhtar \betasrconediph = \Qh^\top \SigmaXhtar \betatarcon$, which allows us to write
		\begin{align}\label{eqn:ft_dip_betatarcon}
			\betatarcon &=  \SigmaXhtar{}^{-1} \Qhdip\Qhdip^\top \SigmaXhtar \betasrconediph +  \SigmaXhtar{}^{-1}\Vhdip \underbrace{\Vhdip^\top \SigmaXhtar\betatarcon}_{\eqqcolon \alphatarcon}.
		\end{align}
		And \(\alphatarcon\) satisfies the following first-order optimality condition
		\begin{align}\label{eqn:ft_dip_lin_system_finite_sample}
			& \underbrace{\EE{\Vhdip^\top \SigmaXhtar{}^{-1}\Xtar(\Ytar - \betasrconediph{}^\top \SigmaXhtar \Qhdip  \Qhdip^\top\SigmaXhtar{}^{-1} \Xtar )}}_{\eqqcolon \btarcon}  \\
      &\quad \quad = {\underbrace{\EE{\Vhdip^\top \SigmaXhtar{}^{-1}\Xtar\Xtar{}^{\top}\SigmaXhtar{}^{-1} \Vhdip}}_{\eqqcolon \Atarcon} }   \alphatarcon. \nonumber
		\end{align}
		Subtracting Equations~\eqnref{ft_dip_lin_system_pop} and~\eqnref{ft_dip_lin_system_finite_sample}, we obtain
		\begin{align}\label{eqn:ft_dip_decom_2}
			&\underbrace{\bstar - \btarcon}_{\eqqcolon \Delta b} \\
			&= \underbrace{(\Astar - \Atarcon)}_{\eqqcolon  \Delta A}  \alphatrue + { \Atarcon} ( \alphatrue - \alphatarcon).\nonumber
		\end{align}
		In order to obtain the error in the $\SigmaXtar{}^{-1} \Vdip$ direction, it suffices to control  $ \alphatrue - \alphatarcon$, which reduces to control $\Delta b$ and $\Delta A$ separately.
		
		\paragraph*{(b) Error in $V$ directions} To bound $\Delta b$, rearranging terms and apply triangle inequality, we obtain
		\begin{multline*}
			\vecnorm{\Delta b}{2} \leq \underbrace{\vecnorm{\EE{\Vdip^\top \SigmaXtar{}^{-1}\Xtar\Ytar} - \EE{\Vhdip^\top \SigmaXhtar{}^{-1}\Xtar\Ytar}}{2}}_{\coloneqq (A_{1})} \\
			+	\underbrace{\vecnorm{{\Vdip^\top \SigmaXtar{}^{-1}\Qdip \Qdip^\top \SigmaXtar \betasrconedip } - {\Vhdip^\top \SigmaXhtar{}^{-1}\SigmaXtar \SigmaXhtar{}^{-1}\Qhdip \Qhdip^\top \SigmaXhtar \betasrconediph }}{2}}_{\coloneqq (A_{2})} .
		\end{multline*}
		To bound $(A_{1})$, note that
		\begin{align*}
			\text{$(A_{1})$}&=\vecnorm{\EE{\Vdip^\top \SigmaXtar{}^{-1}\Xtar\Ytar} - \EE{\Vhdip^\top \SigmaXhtar{}^{-1}\Xtar\Ytar}}{2} \nonumber\\
			&\overset{(i)}{=} \vecnorm{{\Vdip^\top \betatrue -\Vhdip^\top \SigmaXhtar{}^{-1}\SigmaXtar\betatrue} }{2} \nonumber  \\
			&\leq \vecnorm{(\Vdip-\Vhdip)^\top \betatrue}{2} + \vecnorm{\Vhdip^\top (\ident - \SigmaXhtar{}^{-1}\SigmaXtar)\betatrue}{2}\nonumber \\
			&\leq \vecopnorm{ \Vdip - \Vhdip}\vecnorm{\betatrue}{2} +\vecopnorm{\ident - \SigmaXhtar{}^{-1}\SigmaXtar} \vecnorm{\betatrue}{2},
		\end{align*}
		where step $(i)$ follows from $\betatrue=\SigmaXtar{}^{-1}\EE{\Xtar\Ytar}$. Under the event $\Eset_1$, we have
		\begin{align}\label{eqn:ft_dip_EXY}
			\text{$(A_{1})$}\leq c_1 \tsigmagap\vecnorm{\betatrue}{2} \sqrt{\frac{\ddel}{\nsrconeu}} +  2c_1 (\tsigmagap+\tsigmamin)\vecnorm{\betatrue}{2}\sqrt{\frac{\ddel}{\ntaru}}.
		\end{align}
		
		To bound $(A_{2})$, we add and subtract $\Vhdip^\top \SigmaXhtar{}^{-1}\Qdip\Qdip^\top \SigmaXtar \betasrconedip$ and apply triangle inequality to get
		\begin{multline}\label{eqn:ft_dip_continue_1}
			 (A_{2})
			\leq \vecnorm{(\Vdip^\top \SigmaXtar{}^{-1} - \Vhdip^\top \SigmaXhtar{}^{-1}){\Qdip\Qdip^\top \SigmaXtar \betasrconedip }}{2}  \\
			+  \vecnorm{{\Vhdip^\top \SigmaXhtar{}^{-1}\SigmaXtar (\SigmaXtar{}^{-1}\Qdip \Qdip^\top \SigmaXtar \betasrconedip - \SigmaXhtar{}^{-1}\Qhdip \Qhdip^\top \SigmaXhtar \betasrconediph) } }{2}.
		\end{multline}
		For the first term on the right-hand side of \eqref{eqn:ft_dip_continue_1}, we have
		\begin{align}\label{eqn:ft_dip_continue_2}
			& \vecnorm{(\Vdip^\top \SigmaXtar{}^{-1} - \Vhdip^\top \SigmaXhtar{}^{-1}){\Qdip\Qdip^\top \SigmaXtar \betasrconedip }}{2}   \\
			&\overset{(i)}{\leq} \vecopnorm{\Vdip^\top \SigmaXtar{}^{-1} - \Vhdip^\top \SigmaXhtar{}^{-1}}\vecopnorm{\SigmaXtar}\vecnorm{\betatrue}{2} \nonumber\\
			&\overset{(ii)}{\leq} 2c_1\tsigmagap\kappabar\vecnorm{\betatrue}{2}  \sqrt{\frac{\ddel}{\nsrconeu}} +  2c_1\left(\tsigmamin + \tsigmagap\right)\kappabar\vecnorm{\betatrue}{2}  \sqrt{\frac{\ddel}{\ntaru}}, \nonumber
		\end{align}
		where step $(i)$ uses the constraint $\Qdip^\top\SigmaXtar\betasrconedip=\Qdip^\top\SigmaXtar\betatrue$ and the inequality $\vecopnorm{ABx}\leq \vecopnorm{A}\vecopnorm{B}\vecnorm{x}{2}$, and step $(ii)$ uses the inequalities under $\Eset_1$. For the second term, observe that by the assumption $\ntaru\geq 4c_1^2\tsigmamin^2  \ddel$, we have
		\begin{align}\label{eqn:ft_dip_continue_7}
			\vecopnorm{\Vhdip^\top \SigmaXhtar{}^{-1}\SigmaXtar} \leq \vecopnorm{ \SigmaXhtar{}^{-1}\SigmaXtar -\ident} +\vecopnorm{\ident}\leq 2c_1\tsigmamin\sqrt{\frac{\ddel}{\ntaru}} + 1 \leq 2,
		\end{align}
		where the second step follows from~\eqnref{ftdip_ineq4}. Further, using the constraint $\Qdip^\top\SigmaXtar\betasrconedip=\Qdip^\top\SigmaXtar\betatrue$, applying the triangle inequality and the sub-multiplicative property of the operator norm multiple times, we have
		\begin{align*}
			&\underbrace{\vecnorm{\SigmaXtar{}^{-1}\Qdip \Qdip^\top \SigmaXtar \betasrconedip - \SigmaXhtar{}^{-1}\Qhdip \Qhdip^\top \SigmaXhtar \betasrconediph}{2}}_{\eqcolon (D_1)} \\
			&\leq \vecopnorm{\SigmaXtar{}^{-1}\Qdip - \SigmaXhtar{}^{-1}\Qhdip} \vecnorm{\Qdip^\top \SigmaXtar\betatrue}{2} 
			\\&\hspace{.3in}+ \vecopnorm{\SigmaXhtar{}^{-1}\Qhdip} \vecopnorm{\SigmaXtar \Qdip - \SigmaXhtar\Qhdip}\vecnorm{\betasrconedip}{2} 
			+ \vecopnorm{\SigmaXhtar{}^{-1}\Qhdip} \vecopnorm{ \SigmaXhtar\Qhdip}\vecnorm{\betasrconedip-\betasrconediph}{2}.
		\end{align*}
				Under $\Eset$, this yields, with $\eta_1$ in~\eqnref{eta1_2_def} and $\teta_1, \teta_2$ in~\eqnref{teta1_2_def},
		\begin{align}\label{eqn:ft_dip_continue_6}
			(D_1)\leq 4\kappabar\eta_1\sqrt{\frac{d\log(1/\delta)}{\nsrcone}}+ c_6 \teta_1\sqrt{\frac{\ddel}{\nsrconeu}} +c_6 \teta_2 \sqrt{\frac{\ddel}{\ntaru}}.
		\end{align}
		Combining~\eqnref{ft_dip_EXY},~\eqnref{ft_dip_continue_1},~\eqnref{ft_dip_continue_2},~\eqnref{ft_dip_continue_7},~\eqnref{ft_dip_continue_6}, and using $\kappabar\geq 1$, we obtain
		\begin{align}\label{eqn:ft_dip_continue_4}
			\vecnorm{\Delta b}{2} \leq  8\eta_1\kappabar\sqrt{\frac{d\log (1/\delta)}{\nsrcone}} + c_7 \teta_1\sqrt{\frac{\ddel}{\nsrconeu}} + c_7 \teta_2\sqrt{\frac{\ddel}{\ntaru}},
		\end{align}
		where with $\eta_2$ in~\eqnref{eta1_2_def}
		\begin{equation}\label{eqn:teta1_2_def}
			\begin{aligned}
				&\teta_1 \coloneqq \kappabar\left[\tsigmagap\vecnorm{\betatrue}{2} + \tsigmagap \vecnorm{\betasrconedip}{2}+ \eta_2\right], \text{ and }\\
				&\teta_2 \coloneqq \kappabar\left[(\tsigmamin+ \tsigmagap)\vecnorm{\betatrue}{2} + (\tsigmamin+\tsigmagap) \vecnorm{\betasrconedip}{2}+ \eta_2\right]. 
			\end{aligned}
		\end{equation}
		To bound $\Delta A$, applying triangle inequality and the sub-multiplicative property of the operator norm repeatedly, we obtain 
		\begin{align*}
		\vecopnorm{\Delta A} 
			&\leq \vecopnorm{\Astar - \Vhdip^\top \SigmaXhtar{}^{-1}\Vdip } +  \vecopnorm{ \Vhdip^\top \SigmaXhtar{}^{-1}\Vdip - \Atarcon } \nonumber\\
			&\leq \vecopnorm{\SigmaXtar{}^{-1}\Vdip - \SigmaXhtar{}^{-1}\Vhdip}\vecopnorm{\Vdip} + \vecopnorm{\SigmaXhtar{}^{-1}\Vhdip}\vecopnorm{\ident - \SigmaXtar\SigmaXhtar{}^{-1}}  \vecopnorm{\Vdip} \\
			&\hspace{.9in} + \vecopnorm{\SigmaXhtar{}^{-1}\Vhdip}\vecopnorm{\SigmaXtar\SigmaXhtar{}^{-1}}  \vecopnorm{\Vdip - \Vhdip}.
		\end{align*}
		Under $\Eset$, and using~\eqnref{ftdip_ineq4},~\eqnref{ft_dip_continue_7}, this yields the bound
		\begin{align}\label{eqn:ft_dip_continue_5}
			\vecopnorm{\Delta A} \leq 6c_1 \lamminbar^{-1}\tsigmagap \sqrt{\frac{\ddel}{\nsrconeu}} +    6c_1\lamminbar^{-1}(\tsigmamin + \tsigmagap)\sqrt{\frac{\ddel}{\ntaru}}.
		\end{align}
				
		Now in order to bound $\alphatrue - \alphatarcon$, we use~\eqnref{ft_dip_decom_2} together with \eqnref{ft_dip_continue_4},~\eqnref{ft_dip_continue_5},
		\begin{align*}
       \lammin(\Atarcon) \vecnorm{\alphatrue - \alphatarcon}{2}
			&\leq \vecnorm{\Atarcon( \alphatrue - \alphatarcon)}{2}  \\
			&\leq \vecnorm{\Delta b}{2} + \vecopnorm{\Delta A}\vecnorm{\alphatrue}{2} \\
			&\leq \vecnorm{\Delta b}{2} + \lammaxbar\vecopnorm{\Delta A}\vecnorm{\betatrue}{2},
    \end{align*}
		where the last step uses $\vecnorm{\alphatrue}{2} =\vecnorm{\Vdip^\top \SigmaXtar\betatrue}{2} \leq \lammaxbar\vecnorm{\betatrue}{2}$.
		Additionally, applying the inequality \(\sigma_{\textnormal{min}}(AB) \geq \sigma_{\textnormal{min}}(A)\sigma_{\textnormal{min}}(B)\) for matrices $A,B$ with $B$ full column rank, we have
    \begin{align}\label{eqn:Atarcon_bound}
      \lammin(\Atarcon) \geq \sigma_{\textnormal{min}}^2(\Vh)\lammin\left(\SigmaXhtar{}^{-1}\SigmaXtar\SigmaXhtar{}^{-1}\right)=\lammin\left(\SigmaXhtar{}^{-1}\SigmaXtar\SigmaXhtar{}^{-1}\right).
    \end{align}
		We further observe
		\begin{align*}
			\lammin^{-1}\left(\SigmaXhtar{}^{-1}\SigmaXtar\SigmaXhtar{}^{-1}\right) &= \vecopnorm{\SigmaXhtar\SigmaXtar{}^{-1}\SigmaXhtar} \\
      &= \vecopnorm{\SigmaXhtar\SigmaXtar{}^{-1} (\SigmaXhtar  - \SigmaXtar{}) + \SigmaXhtar} \\
			&\overset{(i)}{\leq} 2\lammaxbar\left(c_1\lamminbar^{-1}\sigma_X^2\sqrt{\frac{\ddel}{\ntaru}} + 1\right) \overset{(ii)}{\leq} 4\lammaxbar,
		\end{align*}
		where step $(i)$ follows from the concentration of empirical covariance matrix under $\Eset$, and step $(ii)$ follows from the assumption $\ntaru\geq 4c_1^2\tsigmamin^2 \ddel$. 

		Combining the above and simplifying, we conclude that
		\begin{align}\label{eqn:ft_dip_alpha_diff}
			\vecnorm{\alphatrue-\alphatarcon}{2} \leq 32\lammaxbar\kappabar\eta_1\sqrt{\frac{d\log (1/\delta)}{\nsrcone}} + 4c_8\lammaxbar\teta_1\sqrt{\frac{\ddel}{\nsrconeu}} +4c_8\lammaxbar \teta_2\sqrt{\frac{\ddel}{\ntaru}}.
		\end{align}
		
		\paragraph*{(c) Error in $Q$ directions and final bound of $T_1$}
		Finally, combining~\eqnref{ft_dip_betastar} and~\eqnref{ft_dip_betatarcon}, and proceeding similarly to the preceding derivations, we can decompose $T_1$ as
		\begin{align*}
			&T_1 =	\vecnorm{\betatrue - \betatarcon }{2}  \\
			&\leq \underbrace{\vecnorm{\SigmaXtar{}^{-1} \Qdip\Qdip^\top \SigmaXtar\betasrconedip - \SigmaXhtar{}^{-1}\Qhdip\Qhdip^\top \SigmaXhtar\betasrconediph }{2}}_{= (D_1)} + \underbrace{\vecnorm{\SigmaXtar{}^{-1}\Vdip\alphatrue - \SigmaXhtar{}^{-1} \Vhdip \alphatarcon}{2}}_{\eqqcolon (D_2)}.
		\end{align*}
		We recognize the bound of $(D_1)$ from~\eqnref{ft_dip_continue_6}. To bound $(D_2)$, after applying triangle inequality, observe that everything can be bounded with inequalities in $\Eset$ and \eqnref{ft_dip_alpha_diff},
		\begin{align*} 
			(D_2)
			&\leq\vecnorm{\SigmaXtar{}^{-1}\Vdip\alphatrue - \SigmaXhtar{}^{-1} \Vhdip \alphatrue}{2} + \vecnorm{ \SigmaXhtar{}^{-1} \Vhdip \alphatrue - \SigmaXhtar{}^{-1} \Vhdip \alphatarcon}{2}\\
			&\leq \vecnorm{\SigmaXtar{}^{-1}\Vdip - \SigmaXhtar{}^{-1} \Vhdip}{2}\vecnorm{\alphatrue}{2} + \vecopnorm{ \SigmaXhtar{}^{-1} \Vhdip}\vecnorm{  \alphatrue - \alphatarcon}{2}.
		\end{align*}
    Putting these bounds together and simplifying, we conclude, with $\eta_1$ in~\eqnref{eta1_2_def} and $\teta_1, \teta_2$ in~\eqnref{teta1_2_def},
		\begin{align}\label{eqn:ft_dip_T1_bound}
			T_1   &\leq c_9\kappabar^2\eta_1\sqrt{\frac{d\log (1/\delta)}{\nsrcone}} +c_9\kappabar \teta_1 \sqrt{\frac{\ddel}{\nsrconeu}} + c_9\kappabar \teta_2 \sqrt{\frac{\ddel}{\ntaru}}. 
		\end{align}
		
		\subsubsection{Bounding $T_2$}
		Bounding $T_2$ in~\eqnref{ft_dip_decompose} is about bounding finite-sample risk for constrained least squares. We introduce the shorthand $\Delta=\betatarcon-\betahftdip$. The main steps are outlined as follows.
		\begin{enumerate}[(a)]
			\item (Feasibility and basic inequality) First, after checking the feasibility of $\betatarcon$ in~\eqnref{dip_var_matching_finite}, we derive a basic inequality using the optimality of $\betahftdip$. The basic inequality contains a quadratic term in $\Delta$ (denoted $B_1$) and a linear term in $\Delta$ (denoted $B_2$).
			\item ($B_2$ bound) Second, we bound $B_2$ using sub-exponential concentration and the fact that $\Delta$ lies in an $\rdip$-dimensional subspace.
			\item ($B_1$ bound) Third, we lower bound $B_1$ using sub-Gaussian concentration and a restricted eigenvalue condition. Then we conclude.
		\end{enumerate}
		
		\paragraph*{(a) Feasibility and basic inequality}
		We proceed by conditioning on $\Qhdip,\SigmaXhtar,\betasrconediph$ and by focusing on the randomness over the labeled target dataset $\Dtar$. To check the feasibility of $\betatarcon$ in Equation~\eqnref{dip_var_matching_finite}, note from~\eqnref{ft_dip_betatarcon},~\eqnref{ft_dip_lin_system_finite_sample} that
		\begin{align*}
			\vecnorm{\Vh^\top\SigmaXhtar \betatarcon}{2} = \vecnorm{\alphatarcon}{2} = \vecnorm{\Atarcon^{-1}\btarcon}{2} \leq 4\lammaxbar\vecnorm{\btarcon}{2},
		\end{align*}
		where the last step follows from~\eqnref{Atarcon_bound}. Furthermore, we observe
		\begin{align*}
			\vecnorm{\btarcon}{2} &= \vecnorm{\Vdip^\top\SigmaXhtar{}^{-1}\SigmaXtar\left(\betatrue -  \SigmaXhtar{}^{-1}\Qhdip\Qhdip^\top\SigmaXhtar \betasrconediph\right)}{2} \\
			&\overset{(i)}{\leq} \vecopnorm{\Vhdip^\top \SigmaXhtar{}^{-1}\SigmaXtar}\left(\vecnorm{\betatrue}{2} +\vecnorm{\SigmaXhtar{}^{-1}\Qhdip\Qhdip^\top\SigmaXhtar \betasrconediph}{2} \right)\\
			&\overset{(ii)}{\leq} 2\left(\vecnorm{\betatrue}{2} + 4\kappabar\vecnorm{\betasrconediph}{2}\right)\\
			 &\overset{(iii)}{\leq} 2\left(\vecnorm{\betatrue}{2} + 8\kappabar\sqrt{\kappabar}\vecnorm{\betahsrcone}{2}\right) , 
		\end{align*}
		where (i) follows from triangle inequality together with $\vecnorm{Ax}{2}\leq\vecopnorm{A}\vecnorm{x}{2}$; (ii) uses~\eqnref{ft_dip_continue_7} and the bounds on $\Eset$; and (iii) follows from the closed form expression for $\betasrconediph$ (c.f. see Equation~\eqnref{dip_uhat}) and the fact that the projection operator has unit operator norm. Under event~$\Eset_3$, we also have
		\begin{align*}
			\vecnorm{\betahsrcone}{2} \leq \vecnorm{\betahsrcone-\betasrcone}{2} + \vecnorm{\betasrcone}{2}&\leq c_3\tsigmamin \left(\vecnorm{\betasrcone}{2}+\frac{\sigma_Y}{\sigma_X}\right)\sqrt{\frac{d\log(1/\delta)}{\nsrcone}} + \vecnorm{\betasrcone}{2} \\
			&\leq 2\vecnorm{\betasrcone}{2}+\frac{\sigma_Y}{\sigma_X},
		\end{align*}
		under the assumption $\nsrcone\geq c_3^2\tsigmamin^2\cdot d\log(1/\delta)$. Therefore, as long as 
    \begin{align}\label{eqn:rho_condition}
      \varrho\geq 64\lammaxbar\left(\vecnorm{\betatrue}{2} + \kappabar\sqrt{\kappabar}\vecnorm{\betasrcone}{2} +\kappabar\sqrt{\kappabar}\frac{\sigma_Y}{\sigma_X} \right),
    \end{align}
    $\betatarcon$ satisfies the constraint in~\eqnref{dip_var_matching_finite}. By the optimality of $\betahftdip$, we have
		\begin{align*}
			\frac{1}{\ntar}\vecnorm{\bYtar - \bXtar \betahftdip}{2}^2 \leq \frac{1}{\ntar}\vecnorm{\bYtar- \bXtar \betatarcon}{2}^2.
		\end{align*}	
		Rearranging the terms, we arrive at the following basic inequality
		\begin{align}\label{eqn:dip_finetuning_basic_inequality_1}
			\underbrace{\frac{1}{\ntar}\vecnorm{\bXtar\Delta}{2}^2}_{\eqqcolon (B_1)}\leq   \underbrace{\frac{2}{\ntar}\inner{\bXtar{}^\top ( \bXtar\betatarcon - \bYtar)}{\Delta}}_{\eqqcolon (B_2)}.
		\end{align}		
		
		Before bounding $(B_1)$ and $(B_2)$, we make a preliminary observation. Given the constraint $\Qhdip^\top \SigmaXhtar \betasrconediph = \Qhdip^\top \SigmaXhtar \betahftdip$, we decompose $\betahftdip$ analogously to~\eqnref{ft_dip_betatarcon}:
		\begin{align}\label{eqn:ft_dip_betah}
			\betahftdip &=  \SigmaXhtar{}^{-1} \Qhdip\Qhdip^\top \SigmaXhtar \betasrconediph +  \SigmaXhtar{}^{-1}\Vhdip \underbrace{\Vhdip^\top \SigmaXhtar\betahftdip}_{\coloneqq \alphahdip}.
		\end{align}
		Combining~\eqnref{ft_dip_betatarcon},~\eqnref{ft_dip_betah}, we derive an explicit expression for $\Delta$:
		\begin{equation}\label{eqn:ft_dip_delta}
			\Delta = \betatarcon - \betahftdip = \SigmaXhtar{}^{-1}\Vhdip (\alphatarcon - \alphahdip).
		\end{equation}
		\paragraph*{(b) $(B_2)$ bound} Using notation in~\eqnref{adj_resp_proj_covariate}, observe that the residual can be written as $\Xtar_i{}^\top \betatarcon - \Ytar_i=\Wtar_i{}^\top \alphatarcon - \Ztar_i$. Using this expression, we bound $(B_2)$ as follows:
		\begin{align*}
			&\frac{2}{\ntar}\inner{\bXtar{}^\top ( \bXtar\betatarcon - \bYtar)}{\Delta} \\
			&= \frac{2}{\ntar}\inner{\bXtar{}^\top ( \bWtar\alphatarcon - \bZtar)}{\SigmaXhtar{}^{-1}\Vhdip(\alphatarcon - \alphahdip)}\\
			&= \frac{2}{\ntar}\inner{\bWtar{}^\top ( \bWtar\alphatarcon - \bZtar)}{\alphatarcon - \alphahdip}\\
			&\overset{(i)}{=}\frac{2}{\ntar}\inner{\bWtar{}^\top ( \bWtar\alphatarcon - \bZtar) - \EE{\bWtar{}^\top ( \bWtar\alphatarcon - \bZtar)}}{\alphatarcon - \alphahdip}\\
			&\overset{(ii)}{\leq} \frac{2}{\ntar}\vecnorm{\bWtar{}^\top ( \bWtar\alphatarcon - \bZtar) - \EE{\bWtar{}^\top ( \bWtar\alphatarcon - \bZtar)}}{2}\vecnorm{\alphatarcon - \alphahdip}{2}.
		\end{align*}
    Here step $(i)$ uses the fact that $\EE{\Wtar_1\Wtar_i{}^\top}\alphatarcon = \EE{\Wtar_1\Ztar_1}$ which follows from~\eqnref{ft_dip_lin_system_finite_sample}, and step $(ii)$ follows from Cauchy-Schwarz inequality. 
    
    The first term is further bounded under event~$\Eset_4$ given in~\eqnref{event_thm41_4}. Applying~\lemref{sub_gaussian} and the inequalities under $\Eset$, we bound $\sigma_W\sigma_Z$ as
		\begin{align*}
			\sigma_W\sigma_Z \leq \sigma_X\vecopnorm{\SigmaXhtar{}^{-1}\Vhdip}\left(\sigma_X\vecopnorm{\SigmaXhtar{}^{-1}\Vhdip} \vecnorm{\alphatarcon}{2} + \sigma_Y\right) \leq 2\sigma_X^2\lamminbar^{-1}\left(2\varrho\lamminbar^{-1}+\frac{\sigma_Y}{\sigma_X}\right).
		\end{align*}
		Substituting back, we arrive at the bound for $(B_2)$:
		\begin{align}\label{eqn:ft_dip_upperbound}
			&\frac{2}{\ntar}\inner{\bXtar{}^\top ( \bXtar\betatarcon - \bYtar)}{\Delta}   \\
			&\hspace{1.in} \leq 2c_{10}\tsigmamin\left(2\varrho\lamminbar^{-1}+\frac{\sigma_Y}{\sigma_X}\right)\sqrt{\frac{\rdip\log(1/\delta)}{\ntar}}\vecnorm{\alphatarcon - \alphahdip}{2}. \nonumber
		\end{align}		
		
		\paragraph*{(c) $(B_1)$ bound and final bound of $T_2$}
		Since both $\betahftdip$ and $\betatarcon$ are feasible, we have $\Delta\in \Lambda$ defined in~\eqnref{event_thm41_5_high_prob_set}. Under event~$\Eset_5$ given in~\eqnref{event_thm41_5}, we have
		\begin{align}\label{eqn:ft_dip_lowerbound0}
			(B_1)=\frac{1}{\ntar}\vecnorm{\bXtar \Delta}{2}^2 \geq  \frac{\lamminbar}{2}\vecnorm{\Delta}{2}^2 - \frac{c_5  \lamminbar^{-1}\sigma_X^4\kappabar\gw{\Lambda}^2  \log (1/\delta)}{\ntar}.
		\end{align}
		The following lemma provides an upper bound for \(\gw{\Lambda}\) under the event \(\Eset\). Its proof is deferred to Appendix~\ref{app:proof_of_ft_dip_gaussian_width}.
		\begin{lemma}\label{lem:ft_dip_gaussian_width}
			On the event $\Eset$, the Gaussian width of $\Lambda$ is bounded as
			\begin{align*}
				\gw{\Lambda} \leq 8\varrho \lamminbar^{-1}\sqrt{\rdip}.
			\end{align*}
		\end{lemma}
		Substituting this bound into~\eqnref{ft_dip_lowerbound0}, yields
		\begin{align}\label{eqn:ft_dip_lowerbound}
			\frac{1}{\ntar}\vecnorm{\bXtar \Delta}{2}^2 &\geq \frac{\lamminbar}{2}\vecnorm{\Delta}{2}^2 - 16c_5^2\frac{\lamminbar^{-3}\sigma_X^4\kappabar \log(1/\delta)}{\ntar}\varrho^2 \rdip.
		\end{align}
		Finally, we return to~\eqnref{dip_finetuning_basic_inequality_1} and plug in our results in~\eqnref{ft_dip_upperbound},~\eqnref{ft_dip_lowerbound}, to have
		\begin{align*}
			 &\frac{\lamminbar}{2}\vecnorm{\Delta}{2}^2- 16c_5^2\frac{\lamminbar^{-3}\sigma_X^4\kappabar \log(1/\delta)}{\ntar}\varrho^2 \rdip \\
			 &\hspace{.3in}\leq  2c_9\tsigmamin\left(2\varrho\lamminbar^{-1}+\frac{\sigma_Y}{\sigma_X}\right)\sqrt{\frac{\rdip\log(1/\delta)}{\ntar}}\vecnorm{\alphatarcon - \alphahdip}{2} \nonumber \\
			 &\hspace{.3in}\leq 4c_9\lammaxbar\tsigmamin\left(2\varrho\lamminbar^{-1}+\frac{\sigma_Y}{\sigma_X}\right)\sqrt{\frac{\rdip\log(1/\delta)}{\ntar}}\vecnorm{\Delta}{2}, \nonumber
		\end{align*}
		where the last step uses $\alphatarcon - \alphahdip = \Vhdip^\top \SigmaXhtar \Delta$ from~\eqref{eqn:ft_dip_delta} and the bound $\opnorm{\SigmaXhtar}\leq 2\lammaxbar$ on the event $\Eset$.
		Rearranging terms, using $\sqrt{a+b}\leq \sqrt{a}+\sqrt{b}$ for $a,b\geq 0$, $\tsigmamin=\sigma_X^2\lamminbar^{-1}$, $\kappabar\geq 1$, and simplifying, we can deduce
		\begin{align}\label{eqn:ft_dip_T2_bound}
			T_2=\vecnorm{\Delta}{2} \leq c'\kappabar\tsigmamin\left[\varrho\lamminbar^{-1}+\frac{\sigma_Y}{\sigma_X}  \right] \sqrt{\frac{\rdip\log(1/\delta)}{\ntar}} .
		\end{align}
		
		\subsubsection{Combining $T_1$ and $T_2$ to conclude}
		Finally, combining~\eqnref{ft_dip_T1_bound} and~\eqnref{ft_dip_T2_bound}, we conclude that with probability at least $1-10\delta $ over the randomness of all datasets $\Dsrcone,\Dsrconeu,\Dtar,\Dtaru$, the following bound holds
		\begin{multline*}
			\vecnorm{\betatrue-\betahftdip}{2} \leq T_1 + T_2 \leq c' \kappabar\tsigmamin\left[\varrho\lamminbar^{-1}+\frac{\sigma_Y}{\sigma_X} \right] \sqrt{\frac{\rdip\log(1/\delta)}{\ntar}} \\
			+ c_9\kappabar^2\eta_1\sqrt{\frac{d\log (1/\delta)}{\nsrcone}} +c_9\kappabar^2\left[\tsigmagap\vecnorm{\betatrue}{2} + \tsigmagap \vecnorm{\betasrconedip}{2}+ \eta_2\right]\sqrt{\frac{\ddel}{\nsrconeu}} \\
			+ c_9\kappabar^2\left[(\tsigmamin+ \tsigmagap)\vecnorm{\betatrue}{2} + (\tsigmamin+\tsigmagap) \vecnorm{\betasrconedip}{2}+ \eta_2\right]\sqrt{\frac{\ddel}{\ntaru}}. 
		\end{multline*}
		The excess risk is then bounded as 
		\begin{align*}
			\delta\risktar{f_{\betahftdip}} =(\betahftdip-\betatrue)^\top \SigmaXtar (\betahftdip -\betatrue) \leq \lammaxbar \vecnorm{\betatrue-\betahftdip}{2}^2.
		\end{align*}
		This completes the proof of~\thmref{dip_var_matching_finite}.
		
		\subsection{Proof of~\thmref{connectivity_estimator_finite}}\label{app:proof_connectivity_theorem}
		We introduce the shorthand notation $\Delta=\betatrue - \betahftconn$. To deal with the $\ell_1$ penalty in FT-OLS-Src~\eqnref{connectivity_finite}, our proof strategy to bound $\norm{\Delta}_2$ follows closely the literature on sparse linear regression (see e.g., Chapter 7 of~\cite{wainwright2019high}). The main additional technicality in our proof is due to the fact that we only know $(\betatrue - \betasrcone)$ is sparse in population quantities, however, the $\ell_1$ penalty in~\eqnref{connectivity_finite} is formulated with respect to an empirical estimate $\betahsrcone$. Consequently, we need to make use of the source data samples with size $\nsrcone$ to accurately estimate  $\betahsrcone$, and account for its finite-sample uncertainty when deriving the final risk bound. 
		
		Based on the above intuition, in the regime where $\nsrcone$ is large, we expect the risk bound which scales as ${\rconn}/{\ntar}$ where $\rconn$ is the sparsity level and $\ntar$ is the target sample size, as explained in~\secref{fine_tuning_connectivity_shift}. The proof is organized as follows.
		\begin{enumerate}[(a)]
			\item (Feasibility and basic inequality) First, after verifying the feasibility of $\betatrue$ in~\eqnref{connectivity_finite}, we obtain a basic inequality using the optimality of $\betahftconn$. The basic inequality contains a quadratic term in $\Delta$ (denoted $T_1$) and a linear term in $\Delta$ (denoted $T_2$). 
			\item (Upper bounding $T_2$) Second, we bound $T_2$ using sub-exponential concentration and the sparsity assumption.
			\item (Lower bounding $T_1$) Third, we bound $T_1$ using sub-Gaussian concentration and a restricted eigenvalue condition. 
			\item (Combining) Finally, we arrive at an upper bound on $\vecnorm{\Delta}{2}$ after combining above bounds and rearranging terms in the basic inequality. 
		\end{enumerate}
    Before we proceed, we define high probability events under which empirical 
    estimates concentrate. Define
    \begin{align} \label{eqn:event_thm43_1}
      \Eset_1 \defn \Bigg\{\frac{1}{\ntar}\vecnorm{\bXtar{}^\top ( \bXtar\betatrue - \bYtar) - \EE{\bXtar{}^\top ( \bXtar\betatrue - \bYtar)}}{\infty} \\ \leq c_1 \lamminbar \eta_{\star} \parenth{\sqrt{\frac{\log d + \log (1/\delta)}{\ntar}} } \Bigg\},\nonumber
    \end{align}
    where $\eta_{\star} \defn \tsigmamin \parenth{\frac{\sigma_Y}{\sigma_X} + \vecnorm{\betastar}{2}}$.
    By applying~\lemref{l_infty_norm_bound} for concentration of product of sub-Gaussian vectors around its mean, and using the fact that $\Xtar{}^\top \betatrue - \Ytar$ is sub-Gaussian with parameter $\vecnorm{\betatrue}{2}\sigma_X+\sigma_Y$ by~\lemref{sub_gaussian}, with $\ntar \geq \log d + \log (1/\delta)$, the event $\Eset_1$ holds with probability at least $1-\delta$ over the randomness of $\Dtar$.

    Define
    \begin{align} \label{eqn:event_thm43_2}
      \Eset_2 \defn \braces{\vecnorm{\betasrcone - \betahsrcone}{2} \leq c_2 \eta_{\text{LS}} \sqrt{\frac{d\log(1/\delta)}{\nsrcone}}},
    \end{align}
		where $\eta_{\text{LS}} \defn \tsigmamin \parenth{\frac{\sigma_Y}{\sigma_X} + \vecnorm{\betasrcone}{2}}$. By~\propref{least_squares_bound} with $M=1$ and the assumption $\nsrcone\geq c_1 \max\{\ddel \tsigmamin^2  ,d\log (1/\delta)\}$, the event $\Eset_2$ holds with probability at least $1-2\delta$ over the randomness of $\Dsrcone$. 

    Define the set
		\begin{align}\label{eqn:event_thm43_3_high_prob_set}
			\Lambda \defn \braces{x: \vecnorm{x}{1} \leq 2\sqrt{\rconn}\vecnorm{x}{2} +  2 c_2 \eta_{\text{LS}} d \sqrt{\frac{\log (1/\delta)}{\nsrcone}} } \cap \braces{x: \vecnorm{x}{2}\leq 2\rho },
		\end{align}
		and the event
    \begin{align} \label{eqn:event_thm43_3}
      \Eset_3 \defn \braces{\frac{1}{\ntar}\vecnorm{\bXtar x}{2}^2 \geq \frac{\lamminbar}{2}\vecnorm{x}{2}^2 - \frac{c'^2  \lamminbar \tsigmamin^2 \kappabar\gw{\Lambda}^2  \log (1/\delta)}{\ntar},\quad \forall x \in \Lambda}.
    \end{align}
    Since $\Lambda$ contains origin, applying~\lemref{restricted_eigenvalue}, the event $\Eset_3$ holds with probability at least $1-2\delta$ over the randomness of $\Dtar$. 

    Finally, define the intersection of the events
    \[\Eset \defn \Eset_1 \cap \Eset_2 \cap \Eset_3. \]
	By the probability bounds above and a union bound, the event $E$ holds with probability at least $1-5\delta$.	
	In the following, we prove our result under the event $\Eset$. 
		\paragraph*{(a) Feasibility and basic inequality} We begin by noting that $\betatrue$ satisfies the constraint in Equation~\eqnref{connectivity_finite} by construction: we choose \(\rho \geq \vecnorm{\betatrue}{2}\) and \(\gamma = \|\betatrue - \betahsrcone\|_1\). By the optimality of \(\betahftconn\), it follows that
		\[
		\frac{1}{\ntar}\|\bYtar - \bXtar \betahftconn\|_2^2 \leq \frac{1}{\ntar}\|\bYtar - \bXtar \betatrue\|_2^2.
		\]
		Rearranging the terms, we arrive at the following basic inequality
		\begin{align}\label{eqn:basic_inequality}
			\underbrace{\frac{1}{\ntar}\vecnorm{\bXtar\Delta}{2}^2 }_{\coloneqq T_1} &\leq  \frac{2}{\ntar}\inner{\bXtar{}^\top ( \bXtar\betatrue - \bYtar)}{\Delta} \\
			&\overset{(i)}{=}  \underbrace{\frac{2}{\ntar}\inner{\bXtar{}^\top ( \bXtar\betatrue - \bYtar) - \EE{\bXtar{}^\top ( \bXtar\betatrue - \bYtar)}}{ \Delta}}_{\coloneqq T_2},\nonumber
		\end{align}		
		where step $(i)$ holds because \(\EE{\bXtar{}^\top \bXtar \betatrue} = \EE{\bXtar{}^\top \bYtar}\), by the definition of \(\betatrue\). It remains to bound $T_1$ and $T_2$ to arrive at a bound on $\vecnorm{\Delta}{2}$.
		
		\paragraph*{(b) Upper-bounding $T_2$} We apply H\"older's inequality to obtain
		\begin{align}\label{eqn:connectivity_T2}
			T_2 &\leq \frac{2}{\ntar}\vecnorm{\bXtar{}^\top ( \bXtar\betatrue - \bYtar) - \EE{\bXtar{}^\top ( \bXtar\betatrue - \bYtar)}}{\infty}\vecnorm{\Delta}{1}.
		\end{align}
		To relate $\vecnorm{\Delta}{1}$ with $\vecnorm{\Delta}{2}$, we make use of the sparsity assumption. Let $S$ denote the support of $\betatrue - \betasrcone$ as defined in~\assumpref{connectivity_shift_linear_SCM}. Using the feasibility of $\betahftconn$, we have
		\begin{align*}
			\vecnorm{\betatrue-\betahsrcone}{1} &\geq \vecnorm{\betatrue-\betahsrcone - \Delta}{1} \\
			&= \vecnorm{(\betatrue-\betahsrcone - \Delta)_S}{1} + \vecnorm{(\betatrue-\betahsrcone - \Delta)_{S^c}}{1} \\
			&\overset{(i)}{\geq} \vecnorm{(\betatrue-\betahsrcone)_S}{1} - \vecnorm{\Delta_S}{1} + \vecnorm{\Delta_{S^c}}{1} -  \vecnorm{(\betatrue-\betahsrcone)_{S^c}}{1},
		\end{align*}
		where step $(i)$ uses the triangle inequality twice. Rearranging and simplifying, we obtain
		\begin{align*}
			\vecnorm{\Delta_{S^c}}{1} \leq  \vecnorm{\Delta_S}{1} + 2\vecnorm{(\betatrue-\betahsrcone)_{S^c}}{1} &=  \vecnorm{\Delta_S}{1} + 2\vecnorm{(\betatrue- \betasrcone  + \betasrcone  - \betahsrcone)_{S^c}}{1} \nonumber\\
			&\overset{(ii)}{=} \vecnorm{\Delta_S}{1} + 2\vecnorm{(\betasrcone  - \betahsrcone)_{S^c}}{1},
		\end{align*}
		where step $(ii)$ holds because $(\betatrue - \betasrcone)_{S^c}=0$. Using the fact that $|S|=\rconn$, we have
		\begin{align}\label{eqn:connectivity_ell1_bound}
			\vecnorm{\Delta}{1} = \vecnorm{\Delta_S}{1} + \vecnorm{\Delta_{S^c}}{1} &\leq 2\vecnorm{\Delta_S}{1} + 2\vecnorm{(\betasrcone  - \betahsrcone)_{S^c}}{1} \\ 
			&\leq  2\sqrt{\rconn}\vecnorm{\Delta}{2}  + 2\sqrt{d-\rconn}\vecnorm{\betasrcone - \betahsrcone}{2} \nonumber\\ 
      &\overset{(i)}{\leq} 2\sqrt{\rconn}\vecnorm{\Delta}{2} +  2 c_2 \eta_{\text{LS}}  \sqrt{\frac{d^2\log(1/\delta)}{\nsrcone}}, \nonumber
		\end{align}
    where step (i) follows under event $\Eset_2$ given in~\eqnref{event_thm43_2}. 
		Returning to~\eqnref{connectivity_T2}, under event $\Eset_1 \cap \Eset_2$, we have
		\begin{align}\label{eqn:connectivity_T2_2}
			T_2 \leq c_3 \lamminbar \eta_{\star} \parenth{\sqrt{\frac{\log d + \log (1/\delta)}{\ntar}} } \brackets{\sqrt{\rconn}\vecnorm{\Delta}{2} +  \eta_{\text{LS}}  \sqrt{\frac{d^2\log(1/\delta)}{\nsrcone}} }.
		\end{align}

		\paragraph*{(c) Lower-bounding $T_1$} Next, we derive a lower bound on $T_1$. Due to the feasibility of $\betahftconn$ and $\betatrue$, we have $\vecnorm{\Delta}{2}\leq 2\rho$. Hence, on the event where the inequality~\eqnref{connectivity_ell1_bound} holds, we have $\Delta\in \Lambda$ where $\Lambda$ is defined in~\eqnref{event_thm43_3_high_prob_set}. Under $\Eset_2 \cap \Eset_3$, we have
		\begin{align}\label{eqn:connectivity_re}
			T_1 \geq \frac{\lamminbar}{2}\vecnorm{\Delta}{2}^2 - \frac{c'^2  \lamminbar \tsigmamin^2 \kappabar\gw{\Lambda}^2  \log (1/\delta)}{\ntar}.
		\end{align}
    Furthermore, the Gaussian width of $\Lambda$ is bounded as
			\begin{align}\label{eqn:connectivity_gaussian_width}
				\gw{\Lambda} \leq c_4 \brackets{\rho\sqrt{\rconn\log d} +  \eta_{\text{LS}} \sqrt{\frac{d^2 \log(d) \log(1/\delta)}{\nsrcone}}}.
			\end{align}
    This is because for $g \sim \Normal(0, \Ind_n)$, $
			\gw{\Lambda} = \EE{\sup_{x \in \Lambda} \inner{g}{x}} \leq  \EE{\sup_{x \in \Lambda}\vecnorm{x}{1}\vecnorm{g}{\infty}}$.
		By~\cite[Exercise 2.5.10]{vershynin2018high}, $\EE{\vecnorm{g}{\infty} }\leq c\sqrt{\log d}$. 
		Plugging \eqnref{connectivity_gaussian_width} into~\eqnref{connectivity_re}, and using $(a+b)^2\leq 2a^2 + 2b^2$ and simplifying, we obtain
		\begin{align}\label{eqn:connectivity_lower_bound}
			T_1 \geq  \frac{\lamminbar}{2}\vecnorm{\Delta}{2}^2  - c_5 \lamminbar \tsigmamin^2  \kappabar \parenth{\frac{\rho^2 \rconn \log d}{\ntar} + \eta_{\text{LS}}^2\frac{d^2\log(d) \log(1/\delta)}{\ntar\nsrcone}} \log(1/\delta).
		\end{align}
		
		\paragraph*{(d) Combining and conclude}
		Finally, combining~\eqnref{basic_inequality},~\eqnref{connectivity_T2_2},~\eqnref{connectivity_lower_bound}, and rearranging, leads to
		\begin{multline*}
			\frac{1}{2}\vecnorm{\Delta}{2}^2  \leq c_3  \eta_{\star} \parenth{\sqrt{\frac{\log d + \log (1/\delta)}{\ntar}} } \brackets{\sqrt{\rconn}\vecnorm{\Delta}{2} +  \eta_{\text{LS}}  \sqrt{\frac{d^2\log(1/\delta)}{\nsrcone}} }  \\
      +  c_5  \tsigmamin^2  \kappabar \parenth{\frac{\rho^2 \rconn \log d}{\ntar} + \eta_{\text{LS}}^2\frac{d^2\log(d)\log(1/\delta)}{\ntar\nsrcone}}\log(1/\delta).
		\end{multline*} Using $\sqrt{a+b}\leq \sqrt{a}+\sqrt{b}$ for $a,b\geq 0$ together with $\log d \geq 1$, $\log(1/\delta)\geq 1$, we deduce that under $\Eset$,
		\begin{align*}
      \vecnorm{\Delta}{2}^2 &\leq c \brackets{ \parenth{\eta_{\star}^2 + \tsigmamin^2  \kappabar \rho^2 } \frac{\rconn \left(\log d+ \log (1/\delta)\right) }{\ntar}  \log (1/\delta) } +  \\
      &\quad \quad c\brackets{ \parenth{\eta_{\star} \eta_{\text{LS}} + \tsigmamin^2  \kappabar \cdot \eta_{\text{LS}}^2 } \max\braces{\frac{{d}}{\sqrt{\ntar\nsrcone}}, \frac{d^2 }{{\ntar\nsrcone}} } \log(d) \log^2 (1/\delta)  }.
		\end{align*}

		Given the bound on the squared $\ell_2$ norm of $\Delta$, the excess risk is bounded as
		\begin{align*}
			\delta\risktar{f_{\betahftconn}} &= (\betahftconn-\betatrue)^\top \SigmaXtar (\betahftconn-\betatrue) \leq \lammaxbar \vecnorm{\Delta}{2}^2.
		\end{align*}
		This completes the proof of~\thmref{connectivity_estimator_finite}.		
		
		\subsection{Proof of~\thmref{cip_mean_matching_finite}}\label{app:cip_theorem_proof}
		Since both FT-DIP$^{(1)}$ and FT-CIP are constrained least squares problems on the linear subspace, the proof of this theorem closely follows that of~\thmref{dip_var_matching_finite}. To avoid repeating ourselves, we only provide a sketch here. The main difference in FT-CIP lies in the constraint of~\eqnref{cip_flne_tuning} which is based on different finite-sample estimates---namely, $\Qhcipaug,\Vhcipaug,\SigmaXhsrcone$---whose concentration properties also differ.
		
		From~\eqnref{cip_alternative2}, we know $\Qcipaug^\top\SigmaXsrcone \betacip=\Qcipaug^\top\SigmaXsrcone\betatrue$. This implies that $\betatrue$ solves the following population-level constrained optimization problem
		\begin{equation}\label{eqn:fine_tune_cip_betatrue_constrained}
			\begin{aligned}
				&\betatrue= \argmin_{\beta} \EE{(\Ytar - \beta^\top \Xtar)^2}   \text{ s.t. }\Qcipaug^\top\SigmaXsrcone \betacip=\Qcipaug^\top\SigmaXsrcone \beta.
			\end{aligned}
		\end{equation}
		We also define an intermediate estimator $\betatarcon$ that uses the same population objective but uses empirical version of the constraints
		\begin{equation}\label{eqn:fine_tune_cip_betatarcon_constrained}
			\begin{aligned}
				&\betatarcon= \argmin_{\beta} \EE{(\Ytar - \beta^\top \Xtar)^2}  \text{ s.t. } \Qhcipaug^\top\SigmaXhsrcone \betaciph=\Qhcipaug^\top\SigmaXhsrcone \beta.
			\end{aligned}
		\end{equation}
		We decompose the error into two parts by writing $\betatrue - \betahftcip = (\betatrue - \betatarcon) + (\betatarcon - \betahftcip)$ and bound each term separately
		\begin{equation}\label{eqn:ft_cip_decompose}
			\vecnorm{\betatrue - \betahftcip}{2} \leq {\vecnorm{\betatrue-\betatarcon}{2}} + {\vecnorm{\betatarcon - \betahftcip}{2}}.
		\end{equation}		
		
		As in the proof of~\thmref{dip_var_matching_finite}, we first define high probability events under which empirical estimates from multiple source domains concentrate around their population quantities. In particular, let \(R_q \in \R^{(d-\rcip-1) \times (d-\rcip-1)}\) and \(R_v \in \R^{(\rcip+1) \times (\rcip+1)}\) be orthonormal matrices defined as
		\[
		R_q = \inf_{\widetilde{R}^\top\widetilde{R} = \ident} \vecopnorm{\Qhcipaug - \Qcipaug \widetilde{R}}, \quad \text{and} \quad R_v = \inf_{\widetilde{R}^\top\widetilde{R} = \ident} \vecopnorm{\Vhcipaug - \Vcipaug\widetilde{R}}.
		\]
		Define
		\begin{align*}
			\Eset_1 \defn \Bigg\{ \,
			& \vecopnorm{\SigmaXhsrcone - \SigmaXsrcone} \leq c_1\sigma_X^2\sqrt{\frac{\ddel}{\nsrconeu}};  \;\;\; \frac{\lamminbar}{2} \leq \lammin(\SigmaXhsrcone) \leq \lammax(\SigmaXhsrcone) \leq 2\lammaxbar;\\
			& \vecopnorm{\Qhcipaug - \Qcipaug R_q} \leq c_1 \tsigmagapcip \sqrt{\frac{\ddel}{\nbar}} ; \\  
			&\vecopnorm{\Vhcipaug - \Vcipaug R_v} \leq c_1 \tsigmagapcip \sqrt{\frac{\ddel}{\nbar}}
			\; \Bigg\}.
		\end{align*}
		By~\lemref{bounds_cov_matrices} and Equation~\eqnref{bounds_eig_values}, the first two bounds in $\Eset_1$ hold with probability at least $1- \delta$ over the randomness of $\Dsrconeu$, provided that $\nsrconeu \geq  c_2 \max\{1,\tsigmamin^2\} \ddel$. For the latter two bounds, we can follow the argument in~\lemref{cip_difference_eigenvectors} to show that they hold with probability at least $1- (2M+4) \delta$ provided that $\nbar\geq c_3\frac{\sigma_Y^4}{\nu_Y^4}\log (1/\delta)$ for some universal constant $c_3>0$.
		
		Under the event \(\Eset_1\), we can further establish the following bounds, which are analogous to~\eqnref{ftdip_ineq4}, \eqnref{ftdip_ineq1}, \eqnref{ftdip_ineq2}, \eqnref{ftdip_ineq3}, 
		\begin{align}\label{eqn:ftcip_ineq4}
			\vecopnorm{\ident - \SigmaXhsrcone{}^{-1} \SigmaXsrcone}\leq  2c_1 \tsigmamin\sqrt{\frac{\ddel}{\nsrconeu}};
		\end{align}
		\begin{align}\label{eqn:ftcip_ineq1}
			\vecopnorm{\SigmaXhsrcone{}^{-1} - \SigmaXsrcone{}^{-1}}\leq   2c_1 \lamminbar^{-1}\tsigmamin\sqrt{\frac{\ddel}{\nsrconeu}};
		\end{align}
		\begin{align}\label{eqn:ftcip_ineq2}
			\vecopnorm{\SigmaXhsrcone{}^{-1}\Vhcipaug - \SigmaXsrcone{}^{-1}\Vcipaug R_v}   \leq 2c_1\lamminbar^{-1} \tsigmagapcip\sqrt{\frac{\ddel}{\nbar}} +    2c_1\lamminbar^{-1} \tsigmamin \sqrt{\frac{\ddel}{\nsrconeu}};  
		\end{align}
		\begin{align}\label{eqn:ftcip_ineq3}
			\vecopnorm{\SigmaXhsrcone{}^{-1}\Qhcipaug - \SigmaXsrcone{}^{-1}\Qcipaug R_q}   \leq 2c_1\lamminbar^{-1} \tsigmagapcip\sqrt{\frac{\ddel}{\nbar}} +    2c_1\lamminbar^{-1} \tsigmamin \sqrt{\frac{\ddel}{\nsrconeu}}.
		\end{align}		
		Define the event $\Eset_2$ as
		\begin{align*}
			\Eset_2 \defn \left\{\vecnorm{\betaciph - \betacip}{2} \leq \eta_1\sqrt{\frac{d\log (1/\delta)}{\nbar}} + \eta_2 \sqrt{\frac{\ddel}{\nbar}}\right\},
		\end{align*}
		where $\eta_1= c_4\tsigmamin\left(\frac{\sigma_Y}{\sigma_X}+\sqrt{\kappabar}\vecnorm{\betapool}{2}\right)$ and $\eta_2= c_4\tsigmagapcip\kappabar^{3/2}\vecnorm{\betapool}{2}$. By~\propref{CIP_finite_sample_error_bound} and the assumption that $\nbar\geq c_2\max\{\tsigmamin^2 \ddel,\tsigmamin^2 d\log(1/\delta) , \frac{\sigma_Y^4}{\nu_Y^4}\log (1/\delta)\}$, the event $\Eset_2$ holds with probability at least $1-(4M+4)\delta$.
		
		The remaining high probability events can be defined analogously to those in~\thmref{dip_var_matching_finite}, and we denote their intersection by $\Eset$. 

		Then the event $\Eset$ depends only on the source datasets and is independent of the labeled target dataset. Conditioned on $\Eset$, the rest of the proof follows exactly as in~\thmref{dip_var_matching_finite} and uses the randomness of the labeled target dataset, so we omit the details.
		
		In particular, with probability at least $1- c_5 M \delta $ over the randomness of all datasets, the following bound holds
		\begin{align*}
			\vecnorm{\betatrue-\betahftcip}{2} 
			&\leq c' \kappabar\tsigmamin\left[\varrho\lamminbar^{-1}+\frac{\sigma_Y}{\sigma_X} \right] \sqrt{\frac{\rcip\log(1/\delta)}{\ntar}} \\
			&+c'\kappabar^2\left[\eta_1\sqrt{\log(1/\delta)}+\tsigmagapcip\vecnorm{\betatrue}{2} + \tsigmagapcip \vecnorm{\betacip}{2}+ \eta_2\right]\sqrt{\frac{\ddel}{\nbar}}  \\
			&+ c'\kappabar^2\tsigmamin\left[\vecnorm{\betatrue}{2} + \vecnorm{\betacip}{2}\right]\sqrt{\frac{\ddel}{\nsrconeu}}. 
		\end{align*}
		Translating this bound into excess risk concludes the proof of~\thmref{cip_mean_matching_finite}.

		\section{Proof of proposition and main theorems on risk lower bounds}

		In this section, we prove results related to minimax lower bounds, including~\propref{anticausal_minimax_lowerbound},~\thmref{dip_var_matching_finite_minimax},~\thmref{connectivity_estimator_finite_minimax}, and~\thmref{cip_mean_matching_finite_minimax}. We begin by recalling Fano's inequality, which we use to establish the minimax results.
		\begin{lemma}[Fano's inequality]\label{lem:fano_bound}
			Let $\mathfrak{B}$ be a class of distributions, and suppose there exists a finite subset $\Mset=\{\Pset_1,\ldots,\Pset_N \}\subseteq \mathfrak{B}$ such that $\{\beta(\Pset_1),\ldots,\beta(\Pset_N) \}\subseteq\R^{d}$ is a $2\delta$-separated set in $\ell_2$ norm. If for some universal constant $c>0$, the KL divergences and the number of elements $N$ satisfy
      \begin{align*}
        \kl{\Pset_j}{\Pset_k}\leq c^2 n\delta^2,\ \forall j\neq k, \text{ and } \log N \geq 2(c^2n\delta^2 + \log 2),
      \end{align*}

			then the minimax risk in $\ell_2$ norm is lower bounded as
      \begin{align*}
        \min_{\widehat{\beta}} \sup_{\Pset\in \mathfrak{B}}  \EE{\vecnorm{\widehat{\beta}-\beta(\Pset)}{2}^2} \geq \min_{\widehat{\beta}} \sup_{\Pset\in \Mset}  \EE{\vecnorm{\widehat{\beta}-\beta(\Pset)}{2}^2} \geq \frac{1}{2}\delta^2.
      \end{align*}
		\end{lemma}
		\lemref{fano_bound} restates~\cite[Proposition 15.12]{wainwright2019high} using the $\ell_2$ norm as the semi-metric, together with Equations (15.35a) and (15.35b) of~\cite{wainwright2019high}. To apply this result in our settings, we construct $N$ distinct pairs $\Pset_j$, $j=1,\ldots,N$, of joint (labeled and unlabeled) source and target distributions and define the functional $\beta(\Pset_j)=\betatrue_j$ where $\betatrue_j$ is the oracle estimator under $\Pset_j$, while the specific construction of each $\Pset_j$ depends on the class of source and target distributions that we consider. Based on \lemref{fano_bound}, all upcoming proofs of minimax lower bounds follow the same outline:
    \begin{enumerate}
      \item Designing $\Pset_j$, via imposing conditions on the parameters of SCMs. 
      \item Upper bounding KL-divergence.
      \item Separation of $\beta(\Pset_j)$.
      \item Conclude via Fano's inequality. 
    \end{enumerate}
		
		When designing $\Pset_j$ and bounding KL-divergence between two distributions, the following two lemmas are useful. 
    \begin{lemma}[Varshamov-Gilbert]\label{lem:varshamov_gilbert}
      Let $\Omega = \braces{0, 1}^m$, where $m \geq 8$. Then there exists a pruned hypercube $\widetilde{\Omega} = \braces{\omega_0, \omega_1, \ldots, \omega_N}$ such that $\omega_0 = (0, \ldots, 0)$, 
      \begin{itemize}
        \item $N \geq 2^{m/8}$, and 
        \item $\vecnorm{w_j- w_k}{1} \geq m/8,\;\;\; \forall 0 \leq j < k \leq N$.   
      \end{itemize}
    \end{lemma}
    See, e.g.,~\cite[Chapter 2.6]{tsybakov2009introduction} for a proof. 

		\begin{lemma}\label{lem:trace}
		Let $A,B\in\R^{p\times p}$ be symmetric positive-definite matrices. Then $\ident -AB$ and $\ident - A^{1/2}B A^{1/2}$ have the same eigenvalues, and in particular, they are real. Furthermore, if all eigenvalues of $\ident-AB$ are less than or equal to $\frac{1}{2}$, then
			\[-\log \det(AB) - \textnormal{tr}\left(\ident-AB\right)\leq \textnormal{tr}\left((\ident-AB)^2\right). \]
		\end{lemma}
		 The proof of~\lemref{trace} can be found in~\appref{trace_proof}.

		\subsection{Proof of~\propref{anticausal_minimax_lowerbound}}\label{app:minimax_proof} The goal is to apply Fano's inequality~\lemref{fano_bound} following the outline stated below the lemma. The main differences to the classical minimax lower bound in linear regression are 
    \begin{itemize}
      \item the anticausal SCM setting, and
      \item the use of unlabeled target data. 
    \end{itemize}
    We need to design the unlabeled data generation such that no extra information of parameters of interest is revealed. Throughout the proof, we assume $d\geq 32$, which allows us to apply~\lemref{varshamov_gilbert} to construct the desired data generation models and invoke Fano's inequality. Note that when $d<32$, this approach is no longer applicable, but  a lower bound $\frac{c}{\ntar}$ for some universal constant $c>0$ can still be obtained by manually constructing suitable data generation models---where only a single number of parameters varies---and applying Fano's inequality directly. For brevity, we omit the details for the $d<32$ case, as the argument is a straightforward adaptation of the $d\geq 32$ case.

    \paragraph*{1. Designing $\Pset_j$} Let $N = 2^{d/8}$. Let  $(\omega_0,\omega_1,\ldots,\omega_N)$ be the pruned hypercube from Varshamov-Gilbert~\lemref{varshamov_gilbert} satisfying 
    \begin{align}\label{eqn:prop_32_omega_distance}
      \vecnorm{\omega_j - \omega_k}{1} \geq d/8, \;\;\; \forall 0 \leq j < k \leq N.
    \end{align}
    Take  $\tdelta \defn \min\braces{\frac{1}{4},\sqrt{ \frac{d}{128\ntar}} }$, and set
		\begin{align*}
			\btar_j \defn \frac{\tdelta}{\sqrt{d}} \omega_j, \text{ for } j = 1, \ldots, N.
		\end{align*}
    Let $\tPset_j$ be the joint Gaussian distribution of $(\Xtar,\Ytar)$ which satisfies the linear SCM 
    \begin{align*}
			\begin{bmatrix}
				\Xtar \\ \Ytar
			\end{bmatrix}= 		
			\begin{bmatrix}
				\ident & \btar_j \\ 
				0 & 1
			\end{bmatrix}
			\begin{bmatrix}
				\epstar_X \\ \epstar_Y
			\end{bmatrix},
		\end{align*}
		where $\epstar_X\sim \Nset(0,\Sigma_{\eps_X,j})$ and $\epstar_Y\sim \Nset(0,\vary)$ are independent. Since $\norm{\btar_j}_2< 1$, we can choose the positive-definite matrix $\Sigma_{\eps_X,j}$ such that
		\begin{equation*}\label{eqn:minimax_tar_cov}
			\EE{\Xtar\Xtar{}^\top}= \EE{(\epstar_X+\btar_j\epstar_Y)(\epstar_X+\btar_j\epstar_Y)^\top} = \Sigma_{\eps_X,j} + \nu_Y^2\btar_j\btar_j{}^\top = \nu_Y^2\ident.
		\end{equation*} 
		This choice ensures that the unlabeled target data reveal no information about $\btar_j$. 
		
		Consequently, the joint distribution of $(\Xtar,\Ytar)\sim\tPset_j$ is Gaussian with mean zero and covariance matrix given by
		\[ \Sigma_j\coloneqq \VV{	\begin{bmatrix}
				\Xtar \\ \Ytar
		\end{bmatrix}}  =	\vary\begin{pmatrix}
			\ident & \btar_j  \\ \btar_j{}^\top & 1
		\end{pmatrix} \in\R^{(d+1)\times (d+1)}.\]
		Denote by ${\Pset}_j$ the joint distribution of the $\ntar$ data points from $\Dtar=\{(\Xtar_i,\Ytar_i)\}_{i=1}^{\ntar}\iidsim \tPset_j$ and $\ntaru$ data points from $\Dtaru=\{ \Xtartil_i\}_{i=1}^{\ntaru} \iidsim\tPset_{X,j}$ (the marginal on $X$), and let $\Mset = \{\Pset_1,\ldots,\Pset_N\}$.

    \paragraph*{2. Upper bounding KL-divergence}
		Since the unlabeled distributions $\tPset_{X,j}$ are the same for all $j$, using the formula for the KL-divergence between two Gaussian distributions, we have
		\begin{align}\label{eqn:kl_two_gaussians}
			\kl{\Pset_j}{\Pset_k} = \frac{\ntar}{2}\left(-\log \det (\Sigma_k^{-1}\Sigma_j) + \text{tr}(\Sigma_k^{-1}\Sigma_j) - (d+1)\right),
		\end{align}
		where the unlabeled distributions do not contribute to the KL divergence. We claim that
    \begin{align}\label{eqn:eigenvalues_inequality}
			\lambda_\ell\left(\ident-\Sigma_k^{-1}\Sigma_j\right) \leq \frac{1}{2} \ \text{ for all $\ell=1,\ldots,d+1$},
		\end{align}
		where $\lambda_\ell(A)$ denotes the $\ell$-th largest eigenvalue of $A$. Its proof is deferred to the end. Assuming~\eqnref{eigenvalues_inequality}, and applying~\lemref{trace} to~\eqnref{kl_two_gaussians}, we obtain 
		\begin{align}\label{eqn:kl_two_gaussians_two}
			\kl{\Pset_j}{\Pset_k} \leq \frac{\ntar}{2}\textnormal{tr}\left((\ident-\Sigma_k^{-1}\Sigma_j)^2\right).
		\end{align}
		Let's introduce the shorthand $\Delta\coloneqq \Sigma_k - \Sigma_j$. To upper bound the right-hand side of~\eqnref{kl_two_gaussians_two}, we  observe that
		\begin{align}\label{eqn:trace_fro_ineq}
			\textnormal{tr}\left((\ident-\Sigma_k^{-1}\Sigma_j)^2\right) &= \textnormal{tr}\left((\Sigma_k^{-1}(\Sigma_k-\Sigma_j))^2\right) \\
			&=\textnormal{tr}\left( \Sigma_k^{-1/2}\Delta\Sigma_k^{-1}\Delta\Sigma_k^{-1/2} \right) \nonumber \\ 
			&= \vecnorm{ \Sigma_k^{-1/2}\Delta\Sigma_k^{-1/2}}{\text{F}}^2 \nonumber \\
			&\overset{(i)}{\leq} \vecopnorm{\Sigma_k^{-1/2}}^4 \vecnorm{\Delta}{\text{F}}^2
			\overset{(ii)}{\leq} \frac{2\nu_Y^4\vecnorm{\btar_k-\btar_j}{2}^2}{\lammin^2(\Sigma_k)} , \nonumber
		\end{align}
		where $(i)$ follows from the sub-multiplicative property of the operator norm, and in $(ii)$, we used
		$
			{\Delta} = 	{\vary\begin{pmatrix}
				0 & \btar_k - \btar_j  \\ (\btar_k - \btar_j )^\top& 0
			\end{pmatrix} }.
		$
		Furthermore, since $\lammin(A+B) \geq \lammin(A)+\lammin(B)$ by Weyl's inequality, we have
		\begin{align}\label{eqn:minimax_min_eig}
			\lammin(\Sigma_k) &= \vary \lammin\left(\ident_{d+1} + \begin{bmatrix}
				0 & \btar_k  \\ \btar_k  & 0
			\end{bmatrix} \right)  \\ 
			&\geq \vary \left(\lammin(\ident_{d+1}) + \lammin\left(\begin{bmatrix}
			0 & \btar_k  \\ \btar_k  & 0
			\end{bmatrix} \right)\right) \nonumber\\
			&= \vary \left(1 - \vecnorm{\btar_k}{2}\right). \nonumber
		\end{align}
		Returning to~\eqnref{trace_fro_ineq}, we obtain
		\begin{align*}
			\textnormal{tr}\left((\ident-\Sigma_k^{-1}\Sigma_j)^2\right)\leq \frac{2\vecnorm{\btar_k-\btar_j}{2}^2}{\left(1-\vecnorm{\btar_k}{2}\right)^2} \leq 4\tdelta^2,
		\end{align*}
		where the last inequality holds since $\norm{\btar_k-\btar_j}_2\leq \tdelta$ and $\norm{\btar_k}_2\leq \tdelta \leq 1/4$. Plugging it into~\eqnref{kl_two_gaussians_two}, we have
		\begin{equation}\label{eqn:kl_two_gaussians_two2}
			\kl{\Pset_j}{\Pset_k} \leq 2\ntar\tdelta^2.
		\end{equation}
		
	\paragraph*{3. Separation of $\beta(\Pset_j)$}
	Letting $\beta(\Pset_j)=\EE{\Xtar\Xtar{}^\top}^{-1}\EE{\Xtar\Ytar}=\btar_j$ for $(\Xtar,\Ytar)\sim \tPset_j$, then
	\begin{align}\label{eqn:packing} \min_{j\neq k}  \vecnorm{\beta(\Pset_j)-\beta(\Pset_k)}{2} = \min_{j\neq k}  {\vecnorm{\btar_j - \btar_k}{2}} \overset{(i)}{\geq} \frac{\tdelta}{\sqrt{8}},\end{align}
	where step $(i)$ holds since $\vecnorm{\omega_j - \omega_k}{2} = \sqrt{\vecnorm{\omega_j - \omega_k}{1}} \geq \sqrt{d/8}$ from the construction~\eqnref{prop_32_omega_distance}. 
	
	\paragraph*{4. Conclude via Fano's inequality}
	Taking $\delta=\frac{\tdelta}{2\sqrt{8}}=\min\left\{\frac{1}{8\sqrt{8}},\frac{1}{64} \sqrt{ \frac{d}{\ntar}}\right\}$, it follows from~\eqnref{kl_two_gaussians_two2},~\eqnref{packing} that $\{\beta(\Pset_1),\ldots,\beta(\Pset_N) \}$ is a $2\delta$-packing set and 
	\[\kl{\Pset_j}{\Pset_k}\leq 64\ntar\delta^2. \]
	Since we assume $d\geq 32$, it can be easily checked that
	\[\log N= \frac{d}{8}\log 2 \geq 2(64\ntar\delta^2 + \log 2).\]
	Therefore, applying~\lemref{fano_bound}, the minimax estimation error is lower bounded by
	\begin{align*}
		 \min_{\widehat{\beta}} \sup_{\Pset\in \Mset}  \EE{\vecnorm{\widehat{\beta}-\beta(\Pset)}{2}^2} \geq \frac{1}{2}\delta^2.
	\end{align*}
	Finally, the minimax excess risk bound can be related to the minimax estimation error via
		\begin{align*}
		\EE{\delta\risktar{f_{\betah}} }  = \EE{(\betah-\beta(\Pset))^\top \SigmaXtar (\betah-\beta(\Pset)) } = \vary\EE{\vecnorm{\betah-\beta(\Pset)}{2}^2},
	\end{align*}
	for any $\betah$ and $\Pset\in\Mset$. Hence, we obtain the desired minimax lower bound
	\begin{align*}
		\min_{\widehat{\beta}} \sup_{\Pset\in \Pfraktar}  \EE{\delta\risktar{f_{\betah}} }\geq \min_{\widehat{\beta}} \sup_{\Pset\in \Mset}  \EE{\delta\risktar{f_{\betah}} }  \geq \frac{\vary}{2}\delta^2.
	\end{align*}
	
	\paragraph*{Verifying the eigenvalue condition}
	It remains to verify the condition~\eqnref{eigenvalues_inequality}. Since the eigenvalues of $\ident-\Sigma_k^{-1}\Sigma_j$ and $\ident-\Sigma_k^{-1/2}\Sigma_j\Sigma_k^{-1/2}$ are identical by~\lemref{trace}, it suffices to show that the operator norm of $\ident-\Sigma_k^{-1/2}\Sigma_j\Sigma_k^{-1/2}$ is smaller than or equal to $1/2$. To see this, observe
	\begin{align}\label{eqn:eigenvalue_inequality}
		\vecopnorm{\ident-\Sigma_k^{-1/2}\Sigma_j\Sigma_k^{-1/2}} &= \vecopnorm{\Sigma_k^{-1/2}(\Sigma_k - \Sigma_j)\Sigma_k^{-1/2}} 
		\leq \vecopnorm{\Sigma_k^{-1/2}}^2\vecopnorm{\Sigma_k - \Sigma_j}  \\
		&\leq \frac{\vecopnorm{\Sigma_k - \Sigma_j}}{\lammin(\Sigma_k)} \overset{(i)}{\leq} \frac{\nu_Y^2\vecnorm{\btar_k-\btar_j}{2}}{\vary \left(1-\vecnorm{\btar_k}{2}^2\right)}, \nonumber
	\end{align}
	where step $(i)$ applies the inequality~\eqnref{minimax_min_eig}. Since we know $\norm{\btar_k-\btar_j}_2,\norm{\btar_k}_2\leq \tdelta \leq 1/4$ by definition of $\btar_j$s, it follows that $\vecopnorm{\ident-\Sigma_k^{-1/2}\Sigma_j\Sigma_k^{-1/2}} \leq 1/2$. This completes the proof.
	
  \subsection{Proof of~\thmref{dip_var_matching_finite_minimax}}\label{app:minimax_ftdip_proof} The goal is to apply Fano's inequality~\lemref{fano_bound} following the outline stated below the lemma. The main differences to the setting in~\propref{anticausal_minimax_lowerbound} are 
  \begin{itemize}
    \item the additional use of source data, and
    \item the confounded additive shift interventions.
  \end{itemize}
  We need to design the source data and unlabeled target data generation tailored to CA shift interventions such that no extra information of parameters of interest is revealed. Throughout the proof, we assume $\rdip\geq 32$, which allows us to apply~\lemref{varshamov_gilbert} to construct the desired data generation models and invoke Fano's inequality. For $\rdip<32$, the proof can be easily adapted from the $\rdip\geq 32$ case, so we omit the details (see also the discussion at the beginning of~\appref{minimax_proof}).
  
  \paragraph*{1. Designing $\Pset_j$} Let $N = 2^{\rdip/8}$. Let  $(\omega_0,\omega_1,\ldots,\omega_N)$ be the pruned hypercube from Varshamov-Gilbert~\lemref{varshamov_gilbert} satisfying
  \begin{align}\label{eqn:thm_42_omega_distance}
    \vecnorm{\omega_j - \omega_k}{1} \geq \rdip/8, \;\;\; \forall 0 \leq j < k \leq N.
  \end{align}
  Take $\tdelta \defn \min \braces{\frac{\nu_Y}{4},\frac{\nu_Y}{8}\sqrt{ \frac{\rdip}{5\ntar}} }$, and set 
  \begin{align*}
    w_{Y,j} \defn \frac{\tdelta}{\sqrt{\rdip}} \omega_j, \text{ for } j = 1, \ldots, N.
  \end{align*}
  Let $\tPset_j=\tPset_j^{(1)}\otimes \tPset_j^{\tagtar}$ be the joint Gaussian distribution of source and target where $(\Xsrcone,\Ysrcone)\sim\tPset_j^{(1)}$ and $(\Xtar,\Ytar)\sim\tPset_j^{\tagtar}$ are specified by the following linear SCMs 
  \begin{align}\label{eqn:minimax_linear_SCMs_ftdip}
          \begin{bmatrix}
      \Xsrcone \\ \Ysrcone
    \end{bmatrix}= 		
    \begin{bmatrix}
      \ident & b \\ 
      0 & 1
    \end{bmatrix}
    \begin{bmatrix}
      \epssrcone_X \\ \epssrcone_Y,
    \end{bmatrix},
    \text{ and }
    \begin{bmatrix}
      \Xtar \\ \Ytar
    \end{bmatrix}= 		
    \begin{bmatrix}
      \ident & b \\ 
      0 & 1
    \end{bmatrix}
    \begin{bmatrix}
      \epstar_X \\ \epstar_Y,
    \end{bmatrix} + 			
    \begin{bmatrix}
    bw_{Y,j}^\top + W_j \\ 
    w_{Y,j}^\top
    \end{bmatrix}Z,
  \end{align}
  and $\epssrcone_X,\epstar_X\sim \Nset(0,\Sigma_{\eps_X})$, $\epssrcone_Y,\epstar_Y\sim \Nset(0,\vary)$, and $Z\sim\Nset(0,\ident_{\rdip})$ are all independent. Here, we first pick and fix $b \in \R^d$ and $R\in\R^{d\times \rdip}$ such that $\vecnorm{b}{2}\leq 1/4$, $R$ is orthonormal matrix (i.e., $R^\top R=\ident_{\rdip}$), and $b$ is orthogonal to the columns of $R$. We then set $\Sigma_{\eps_X}=\vary (2 \ident_d -  bb^\top - RR^\top)$ which is guaranteed to be positive-definite by the construction of $b$ and $R$. Finally, we set $W_j = \nu_Y R -  bw_{Y,j}^\top$, which is full rank because $\vecnorm{w_{Y,j}}{2}\leq \nu_Y$. With these choices, we observe
  \begin{align*}
    \EE{\Xtar\Xtar{}^\top} &= \vary bb^\top + (bw_{Y,j}^\top + W_j)(bw_{Y,j}^\top + W_j)^\top + \Sigma_{\eps_X} \\
    &= \vary bb^\top +  \vary RR^\top + \Sigma_{\eps_X} = 2\vary\ident_d.
  \end{align*}
  In words, the unlabeled target data have the same covariance matrices across different $\tPset_j$, and thus have the same distribution.
  
  With the above design of $\tPset_j=\tPset_j^{(1)}\otimes \tPset_j^{\tagtar}$, observe that $\tPset_j^{(1)}$ is same for all $j$ while the joint distribution of $(\Xtar,\Ytar)\sim \tPset^{\tagtar}_j$ is Gaussian with mean zero and covariance matrix
  \begin{align*}
        \Sigma_j\coloneqq \VV{	\begin{bmatrix}
        \Xtar \\ \Ytar
    \end{bmatrix}}  &=	\begin{bmatrix}
      2\vary \cdot\ident_d & b(\vary + \vecnorm{w_{Y,j}}{2}^2) + W_j w_{Y,j}  \\ \left( b(\vary + \vecnorm{w_{Y,j}}{2}^2) + W_j w_{Y,j} \right)^\top &  \vary + \vecnorm{w_{Y,j}}{2}^2
    \end{bmatrix} \\
    &= \begin{bmatrix}
      2\vary \cdot\ident_d & b\vary + \nu_Y R w_{Y,j} \\ \left( b\vary + \nu_Y R w_{Y,j}  \right)^\top &  \vary + \vecnorm{w_{Y,j}}{2}^2
    \end{bmatrix},
  \end{align*}
  where we substitute $W_j=\nu_Y R - b w_{Y,j}^\top$ in the last step. 
  
  Now let ${\Pset}_j$ denote the joint distribution of 
  \begin{itemize}
    \item $\nsrcone$ data points from $\Dsrcone=\{(\Xsrcone_i,\Ysrcone_i)\}_{i=1}^{\nsrcone}\iidsim \tPset^{(1)}_j$,
    \item $\nsrconeu$ data points from $\Dsrconeu=\{ \Xsrconetil_i \}_{i=1}^{\nsrconeu}\iidsim \tPset^{(1)}_{X, j}$,
    \item $\ntar$ data points from $\Dtar=\{(\Xtar_i,\Ytar_i)\}_{i=1}^{\ntar}\iidsim \tPset^{\tagtar}_j$,
    \item $\ntaru$ data points from $\Dtaru=\{ \Xtartil_i\}_{i=1}^{\ntaru} \iidsim\tPset^{\tagtar}_{X,j}$.
  \end{itemize} 
  Let $\Mset = \{\Pset_1,\ldots,\Pset_N\}$. We verify that $\Mset$ satisfies the conditions for applying Fano's inequality~\lemref{fano_bound}.
  
  \paragraph*{2. Upper bounding KL-divergence} Since the source and unlabeled (both source and target) distributions $\tPset_j^{(1)}, \tPset_{X, j}^{(1)}$, and $\tPset^{\tagtar}_{X,j}$ are the same for all $j$ and do not contribute to KL-divergence, we get the following expression for KL-divergence between two Gaussian distributions $\Pset_j$ and $\Pset_k$
  \begin{equation}\label{eqn:kl_two_gaussians_ftdip}
    \kl{\Pset_j}{\Pset_k} = \frac{\ntar}{2}\left(-\log \det (\Sigma_k^{-1}\Sigma_j) + \text{tr}(\Sigma_k^{-1}\Sigma_j) - (d+1)\right).
  \end{equation}
  Assuming that all eigenvalues of $\ident-\Sigma_k^{-1}\Sigma_j$ are less than or equal to $\frac{1}{2}$, for which the proof is deferred to the end, applying~\lemref{trace} to inequality~\eqnref{kl_two_gaussians_ftdip}, we obtain
  \begin{equation*}\label{eqn:kl_two_gaussians_two_ftdip}
    \kl{\Pset_j}{\Pset_k} \leq \frac{\ntar}{2}\textnormal{tr}\left((\ident-\Sigma_k^{-1}\Sigma_j)^2\right).
  \end{equation*}
  Let's introduce the shorthand $\Delta\coloneqq \Sigma_k - \Sigma_j$. Following the same calculation in~\eqnref{trace_fro_ineq}, we deduce that
  \begin{align*}
    \kl{\Pset_j}{\Pset_k} \leq \frac{\ntar}{2} \frac{ \vecnorm{\Delta}{\text{F}}^2}{\lammin^2(\Sigma_k)}.
  \end{align*}
  Further, we observe
  \begin{align}\label{eqn:kl_bound_ftdip}
    \vecnorm{\Delta}{\text{F}}^2 &= 2\vary\vecnorm{ R (w_{Y,j} - w_{Y,k})}{2}^2 + \left(\vecnorm{w_{Y,j}}{2}^2 - \vecnorm{w_{Y,k}}{2}^2\right)^2 \\
    &= 2\vary\vecnorm{ w_{Y,j} - w_{Y,k}}{2}^2 + \left((w_{Y,j} - w_{Y,k})^\top(w_{Y,j}+w_{Y,k}) \right)^2\nonumber\\
    &\leq 2\vary\vecnorm{ w_{Y,j} - w_{Y,k}}{2}^2 + \vecnorm{w_{Y,j} + w_{Y,k}}{2}^2 \vecnorm{w_{Y,j} - w_{Y,k}}{2}^2 \nonumber\\
    &\leq 2\vary\tdelta^2 + \frac{\vary}{4}\tdelta^2,\nonumber
  \end{align}
  where the last step holds since $\vecnorm{ w_{Y,j} - w_{Y,k}}{2}\leq \tdelta$ and $\vecnorm{w_{Y,j}}{2},\vecnorm{w_{Y,k}}{2}\leq \tdelta\leq \nu_Y/4$. By Weyl's inequality, we also have
  \begin{align}\label{eqn:lammin_lowerbound_ftdip}
    \lammin(\Sigma_k) &= \lammin\left( \begin{bmatrix}
      2\vary\cdot \ident_d & 0  \\ 0  & \vary +\vecnorm{w_{Y,k}}{2}^2
    \end{bmatrix} + \begin{bmatrix}
      0 & b\vary + \nu_Y R w_{Y,k} \\ (b\vary + \nu_Y R w_{Y,k})^\top  & 0
    \end{bmatrix} \right)  \\ 
    &\geq \lammin\left(\begin{bmatrix}
      2\vary\cdot \ident_d & 0  \\ 0  & \vary +\vecnorm{w_{Y,k}}{2}^2
    \end{bmatrix} \right) + \lammin\left(\begin{bmatrix}
    0 & b\vary + \nu_Y R w_{Y,k} \\ (b\vary + \nu_Y R w_{Y,k})^\top  & 0
    \end{bmatrix}\right) \notag\\
    &\geq \vary - \vecnorm{b\vary + \nu_Y R w_{Y,k}}{2} \geq \vary - \vary \vecnorm{b}{2} - \nu_Y\vecnorm{w_{Y,k}}{2}\notag \\
      &\geq \frac{\vary}{2} \notag .
  \end{align}
  Combining the inequalities, we obtain
  \begin{align}\label{eqn:kl_two_gaussians_two2_ftdip}
    \kl{\Pset_j}{\Pset_k} \leq \frac{5\ntar\tdelta^2}{\vary}.
  \end{align}
  
  \paragraph*{3. Separation of $\beta(\Pset_j)$}
  Letting $\beta(\Pset_j)=\EE{\Xtar\Xtar{}^\top}^{-1}\EE{\Xtar\Ytar}=\frac{b}{2} + \frac{Rw_{Y,j}}{2\nu_Y}$ for $(\Xtar,\Ytar)\sim \tPset^{\tagtar}_j$, then
  \begin{align}\label{eqn:packing_ftdip} \min_{j\neq k}  \vecnorm{\beta(\Pset_j)-\beta(\Pset_k)}{2} =  \min_{j\neq k}\frac{1}{2\nu_Y}{\vecnorm{R w_{Y,j} - R w_{Y,k}}{2}} \overset{(i)}{\geq} \frac{\tdelta}{2\sqrt{8}\nu_Y},\end{align}
  where $(i)$ holds since $\vecnorm{\omega_j - \omega_k}{2} \geq \sqrt{\rdip/8}$ by its construction~\eqnref{thm_42_omega_distance}. 
  
  \paragraph*{4. Conclude via Fano's inequality}
  Taking $\delta\defn\frac{\tdelta}{4\sqrt{8}\nu_Y}=\min\left\{\frac{1}{16\sqrt{8}}, \sqrt{ \frac{\rdip}{40960\ntar}}\right\}$, it follows from~\eqnref{kl_two_gaussians_two2_ftdip},~\eqnref{packing_ftdip} that $\{\beta(\Pset_1),\ldots,\beta(\Pset_N) \}$ is a $2\delta$-packing set and 
  \[\kl{\Pset_j}{\Pset_k}\leq 640\ntar\delta^2. \]
  Since we assume $\rdip\geq 32$, it can be easily checked that
  \[\log N= \frac{\rdip}{8}\log 2 \geq 2(640\ntar\delta^2 + \log 2).\]
  Therefore, applying~\lemref{fano_bound}, the minimax estimation error is lower bounded by
  \begin{align*}
    \min_{\widehat{\beta}} \sup_{\Pset\in \Mset}  \EE{\vecnorm{\widehat{\beta}-\beta(\Pset)}{2}^2} \geq \frac{1}{2}\delta^2.
  \end{align*}
  Finally, the minimax excess risk bound can be related to the minimax estimation error via
  \begin{align*}
    \EE{\delta\risktar{f_{\betah}} }  = \EE{(\betah-\beta(\Pset))^\top \SigmaXtar (\betah-\beta(\Pset)) } = 2\sigma_Y^2\EE{\vecnorm{\betah-\beta(\Pset)}{2}^2},
  \end{align*}
  for any $\betah$ and $\Pset\in\Mset$, where we use the fact that $\nu_Y=\sigma_Y$ since $\epssrcone_Y,\epstar_Y$ are Gaussian. Hence we have
  \begin{align*}
    \min_{\widehat{\beta}} \sup_{\Pset\in \Pfrak}  \EE{\delta\risktar{f_{\betah}} }\geq \min_{\widehat{\beta}} \sup_{\Pset\in \Mset}  \EE{\delta\risktar{f_{\betah}} }  \geq \sigma_Y^2\delta^2,
  \end{align*}
  proving the desired minimax lower bound.
  
  \paragraph*{Verifying the eigenvalue condition}
  To verify that all eigenvalues of  $\ident-\Sigma_k^{-1}\Sigma_j$ are less than or equal to $1/2$, we proceed similarly to~\eqnref{eigenvalue_inequality} and observe
  \begin{align*}
    \vecopnorm{\ident-\Sigma_k^{-1/2}\Sigma_j\Sigma_k^{-1/2}} \leq \frac{\vecopnorm{\Sigma_k - \Sigma_j}}{\lammin(\Sigma_k)} \overset{(i)}{\leq} \frac{\vecfronorm{\Sigma_k - \Sigma_j}}{\lammin(\Sigma_k)} \overset{(ii)}{\leq} \frac{\tdelta}{\nu_Y} ,
  \end{align*}
  where $(i)$ applies the inequality $\vecopnorm{A}\leq\vecfronorm{A}$ for any matrix $A$ and $(ii)$ applies~\eqnref{kl_bound_ftdip},~\eqnref{lammin_lowerbound_ftdip}. Since $\tdelta\leq \frac{\nu_Y}{2}$, this completes the proof.

  \subsection{Proof of~\thmref{connectivity_estimator_finite_minimax}}\label{app:minimax_ftsparse_proof} The goal is to apply Fano's inequality~\lemref{fano_bound} following the outline stated below the lemma. The main difference to the settings in~\propref{anticausal_minimax_lowerbound} and~\thmref{dip_var_matching_finite_minimax} is that
  \begin{itemize}
    \item the distribution shift now lies in the connectivity matrix (the sparse connectivity shift interventions). 
  \end{itemize}
  We need to design the source data and unlabeled target data generation tailored to SC shift interventions such that no extra information of parameters of interest is revealed. Throughout the proof, we assume $\rconn\geq 32$, which allows us to apply~\lemref{varshamov_gilbert} to construct the desired data generation models and invoke Fano's inequality. For $\rconn<32$, the proof can be easily adapted from the $\rconn\geq 32$ case, so we omit the details (see also the discussion at the beginning of~\appref{minimax_proof}).

  \paragraph*{1. Designing $\Pset_j$} Let $N=2^{\rconn/8}$. Let $(\omega_0,\omega_1,\ldots,\omega_N)$ be the pruned hypercube from Varshamov-Gilbert~\lemref{varshamov_gilbert} satisfying 
  \begin{align}\label{eqn:thm_44_omega_distance}
    \vecnorm{\omega_j - \omega_k}{1} \geq \rconn/8, \;\;\; \forall 0 \leq j < k \leq N.
  \end{align}
  Take $\tdelta \defn \min \braces{\frac{1}{2\sqrt{2}},\frac{1}{16}\sqrt{\frac{\rconn}{\ntar}} }$.

  Let $\tPset_j=\tPset_j^{(1)}\otimes \tPset_j^{\tagtar}$ be the joint Gaussian distribution of source and target where $(\Xsrcone,\Ysrcone)\sim\tPset_j^{(1)}$ and $(\Xtar,\Ytar)\sim\tPset_j^{\tagtar}$ are specified by the following linear SCMs
  \begin{align}\label{eqn:minimax_linear_SCMs_ftsparse}
			\begin{bmatrix}
				\Xsrcone \\ \Ysrcone
			\end{bmatrix}= 
      \begin{bmatrix}
				0 & b \\ 
				0 & 0
			\end{bmatrix}	
      \begin{bmatrix}
				\Xsrcone \\ \Ysrcone
			\end{bmatrix} + 	
			\begin{bmatrix}
				\epssrcone_X \\ \epssrcone_Y,
			\end{bmatrix},
			\text{ and }
			\begin{bmatrix}
				\Xtar \\ \Ytar
			\end{bmatrix}= 		
			\begin{bmatrix}
				\Btar_j & b \\ 
				0 & 0
			\end{bmatrix}
			\begin{bmatrix}
				\Xtar \\ \Ytar
			\end{bmatrix}  + 			
			\begin{bmatrix}
			\epstar_X \\ \epstar_Y
			\end{bmatrix} ,
		\end{align}
		and $\epssrcone_X,\epstar_X\sim \Nset(0,\Sigma_{\eps_X})$, $\epssrcone_Y,\epstar_Y\sim \Nset(0,\vary)$ are independent. Here, we first pick and fix $b$ as 
    \begin{align*}
      b\defn\frac{\tdelta}{\sqrt{\rconn}}\begin{pmatrix}
			1, & 1, & \cdots & 1, & 0, & \cdots & 0
		\end{pmatrix}^\top.
    \end{align*}
    That is, the first $\rconn$ entries are $\frac{\tdelta}{\sqrt{\rconn}}$ and the remaining entries are zero, so that $\vecnorm{b}{2}\leq \tdelta$. Next, we set $\Btar_j$ as a diagonal matrix
    \begin{align*}
      \Btar_j \defn \diag\left(2\omega_j, 0,\ldots,0\right),
    \end{align*}
		with the first $\rconn$ diagonal entries equal to $2\omega_j$ and the rest zero. We then set $\Sigma_{\eps_X}=\vary (\ident - bb^\top)$ which is positive-definite because $\vecnorm{b}{2}\leq \tdelta\leq 1/2$ by construction. It follows that
		\begin{align*}
			\EE{\Xtar\Xtar{}^\top} &= (\ident-\Btar_j)^{-1}(\vary bb^\top + \Sigma_{\eps_X})(\ident-\Btar_j)^{-\top} =\vary\cdot\ident,
		\end{align*}
		i.e., the unlabeled target data share the same covariance matrix across $j$ and do not reveal any information about the prediction of label.
		
		With the above design of $\tPset_j=\tPset_j^{(1)}\otimes \tPset_j^{\tagtar}$, observe that $\tPset_j^{(1)}$ is same for all $j$ while the joint distribution of $(\Xtar,\Ytar)\sim \tPset^{\tagtar}_j$ is Gaussian with mean zero and covariance matrix
		\begin{align*}
			\Sigma_j\coloneqq \VV{	\begin{bmatrix}
					\Xtar \\ \Ytar
			\end{bmatrix}}  
			&= 	\begin{bmatrix}
				\vary \cdot\ident_d & \vary (\ident-\Btar_j)b  \\ \left( \vary (\ident-\Btar_j)b \right)^\top &  \vary 
			\end{bmatrix}.
		\end{align*}
		Let ${\Pset}_j$ denote the joint distribution of 
    \begin{itemize}
      \item $\nsrcone$ data points from $\Dsrcone=\{(\Xsrcone_i,\Ysrcone_i)\}_{i=1}^{\nsrcone}\iidsim \tPset^{(1)}_j$,
      \item $\nsrconeu$ data points from $\Dsrconeu=\{ \Xsrconetil_i \}_{i=1}^{\nsrconeu}\iidsim \tPset^{(1)}_{X, j}$,
      \item $\ntar$ data points from $\Dtar=\{(\Xtar_i,\Ytar_i)\}_{i=1}^{\ntar}\iidsim \tPset^{\tagtar}_j$,
      \item $\ntaru$ data points from $\Dtaru=\{ \Xtartil_i\}_{i=1}^{\ntaru} \iidsim\tPset^{\tagtar}_{X,j}$.
    \end{itemize}
    Let $\Mset \defn \{\Pset_1,\ldots,\Pset_N\}$. We show that $\Mset$ satisfies the conditions for applying Fano's inequality~\lemref{fano_bound}.
		
		\paragraph*{2. Upper bounding KL-divergence} Since the source and unlabeled (both source and target) distributions $\tPset_j^{(1)}, \tPset_{X,j}^{(1)}$, and $\tPset^{\tagtar}_{X,j}$ do not contribute to the calculation of KL-divergence, we get
		\begin{equation}\label{eqn:kl_two_gaussians_ftsparse}
			\kl{\Pset_j}{\Pset_k} = \frac{\ntar}{2}\left(-\log \det (\Sigma_k^{-1}\Sigma_j) + \text{tr}(\Sigma_k^{-1}\Sigma_j) - (d+1)\right).
		\end{equation}
		Assuming that all eigenvalues of $\ident-\Sigma_k^{-1}\Sigma_j$ are less than or equal to $\frac{1}{2}$, for which the proof is deferred to the end, applying~\lemref{trace} to inequality~\eqnref{kl_two_gaussians_ftsparse}, we obtain
		\begin{equation*}\label{eqn:kl_two_gaussians_two_ftsparse}
			\kl{\Pset_j}{\Pset_k} \leq \frac{\ntar}{2}\textnormal{tr}\left((\ident-\Sigma_k^{-1}\Sigma_j)^2\right).
		\end{equation*}
		Let's introduce the shorthand $\Delta\coloneqq \Sigma_k - \Sigma_j$. From~\eqnref{trace_fro_ineq}, we deduce 
		\begin{align*}
			\kl{\Pset_j}{\Pset_k} \leq \frac{\ntar}{2} \frac{ \vecnorm{\Delta}{\text{F}}^2}{\lammin^2(\Sigma_k)}.
		\end{align*}
		Further, we observe
		\begin{align}\label{eqn:kl_bound_ftsparse}
			\vecnorm{\Delta}{\text{F}}^2 &= 2\nu_Y^4\vecnorm{ (\Btar_j - \Btar_k)b }{2}^2 \overset{(i)}{=} \frac{2\nu_Y^4\tdelta^2}{{\rconn}}\vecnorm{\omega_j - \omega_k}{2}^2 \overset{(ii)}{\leq} 2\nu_Y^4\tdelta^2,
		\end{align}
		where $(i)$ follows from the definitions of $b$ and $\Btar_j$, and step $(ii)$ uses the fact that $\vecnorm{\omega_j - \omega_k}{2}\leq \sqrt{\rconn}$. And 
		\begin{align}\label{eqn:lammin_lowerbound_ftsparse}
			\lammin(\Sigma_k) &= \lammin\left( \begin{bmatrix}
				\vary\cdot \ident_d & 0  \\ 0  & \vary 
			\end{bmatrix} + \begin{bmatrix}
				0 &  \vary (\ident-\Btar_k)b \\ ( \vary (\ident-\Btar_k)b)^\top  & 0
			\end{bmatrix} \right)  \\ 
			&\geq \lammin\left(\begin{bmatrix}
				\vary\cdot \ident_d & 0  \\ 0  & \vary 
			\end{bmatrix} \right) + \lammin\left(\begin{bmatrix}
				0 &  \vary (\ident-\Btar_k) b\\ ( \vary (\ident-\Btar_k)b)^\top  & 0
			\end{bmatrix}\right) \nonumber\\
			&= \vary - \vecnorm{ \vary (\ident-\Btar_k)b}{2} = \vary  - \frac{\vary\tdelta}{\sqrt{\rconn}}\vecnorm{\omega_k}{2} \overset{(i)}{\geq} \frac{\vary}{2}, \nonumber
		\end{align}
		where $(i)$ follows since $\tdelta\leq 1/2$ and $\vecnorm{\omega_k}{2}\leq \sqrt{\rconn}$.
		Combining the above, we obtain
		\begin{align}\label{eqn:kl_two_gaussians_two2_ftsparse}
			\kl{\Pset_j}{\Pset_k} \leq {4\ntar\tdelta^2}.
		\end{align}
		
		\paragraph*{3. Separation of $\beta(\Pset_j)$}
		Letting $\beta(\Pset_j)=\EE{\Xtar\Xtar{}^\top}^{-1}\EE{\Xtar\Ytar}=(\ident-\Btar_j)b$ for $(\Xtar,\Ytar)\sim \tPset^{\tagtar}_j$, then
		\begin{align}\label{eqn:packing_ftsparse} \min_{j\neq k}  \vecnorm{\beta(\Pset_j)-\beta(\Pset_k)}{2} =\min_{j\neq k} {\vecnorm{(\Btar_j - \Btar_k)b }{2}} = \frac{\tdelta}{\sqrt{\rconn}}\vecnorm{\omega_j-\omega_k}{2} \geq \frac{\tdelta}{2\sqrt{2}},\end{align}
		where the last step follows from~\eqnref{thm_44_omega_distance}. 
		
		\paragraph*{4. Conclude via Fano's inequality}
	
		Taking $\delta\defn\frac{\tdelta}{4\sqrt{2}}=\min \braces{\frac{1}{16},\frac{1}{64} \sqrt{ \frac{\rconn}{2\ntar}} }$, it follows from~\eqnref{kl_two_gaussians_two2_ftsparse}, \eqnref{packing_ftsparse} that $\{\beta(\Pset_1),\ldots,\beta(\Pset_N) \}$ is a $2\delta$-packing set and
		\begin{align*}
		 \kl{\Pset_j}{\Pset_k}\leq 128\ntar\delta^2.
		\end{align*}
		Since we assume $\rconn\geq 32$, it can be easily checked that
		\begin{align*}
		\log N= \frac{\rconn}{8}\log 2 \geq 2(128\ntar\delta^2 + \log 2).
		\end{align*}
		Therefore, applying~\lemref{fano_bound}, the minimax estimation error is lower bounded by
		\begin{align*}
			\min_{\widehat{\beta}} \sup_{\Pset\in \Mset}  \EE{\vecnorm{\widehat{\beta}-\beta(\Pset)}{2}^2} \geq \frac{1}{2}\delta^2.
		\end{align*}
		Finally, the minimax excess risk bound can be related to the minimax estimation error via
		\begin{align*}
			\EE{\delta\risktar{f_{\betah}} }  = \EE{(\betah-\beta(\Pset))^\top \SigmaXtar (\betah-\beta(\Pset)) } = \sigma_Y^2\EE{\vecnorm{\betah-\beta(\Pset)}{2}^2},
		\end{align*}
		for any $\betah$ and $\Pset\in\Mset$, where we use the fact that $\nu_Y=\sigma_Y$ since $\epssrcone_Y,\epstar_Y$ are Gaussian. Hence, we have
		\begin{align*}
			\min_{\widehat{\beta}} \sup_{\Pset\in \Pfrak}  \EE{\delta\risktar{f_{\betah}} }\geq \min_{\widehat{\beta}} \sup_{\Pset\in \Mset}  \EE{\delta\risktar{f_{\betah}} }  \geq \frac{\sigma_Y^2}{2}\delta^2,
		\end{align*}
		proving the desired minimax lower bound.
		
		\paragraph*{Verifying the eigenvalue condition}
		To verify that all eigenvalues of  $\ident-\Sigma_k^{-1}\Sigma_j$ are less than or equal to $1/2$, we proceed similarly to~\eqnref{eigenvalue_inequality} and observe
		\begin{align*}
			\vecopnorm{\ident-\Sigma_k^{-1/2}\Sigma_j\Sigma_k^{-1/2}} \leq \frac{\vecopnorm{\Sigma_k - \Sigma_j}}{\lammin(\Sigma_k)} \leq \frac{\vecfronorm{\Sigma_k - \Sigma_j}}{\lammin(\Sigma_k)} \overset{(i)}{\leq} 2\sqrt{2}\tdelta ,
		\end{align*}
		where $(i)$ applies~\eqnref{kl_bound_ftsparse},~\eqnref{lammin_lowerbound_ftsparse}. Since $\tdelta\leq \frac{1}{2\sqrt{2}}$, this completes the proof.

  \subsection{Proof of~\thmref{cip_mean_matching_finite_minimax}}\label{app:minimax_ftcip_proof}
  The goal is to apply Fano's inequality~\lemref{fano_bound} following the outline stated below the lemma. The main difference to the settings in~\propref{anticausal_minimax_lowerbound},~\thmref{dip_var_matching_finite_minimax}, and~\thmref{connectivity_estimator_finite_minimax} is that
  \begin{itemize}
    \item the distribution shift now happens to $b$, which is the anticausal weights connecting $Y$ and $X$. 
    \item no unlabeled target data is used.

  \end{itemize}
  We need to design the source data generation tailored to AW shift interventions such that no extra information of parameters of interest is revealed. Throughout the proof, we assume $\rcip\geq 32$, which allows us to apply~\lemref{varshamov_gilbert} to construct the desired data generation models and invoke Fano's inequality. For $\rcip<32$, the proof can be easily adapted from the $\rcip\geq 32$ case, so we omit the details (see also the discussion at the beginning of~\appref{minimax_proof}).

  \paragraph*{1. Designing $\Pset_j$} Let $N=2^{\rcip/8}$. Let $(\omega_0,\omega_1,\ldots,\omega_N)$ be the pruned hypercube from Varshamov-Gilbert~\lemref{varshamov_gilbert} satisfying 
  \begin{align}\label{eqn:thm_46_omega_distance}
    \vecnorm{\omega_j - \omega_k}{1} \geq \rcip/8, \;\;\; \forall 0 \leq j < k \leq N.
  \end{align}
  Take $\tdelta\defn\min\{\frac{1}{8},64\sqrt{\frac{\rcip}{2510\ntar}}\}$, and set 
  \begin{align}\label{eqn:b_choice_ftcip}
  \binv \defn \frac{\tdelta}{\sqrt{d-\rcip}}\ones{d-\rcip} \text{ and } \tbtar_j = \frac{\tdelta}{\sqrt{\rcip}}\omega_j, 
  \end{align}
  where $\ones{d-\rcip}\in\R^{d-\rcip}$ represents the vector with all entries $1$.
		
  Let $\tPset_j=\tPset_j^{(1)}\otimes \cdots \otimes \tPset_j^{(M)}\otimes \tPset_j^{\tagtar}$ denote the joint Gaussian distribution of $M=\rcip+1$ source and target distributions. The source distributions $\tPset_j^{(1)}\otimes \cdots \otimes \tPset_j^{(M)}$ are specified by the following linear SCMs
  \begin{align}\label{eqn:minimax_linear_SCMs_ftcip}
    \begin{bmatrix}
      \Xsrc \\ \Ysrc
    \end{bmatrix}= 		
    \begin{bmatrix}
      \ident & \bsrc \\ 
      0 & 1
    \end{bmatrix}
    \begin{bmatrix}
      \epssrc_X \\ \epssrc_Y
    \end{bmatrix} \text{ for $m=1,\ldots,M$},
  \end{align}
  where $\epssrc_X\sim \Nset(0,\vary\cdot\ident_d)$, $\epssrc_Y \sim \Nset(0,\vary)$ are independent and for $\bsrcnoninv \in \R^{\rcip}$,
  \[\bsrc = \begin{pmatrix}
    \binv \\
    \bsrcnoninv
  \end{pmatrix}, \]
  where $\bsrcnoninv$'s are chosen to satisfy $\spn\left( b_{\text{non}}^{(2)}-b_{\text{non}}^{(1)},  \ldots, b_{\text{non}}^{(M)}-b_{\text{non}}^{(1)}\right)=\R^{\rcip}$ and then fixed. The target distribution $\tPset^{\tagtar}_j$ is  specified by the following linear SCM
  \begin{align}\label{eqn:minimax_linear_SCMs_target_ftcip}
    \begin{bmatrix}
      \Xtar \\ \Ytar
    \end{bmatrix}= 		
    \begin{bmatrix}
      \ident & \btar_j \\ 
      0 & 1
    \end{bmatrix}
    \begin{bmatrix}
      \epstar_X \\ \epstar_Y
    \end{bmatrix} ,
  \end{align}		
  where $\epstar_X\sim \Nset(0,\vary\cdot\ident)$, $\epstar_Y \sim \Nset(0,\vary)$ are independent, and
  \[\btar_j = \begin{pmatrix}
    \binv \\
    \tbtar_j
  \end{pmatrix}. \]		
		
  With the above design of $\tPset_j=\tPset_j^{(1)}\otimes \cdots \otimes \tPset_j^{(M)}\otimes \tPset_j^{\tagtar}$, we observe that $\tPset_j^{(m)}$, $m=1,\ldots,M$, are the same for all $j$ while the joint distribution of $(\Xtar,\Ytar)\sim \tPset^{\tagtar}_j$ is Gaussian with mean zero and covariance matrix
  \begin{align*}
    \Sigma_j\coloneqq \VV{	\begin{bmatrix}
        \Xtar \\ \Ytar
    \end{bmatrix}}  &=	\vary\begin{bmatrix}
        \btar_j \btar_j{}^\top + \ident_d &  \btar_j  \\   \btar_j{}^\top &  1 
    \end{bmatrix}.
  \end{align*}
  Let ${\Pset}_j$ be the joint distribution of 
  \begin{itemize}
    \item $\nsrc$ data points from $\Dsrc=\{(\Xsrc_i,\Ysrc_i)\}_{i=1}^{\nsrc}\iidsim \tPset_j^{(m)}$, $m=1,\ldots,M$,
    \item $\nsrcu$ data points from $\Dsrcu=\{(\Xsrc_i)\}_{i=1}^{\nsrcu}\iidsim \tPset_{X,j}^{(m)}$, $m=1,\ldots,M$,
    \item $\ntar$ data points from $\Dtar=\{(\Xtar_i,\Ytar_i)\}_{i=1}^{\ntar}\iidsim \tPset^{\tagtar}_j$.
  \end{itemize}
  Let $\Mset = \{\Pset_1,\ldots,\Pset_N\}$. We show that $\Mset$ satisfies the conditions for applying Fano's inequality~\lemref{fano_bound}.
		
  \paragraph*{Upper bounding KL-divergence} 
  Since the source distributions $\tPset_j^{(m)}$, $m\geq 1$, are kept the same, using the formula for KL-divergence between two Gaussian distributions, we have
  \begin{equation}\label{eqn:kl_two_gaussians_ftcip}
    \kl{\Pset_j}{\Pset_k} = \frac{\ntar}{2}\left(-\log \det (\Sigma_k^{-1}\Sigma_j) + \text{tr}(\Sigma_k^{-1}\Sigma_j) - (d+1)\right),
  \end{equation}
  where the source distributions do not contribute to the KL divergence.
  Assuming that all eigenvalues of $\ident-\Sigma_k^{-1}\Sigma_j$ are less than or equal to $\frac{1}{2}$, for which the proof is deferred to the end, applying~\lemref{trace} to inequality~\eqnref{kl_two_gaussians_ftcip}, we obtain
  \begin{equation*}
    \kl{\Pset_j}{\Pset_k} \leq \frac{\ntar}{2}\textnormal{tr}\left((\ident-\Sigma_k^{-1}\Sigma_j)^2\right).
  \end{equation*}
		
		Let us introduce the shorthand $\Delta\coloneqq \Sigma_k - \Sigma_j$. From~\eqnref{trace_fro_ineq}, we deduce 
		\begin{align}
			\label{eqn:kl_two_gaussians_two_ftcip}
			\kl{\Pset_j}{\Pset_k} \leq \frac{\ntar}{2} \frac{ \vecnorm{\Delta}{\text{F}}^2}{\lammin^2(\Sigma_k)}.
		\end{align}
		Further, note that 
		\begin{align*}
			\vecnorm{\Delta}{\text{F}}^2 &=\nu_Y^4\vecfronorm{\btar_j\btar_j{}^\top - \btar_k\btar_k{}^\top}^2 +  2\nu_Y^4\vecnorm{ \btar_j - \btar_k}{2}^2 \\
			&\leq \nu_Y^4 \left(\vecnorm{\btar_j}{2} + \vecnorm{\btar_k}{2} \right)^2\vecnorm{ \btar_j - \btar_k}{2}^2 +  2\nu_Y^4\vecnorm{ \btar_j - \btar_k}{2}^2 \nonumber,
		\end{align*}
		where for the last step, we use the inequality that for two vectors $u$ and $v$, 
		\begin{align*}
			\vecfronorm{uu^\top - vv^\top} = \vecfronorm{uu^\top - uv^\top + uv^\top - vv^\top} &\leq \vecfronorm{uu^\top - uv^\top} + \vecfronorm{uv^\top - vv^\top} \\
			&= \left(\vecnorm{u}{2} + \vecnorm{v}{2} \right)\vecnorm{u-v}{2}.
		\end{align*}
		Using the construction of $\omega_j$ in~\eqnref{thm_46_omega_distance} and $\btar_j$ in~\eqnref{b_choice_ftcip}, it follows
		\[\vecnorm{\btar_j}{2} \leq \sqrt{2}\tdelta \leq \frac{1}{2}, \text{ and } \vecnorm{\btar_j-\btar_k}{2}\leq \tdelta. \]
		Plugging into the inequality above, we have
		\begin{align}\label{eqn:kl_bound_ftcip}
			\vecnorm{\Delta}{\text{F}}^2  \leq 3\nu_Y^4 \tdelta^2.
		\end{align}
		Additionally, to lower bound the minimum eigenvalue of $\Sigma_k$, we have 
		\begin{align}\label{eqn:lammin_lowerbound_ftcip}
			\lammin(\Sigma_k) &= \lammin\left(\vary\ident_{d+1}+ \vary\begin{bmatrix}
			\btar_k\btar_k{}^\top &  \btar_k \\ \btar_k{}^\top  & 0
			\end{bmatrix} \right) \\ 
			&\geq \lammin\left(\vary \ident_{d+1} \right) + \lammin\left(  \vary\begin{bmatrix}
				\btar_k\btar_k{}^\top &  \btar_k \\ \btar_k{}^\top  & 0
			\end{bmatrix}  \right) \nonumber\\
			&=\vary +  \lammin\left(  \vary\begin{bmatrix}
				\btar_k\btar_k{}^\top &  0 \\ 0  & 0
			\end{bmatrix}  \right) + \lammin\left(  \vary\begin{bmatrix}
			0 &  \btar_k \\ \btar_k{}^\top  & 0
			\end{bmatrix}  \right)  \nonumber \\
			&\geq \vary - \vary\vecnorm{\btar_k}{2} \geq \frac{\vary}{2}, \nonumber
		\end{align}
		where the last step holds since $\norm{\btar_k}_{2}\leq 1/2$. Combining~\eqnref{kl_two_gaussians_two_ftcip} with~\eqnref{kl_bound_ftcip},~\eqnref{lammin_lowerbound_ftcip}, we obtain
		\begin{align}\label{eqn:kl_two_gaussians_two2_ftcip}
			\kl{\Pset_j}{\Pset_k} \leq {6\ntar\tdelta^2}.
		\end{align}
		
		\paragraph*{3. Separation of $\beta(\Pset_j)$}
		Let $\beta(\Pset_j)=\EE{\Xtar\Xtar{}^\top}^{-1}\EE{\Xtar\Ytar}=(\ident + \btar_j\btar_j{}^\top)^{-1}\btar_j$ for $(\Xtar,\Ytar)\sim \tPset^{\tagtar}_j$. Applying the Sherman-Morrison inversion formula, we have 
		\[\beta(\Pset_j) =\frac{\btar_j}{1+\vecnorm{\btar_j}{2}^2}.  \]
		For $j\neq k$, note that
		\begin{align*}
			\vecnorm{\beta(\Pset_j)-\beta(\Pset_k)}{2} &= \vecnorm{\frac{\btar_j}{1+\vecnorm{\btar_j}{2}^2} - \frac{\btar_k}{1+\vecnorm{\btar_k}{2}^2}}{2} \\
			&= \vecnorm{ \frac{\btar_j - \btar_k + \btar_j \vecnorm{\btar_k}{2}^2 - \btar_k\vecnorm{\btar_j}{2}^2 }{\left(1+\vecnorm{\btar_j}{2}^2\right)\left(1+\vecnorm{\btar_k}{2}^2\right) } }{2} \nonumber \\
			&\geq \frac{\vecnorm{\btar_j - \btar_k}{2} -\vecnorm{\btar_j }{2}\vecnorm{\btar_k }{2}\left(\vecnorm{\btar_j}{2} + \vecnorm{\btar_k}{2} \right)  }{\left(1+\vecnorm{\btar_j}{2}^2\right)\left(1+\vecnorm{\btar_k}{2}^2\right) } ,  \nonumber
		\end{align*}
		where the last step applies the triangle inequality. 
		Since $\norm{\btar_j}_{2}\leq \sqrt{2}\tdelta \leq \frac{1}{4}$ and $\norm{\btar_j-\btar_k}_{2}\geq  \frac{\tdelta}{2\sqrt{2}}$ by~\eqnref{thm_46_omega_distance}, we then have
		\begin{align}\label{eqn:packing_ftcip}
			\vecnorm{\beta(\Pset_j)-\beta(\Pset_k)}{2}  \geq \frac{\sqrt{2}}{32}\tdelta.
		\end{align}
		
		\paragraph*{4. Conclude via Fano's inequality}
		
		Taking $\delta \defn \frac{\sqrt{2}}{64}\tdelta=\min\left\{\frac{\sqrt{2}}{512}, \sqrt{ \frac{\rcip}{1255\ntar}}\right\}$, it follows from~\eqnref{kl_two_gaussians_two2_ftcip},~\eqnref{packing_ftcip} that $\{\beta(\Pset_1),\ldots,\beta(\Pset_N) \}$ is a $2\delta$-packing set and 
		\[\kl{\Pset_j}{\Pset_k}\leq 24576\ntar\delta^2. \]
		Since we assume $\rcip\geq 32$, it can be easily checked that
		\[\log N= \frac{\rcip}{8}\log 2 \geq 2(24576\ntar\delta^2 + \log 2).\]
		Therefore, applying~\lemref{fano_bound}, the minimax estimation error is lower bounded by
		\begin{align*}
			\min_{\widehat{\beta}} \sup_{\Pset\in \Mset}  \EE{\vecnorm{\widehat{\beta}-\beta(\Pset)}{2}^2} \geq \frac{1}{2}\delta^2.
		\end{align*}
		Finally, the minimax excess risk bound can be related to the minimax estimation error via
		\begin{align*}
			\EE{\delta\risktar{f_{\betah}} }  = \EE{(\betah-\beta(\Pset))^\top \SigmaXtar (\betah-\beta(\Pset)) } \geq \sigma_Y^2\EE{\vecnorm{\betah-\beta(\Pset)}{2}^2},
		\end{align*}
		since $\lammin(\SigmaXtar) \geq \vary$ 	for any $\betah$ and $\Pset\in\Mset$, and we have $\vary=\sigma_Y^2$ under Gaussianity. Hence we have
		\begin{align*}
			\min_{\widehat{\beta}} \sup_{\Pset\in \Pfrak}  \EE{\delta\risktar{f_{\betah}} }\geq \min_{\widehat{\beta}} \sup_{\Pset\in \Mset}  \EE{\delta\risktar{f_{\betah}} }  \geq \frac{\sigma_Y^2}{2}\delta^2,
		\end{align*}
		proving the desired minimax lower bound.
		
		\paragraph*{Verifying the eigenvalue condition}
		To verify that all eigenvalues of  $\ident-\Sigma_k^{-1}\Sigma_j$ are less than or equal to $1/2$, we proceed similarly to~\eqnref{eigenvalue_inequality} and observe
		\begin{align*}
			\vecopnorm{\ident-\Sigma_k^{-1/2}\Sigma_j\Sigma_k^{-1/2}} \leq \frac{\vecopnorm{\Sigma_k - \Sigma_j}}{\lammin(\Sigma_k)} \leq \frac{\vecfronorm{\Sigma_k - \Sigma_j}}{\lammin(\Sigma_k)} \overset{(i)}{\leq} \frac{\sqrt{3}\vary\tdelta}{\vary/2} ,
		\end{align*}
		where step $(i)$ applies~\eqnref{kl_bound_ftcip},~\eqnref{lammin_lowerbound_ftcip}. Since $\tdelta\leq \frac{1}{8}$, this completes the proof.

		\section{Proof of propositions}
		
		In this section, we provide technical proofs for the propositions presented in~\appref{UDA_estimator_finite_sample_guarantees} and~\appref{model_selection_MASFT}.
		
		\subsection{Proof of~\propref{least_squares_bound}}\label{app:ols_pool_proof}

		Denote the average population and empirical quantities across $M$ source domains by
		\[\begin{cases}
			\SigmaXbar \defn \frac{1}{M}\sum_{m=1}^{M}\SigmaXsrc, \;\; \Gbar \defn \frac{1}{M}\sum_{m=1}^{M}\EE{\Xsrc\Ysrc}, \text{ and }\\
			\SigmaXhbar \defn  \frac{1}{M}\sum_{m=1}^{M}\frac{1}{\nsrc}\sum_{i=1}^{\nsrc}\Xsrc_i\Xsrc_i{}^\top, \;\; \Ghbar \defn \frac{1}{M}\sum_{m=1}^{M}\frac{1}{\nsrc}\sum_{i=1}^{\nsrc}\Xsrc_i\Ysrc_i.
		\end{cases} \]		
		Since $\betahpool$ is optimal for the quadratic program~\eqnref{OLSPool_finite}, we have the inequality 
		\[ \frac{1}{M}\sum_{m=1}^{M}\frac{1}{\nsrc}\vecnorm{\bYsrc - \bXsrc\betahpool}{2}^2 \leq \frac{1}{M}\sum_{m=1}^{M}\frac{1}{\nsrc}\vecnorm{\bYsrc - \bXsrc\betapool}{2}^2.\]
		Defining $\Delta\defn\betahpool-\betapool$ and performing some algebra yield the basic inequality 
		\begin{align}\label{eqn:pooled_l2_basic_inequality}
			\underbrace{\frac{1}{M}\sum_{m=1}^{M}\frac{1}{\nsrc}\vecnorm{\bXsrc\Delta}{2}^2}_{\eqqcolon T_1} &\leq \frac{1}{M}\sum_{m=1}^{M}\frac{2}{\nsrc}\inner{\bXsrc{}^\top(\bYsrc - \bXsrc\betapool)}{\Delta} \\
			&=2\inner{\Ghbar -  \SigmaXhbar \betapool}{\Delta} \nonumber\\
			&\overset{(i)}{=}2\underbrace{\inner{\Ghbar-\Gbar - (\SigmaXhbar-\SigmaXbar) \betapool}{\Delta}}_{\eqqcolon T_2}, \nonumber
		\end{align}
		where step $(i)$ holds since $\SigmaXbar\betapool = \Gbar$. We can lower bound $T_1$ as follows: if $\nbar\geq c_1 \sigma_X^4\lamminbar^{-2}\cdot \max\braces{d, \log(1/\delta)}$, then applying~\eqnref{bounds_eig_values} together with union bound yields
		\begin{multline}\label{eqn:pooled_l2_T1_bound}
			T_1\geq \frac{1}{M}\sum_{m=1}^{M}\lammin\left(\frac{1}{\nsrc}\bXsrc{}^\top\bXsrc\right)\vecnorm{\Delta}{2}^2 \geq \frac{1}{M}\sum_{m=1}^{M}\frac{\lammin(\SigmaXsrc)}{2}\vecnorm{\Delta}{2}^2  \\ \geq \frac{\lamminbar}{2}\vecnorm{\Delta}{2}^2,
		\end{multline}
		with probability at least $1- M \delta$.  For $T_2$, we use the Cauchy-Schwarz inequality and triangle inequality to get
		\begin{align*}
			&T_2 \leq \vecnorm{\Ghbar-\Gbar - (\SigmaXhbar-\SigmaXbar) \betapool}{2}\vecnorm{\Delta}{2} \\
			&\leq \frac{1}{M}\sum_{m=1}^{M}\frac{1}{\nsrc}\vecnorm{\bXsrc{}^\top \bYsrc -  \bXsrc{}^\top \bXsrc\betapool - \EE{\Xsrc_1\Ysrc_1 - \Xsrc_1\Xsrc_1{}^\top \betapool}}{2}\vecnorm{\Delta}{2}.
		\end{align*}
		For each $m\geq 1$, $\Ysrc_1 - \Xsrc_1{}^\top\betapool$ is sub-Gaussian with parameter $\tsigmapool \defn \sigma_Y+\sigma_X\norm{\betapool}_2$ according to~\lemref{sub_gaussian}. Then, for $\nsrc\geq d\log(1/\delta)$,  applying~\lemref{l2_norm_bound} yields
		\begin{align*}
			&\frac{1}{\nsrc}\vecnorm{\bXsrc{}^\top \bYsrc -  \bXsrc{}^\top \bXsrc\betapool - \EE{\Xsrc_1\Ysrc_1 - \Xsrc_1\Xsrc_1{}^\top \betapool}}{2} \\
			&\leq c_2\sigma_X \tsigmapool \parenth{ \sqrt{\frac{d\log(1/\delta)}{\nsrc}} }, \quad \forall m\geq 1,
		\end{align*}
		with probability at least $1-M\delta$. It follows
		\begin{align*}
			T_2 \leq \frac{c_2\sigma_X \tsigmapool}{M}\sum_{m=1}^{M}\sqrt{\frac{d\log(1/\delta)}{\nsrc}}\vecnorm{\Delta}{2} \leq c_2\sigma_X\tsigmapool\max_{m=1,\ldots,M}\sqrt{\frac{d\log(1/\delta)}{\nsrc}}\vecnorm{\Delta}{2}.
		\end{align*}
		Combining with~\eqnref{pooled_l2_basic_inequality},~\eqnref{pooled_l2_T1_bound}, rearranging, and simplifying, we conclude
		\begin{align*}
			\vecnorm{\Delta}{2}\leq c_3 \lamminbar^{-1} \sigma_X \tsigmapool \max_{m=1,\ldots,M}\sqrt{\frac{d\log(1/\delta)}{\nsrc}},
		\end{align*}
		with probability at least $1-2M\delta$.

		\subsection{Proof of~\propref{DIP_finite_sample_error_bound}}\label{app:dip_finite_proof}

		First recall from~\eqnref{pop_dip_estimator} that the population DIP$^{(1)}$-cov estimator has the following explicit form
		\begin{align*}
			\betasrconedip=\Qdip\left(\Qdip^\top \SigmaXsrcone \Qdip\right)^{-1} \Qdip^\top \EE{\Xsrcone\Ysrcone}.
		\end{align*}
		Using the similar reasoning, the finite-sample DIP$^{(1)}$-cov estimator is given by
		\begin{equation}\label{eqn:dip_uhat}
			\betasrconediph = \Qhdip\underbrace{\left(\frac{1}{\nsrcone} \Qhdip^\top \bXsrcone{}^\top \bXsrcone \Qhdip \right)^{-1} \left( \frac{1}{\nsrcone} \Qhdip^\top \bXsrcone{}^\top \bYsrcone \right)}_{\eqqcolon \usrconediph}. 
		\end{equation}
		We also define an intermediate population quantity
		\begin{equation}\label{eqn:dip_util}
			\betasrconedipt= \Qhdip  \underbrace{\left(\Qhdip^\top \SigmaXsrcone \Qhdip\right)^{-1}\left(\Qhdip^\top \EE{\Xsrcone\Ysrcone}\right)}_{\eqqcolon \usrconedipt}.
		\end{equation}
		By the triangle inequality, we decompose the estimation error as
		\begin{equation}\label{eqn:dip_decompose}
			\vecnorm{\betasrconediph  - \betasrconedip}{2} \leq \underbrace{ \vecnorm{\betasrconediph  - \betasrconedipt }{2}}_{\eqqcolon T_1} + \underbrace{\vecnorm{\betasrconedipt - \betasrconedip}{2}}_{\eqqcolon T_2} .
		\end{equation}
		We proceed to bound each term $T_1$ and $T_2$ separately. 
		
		\paragraph*{Bounding $T_1$} 
		We treat $\Qhdip$ as fixed and analyze the randomness of $(\bXsrcone$, $\bYsrcone)$. Since \( \Qhdip \) is orthonormal, controlling $T_1$  reduces to bounding \( \vecnorm{\usrconediph - \usrconedipt}{2} \).
		
		Note that $\usrconedipt$ is the population OLS$^{(1)}$ estimator regressing $\Ysrcone$ onto $\Qhdip^\top\Xsrcone$, and $\usrconediph$ is its plug-in finite-sample estimator. To apply~\propref{least_squares_bound} with $M=1$, we verify two conditions: (1) By~\lemref{sub_gaussian} and \( \opnorm{\Qhdip} = 1 \), the random vector \( \Qhdip^\top \Xsrcone \) is sub-Gaussian with parameter $\sigma_X$. (2) Since $\sigma_{\text{min}}(AB)\geq \sigma_{\text{min}}(A)\sigma_{\text{min}}(B)$ for any matrices $A$ and $B$ with $B$ having full column rank, we get
		\begin{equation}\label{eqn:dip_min_eig_ineq}
			\lammin(\Qhdip^\top \SigmaXsrcone \Qhdip) \geq \sigma^2_{\text{min}}(\Qhdip)\lammin(\SigmaXsrcone) \geq \lamminbar.
		\end{equation}
		Since $\nsrcone\geq c \max\braces{\tsigmamin^2 d_\delta, d \log(1/\delta) }$,~\propref{least_squares_bound} with $M=1$ thus gives
		\[
		\vecnorm{\usrconedipt-\usrconediph}{2} \leq  {c_1 \tsigmamin  \parenth{\frac{\sigma_Y}{\sigma_X}+\vecnorm{\usrconedipt}{2}} }\sqrt{\frac{d\log(1/\delta)}{\nsrcone}},
		\]
		with probability at least $1-2 \delta$ over the randomness of $\Dsrcone$. Since $\vecnorm{\usrconedipt}{2}=\vecnorm{\betasrconedipt}{2}\leq \sqrt{\kappa}\vecnorm{\betasrcone}{2}$, defining 
		\[\eta_1 \defn c_1 \tsigmamin \parenth{\frac{\sigma_Y}{\sigma_X}+ \sqrt{\kappa}\vecnorm{\betasrcone}{2} }, \]
		it follows
		\begin{equation}\label{eqn:dip_bound1}
			\vecnorm{\betasrconediph - \betasrconedipt}{2} = \vecnorm{\Qhdip\usrconediph - \Qhdip\usrconedipt}{2}  = \vecnorm{\usrconedipt-\usrconediph}{2} \leq  \eta_1\sqrt{\frac{d\log(1/\delta)}{\nsrcone}}.
		\end{equation}
		
		\paragraph*{Bounding $T_2$} 
		Now we treat $\Qhdip$ as random and bound $T_2$ due to the randomness of $\Dsrconeu$ and $\Dtaru$. Let 
		\[\Pi_{\Qdip}\defn\sqrt{\SigmaXsrcone}\Qdip\left(\Qdip^\top \SigmaXsrcone \Qdip\right)^{-1}\Qdip^\top \sqrt{\SigmaXsrcone},\]
		be the projection matrix onto $\col{\sqrt{\SigmaXsrcone}\Qdip}$, and define $\Pi_{\Qhdip}$ analogously. Then we have
		\begin{align}\label{eqn:dip_bound2}
			\vecnorm{\betasrconedipt - \betasrconedip}{2} 
			&= \vecnorm{ \left(\Qhdip\left(\Qhdip^\top \SigmaXsrcone \Qhdip\right)^{-1}\Qhdip^\top - \Qdip\left(\Qdip^\top \SigmaXsrcone \Qdip\right)^{-1}\Qdip^\top\right) \SigmaXsrcone \betasrcone}{2}  \\ 
			&\leq {\vecopnorm{ \Qhdip\left(\Qhdip^\top \SigmaXsrcone \Qhdip\right)^{-1}\Qhdip^\top\SigmaXsrcone - \Qdip\left(\Qdip^\top \SigmaXsrcone \Qdip\right)^{-1}\Qdip^\top\SigmaXsrcone }}\vecnorm{\betasrcone}{2} \nonumber \\
			&= \vecopnorm{  \SigmaXsrcone{}^{-1/2} \Pi_{\Qhdip} \SigmaXsrcone{}^{1/2} - \SigmaXsrcone{}^{-1/2} \Pi_{\Qdip}  \SigmaXsrcone{}^{1/2} } \vecnorm{\betasrcone}{2}  \nonumber \\
			&\leq \vecopnorm{\Pi_{\Qhdip} - \Pi_{\Qdip}}\cdot  \sqrt{\kappabar}\vecnorm{\betasrcone}{2} . \nonumber
		\end{align}
		Observe that both $\sqrt{\SigmaXsrcone} \Qhdip \Qhdip^\top \sqrt{\SigmaXsrcone}$ and $\sqrt{\SigmaXsrcone} \Qdip\Qdip^\top \sqrt{\SigmaXsrcone}$ have rank $d-\rdip$, and their smallest nonzero eigenvalues are lower bounded by $\lamminbar$, as shown in~\eqnref{dip_min_eig_ineq}. Therefore, the eigen-gap between the $d-\rdip$ leading eigenvalues and the remaining eigenvalues (i.e., zeros) is at least $\lamminbar$. Applying the Davis-Kahan theorem~\cite[Theorem 1]{yu2015useful}, we can control the subspace angle between the column space of the two projection matrices 

		\begin{align}\label{eqn:dip_sin_theta}
			\vecopnorm{\sin \Theta(U,\Uh)} &\leq \frac{\vecopnorm{\sqrt{\SigmaXsrcone} \Qdip\Qdip^\top \sqrt{\SigmaXsrcone} -  \sqrt{\SigmaXsrcone} \Qhdip\Qhdip^\top \sqrt{\SigmaXsrcone}} }{\lamminbar}   \\&\leq \frac{\lammaxbar\vecopnorm{\Qdip\Qdip^\top - \Qhdip\Qhdip^\top}}{\lamminbar} \nonumber \\&= \kappabar\vecopnorm{\Qdip\Qdip^\top - \Qhdip\Qhdip^\top},\nonumber
		\end{align}
		where $U,\Uh\in\R^{d\times (d-\rdip)}$ are orthonormal bases for $\col{\sqrt{\SigmaXsrcone}\Qdip}$, $\col{\sqrt{\SigmaXsrcone}\Qhdip}$, respectively.
		
		By~\cite[Exercise VII.1.11]{bhatia2013matrix}, we further have the equivalence
		\begin{equation}\label{eqn:dip_left_hand_side}
			\vecopnorm{\sin \Theta(U,\Uh)} =\vecopnorm{U U^\top - \Uh \Uh^\top}\overset{(i)}{=}\vecopnorm{\Pi_{\Qdip} - \Pi_{\Qhdip}},
		\end{equation}
		where step $(i)$ holds because $U$ and $\sqrt{\SigmaXsrcone}\Qdip$ span the same column spaces, so $U U^\top = \Pi_{\Qdip}$, and similarly, $\Uh\Uh^\top = \Pi_{\Qhdip}$. Additionally, we have
		\[\kappabar\vecopnorm{\Qdip\Qdip^\top - \Qhdip\Qhdip^\top} = \kappabar\vecopnorm{\sin \Theta(\Qdip,\Qhdip)},  \]
		 so	under the assumption that $\nsrconeu, \ntaru \geq c_2' \tsigmagap^2 \max\braces{d, \log(1/\delta)}$,~\lemref{error_orthogonal_space} (with the eigen-gap $\lamdipgap$ defined in~\eqnref{ftdip_eigengap}) implies
		\begin{align}\label{eqn:dip_right_hand_side}
			\kappabar\vecopnorm{\Qdip\Qdip^\top - \Qhdip\Qhdip^\top}\leq c_2 \kappabar \tsigmagap  \parenth{\sqrt{\frac{d}{\nsrconeu}} + \sqrt{\frac{\log(1/\delta)}{\nsrconeu}} + \sqrt{\frac{d}{\ntaru}} + \sqrt{\frac{\log(1/\delta)}{\ntaru}} },
		\end{align}
		with probability at least $1- \delta$ over randomness of  $\Dsrconeu$, $\Dtaru$. Combining~\eqnref{dip_sin_theta}~\eqnref{dip_left_hand_side},~\eqnref{dip_right_hand_side} with~\eqref{eqn:dip_bound2}, we get
		\begin{equation}\label{eqn:dip_bound3}
			\vecnorm{\betasrconedipt - \betasrconedip}{2} \leq \underbrace{c_2 \kappabar^{3/2} \tsigmagap \vecnorm{\betasrcone}{2} }_{\eqqcolon \eta_2}\parenth{\sqrt{\frac{d}{\nsrconeu}} + \sqrt{\frac{\log(1/\delta)}{\nsrconeu}} + \sqrt{\frac{d}{\ntaru}} + \sqrt{\frac{\log(1/\delta)}{\ntaru}} }.
		\end{equation}
		\paragraph*{Combining} Finally, combining~\eqnref{dip_decompose},~\eqnref{dip_bound1},~\eqnref{dip_bound3} together, we conclude that with probability at least $1-3 \delta$ over the randomness of all datasets, $\betasrconediph$ satisfies 
		\begin{equation}\label{eqn:dip_final_bound}
			\vecnorm{\betasrconediph-\betasrconedip}{2} \leq \eta_1\sqrt{\frac{d\log(1/\delta)}{\nsrcone}} + \eta_2\parenth{\sqrt{\frac{d}{\nsrconeu}} + \sqrt{\frac{\log(1/\delta)}{\nsrconeu}} + \sqrt{\frac{d}{\ntaru}} + \sqrt{\frac{\log(1/\delta)}{\ntaru}} },
		\end{equation}
		proving the desired result.
		
		\subsection{Proof of~\propref{CIP_finite_sample_error_bound}}\label{app:proof_of_CIP_finite_sample_error_bound}
		The proof proceeds similarly to that of Proposition~\ref{prop:DIP_finite_sample_error_bound}. Define the population and empirical average quantities
		\[\begin{cases}
			\SigmaXbar \defn \frac{1}{\mcip}\sum_{m=1}^{\mcip}\SigmaXsrc, \;\; \Gbar \defn \frac{1}{\mcip}\sum_{m=1}^{\mcip}\EE{\Xsrc\Ysrc}, \text{ and }\\
			\SigmaXhbar \defn  \frac{1}{\mcip}\sum_{m=1}^{\mcip}\frac{1}{\nsrc}\sum_{i=1}^{\nsrc}\Xsrc_i\Xsrc_i{}^\top, \;\; \Ghbar \defn \frac{1}{\mcip}\sum_{m=1}^{\mcip}\frac{1}{\nsrc}\sum_{i=1}^{\nsrc}\Xsrc_i\Ysrc_i.
		\end{cases} \]
		From Appendix~\ref{app:pop_cip_estimator}, we know the population CIP-mean has the explicit form
		\[\betacip = \Qcip(\Qcip^\top \SigmaXtar \Qcip)^{-1} \Qcip^\top \EE{\Xtar\Ytar}= \Qcip\left(\Qcip^\top \SigmaXbar \Qcip\right)^{-1} \Qcip^\top \Gbar.\]
		Similarly, the finite-sample CIP-mean can be written as
		\begin{align*}
			\betahcip = \Qhcip\underbrace{\left(\Qhcip^\top \SigmaXhbar \Qhcip\right)^{-1} \Qhcip^\top \Ghbar}_{\eqqcolon \uciph},
		\end{align*}
		where $\Qhcip$ is an orthonormal matrix spanning the orthogonal complement of \( \col{\Vhcip} \). To facilitate the analysis, we introduce an intermediate estimator
		\begin{align*}
			\betaciptil=\Qhcip\underbrace{\left(\Qhcip^\top \SigmaXbar \Qhcip\right)^{-1} \Qhcip^\top \Gbar}_{\eqqcolon \uciptil}.
		\end{align*}
		We then decompose the error as
		\begin{align*}
			\vecnorm{\betaciph -\betacip}{2} \leq \underbrace{\vecnorm{\betaciph -\betaciptil}{2}}_{\eqqcolon T_1} + \underbrace{\vecnorm{\betaciptil -\betacip}{2}}_{\eqqcolon T_2}.
		\end{align*}
		In the following, the terms  $T_1$ and $T_2$ are bounded separately.
		
		\paragraph*{Bounding $T_1$} We first treat $\Qhcip$ as fixed. Observe
		\[\vecnorm{\betaciph - \betaciptil}{2} = \vecnorm{\Qhcip\uciph - \Qhcip\uciptil}{2}  \leq \vecnorm{\uciph - \uciptil}{2} . \]
		As in the proof of~\propref{DIP_finite_sample_error_bound}, we apply~\propref{least_squares_bound}, and so by the assumption $\nbar\geq c' \max\braces{\tsigmamin^2 \ddel, d\log(1/\delta) }$, we have
		\begin{align*}
			\vecnorm{\uciph - \uciptil}{2} \leq c\tsigmamin\left(\frac{\sigma_Y}{\sigma_X} +\vecnorm{\uciptil}{2}\right)\max_{m=1,\ldots,\mcip}\sqrt{\frac{d\log(1/\delta)}{\nsrc}},
		\end{align*}
		with probability at least $1-2\mcip\delta$ over the randomness of $\Dsrc_2$, $m=1,\ldots,\mcip$. Since $\vecnorm{\uciptil}{2} = \vecnorm{\betaciptil}{2}\leq \sqrt{\kappabar} \vecnorm{\betapool}{2}$, defining 
		\[ \eta_1 =c_1\tsigmamin\left(\frac{\sigma_Y}{\sigma_X} + \sqrt{\kappabar}\vecnorm{\betapool}{2}\right),  \]
		it follows
		\begin{equation}\label{eqn:cip_bound_1}
			\vecnorm{\betaciph - \betaciptil}{2} \leq \eta_1\max_{m=1,\ldots,\mcip}\sqrt{\frac{d\log(1/\delta)}{\nsrc}}. 
		\end{equation}
		
		\paragraph*{Bounding $T_2$} Next we treat \( \Qhcip \) as random with respect to the datasets \( \Dsrc_1 \), \( m = 1, \ldots, \mcip \). As in the proof of~\propref{DIP_finite_sample_error_bound} (see Equations~\eqnref{dip_bound2},\eqnref{dip_sin_theta},\eqnref{dip_left_hand_side}), we can deduce that
		\begin{align*}
			\vecnorm{\betaciptil - \betacip}{2} \leq \vecopnorm{\Pi_{\Qhcip} - \Pi_{\Qcip}} \cdot \sqrt{\kappabar} \vecnorm{\betapool}{2}\leq \vecopnorm{\Qcip \Qcip^\top - \Qhcip\Qhcip^\top} \cdot \kappabar^{3/2}  \vecnorm{\betapool}{2}.
		\end{align*}
		Further, the following lemma controls the deviation of $\Qhcip$ from $\Qcip$.
		\begin{lemma}\label{lem:cip_difference_eigenvectors}
			Suppose~\assumpsref{linear_SCMs},~\assumpssref{sub_gaussianity},~\assumpssref{coefficient_shift_linear_SCM} hold, and let $\Pw$ and $\Ph$ be defined in~\eqnref{def_Pw} and~\eqnref{def_Pwh}, respectively.

			There exist universal constants $c,c',c''>0$ such that for $\delta \in (0, 1/(2M+4))$, if $\nbar \geq c\frac{\sigma_Y^4}{\nu_Y^4} \log (1/\delta)$, the following bounds hold with probability at least $1- (2M+4) \delta$:
			\begin{align*}
				\inf_{R:R^\top R=\ident}\vecopnorm{\Vhcip- \Vcip R}&\leq \sqrt{2}\vecopnorm{\sin \Theta(\Vcip,\Vhcip)}  \leq c'\tsigmagapcip\sqrt{\frac{\ddel}{\nbar}} , \text{ and }\\
				\inf_{R:R^\top R=\ident}\vecopnorm{\Qhcip - \Qcip R} &\leq \sqrt{2}\vecopnorm{\sin \Theta(\Qcip,\Qhcip)} \leq c'\tsigmagapcip\sqrt{\frac{\ddel}{\nbar}},
			\end{align*}
			where the randomness is over $\Dsrc_1$, $m=1,\ldots,\mcip$.
		\end{lemma}

		The proof is provided in~\appref{proof_of_cip_difference_eigenvectors}.  By~\lemref{cip_difference_eigenvectors}, with probability at least  $1-(2M+4)\delta$, we have
		\[\vecopnorm{\Qcip\Qcip^\top - \Qhcip\Qhcip^\top} = \vecopnorm{\sin \Theta(\Qcip,\Qhcip)} \leq c_2\tsigmagapcip\sqrt{\frac{\ddel}{\nbar}}.  \]
		Putting the pieces together, we obtain
		\begin{equation}\label{eqn:cip_bound_2}
			\vecnorm{\betaciptil - \betacip}{2} \leq \underbrace{ {c_2\tsigmagapcip\cdot \kappabar^{3/2} \norm{\betapool}_2  } }_{\eqqcolon \eta_2} \sqrt{\frac{\ddel}{\nbar}}.
		\end{equation}
		
		Finally, combining~\eqnref{cip_bound_1} and~\eqnref{cip_bound_2}, we arrive at
		\begin{align*}
			\vecnorm{\betaciph -\betacip}{2}  \leq \eta_1\sqrt{\frac{d\log(1/\delta)}{\nbar}} +  \eta_2 \sqrt{\frac{\ddel}{\nbar}},
		\end{align*}
		with probability at least $1- (4\mcip+4) \delta$. This completes the proof of the proposition.		
		
		\subsection{Proof of~\propref{model_selection}}\label{app:model_selection_proof}

		The proof follows a standard technique for model selection (e.g.,~\cite[Section 4.6]{bach2024learning}) with minor modifications. For a fixed \(\{\thetahfti{1}, \ldots, \thetahfti{L}\}\), define $\ell_i(\theta) \defn \ell(f_\theta(\Xtarval_i),\Ytarval_i)=(\Ytarval_i-f_{\theta}(\Xtarval_i))^2$, and the empirical target risk over the hold-out validation dataset $\Dtarval$ as
		\[\riskhtar{f_{\theta};\Dtarval} = \frac{1}{\ntarval}\sum_{i=1}^{\ntarval}\ell_i(\theta)= \frac{1}{\ntarval}\sum_{i=1}^{\ntarval} (\Ytarval_i-f_{\theta}(\Xtarval_i))^2. \]
		Since by assumption of the proposition, $\Ytarval_i-f_{\theta}(\Xtarval_i)$ is a sub-Gaussian random variable with parameter $\sigma_\ell$, it follows from~\lemref{sub_exp} that $\ell_i(\theta)$ is a sub-exponential random variable with parameter $K=\sigma_\ell^2$. Applying Bernstein's inequality~\cite[Theorem 2.8.1]{vershynin2018high}, we then have
		\[\PP{\abs{\riskhtar{f_{\theta};\Dtarval}  - \risktar{f_\theta}} \geq \frac{t}{\ntarval} } \leq 2\exp\left( -c_1\min\left\{ \frac{t^2}{\ntarval K^2}, \frac{t}{K} \right\}\right), \]
		where $c_1>0$ is a universal constant and the randomness is with respect to \(\Dtarval\). Using the union bound, with probability at least $1-\delta$, 
		\begin{equation}\label{eqn:model_selection_union_bound} 
			\PP{\sup_{\theta\in \{\thetahfti{1}, \ldots, \thetahfti{L}\} }\abs{ \riskhtar{f_{\theta};\Dtarval}  - \risktar{f_\theta}} \geq c_2K \sqrt{\frac{\log(2L/\delta)}{\ntarval}} } \leq \delta, 
		\end{equation}
		provided that $\ntarval \geq \frac{\log (2L/\delta)}{c_1}$, where $c_2>0$ is a universal constant. It follows that with probability at least $1-\delta$,
		\begin{align*}
			\risktar{f_{\thetahatfinal}} &\overset{(i)}{\leq} \riskhtar{f_{\thetahatfinal};\Dtarval} + c_2K \sqrt{\frac{\log(2L/\delta)}{\ntarval}} \\
			&\overset{(ii)}{\leq} \riskhtar{f_{\thetahfti{j}};\Dtarval} + c_2K \sqrt{\frac{\log(2L/\delta)}{\ntarval}}  \text{ for all $j=1,\ldots,L$} \\
			&\overset{(iii)}{\leq}\risktar{f_{\thetahfti{j}}} + 2c_2K \sqrt{\frac{\log(2L/\delta)}{\ntarval}} \text{ for all $j=1,\ldots,L$} ,
		\end{align*}
		where step $(i)$ uses the inequality~\eqnref{model_selection_union_bound}, step $(ii)$ follows from the definition of \(\thetahatfinal\) as the minimizer of the empirical validation risk, and step $(iii)$ applies~\eqnref{model_selection_union_bound} again. Subtracting \(\min_{f\in\Fset}\risktar{f}\) from both sides and taking the minimum over $j$, we obtain the desired result.				

		\section{Proof of lemmas}\label{app:proof_of_lemmas}
		In this section, we provide proofs for all lemmas.
		\subsection{Proof of~\lemref{error_orthogonal_space}}\label{app:proof_error_orthogonal_space}
		Let $\lambda_1\geq \ldots \geq \lambda_d$ denote the eigenvalues of $\SigmaW - \SigmaWt$, and let $\widehat{\lambda}_1\geq \ldots \geq \widehat{\lambda}_d$ denote the eigenvalues of $\SigmaWh - \SigmaWht$. Note that $\SigmaW - \SigmaWt$ is not positive definite, so $\lambda_d$ and $\widehat{\lambda}_d$ can take negative values. \lemref{bounds_cov_matrices} together with Weyl's inequality implies that for all $j=1,\ldots,d$, with probability at least $1-2\delta$,
		\begin{align*}
			\abs{\lambda_j - \widehat{\lambda}_j} &\leq \vecopnorm{\SigmaW - \SigmaWt - (\SigmaWh - \SigmaWht)} \\
			&\leq \vecopnorm{\SigmaW - \SigmaWt} + \vecopnorm{\SigmaWh - \SigmaWht}  \\
			&\leq c_1\sigma_W^2 \parenth{ \sqrt{\frac{d}{n}} + \sqrt{\frac{\log(1/\delta)}{n}} + \sqrt{\frac{d}{\nt}} + \sqrt{\frac{\log(1/\delta)}{\nt}} }.
		\end{align*}		
		For any $j$ with $\abs{\lambda_j}<\frac{1}{2}\lamgap$, then
		\begin{align}\label{eqn:davis_kahan_gap_bound}
			\abs{ \widehat{\lambda}_j} \leq \abs{ \lambda_j - \widehat{\lambda}_j}  + \abs{ \lambda_j}   \leq \frac{3}{4}\lamgap,
		\end{align}			
		as long as $n,\nt \geq c_2{\sigma_W^4\lamgap^{-2}\cdot \max\braces{d, \log(1/\delta)} }$ for a large enough $c_2>0$.
		
		Let $S_1=\{\lambda\in\R: |\lambda| \geq \lamgap \}$ and $S_2=\{\lambda\in\R: |\lambda| \leq \frac{3}{4}\lamgap \}$. Note that $\text{dist}(S_1,S_2)=\lamgap/4$, which together with Eq~\eqnref{davis_kahan_gap_bound} implies that the eigenvalues of $\SigmaW - \SigmaWt$ and $\SigmaWh - \SigmaWht$ corresponding to $V$, $\Qh$ lie in $S_1$ and $S_2$ respectively. Therefore, the generalization of Davis-Kahan theorem~\cite[Theorem VII.3.2]{bhatia2013matrix} implies that 
		\begin{align}\label{eqn:sin_theta}
			\vecopnorm{\sin \Theta(V,\widehat{V})} \leq \frac{c_3\vecopnorm{\SigmaW - \SigmaWt - \left(\SigmaWh - \SigmaWht\right) } }{\lamgap},
		\end{align}
		for some universal constant $c_3>0$. Combining with~\cite[Lemma 2.5, Lemma 2.6]{chen2021spectral}, it follows that
		\begin{align*}
			\inf_{R:R^\top R=\ident}\vecopnorm{\widehat{V} - VR} \leq \sqrt{2}\vecopnorm{VV^\top - \widehat{V}\widehat{V}^\top} &= \sqrt{2}\vecopnorm{\sin \Theta(V,\widehat{V})}\\
			 &\leq \frac{c_3\sqrt{2}\vecopnorm{\SigmaW - \SigmaWt - \left(\SigmaWh - \SigmaWht\right) } }{\lamgap}.
		\end{align*}
		On the same event where~\lemref{bounds_cov_matrices} holds, we then have
		\begin{align*}
			\inf_{R:R^\top R=\ident}\vecopnorm{\widehat{V} - VR} \leq c_4\frac{\sigma_W^2}{\lamgap}\parenth{ \sqrt{\frac{d}{n}} + \sqrt{\frac{\log(1/\delta)}{n}} + \sqrt{\frac{d}{\nt}} + \sqrt{\frac{\log(1/\delta)}{\nt}} }.
		\end{align*}
		Similarly, we can obtain the bound for $\inf_{R:R^\top R=\ident}\vecopnorm{\widehat{Q} - QR}$, completing the proof.

		\subsection{Proof of~\lemref{restricted_eigenvalue}}\label{app:proof_restricted_eigenvalue}
		
		Let $\gamma(T) = \EE{\sup_{x \in T} |\inner{g}{x}}|$ denote the Gaussian complexity of $T$. 
		When $W_i$ is an isotropic sub-Gaussian random vector (i.e., $\Sigma_W=\ident_d$), the result of the lemma follows from the matrix deviation inequalities given in~\cite[Exercise 9.1.8, 9.1.9]{vershynin2018high}. Without the assumption that $W_i$ is isotropic, we can use the arguments from~\cite[Remark 1.2]{liaw2017simple} to derive:
		\begin{align*}
			\sup_{x\in T}\abs{\vecnorm{\bW x}{2} - \sqrt{n}\vecnorm{\sqrt{\Sigma_W}x}{2} } = \sup_{x\in \sqrt{\Sigma_W}T}\abs{\vecnorm{\bZ x}{2} - \sqrt{n}\vecnorm{x}{2} }  \leq cK^2 \sqrt{\log(1/\delta)} \gamma(\sqrt{\Sigma_W}T),
		\end{align*}
		with probability at least $1-2 \delta$. Here $\bZ=\bW\sqrt{\Sigma_W^{-1}}$ is the whitened version of $\bW$, and $K$ is the sub-Gaussian parameter of $\sqrt{\Sigma_W^{-1}}W_i$.
		
		Since $\sqrt{\Sigma_W}T$ contains the origin, it follows from~\cite[Exercise 7.6.9]{vershynin2018high} that $\gamma(\sqrt{\Sigma_W}T)=\gw{\sqrt{\Sigma_W}T}$. Furthermore, by~\cite[Exercise 7.5.4]{vershynin2018high} we have
		\[\gw{\sqrt{\Sigma_W}T}\leq \vecopnorm{\sqrt{\Sigma_W}}\gw{T} = \sqrt{\vecopnorm{{\Sigma_W}}}\gw{T}.\]
		Together with the bound $K\leq \sigma_W\sqrt{\lammin^{-1}(\Sigma_W)}$, for any $x\in T$, we have
		\begin{align*}
			\frac{1}{\sqrt{n}}\vecnorm{\bW x}{2} &\geq \vecnorm{\sqrt{\Sigma_W}x}{2}  - \frac{c \sigma_W^2 \lammin^{-1}(\Sigma_W) \sqrt{\log(1/\delta)}  \sqrt{\vecopnorm{{\Sigma_W}}}\gw{T}}{\sqrt{n}} \\
			&\geq \sqrt{\lammin(\Sigma_W)}\vecnorm{x}{2}  - \frac{c \sigma_W^2 \lammin^{-1}(\Sigma_W) \sqrt{\log(1/\delta)}  \sqrt{\vecopnorm{{\Sigma_W}}}\gw{T}}{\sqrt{n}}.
		\end{align*}
		Using the inequality $(a-b)^2\geq \frac{1}{2}a^2 - b^2$, we conclude
		\begin{align*}
			\frac{1}{n}\vecnorm{\bW x}{2}^2 \geq \frac{\lammin(\Sigma_W)}{2}\vecnorm{x}{2}^2  - \frac{c^2 \sigma_W^4 \lammin^{-2}(\Sigma_W) \log(1/\delta)  {\vecopnorm{{\Sigma_W}}}\gw{T}^2}{{n}}.
		\end{align*}	
		
		\subsection{Proof of~\lemref{l2_norm_bound}}\label{app:proof_l2_norm_bound}
		
		Let us introduce the shorthand $Q_i = W_i Z_i$. By~\cite[Proposition 7]{maurer2021concentration}, with probability at least $1-\delta$, we have
		\begin{multline*}
			\vecnorm{\frac{1}{n}\left(\bW^\top\bZ - \EE{\bW^\top \bZ } \right)}{2} = \vecnorm{\frac{1}{n}\sum_{i=1}^{n}Q_i - \EE{Q_1}}{2} \\
			\leq  c_1\underbrace{\EE{\vecnorm{Q_1 - \EE{Q_1}}{2}^4}^{1/4}}_{\text{Term (A)}}\sqrt{\frac{2\log(1/\delta)}{n}}  +c_1\underbrace{\vecnorm{\vecnorm{Q_1}{2}}{\psi_1}}_{\text{Term (B)}}{\frac{\log(1/\delta)}{n}},
		\end{multline*}
		where $c_1>0$ is a universal constant. Since $Q_1$ is sub-exponential with parameter $\sigma_W\sigma_Z$ by~\lemref{sub_exp}, the centering lemma~\cite[Exercise 2.7.10]{vershynin2018high} implies that $Q_1 - \EE{Q_1}$ is also sub-exponential with parameter $c_2\sigma_W\sigma_Z$ for some universal constant $c_2>0$. The following lemma bounds the $4$-th moment of the random variable $\vecnorm{Q_1-\EE{Q_1}}{2}$.
		\begin{lemma}\label{lem:fourth_moment}
			There is a universal constant $c_3>0$ such that
			\[ \mathbb{E}\left[\vecnorm{Q_1-\EE{Q_1}}{2}^4\right]\leq c_3 d^2\sigma_W^4\sigma_Z^4.\]
		\end{lemma}	
		Applying~\lemref{fourth_moment}, Term (A) can be bounded as 
		\begin{align*}
			\text{Term (A)} = \EE{\vecnorm{Q_1 - \EE{Q_1}}{2}^4}^{1/4} &\leq c_3\sqrt{d}\cdot \sigma_W\sigma_Z.
		\end{align*}	
		To bound Term (B), we use the variational form of $\norm{Q_1}_2$, i.e.,
		\[\norm{Q_1}_2 =\sup_{v\in \mathbb{B}} \abs{v^\top Q_1}, \]
		where $\mathbb{B}=\{v\in\R^{d}:\norm{v}_2\leq 1 \}$ is a $\ell_2$ ball with unit-norm. Using~\cite[Example 5.8]{wainwright2019high}, the metric entropy of $\mathbb{B}$ is bounded by \[\log\cover{\eps}{\mathbb{B}}\leq d\log\left(1+\frac{2}{\eps}\right).\] 
		Additionally, from~\lemref{sub_exp}, for every $u,v\in\mathbb{B}$, $\frac{(u-v)}{\norm{u-v}_2}^\top Q_1$ is a sub-exponential random variable with parameter $\sigma_W\sigma_Z$, that is, 
		\[\vecnorm{(u-v)^\top Q_1}{\psi_1} \leq  \sigma_W\sigma_Z \norm{u-v}_2.\]
		This shows that the condition of~\cite[Theorem 2.2.4]{vaart1997weak} is satisfied with the constant $\sigma_W\sigma_Z$. Therefore, applying~\cite[Corollary 2.2.5]{vaart1997weak} we get
		\begin{align*}
			\vecnorm{\vecnorm{Q_1}{2}}{\psi_1} &= \vecnorm{\sup_{v\in \mathbb{B}} \abs{v^\top Q_1}}{\psi_1} \leq c_4\sigma_W\sigma_Z\int_{0}^{\diam{\mathbb{B}}}\psi_1^{-1}\left(\cover{\eps/2}{\mathbb{B}}\right) d\eps,
		\end{align*}	
		where $\psi_1^{-1}(x)=\log(1+x)$, and $c_4>0$ is an universal constant. Since $\diam{\mathbb{B}}=2$, it follows 
		\begin{align*}
			\vecnorm{\vecnorm{Q_1}{2}}{\psi_1} &\leq c_4 \sigma_W\sigma_Z \int_{0}^{2}\log\left(1+\cover{\eps/2}{\mathbb{B}}\right) d\eps \\
			&\leq c_4\sigma_W\sigma_Z \int_{0}^{2}\log\left(1+ \left(1+\frac{4}{\eps}\right)^d\right) d\eps \\
			&\leq c_4\sigma_W\sigma_Z \int_{0}^{2}\log \left(1+\frac{4}{\eps}\right)^{2d} d\eps \\
			&\leq c_4\sigma_W\sigma_Z \int_{0}^{2}2d\log\left(\frac{8}{\eps}\right)d\eps \leq 10 c_4\sigma_W\sigma_Z\cdot d,
		\end{align*}
		where the third step uses the inequality $1+x \leq x^2$ for all $x\geq 3$. Putting everything together, we have that with probability at least $1-\delta$, 
		\begin{align*}
			\vecnorm{\frac{1}{n}\left(\bW^\top\bZ - \EE{\bW^\top \bZ } \right)}{2}  \leq c\sigma_W\sigma_Z\left\{\sqrt{\frac{d\log(1/\delta)}{n}} + \frac{d\log(1/\delta)}{n}\right\},
		\end{align*}
		where $c>0$ is a universal constant.
		
		\begin{proof}[Proof of~\lemref{fourth_moment}] Let $\widetilde{Q}=Q_1-\EE{Q_1}$ and denote by $\widetilde{Q}_{[i]}$ the $i$-th component of $\widetilde{Q}$. Then we can compute
			\begin{multline*}
				\mathbb{E}\left[\|Q_1 - \mathbb{E}[Q_1]\|_2^4\right] = \mathbb{E}\left[\left(\sum_{i=1}^{d}\widetilde{Q}_{[i]}^2\right)^2\right] = \sum_{i=1}^{d}\mathbb{E}[\widetilde{Q}_{[i]}^4] + \sum_{i \neq j}\mathbb{E}[\widetilde{Q}_{[i]}^2 \widetilde{Q}_{[j]}^2] \\
				\leq \sum_{i=1}^{d}\mathbb{E}[\widetilde{Q}_{[i]}^4] + \frac{1}{2}\sum_{i \neq j}\left(\mathbb{E}[\widetilde{Q}_{[i]}^4] + \mathbb{E}[\widetilde{Q}_{[j]}^4]\right) = d\sum_{i=1}^{d} \mathbb{E}[\widetilde{Q}_{[i]}^4],
			\end{multline*}
			where the inequality uses the identity $ab \leq \frac{a^2 + b^2}{2}$. Since $\widetilde{Q}_{[i]}$ is a sub-exponential random variable with parameter $c_2\sigma_W\sigma_Z$, by~\cite[Proposition 2.7.1]{vershynin2018high}, we have $\mathbb{E}[\widetilde{Q}_{[i]}^4]\leq c_3\sigma_W^4\sigma_Z^4$ for some constant $c_3>0$. Plugging into the above inequality proves the lemma.
		\end{proof}
		
		\subsection{Proof of~\lemref{l_infty_norm_bound}}\label{app:proof_l_infty_norm_bound}
		
		We use the shorthand $Q_i=W_iZ_i$. From the proof of~\lemref{l2_norm_bound}, there is an universal constant $c_1>0$ such that $Q_i-\EE{Q_i}$ is sub-exponential with parameter $c_1\sigma_W\sigma_Z$. By the definition of the sub-exponential random vector, it follows that for any standard basis vector $e_j\in\R^d$, $j=1,\ldots,d$, $e_j^\top (Q_i - \EE{Q_i})$ is also a sub-exponential random variable with parameter $c_1\sigma_W\sigma_Z$. 
		
		Let $K=\max_{j=1,\ldots,d}\vecnorm{e_j^\top (Q_i-\EE{Q_i})}{\psi_1}$. Bernstein's inequality~\cite[Corollary 2.8.3]{vershynin2018high} provides that for all $t\geq 0$, we have
		\[\PP{\abs{\frac{1}{n}\sum_{i=1}^{n}e_j^\top (Q_i-\EE{Q_i})} \geq t} \leq 2\exp\left[-c\min \left(\frac{t^2}{K^2}, \frac{t}{K}\right)n\right]. \]
		Applying the union bound, we have
		\[\PP{\max_{j=1,\ldots,d}\abs{\frac{1}{n}\sum_{i=1}^{n}e_j^\top (Q_i-\EE{Q_i})} \geq t} \leq 2d\exp\left[-c\min \left(\frac{t^2}{K^2}, \frac{t}{K}\right)n\right],\]
		which implies that with probability at least $1-\delta$, 
		\[\max_{j=1,\ldots,d}\abs{\frac{1}{n}\sum_{i=1}^{n}e_j^\top (Q_i-\EE{Q_i})} \leq c_2K \left(\frac{\log d + \log(1/\delta)}{n} + \sqrt{\frac{\log d+ \log(1/\delta)}{n}}\right), \]
		for some universal constant $c_2>0$. Since $K\leq c_1\sigma_W\sigma_Z$, this completes the proof.

		\subsection{Proof of~\lemref{ft_dip_gaussian_width}}\label{app:proof_of_ft_dip_gaussian_width}

		Let $g\sim \mathcal{N}(0,\ident_d)$ be a standard Gaussian random vector. Since $\SigmaXhtar{}^{-1}(\Qhdip\Qhdip^\top + \Vhdip\Vhdip^\top )\SigmaXhtar=\ident_d$, we have the decomposition
		\begin{align*}
			\inner{g}{x} = \inner{\Qhdip^\top \SigmaXhtar{}^{-1} g}{\Qhdip^\top \SigmaXhtar x} + \inner{\Vhdip^\top \SigmaXhtar{}^{-1} g}{\Vhdip^\top \SigmaXhtar x}.
		\end{align*}
		By taking the supremum over $x\in \Lambda$ on both sides and taking the expectation with respect to $g$, we obtain
		\begin{align*}
			\gw{\Lambda} = \EE{\sup_{x\in \Lambda}\inner{g}{x}} &\leq \EE{\sup_{x\in \Lambda}\inner{\Qhdip^\top \SigmaXhtar{}^{-1} g}{\Qhdip^\top \SigmaXhtar x}} + \EE{\sup_{x\in \Lambda}\inner{\Vhdip^\top \SigmaXhtar{}^{-1} g}{\Vhdip^\top \SigmaXhtar x}}\\
			&=  \EE{\sup_{x\in\mathbb{B}_{\rdip}(2\varrho)}\inner{\Vhdip^\top \SigmaXhtar{}^{-1} g}{x}},
		\end{align*}
		where $\mathbb{B}_{\rdip}(2\varrho)$ represent $\ell_2$ ball in $\R^{\rdip}$ with radius $2\varrho$.
		
		Let $\Sigma=\Vhdip^\top \SigmaXhtar{}^{-2}\Vhdip$. Then we can check that $\Vhdip^\top \SigmaXhtar{}^{-1} g\sim \mathcal{N}(0,\Sigma)$, so
		\begin{align*}
			\EE{\sup_{x\in \mathbb{B}_{\rdip}(2\varrho)}\inner{\Vhdip^\top \SigmaXhtar{}^{-1} g}{x}} &= \EE{\sup_{x\in\mathbb{B}_{\rdip}(2\varrho)}\inner{\sqrt{\Sigma} g}{x}} \\
			&= \EE{\sup_{x\in \mathbb{B}_{\rdip}(2\varrho)}\inner{ g}{\sqrt{\Sigma}x}} \\
			&= \EE{\sup_{x\in\sqrt{\Sigma}\mathbb{B}_{\rdip}(2\varrho)}\inner{ g}{x}}  = \gw{\sqrt{\Sigma}\mathbb{B}_{\rdip}(2\varrho)}.
		\end{align*}
		Applying~\cite[Exercise 7.5.4]{vershynin2018high}, we have 
		\begin{align*}
			\gw{\sqrt{\Sigma}\mathbb{B}_{\rdip}(2\varrho)}\leq \vecopnorm{\sqrt{\Sigma}}\gw{\mathbb{B}_{\rdip}(2\varrho)} = \sqrt{\vecopnorm{\Sigma}}\gw{\mathbb{B}_{\rdip}(2\varrho)}.
		\end{align*}
		Since $\gw{\mathbb{B}_{\rdip}(2\varrho)}\leq 2\varrho \sqrt{\rdip}$ (e.g.,~\cite[Example 5.13]{wainwright2019high}) and on the event $\Eset$, $\vecopnorm{\Sigma}\leq \opnorm{\SigmaXhtar{}^{-2}}\leq 4\lamminbar^{-2}$, it follows that
		\begin{align*}
			\gw{\Lambda} \leq \gw{\sqrt{\Sigma}\mathbb{B}_{\rdip}(2\varrho)} \leq 8\varrho \lamminbar^{-1}\sqrt{\rdip},
		\end{align*}
		which completes the proof of the lemma.
		
		\subsection{Proof of~\lemref{trace}}\label{app:trace_proof}
		
		Set $C:=A^{1/2}BA^{1/2}\succ0$. Since $AB=A^{-1/2}CA^{1/2}$, the matrices $AB$ and $C$ are similar, so they have the same eigenvalues. In particular, the eigenvalues of $AB$ (hence of $\ident-AB$) are real. Then it follows
		\[
		\det(AB)=\det(C),\text{ and }
		\textnormal{tr}\left(\ident-AB\right)=\textnormal{tr}\left(\ident-C\right),\text{ and }
		\textnormal{tr}\left((\ident-AB)^2\right)=\textnormal{tr}\left((\ident-C)^2\right).
		\]
		Thus, it suffices to prove
		\begin{equation}\label{eqn:log_det_trace}
			-\log \det\left(C\right) - \textnormal{tr}\left(\ident-C\right)\leq \textnormal{tr}\left((\ident-C)^2\right).
		\end{equation}
		Let $\lambda_j$, $j=1,\ldots,p$, denote the eigenvalues of $\ident-C$. Then 
		\[\det(C) = \prod_{j=1}^{p}(1-\lambda_j), \text{ and } \textnormal{tr}\left(\ident -C\right) = \sum_{j=1}^{p}\lambda_j.\] 
		Substituting into~\eqnref{log_det_trace}, the inequality becomes
		\[ -\sum_{j=1}^{p}\left(\log(1-\lambda_j) + \lambda_j\right) \leq \sum_{j=1}^{p}\lambda_j^2. \]
		Now, the inequality $-\log(1-x)-x \leq x^2$ holds for any ${x}\leq \frac{1}{2}$. Since the assumption of the lemma guarantees ${\lambda_j}\leq \frac{1}{2}$ for all $j$, the result follows.

		\subsection{Proof of~\lemref{cip_difference_eigenvectors}}\label{app:proof_of_cip_difference_eigenvectors}

    First, we claim that for $\nbar \geq c\frac{\sigma_Y^4}{\nu_Y^4}\log (1/\delta)$,
    \begin{align}\label{eqn:cip_difference_matrix_bound}
      \vecopnorm{\Ph - \Pw} \leq  c'\frac{\sigma_X}{\nu_Y} \cdot \sqrt{\frac{\ddel }{\nbar}},
    \end{align}
    with probability at least $1-(2M+4)\delta$. Its proof is deferred to the end.

		Given the claim~\eqnref{cip_difference_matrix_bound}, applying the modified Davis-Kahan-Wedin sine theorem (e.g.,~\cite[Theorem 19]{o2018random}), we obtain
		\[\vecopnorm{\sin \Theta(\Vcip,\Vhcip)} \leq \frac{2\vecopnorm{\Ph - \Pw}}{\lamcipgap} \leq \frac{2c' \sigma_X }{\lamcipgap\nu_Y}\sqrt{\frac{\ddel }{\nbar}} =2c'\tsigmagapcip\sqrt{\frac{\ddel }{\nbar}}  . \]
		Combining with~\cite[Lemma 2.5, Lemma 2.6]{chen2021spectral}, it follows 
		\begin{align*}
			&\inf_{R:R^\top R=\ident}\vecopnorm{\Vhcip - \Vcip R}  \leq \sqrt{2}\vecopnorm{\sin \Theta(\Vcip,\Vhcip)}\leq 2\sqrt{2}c'\tsigmagapcip\sqrt{\frac{\ddel }{\nbar}},\\
			&\inf_{R:R^\top R=\ident}\vecopnorm{\Qhcip - \Qcip R}  \leq \sqrt{2}\vecopnorm{\sin \Theta(\Vcip,\Vhcip)}\leq 2\sqrt{2}c'\tsigmagapcip\sqrt{\frac{\ddel }{\nbar}}.
		\end{align*}

    Next, we turn back to prove the claim~\eqnref{cip_difference_matrix_bound}.  We decompose \( \Ph - \Pw \) into two parts:
		\begin{multline*}
			\Ph - \Pw =\underbrace{\begin{pmatrix}
					\bhsrctwo - H\bsrctwo , \cdots, \bhsrcend - H\bsrcend 
			\end{pmatrix}}_{\coloneqq P_1}
			\\-
			\underbrace{\begin{pmatrix}
					\bhsrcone - H\bsrcone , \cdots, \bhsrcendprev - H\bsrcendprev
			\end{pmatrix}}_{\coloneqq P_2}.
		\end{multline*}
		The proofs for $P_1$ and $P_2$ are similar, we focus on bounding $P_1$. 

    Define the event
    \begin{align*}
      \Eset \defn \braces{\vecnorm{\bYsrc}{2}^2 \geq \frac{\vary}{2}\nsrc, \quad \forall m \in \braces{1, \ldots, M}}. 
    \end{align*}
    Since each coordinate of $\bYsrc$ is sub-Gaussian, applying Hanson-Wright inequality~\cite[Theorem 1.1]{rudelson2013hanson} and the bound $\vary\leq 2\sigma_Y^2$ from~\cite[Proposition 2.5.2]{vershynin2018high}, for $\nsrc \geq c \frac{\sigma_Y^4}{\nu_Y^4} \log (1/\delta)$, the event $\Eset$ holds with probability at least $1-2 M\delta$. 
		
		We claim that $\bhsrc$ defined in~\eqnref{cip_ols_estimator} has an explicit expression, $m=1,\ldots,\mcip$:
    \begin{align}\label{eqn:cip_regressor_bound}
      \bhsrc = H\bsrc + \frac{H \sum_{i=1}^{\nsrc}\epssrc_{X,i}\Ysrc_i }{\vecnorm{\bYsrc}{2}^2}.
    \end{align}
    Its proof is deferred to the end. 
		
		Now let $\Mset, \Nset$ denote the $1/4$-covers of the spheres $\mathbb{S}^{d-1},\mathbb{S}^{\mcip-2}$, respectively. By~\cite[Equation (4.13)]{vershynin2018high}, the operator norm of $P_1$ is bounded as
		\[\vecopnorm{P_1}\leq 2\sup_{x\in\Mset,y\in \Nset}x^\top P_1 y. \]
		Fix $x\in\Mset,y\in\Nset$. Using~\eqnref{cip_regressor_bound}, we can write $x^\top P_1 y$ as
		\begin{align}\label{eqn:cip_quadratic_form}
			x^\top P_1 y &=  \begin{pmatrix}
				\frac{x^\top H \sum_{i=1}^{\nsrctwo}\epssrctwo_{X,i}\Ysrctwo_i }{\vecnorm{\bYsrctwo}{2}^2},& \cdots, &  \frac{x^\top H \sum_{i=1}^{\nsrcend}\epssrcend_{X,i}\Ysrcend_i }{\vecnorm{\bYsrcend}{2}^2}
			\end{pmatrix} y \\
			&= \sum_{m=2}^{\mcip} y_{[m-1]} \frac{x^\top H \sum_{i=1}^{\nsrc}\epssrc_{X,i}\Ysrc_i }{\vecnorm{\bYsrc}{2}^2} \nonumber\\
			&= \sum_{m=2}^{\mcip} \sum_{i=1}^{\nsrc} y_{[m-1]} \frac{x^\top H \epssrc_{X,i}\Ysrc_i }{\vecnorm{\bYsrc}{2}^2}. \nonumber
		\end{align}
		Conditioning on $\Ysrc_i$ for all $i$ and $m=2,\ldots,\mcip$, each summand in~\eqnref{cip_quadratic_form} is sub-Gaussian and independent to each other. We have 
		\begin{align*}
			\vecnorm{x^\top P_1 y}{\psi_2}^2 &= \vecnorm{\sum_{m=2}^{\mcip} \sum_{i=1}^{\nsrc} y_{[m-1]} \frac{x^\top H \epssrc_{X,i}\Ysrc_i }{\vecnorm{\bYsrc}{2}^2}}{\psi_2}^2\\
			&\overset{(i)}{\leq} c_1\sum_{m=2}^{\mcip}\sum_{i=1}^{\nsrc}\vecnorm{y_{[m-1]} \frac{x^\top H \epssrc_{X,i}\Ysrc_i }{\vecnorm{\bYsrc}{2}^2}}{\psi_2}^2 \\
			&= c_1\sum_{m=2}^{\mcip}y_{[m-1]}^2\sum_{i=1}^{\nsrc}\frac{(\Ysrc_i)^2}{\vecnorm{\bYsrc}{2}^4}\vecnorm{x^\top H  \epssrc_{X,i}}{\psi_2}^2 \\
			&\overset{(ii)}{\leq}  c_2\sum_{m=2}^{\mcip}y_{[m-1]}^2\sum_{i=1}^{\nsrc}\frac{(\Ysrc_i)^2}{\vecnorm{\bYsrc}{2}^4}\vecopnorm{H}^2\sigma_{\eps_X}^2 =c_2\sum_{m=2}^{\mcip}y_{[m-1]}^2\frac{\vecopnorm{H}^2\sigma_{\eps_X}^2}{\vecnorm{\bYsrc}{2}^2},
		\end{align*}
    where steps (i) and (ii) follow from \cite[Proposition 2.7.1]{vershynin2018high}.

  Under event~$\Eset$, it follows
		\begin{align*}
			\vecnorm{x^\top P_1 y}{\psi_2}^2 \leq 2c_2\max_{m=2,\ldots,\mcip}\frac{\varyinv\vecopnorm{H}^2\sigma_{\eps_X}^2}{\nsrc }\leq \underbrace{2c_2\max_{m=2,\ldots,\mcip}\frac{\varyinv\sigma_X^2}{\nsrc }}_{\eqqcolon K^2},
		\end{align*}
		where the last inequality uses the fact that \( \vecopnorm{H} \sigma_{\eps_X} \leq \sigma_X \) from~\eqnref{def_sigma_X}. Conditioned on $\Eset$, applying concentration of sub-Gaussian random variables, for any $u\geq 0$, 
    \begin{align*}
      \PP{\abss{x\tp P_1 y} \geq u} \leq 2 \exp(-c u^2/K^2). 
    \end{align*}
		Now it remains to unfix $x$ and $y$ and take a union bound over $\Mset$ and $\Nset$. Specifically, following the same argument as Step 3 in the proof of~\cite[Theorem 4.4.5]{vershynin2018high}, we have, on the event $\Eset$, 
    \begin{align*}
      \vecopnorm{P_1} \leq c_{3}\max_{m=2,\ldots,\mcip}\frac{\varyinvone \sigma_X}{\sqrt{\nsrc}} \parenth{\sqrt{\ddel} +\sqrt{\mcip-1}}. 
    \end{align*}
		with probability at least $1-2\delta$. Similarly, on $\Eset$,
    \begin{align*}
      \vecopnorm{P_2} \leq c_{3}\max_{m=1,\ldots,\mcip-1}\frac{\varyinvone\sigma_X}{\sqrt{\nsrc}} \parenth{\sqrt{\ddel} +\sqrt{\mcip-1}},
    \end{align*}
		with probability at least $1-2 \delta$. Taking union bound, along with the fact that $\mcip\leq d$, we conclude that
		\begin{align*}
			\vecopnorm{\Ph - \Pw} \leq \vecopnorm{P_1} + \vecopnorm{P_2} \leq c_4\frac{\sigma_X}{\nu_Y}  \cdot \max_{m=1,\ldots,\mcip} \sqrt{\frac{\ddel  }{\nsrc}} ,
		\end{align*}
		with probability at least $1- (2M+4)\delta$, given that the source sample sizes satisfy $\nbar \geq c\frac{\sigma_Y^4}{\nu_Y^4}\log (1/\delta)$.
		
    Finally, we prove the claim~\eqnref{cip_regressor_bound}. By regressing $\bXsrc$ onto $\bYsrc$, we obtain 
    \begin{align*}
      \bhsrc = \frac{\bXsrc{}^\top \bYsrc}{\vecnorm{\bYsrc}{2}^2} =\frac{\sum_{i=1}^{\nsrc}\Xsrc_i \Ysrc_i }{\vecnorm{\bYsrc}{2}^2} .
    \end{align*}	
    Since $\Xsrc_{i}$ can be expressed as $H\bsrc \Ysrc_i + H\epssrc_{X,i}$ under~\assumpref{coefficient_shift_linear_SCM}, substituting this expression, we obtain 
    \begin{align*}
      \bhsrc = H\bsrc + \frac{H \sum_{i=1}^{\nsrc} \epssrc_{X,i}\Ysrc_i }{\vecnorm{\bYsrc}{2}^2}.
    \end{align*}
    
    \section{Failure of DIP under anticausal weight shift interventions}\label{app:failure_DIP_AW_shift}

    When anticausal weight (AW) shift interventions exist (\assumpref{coefficient_shift_linear_SCM}), the interventions depend on the label $Y$ which can cause DIP to fail for achieving good target prediction performance. In such cases, matching the marginal covariate distributions between the source and target domains suffers from an identifiability issue, where multiple projection operators can yield the same marginal distributions, but some projections may result in a large target risk. We illustrate this issue with an example below.
    \paragraph*{Example of DIP's failure case} Consider data generated under~\assumpref{coefficient_shift_linear_SCM} with $M=2$ source domains and one target domain. Let $\bsrcone=(b_{[1]}, b_{[2]})^\top \in\R^2$, $\bsrctwo=(b_{[1]}, 0)^\top \in\R^2$, and $\btar=(b_{[1]},-b_{[2]})^\top\in\R^2$. The source and target data are generated from the following SCMs:
    \begin{align*}
    	\begin{cases}
    		\Ysrcone= \epssrcone_Y, \\
    		\Xsrcone =\bsrcone \Ysrcone + \epssrcone_X,
    	\end{cases}
    	\begin{cases}
    		\Ysrctwo = \epssrctwo_Y, \\
    		\Xsrctwo = \bsrctwo\Ysrctwo + \epssrctwo_X,
    	\end{cases}
    	\begin{cases}
    		\Ytar = \epstar_Y, \\
    		\Xtar = \btar \Ytar + \epstar_X,
    	\end{cases}
    \end{align*}
    where $\epssrcone_Y,\epssrctwo_Y,\epstar_Y\sim \Nset(0,1)$ and $\epssrcone_X,\epssrctwo_X,\epstar_X\sim \Nset(0,\ident_2)$. Under this data generation model, consider the projection $v^\top X$ with $v=(1,0)^\top$. Since the conditional distribution of $v^\top X=X_{[1]}$ given $Y=y$ is $\Nset(b_{[1]}y, 1)$ in all three domains, $X_{[1]}$ corresponds to a conditionally invariant component. 
    
    Now suppose we apply DIP$^{(1)}$-cov by matching the first source and target covariates via the covariance matching penalty~\eqnref{cov_distance}. Then it imposes the constraint
    \begin{align*}
    	\EE{\beta^\top \Xsrcone\Xsrcone{}^\top \beta} = \EE{\beta^\top \Xtar\Xtar{}^\top \beta}  &\iff \beta^\top (\bsrcone \bsrcone{}^\top-\btar \btar{}^\top) \beta =0,\\
    	&\iff \beta^\top \begin{pmatrix}
    		0 & b_{[1]}b_{[2]} \\
    		b_{[1]}b_{[2]} & 0
    	\end{pmatrix} \beta = 0.
    \end{align*}
    Both $\beta=(1,0)^\top$ and $\beta=(0,1)^\top$ are valid solutions to the above equation. However, while  $\beta=(1,0)^\top$ correctly identifies the conditionally invariant component $X_{[1]}$, choosing $\beta=(0,1)^\top$ selects only the second covariate $X_{[2]}$, which reverses the sign of the correlation with $Y$ across the first source and target domains. This results in significantly degraded target prediction performance. A similar issue arises with DIP$^{(2)}$-cov, which matches the second source and target covariates, where other solutions besides the conditionally invariant component can have larger target risk.

	\end{appendix}

\bibliographystyle{plainnat} 
\bibliography{reference.bib}

\end{document}